\documentclass[12pt]{ociamthesis}  

\usepackage{setspace}
\usepackage{latexsym}
\usepackage{lipsum}

\usepackage{amsmath}
\usepackage{amsfonts}
\usepackage{amssymb} 
\usepackage{amsthm}

\usepackage{tikz}

\usepackage{subfigure}
\usepackage{lipsum}
\usepackage{gb4e}
\usepackage{qtree}
\usepackage{epigraph}
\usepackage[linktocpage]{hyperref}
\usepackage{ccg}
\usepackage{multirow}

\usepackage{algorithmic}
\usepackage[boxed,chapter]{algorithm}
\usepackage{makeidx}

\newcommand{\ten}{\otimes}
\newcommand{\sem}[1]{[\![ #1 ]\!]}
\newcommand{\ov}{\overrightarrow}
\newcommand{\ol}{\overline}
\newcommand{\op}{\operatorname}

\newcommand{\ket}[1]{| #1 \rangle}
\newcommand{\bra}[1]{\langle #1 |}
\newcommand{\name}[1]{\ulcorner #1 \urcorner}

\newcommand{\M}[1]{\mathcal{#1}}

\newenvironment{chabstract}{\begin{quote}\begin{center}\textbf{Chapter Abstract}\end{center}\singlespace\it}{\begin{center}\rule{0.7\textwidth}{1px}\end{center}\end{quote}}
\theoremstyle{definition}
\newtheorem{theorem}{Theorem}[section]

\newtheorem{definition}[theorem]{Definition}

\pgfdeclarelayer{edgelayer}
\pgfdeclarelayer{nodelayer}
\pgfsetlayers{edgelayer,nodelayer,main}

\tikzstyle{none}=[inner sep=0pt]
\tikzstyle{plain}=[inner sep=0pt]
\tikzstyle{every picture}=[baseline=(current bounding box).east,scale=0.5,node distance=5mm]

\newcommand{\ctikzfig}[1]{%
\begin{center}\rm
  
\InputIfFileExists{./tikz/#1.tikz}{}{\input{./tikz/#1.tikz}}

\end{center}}

\title{Compositional Distributional\\[1ex]Semantics with Compact Closed\\[1ex]Categories and Frobenius Algebras}   

\makeindex

\author{Dimitrios Kartsaklis}     
\college{Wolfson College}         

\degree{Doctor of Philosophy}     
\degreedate{Michaelmas 2014}      

\begin{document}


\setcounter{secnumdepth}{3}
\setcounter{tocdepth}{3}
\pagenumbering{gobble}

\maketitle                  

\thispagestyle{empty}
\begin{dedication}
~\\To my mother, Chryssa, who fought all her life for her children\\to work ``some place with a roof over their heads''\\ \vspace{0.5cm}
To my father, Euripides, who taught me to love technology\\and passed me down the genes to work efficiently with it
\end{dedication}        
\mbox{}
\thispagestyle{empty}
\newpage
\begin{acknowledgements}

This thesis would not be possible without the guidance and support of my immediate supervisor Mehrnoosh Sadrzadeh. I am grateful to her for giving me the opportunity to work in one of the top universities in the world; for introducing me to this new perspective of NLP, full of objects and arrows; for her patience and guidance; for treating me not as a student, but as a collaborator; above all, though, I thank her for being to me a true friend. 

Just being around to people like Bob Coecke and Stephen Pulman (my other two supervisors) is sufficient reason for anyone to quit his job and move 3,500 kilometers from home to a foreign country. You have so many things to learn from them, but what struck me the most was this simple fact of life: the \textit{really} important people don't have to try hard to prove who they are. I am grateful to both of them for their invaluable help during the course of this project.

I owe a big and heartfelt ``thank you'' to Samson Abramsky, Phil Blunsom, Steve Clark, Mirella Lapata, Anne Preller and Matthew Purver for all the advice,  suggestions and comments they kindly offered to me. I would also like to thank my colleagues in Oxford Nal Kalchbrenner, Pengyu Wang and Ed Grefenstette. It would be an obvious omission not to mention my good friend Andreas Chatzistergiou who, although in Edinburgh, kept me company all this time via Google chat; I'm really grateful for his friendship and support. 

Last, I would like to thank my family for their unconditional support to my decision to become a researcher. If there is any reason at all to make me regret the remarkable experience I lived in Oxford, this would be that my place during these difficult times was there with them, offering my help by any means I could use.

\vspace{0.5cm}
This research was supported by EPSRC grant EP/F042728/1.

\end{acknowledgements}
\mbox{}
\thispagestyle{empty}
\newpage
\begin{abstract}

The provision of compositionality in distributional models of meaning, where a word is represented as a vector of co-occurrence counts with every other word in the vocabulary, offers a solution to the fact that no text corpus, regardless of its size, is capable of providing reliable co-occurrence statistics for anything but very short text constituents. The purpose of a \textit{compositional distributional model} is to provide a function that composes the vectors for the words within a sentence, in order to create a vectorial representation that reflects its meaning. Using the abstract mathematical framework of category theory, Coecke, Sadrzadeh and Clark showed that this function can directly depend on the grammatical structure of the sentence, providing an elegant mathematical counterpart of the formal semantics view. The framework is general and compositional but stays abstract to a large extent.

This thesis contributes to ongoing research related to the above categorical model in three ways: Firstly, I propose a concrete instantiation of the abstract framework based on Frobenius algebras (joint work with Sadrzadeh). The theory improves shortcomings of previous proposals, extends the coverage of the language, and is supported by experimental work that improves existing results. The proposed framework describes a new class of compositional models that find intuitive interpretations for a number of linguistic phenomena.

Secondly, I propose and evaluate in practice a new compositional methodology which explicitly deals with the different levels of lexical ambiguity (joint work with Pulman). A concrete algorithm is presented, based on the separation of vector disambiguation from composition in an explicit prior step. Extensive experimental work shows that the proposed methodology indeed results in more accurate composite representations for the framework of Coecke et al. in particular and every other class of compositional models in general.

As a last contribution, I formalize the explicit treatment of lexical ambiguity in the context of the categorical framework by resorting to categorical quantum mechanics (joint work with Coecke). In the proposed extension, the concept of a distributional vector is replaced with that of a density matrix, which compactly represents a probability distribution over the potential different meanings of the specific word. Composition takes the form of quantum measurements, leading to interesting analogies between quantum physics and linguistics. 

\end{abstract}

\mbox{}
\thispagestyle{empty}
\newpage

\begin{romanpages}          
\tableofcontents            
\mbox{}
\newpage
\mbox{}
\newpage
\listoffigures              
\mbox{}\newpage
\mbox{}\newpage
\listoftables
\end{romanpages}            

\onehalfspacing

\mbox{}
\thispagestyle{empty}
\newpage

\clearpage
\pagenumbering{arabic}
\setcounter{page}{1}

\chapter{Introduction}
\label{ch:intro}

\setlength{\epigraphwidth}{12cm}
\epigraph{
``When \textit{I} use a word,'' Humpty Dumpty said, in rather a scornful tone, ``it means just what I choose it to mean---neither more nor less.''}
{Lewis Carroll, \textit{Through the Looking-Glass} (1871)}

\noindent 
Language serves to convey \textit{meaning}. Above anything else, therefore, the ultimate and long-standing goal of every computational linguist is to devise an efficient way to represent this meaning as faithfully as possible within a computer's memory. Every manifestation of what we call natural language processing---sentiment analysis, machine translation, paraphrase detection, to name just a few---culminates in some form of this endeavour. Not a trivial task, to say the least, made harder to achieve by the fact that language is dominated by ambiguity. The means of verbal or written communication---words combined in phrases and sentences according to the rules of some grammar---are rarely monosemous; almost every word has more than one dictionary definition, while in many cases the true meaning of a specific utterance can be intimately related to subjective factors such as beliefs, education, or social environment of the sayer. In order to cope with these difficulties, computational linguists employ a large arsenal of clever techniques, but, first and foremost, they have to provide an answer for a question much more philosophical than technical: \textit{what is meaning} after all?\index{meaning}

At the sentence level, the traditional way of approaching this important problem is a syntax-driven compositional method that follows Montague's tradition: every word in the sentence is associated with a primitive symbol or a predicate, and these are combined into larger and larger logical forms based on the syntactical rules of the grammar. At the end of the syntactical analysis, the logical representation of the whole sentence is a complex formula that can be fed to a theorem prover for further processing. Although such an approach seems intuitive, it has two important drawbacks: First, the resulting composite formula can only state whether the sentence in question is true or false, based on the initial values of the constituents therein; second, it captures the meaning of the atomic units (words) in an axiomatic way, namely by ad-hoc unexplained primitives that have nothing to say about the real semantic value of the specific words. 

On the other hand, distributional models of meaning work by building co-occurren\-ce \textit{vectors} for every word in a corpus based on its context, following Firth's\index{Firth, John Rupert} famous intuition that ``you should know a word by the company it keeps'' \cite{Firth}. These models have been proved useful in many natural language tasks \cite{Curran,Landauer,Manning,Schutze} and can provide concrete information for the words of a sentence, but they do not naturally scale up to larger constituents of text, such as phrases or sentences. The reason behind this is that a sentence or phrase is something much more unique than a word; as a result, there is currently no text corpus available that can provide reliable co-occurrence statistics for anything larger than two- or three-word text segments. 

Given the complementary nature of these two distinct approaches, it is not a surprise that compositional abilities of distributional models have been the subject of much discussion and research in recent years.  Towards this purpose researchers exploit a wide variety of techniques, ranging from simple mathematical operations like vector addition and multiplication to neural networks and even category theory. No matter what means is utilized, the goal is always the same: although it is not possible to directly construct a sentence vector, we can still \textit{compose} the vectors for the words within the sentence in question, providing a vectorial representation that approximates some idealistic behaviour from a semantic perspective. In other words, the purpose of a \textit{compositional distributional model} (CDM)\index{CDMs}\index{compositional distributional models|see {CDMs}} of meaning is, given a sentence $w_1w_2\hdots w_n$, to provide a function $f$ as follows: 

\begin{equation}
  \ov{s} = f(\ov{w_1},\ov{w_2},\hdots,\ov{w_n})
  \label{equ:main}
\end{equation}

\noindent
where $\ov{w_i}$ refers to the vector of the $i$th word and $\ov{s}$ is a vector serving as a semantic representation for the sentence.

The subject of this thesis is CDMs in general, and a specific class of models based on category theory in particular. Using the abstract setting of compact closed categories, Coecke, Sadrzadeh and Clark \cite{Coeckeetal} showed that under certain conditions grammar and vector spaces are structurally homomorphic; in other words, a grammatical derivation can be translated to some algebraic manipulation between vector spaces, which provides the desired passage that Eq. \ref{equ:main} aims to achieve. The categorical framework provides a theoretical justification for a certain class of CDMs that is now known as \textit{tensor-based} models, the main characteristic of which is that they are \textit{multi-linear} in nature. In a tensor-based model, relational words such as verbs and adjectives are functions represented by multi-linear maps (\textit{tensors} of various orders) that act on their arguments (usually vectors representing nouns) through a process known as tensor contraction. In this sense, the framework can be seen as an elegant mathematical counterpart of the formal semantics perspective, as this was expressed by Montague \cite{Mon1}, Lambek \cite{lambek1958} and other pioneers of the formal approaches to language. 

To this date, the categorical structures of \cite{Coeckeetal} remain largely abstract. The only extensive study on the subject comes from Grefenstette \cite{GrefenstetteThesis2013}, who provided valuable insights regarding the construction of tensors for transitive verbs, as well as a first implementation for simple text constructs. In many aspects, the present work is a direct extension of that original contribution. It essentially starts where Grefenstette's work stops, having as its main purpose to help realizing the abstract categorical structures in terms of concrete instantiations and make them applicable to mainstream natural language processing tasks. 

The first contribution of this work is that it provides a robust and scalable framework for instantiating the tensors of relational and functional words based on the application of Frobenius algebras which, as shown in \cite{CoeckeVic}, offer a canonical way for uniformly copying or deleting the basis of a vector space. This proposal provides a way to overcome certain shortcomings of the earlier implementation, such as the fact that sentences with nested grammatical structures could not be assigned a meaning. Furthermore, we will see how the Frobenius operators can be used for modelling a number of functional words (for example prepositions, conjunctions and so on) which until now were considered as semantically vacuous and ignored by most CDMs. The introduction of Frobenius algebras in language results in a new class of CDMs, in which features from both tensor-based models and element-wise compositional models relying on vector multiplication are combined into a unified framework. As will become evident, this unique characteristic finds intuitive interpretations for a number of linguistic phenomena, such as intonation and coordination. The Frobenius framework builds on collaboration with Mehrnoosh Sadrzadeh and our published work in \cite{kartsaklis2012,kartsaklis2014}.

An equally important contribution of the current thesis is that it presents the first large-scale study to date regarding the behaviour of CDMs under the presence of lexical ambiguity. On the theory side, I discuss the different levels of ambiguity in relation to compositionality, and make certain connections with psycholinguistic models that attempt to describe how the human brain behaves in similar settings. One of the main arguments of this thesis is that in general a CDM cannot handle every degree of ambiguity in the same way; a certain distinction must be made between cases of homonymy (genuinely ambiguous words with two or more unrelated meanings) and polysemy (small deviations between the senses of a word), which can be realized by the introduction of an explicit disambiguation step on the word vectors of a sentence before the actual composition. The validity of the proposed methodology is verified by extensive experimental work, which provides solid evidence that the separation of disambiguation from composition in two distinct steps leads to more accurate composite representations. Furthermore, as we will see, the applicability of this method is not restricted to tensor-based models, but it has very positive effects in every class of CDMs. The prior disambiguation hypothesis originated and took shape mainly from discussions and collaboration with Stephen Pulman \cite{kartsaklis:2013:CoNLL}; the experimental evidence was presented in various subsequent publications \cite{kartsaklis:2013:EMNLP,kartsaklis:2014:ACL,cheng2014}. 

The fact that lexical ambiguity (read: uncertainty regarding the true meaning under which a word is used in a specific context) dominates language to a such a great extent implies that any semantic model should have some means of incorporating this notion. Interestingly, the framework of compact closed categories that comprises the basis of the categorical model has been used in the past by Abramsky and Coecke \cite{abramsky2004} for reasoning in a different field with vector space semantics, that of quantum mechanics. The relation between quantum mechanics and linguistics implied by this connection becomes concrete in Chapter \ref{ch:ambiguity}, where the informal discussion about composition and lexical ambiguity takes shape in quantum-theoretic terms. Specifically, the categorical model of \cite{Coeckeetal} is advanced such that the notion of a state tensor is replaced by that of a density matrix, an elegant compact way to represent a probability distribution. The new model is capable of handling the different levels of ambiguity imposed by homonymy and polysemy in a unified and linguistically motivated way, offering overall a richer semantic representation for language. The formalization of lexical ambiguity in categorical quantum mechanics terms was collaborative work with Bob Coecke \cite{piedeleu2015}.

\vspace{0.2cm}
The dissertation is structured in the following way:
\vspace{0.2cm}

Part \ref{prt:background} aims at providing the necessary background to the reader. Chapter \ref{ch:litreview} starts with a concise introduction to compositional and distributional models of meaning. Furthermore, the most important attempts at unifying the two models are briefly presented, and a hierarchy of CDMs is devised according to their theoretical power and various other characteristics. Chapter \ref{ch:framework} explains in some detail the categorical compositional framework of \cite{Coeckeetal} which is the main subject of this work. The reader is introduced to the notions of compact closed categories, pregroup grammars, and Frobenius algebras; furthermore, I present the convenient graphical language of monoidal categories, which will serve as one of my main tools for expressing the ideas in this thesis.

Part \ref{prt:theory} presents the main theoretical contributions of the current thesis. Chapter \ref{ch:frobverbs} introduces Frobenius algebras in language, discusses the theoretical intuition behind it, and examines the effect of the resulting model in relation to a number of linguistic phenomena. Chapter \ref{ch:extend} extends this discussion to functional words such as prepositions, conjunctions, and infinitive particles. Chapter \ref{ch:ambiguity} deals with the vast topic of lexical ambiguity in relation to composition. After an informal discussion that explains the linguistic motivation, section \ref{sec:quantum} proceeds to incorporate ambiguity in the framework of \cite{Coeckeetal} in terms of categorical quantum mechanics and density matrices. 

Part \ref{prt:practice} is dedicated to practical aspects of this research and experimental work. Chapter \ref{ch:frobexp} provides a first evaluation of the Frobenius models in three tasks involving head verb disambiguation, verb phrase similarity, and term/definition classification with promising results. The purpose of Chapter \ref{ch:wsdexp} is two-fold: first, a concrete methodology for implementing the prior disambiguation step for relational tensors is detailed and evaluated; second, the prior disambiguation hypothesis is extensively tested and verified for every class of CDMs. 

Finally, Chapter \ref{ch:conclusions} summarizes the main contributions of this thesis, and briefly discusses the (many) opportunities for future research on the topic of tensor-based compositional models. 


\vspace{-0.2cm}
\subsection*{Related published work}
\index{related published work|(}
\vspace{-0.1cm}

The material I am going to present in this thesis builds on a large number of publications that took place in the course of three years, and inevitably some passages (especially describing background material in Part \ref{prt:background} or experimental settings in Part \ref{prt:practice}) have been re-used verbatim from these original sources. The publications that are relevant to a specific chapter are cited in the abstract of that chapter. Furthermore, I provide here a list of the related published work in chronological order:

\vspace{-0.1cm}
\begin{enumerate}

\item Kartsaklis, Sadrzadeh, and Pulman. A Unified Sentence Space for Categorical Distributional Compositional Semantics: Theory and Experiments. In {\em Proceedings of 24th International Conference on Computational Linguistics (COLING 2012): Posters}, Mumbai, India, December  2012. 

\item Kartsaklis, Sadrzadeh, Pulman, and Coecke. Reasoning about Meaning in Natural Language with Compact Closed Categories and Frobenius Algebras. In J.~Chubb, A.~Eskandarian, and V.~Harizanov, editors, {\em Logic and Algebraic Structures in Quantum Computing and Information}, Association for Symbolic Logic Lecture Notes in Logic. Cambridge University Press, 2015.

\item Kartsaklis, Sadrzadeh, and Pulman. Separating {D}isambiguation from {C}omposition in {D}istributional {S}emantics. In {\em Proceedings of 17th Conference on Computational Natural Language Learning (CoNLL-2013)}, Sofia, Bulgaria, August 2013.

\item Kartsaklis and Sadrzadeh. Prior Disambiguation of Word Tensors for Constructing Sentence Vectors. In {\em Proceedings of the 2013 Conference on Empirical Methods in Natural Language Processing (EMNLP)}, Seattle, USA, October 2013.

\item Kartsaklis. Compositional Operators in Distributional Semantics. {\em Springer Science Reviews}, 2(1-2):161-177, 2014. 

\item Kartsaklis, Kalchbrenner, and Sadrzadeh. Resolving Lexical Ambiguity in Tensor Regression Models of Meaning. In {\em Proceedings of the 52nd Annual Meeting of the Association for Computational Linguistics (ACL): Short Papers}, Baltimore, USA, June 2014.

\item Kartsaklis and Sadrzadeh. A Study of Entanglement in a Categorical Framework of Natural Language. In B. Coecke, I.Hasuo and P. Panangaden, editors, {\em Proceedings of the 11th Workshop of Quantum Physics and Logic (QPL)}, EPTCS 172, Kyoto, Japan, June 2014.

\item Milajevs, Kartsaklis, Sadrzadeh and Purver. Evaluating Neural Word Representations in Tensor-Based Compositional Settings. In {\em Proceedings of the 2014 Conference on Empirical Methods in Natural language Processing (EMNLP)}, Doha, Qatar, October 2014.

\item Cheng, Kartsaklis and Grefenstette. Investigating the Effect of Prior Disambiguation in Deep-Learning Compositional Models of Meaning. \textit{Learning Semantics} workshop, NIPS 2014, Montreal, Canada, December 2014.

\end{enumerate}

\index{related published work|)}


\part{Background}
\label{prt:background}
\mbox{}\newpage

\chapter{Compositionality in Distributional Models of Meaning}
\label{ch:litreview}

\begin{chabstract}
In this chapter the reader is introduced to the two prominent semantic paradigms of natural language: compositional semantics and distributional models of meaning. Then I proceed to the central topic of this thesis, CDMs, presenting a survey of the most important attempts to date for providing compositionality in distributional models. I conclude by putting together a taxonomy of CDMs based on composition function and various other characteristics. Material is based on my review article in \cite{KartsaklisSpringer}.
\end{chabstract}

\section{Compositionality in language}
\label{sec:compsem}

Compositionality in semantics offers an elegant way to address the inherent property of natural language to produce an infinite number of structures (phrases and sentences) from finite resources (words). The \textit{principle of compositionality}\index{principle of compositionality} states that the meaning of a complex expression can be determined by the meanings of its constituents and the rules used for combining them. This idea is quite old, and glimpses of it can be spotted even in works of Plato\index{Plato}. In his dialogue \textit{Sophist}, Plato argues that a sentence consists of a noun and a verb, and that the sentence is true if the verb denotes the action that the noun is currently performing. In other words, Plato argues that (a) a sentence has a structure; (b) the parts of the sentence have different functions; (c) the meaning of the sentence is determined by the function of its parts. Nowadays, this intuitive idea is often attributed to Gottlob Frege,\index{Frege, Gottlob} although it was never explicitly stated by him in some of his published works. In an undated letter to Philip Jourdain, though, included in ``Philosophical and Mathematical Correspondence'' \cite{fregeletter}, Frege justifies the reasoning behind compositionality by saying:

\begin{quote}
The possibility of our understanding propositions which we have never heard before rests evidently on this, that we can construct the sense of a proposition out of parts that correspond to words.
\end{quote}

This forms the basis of the \textit{productivity} argument,\index{productivity argument} often used as a proof for the validity of the principle: humans only know the meaning of words, and the rules to combine them in larger constructs; yet, being equipped with this knowledge, we are able to produce new sentences that we have never uttered or heard before. Indeed, this task seems natural even for a 3-year old child---however, its formalization in a way reproducible by a computer has been proven anything but trivial. The modern compositional models owe a lot to the seminal work of Richard Montague (1930-1971),\index{Montague, Richard} who managed to present a systematic way of processing fragments of the English language in order to get semantic representations capturing their ``meaning'' \cite{Mon1,Mon2,Mon3}. 

%

In ``Universal Grammar'' \cite{Mon2}, Montague starts detailing a systematization of the natural language, an approach that became known as Montague grammar.\index{Montague grammar} To use Montague's method, one would need two things: first, a resource which will provide the logical forms of each specific word (a lexicon); and second, a way to determine the correct order in which the elements in the sentence should be combined in order to end up with a valid semantic representation. A natural way to address the latter is to use the syntactic structure as a means of driving the semantic derivation (an approach called \textit{syntax-driven semantic analysis}). In other words, we assume that there exists a mapping from syntactic to semantic types, and that the composition at the syntax level implies a similar composition at the semantic level. This is known as the \textit{rule-to-rule hypothesis} \cite{Bach:76}.\index{rule-to-rule hypothesis}\index{semantics, correspondence to syntax} 

In order to provide an example, I will use the sentence `Every man walks'. I begin from the lexicon, the job of which is to assign a grammar type and a logical form to every word in the sentence:

\singlespacing
\begin{exe}
\ex 
\begin{xlist}
\ex\label{every} every $\vdash Dt : \lambda P.\lambda Q.\forall x[P(x) \to Q(x)]$
\ex\label{man} man $\vdash N : \lambda y.man(y)$
\ex\label{walks} walks $\vdash  V_{I} : \lambda z.walk(z)$
\end{xlist}
\end{exe}
\onehalfspacing

The above use of formal logic (especially higher-order) in conjunction with $\lambda$-calculus\index{$\lambda$-calculus} was first introduced by Montague, and from then on it constitutes the standard way of providing logical forms to compositional models. In the above lexicon, predicates of the form $man(y)$ and $walk(z)$ are true if the individuals denoted by $y$ and $z$ carry the property (or, respectively, perform the action) indicated by the predicate. From an extensional perspective,\index{extensional semantics} the semantic value of a predicate can be seen as the set of all individuals that carry a specific property. Hence, given a universe of elements $\mathcal{U}$ we define the semantic value of \textit{walk} as follows:

\begin{equation}
   \sem{walk} = \{x | x \in \mathcal{U} \wedge x~\text{walks} \}
\end{equation}

The expression $walk(john)$ will be true if (and only if) the individual denoted by $john$ belongs to the set of all individuals who perform the action of walking; that is, iff $\sem{john} \in \sem{walk}$. Furthermore, $\lambda$-terms like $\lambda x$ or $\lambda Q$ have the role of placeholders that remain to be filled. The logical form $\lambda y.man(y)$, for example, reflects the fact that the entity which is going to be tested for the property of manhood is still unknown and it will be later specified based on the syntactic combinatorics. Finally, the form in (\ref{every}) reflects the traditional way for representing a universal quantifier in natural language, where the still unknown parts are actually the predicates acting over a range of  entities.

In $\lambda$-calculus, function application is achieved via the process of $\beta$-reduction:\index{$\beta$-reduction} given two logical forms $\lambda x.t$ and $s$, the application of the former to the latter will produce a version of $t$ where all the free occurrences of $x$ in $t$ have been replaced by $s$. More formally:

\begin{equation}
  (\lambda x.t)s \to_\beta t[x:=s]
\end{equation}

\begin{figure}[h!]
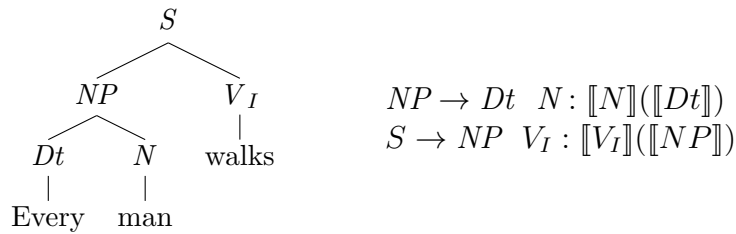

\begin{center}
\begin{minipage}{0.40\linewidth}
\small
\Tree [ .\textit{S} [ .\textit{NP} [ .\textit{Dt} Every ] [.\textit{N} man ] ] [.\textit{V}_{I} walks ] ] 
\normalsize
\end{minipage}
\begin{minipage}{0.40\linewidth}
  $\textit{NP} \to \textit{Dt~~N}: \sem{N}(\sem{Dt})$ \\
  $S \to \textit{NP}~~V_{I}: \sem{V_I}(\sem{NP})$
\end{minipage}
\end{center}
\caption{Syntax and semantics correspondence.}
\label{fig:comp}
\end{figure}

Let us see how we can use the principle of compositionality to get a logical form for the above example sentence, by applying $\beta$-reduction between the semantic forms of text constituents following the grammar rules. Fig. \ref{fig:comp} provides us a syntactic analysis and the corresponding grammar rules augmented with semantic applications.

\singlespacing
\onehalfspacing

The rules on the right part of Fig. \ref{fig:comp} will drive the semantic derivation, the full form of which can be found in Fig. \ref{fig:semder}. At the top of the tree you can see the logical form that is produced by the $\beta$-reduction process and serves as a semantic representation for the whole sentence.

\singlespacing
\begin{figure}[h!]
\footnotesize
\Tree [ .{\textit{S}\\$\forall x[man(x) \to walk(x)]$} [ .{\textit{NP}\\$\lambda Q.\forall x[man(x) \to Q(x)]$} [ .{\textit{Dt}\\$\lambda P.\lambda Q.\forall x[P(x) \to Q(x)]$} Every ] [.{\textit{N}\\$\lambda y.man(y)$} man ] ] [.{\textit{V}_{I}\\$\lambda z.walk(z)$} walks ] ] 
\normalsize
\caption{A semantic derivation for a simple sentence.}
\label{fig:semder}
\end{figure}
\onehalfspacing

A logical form such as $\forall x[man(x) \to walk(x)]$ simply states the truth (or falsity) of the expression given the sub-expressions and the rules for composing them. It does not provide any quantitative interpretation of the result (e.g. grades of truth) and, more importantly, leaves the meaning of words as unexplained primitives ($man$, $walk$ etc). In the next section I will show how distributional semantics can fill this gap.

\section{Distributional models of meaning}
\label{sec:dissem}

\index{distributional models|(}

\subsection{Meaning is context}
\label{sec:disthyp}

The distributional paradigm is based on the \textit{distributional hypothesis}\index{distributional hypothesis} (Harris \cite{Harris}),\index{Harris, Zellig} stating that words that occur in the same context have similar meanings. Various forms of this popular idea keep recurring in the literature: Firth\index{Firth, John Rupert} calls it collocation \cite{Firth}, while Frege\index{Frege, Gottlob} himself states that ``never ask for the meaning of a word in isolation, but only in the context of a proposition''\footnote{Interestingly, a literal interpretation of this statement leads to the conclusion that the meaning of an isolated word cannot be defined. This idea, a more generic form of which is referred to as \textit{principle of contextuality}, contradicts compositionality, which requires the assignment of meaning to each part of a whole. The question regarding which of the two principles Frege actually embraced and to what extent has not yet found a satisfactory answer, as the study of \cite{pelletier} shows.} \cite{frege1980foundations}. The attraction of this principle in the context of Computational Linguistics is that it provides a way of concretely representing the meaning of a word via mathematics: each word is a vector whose elements show how many times this word occurred in some corpus in the same context with every other word in the vocabulary.\index{words!as vectors} If, for example, our basis is $\{cute,sleep,finance,milk\}$, the vector for word `cat' could have the form $(15,7,0,22)$ meaning that `cat' appeared 15 times together with `cute', 7 times with `sleep' and so on. More formally, given an orthonormal basis $\{\ov{n_i}\}_i$ for our vector space, a word is represented as:

\begin{equation}
  \ov{word} = \sum_i c_i\ov{n_i}
\end{equation}

\noindent
where $c_i$ is the coefficient for the $i$th basis vector. As mentioned above, in their simplest form these coefficients can be just co-occurrence counts, although in practice a function on raw counts is often used in order to remove some of the unavoidable frequency bias. A well-known measure is the information-theoretic \textit{point-wise mutual information} (PMI),\index{point-wise mutual information} which can reflect the relationship between a context word $c$ and a target word $t$ as follows:

\begin{equation}
  \text{PMI}(c,t) = \log\frac{p(c|t)}{p(c)}
\end{equation}

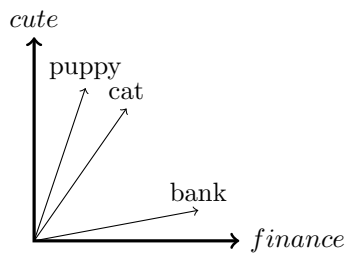
\begin{figure}[h]
 \centering
 \footnotesize
 \begin{tikzpicture}[scale=2.7]
    \draw [<->,very thick] (0,2) node (yaxis) [above] {$cute$}
        |- (2,0) node (xaxis) [right] {$finance$};
    \draw [->] (0,0) -- (0.5,1.5) node (puppy) [above] {puppy};
    \draw [->] (0,0) -- (0.9,1.3) node (cat) [above] {cat};
    
    \draw [->] (0,0) -- (1.6,0.3) node (bank) [above] {bank};
 \end{tikzpicture}
 \normalsize
\caption{A toy ``vector space'' for demonstrating semantic similarity between words.}
\label{fig:sim}
\end{figure}

In contrast to compositional semantics which leaves the meaning of lexical items unexplained, a vector like the one used above for `cat' provides some concrete information about the meaning of the specific word: cats are cute, sleep a lot, they really like milk, and they have nothing to do with finance. Additionally, this quantitative representation allows us to compare the meanings of two words, e.g. by computing the cosine distance of their vectors, and evaluate their semantic similarity. The cosine distance\index{cosine distance} is a popular choice for this task (but not the only one) and is given by the following formula:

\begin{equation}
  \text{sim}(\ov{v},\ov{u}) = \cos({\ov{v},\ov{u}}) = \frac{ \langle \ov{v} | \ov{u} \rangle}{\Vert\ov{v}\Vert \Vert\ov{u}\Vert}
\end{equation}

\noindent
where $\langle \ov{v} | \ov{u} \rangle$ denotes the dot product between $\ov{v}$ and $\ov{u}$, and $\Vert \ov{v} \Vert$ the magnitude of $\ov{v}$. As an example, in the 2-dimensional vector space of Fig. \ref{fig:sim} we see that `cat' and `puppy' are close together (and both of them closer to the basis `cute'), while `bank' is closer to basis vector `finance'.

\begin{figure}[h!]
 \centering
 \fbox{\includegraphics[scale=0.51,trim = 5.7cm 7cm 8cm 3.5cm, clip]{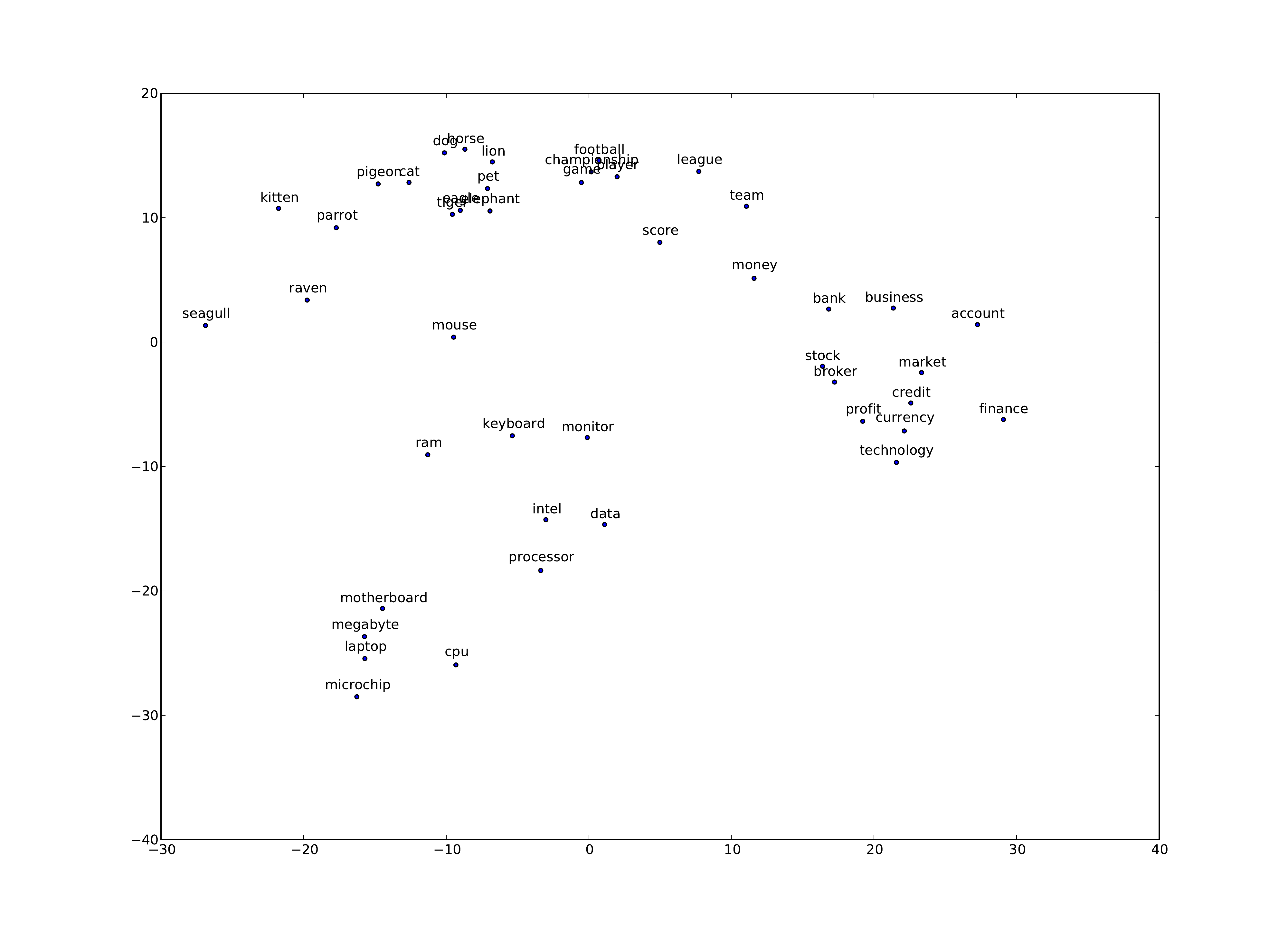}}
 \caption[Visualization of a word space in two dimensions]{Visualization of a word space in two dimensions (original vectors are 2000-dimensional vectors created from BNC).}
 \label{fig:proj}
\end{figure}

A more realistic example is shown in Fig. \ref{fig:proj}. The points in this space represent real distributional word vectors created from the British National Corpus (BNC)\footnote{The BNC is a 100 million-word text corpus consisting of samples of written and spoken English. It can be found online at \texttt{http://www.natcorp.ox.ac.uk/}.},\index{British National Corpus (BNC)} originally 2,000-dimensional and projected onto two dimensions for visualization. Note how words form distinct groups of points according to their semantic correlation. Furthermore, it is interesting to see how ambiguous words\index{ambiguity} behave in these models: the ambiguous word `mouse' (with the two meanings to be that of a rodent and of a computer pointing device), for example, is placed almost equidistantly from the group related to IT concepts (lower left part of the diagram) and the animal group (top left part of the diagram), having a meaning that can be indeed seen as the average of both senses.

\subsection{Forms of word spaces}
\label{sec:vectorspaces}

\index{vector spaces|see {semantic spaces}}
\index{semantic spaces|(}

In the simplest form of a word space (a vector space for words), the context of a word is a set containing all the words that occur within a certain distance from the target word, for example within a 5-word window. Models like these have been extensively studied and implemented the past years, see for example \cite{lowe2001towards,lowe2000direct}. However, although such an approach is simple, intuitive and computationally efficient, it is not optimal, since it assumes that every word within this window will be semantically relevant to the target word, while treating every word outside of the window as irrelevant. Unfortunately, this is not always the case. Consider for example the following phrase: 

\begin{exe}
\ex The movie I saw and John said he really likes
\end{exe}

Here, a word-based model cannot efficiently capture the long-range dependency between `movie' and `likes'; on the other hand, since `said' is closer to `movie' it is more likely to be considered as part of its context, despite the fact that the semantic relationship between the two words is actually weak. Cases like the above suggest that a better approach for the construction of the model would be to take into account not just the surface form of context words, but also the specific grammatical relations that hold between them and the target word. A model like this, for example, will be aware that `movie' is the object of `likes', so it could safely consider the latter as part of the context for the former, and vice versa. This kind of observation motivated many researchers to experiment with vector spaces based not solely on the surface forms of words but also on various syntactical properties of the text.

One of the earliest attempts to add syntactical information in a word space was that of Grefenstette \cite{grefenstette1994}, who used a structured vector space with a basis constructed by grammatical properties such as `subject-of-buy' or `argument-of-useful', denoting that the target word has occurred in the corpus as subject of the verb `buy' or as argument of the adjective `useful'. The weights of a word vector were binary, either 1 for at least one occurrence or 0 otherwise. Lin \cite{lin1998} moves one step further, replacing the binary weights with frequency counts. Following a different path, Erk and Pad\'o \cite{ErkPado} argue that a single vector is not enough to catch the meaning of a word; instead, the vector of a word is accompanied by a set of vectors $R$ representing the lexical preferences of the word for its arguments positions, and a set of vectors $R^{-1}$, denoting the inverse relationship, that is, the usage of the word as argument in the lexical preferences of other words. Subsequent works by Thater et al. \cite{thater2010,thater2011word} present a version of this idea extended with the inclusion of grammatical dependency contexts. Finally, in an attempt to provide a generic distributional framework, Pad\'o and Lapata \cite{pado2007} presented a model based on dependency relations between words. The interesting part of this work is that, given the proper parametrization, it is indeed able to essentially subsume a large amount of other works on the same subject.

\index{semantic spaces|)}
\index{distributional models|)}

\subsection{Neural word embeddings}
\label{sec:neural-embeddings}

\index{neural word embeddings|(}

Recently, a new class of distributional models in which the training of word representations is performed by some form of neural network architecture has received a lot of attention, showing state-of-the-art performances in many natural language processing tasks. \textit{Neural word embeddings} \cite{bengio2006,collobert2008,mikolov2013efficient} interpret the distributional hypothesis differently: instead of relying on observed co-occurrence frequencies, a neural model is trained to maximise some probabilistic objective function; in the skip-gram model of Mikolov et al. \cite{mikolov2013distributed}, for example, this function is related to the probability of observing the surrounding words in some context, and its objective is to maximize the following quantity:

\vspace{-0.5cm}
\begin{align}
 \frac{1}{T}\sum^{T}_{t=1}\sum_{-c \leq j \leq c, j\neq0} \log p(w_{t+j}|w_t)
  \label{eq:objective-func}
\end{align}

Specifically, the function maximises the conditional probability of observing words in a context around the target word $w_t$, where $c$ is the size of the training context, and $w_1 w_2\cdots w_T$ is a sequence of training words. A model like this therefore captures the distributional intuition and can express degrees of lexical similarity. However, it works differently from a typical co-occurrence distributional model in the sense that, instead of simply counting the context words, it actually \textit{predicts} them. Compared to the co-occurrence method, this has the obvious advantage that the model in principle can be much more robust to data sparsity problems, which is always an important issue for traditional word spaces. Indeed, a number of studies \cite{baroni2014don,levy2014linguistic,milajevs2014} suggest that neural vectors reflect better than their ``counting'' counterparts the semantic relationships between the various words. 

\index{neural word embeddings|)}

\subsection{A philosophical digression}
\label{sec:meaning}

\index{Wittgenstein, Ludwig|(}
\index{meaning|(}

To what extent is the distributional hypothesis correct? Even if we accept that a co-occurrence vector can indeed capture the ``meaning'' of a word, it would be far too simplistic to assume that this holds for \textit{every} kind of word. The meaning of some words can be determined by their denotations; one could claim, for example, under a Platonic view of semantics, that the meaning of the word `tree' is the set of all trees, and we are even able to answer the question ``what is a tree?'' by pointing to a member of this set, a technique known as \textit{ostensive} definition.\index{ostensive definition} But this is not true for all words. In ``Philosophy'' (published in ``Philosophical Occasions: 1912-1951'', \cite{witphilosophy}), Ludwig Wittgenstein notes that there exist certain words, like `time', the meaning of which is quite clear to us until the moment we have to explain it to someone else; then we realize that suddenly we are not able any more to express in words what we certainly know---it is like we have forgotten what that specific word really means. Wittgenstein claims that ``if we have this experience, then we have arrived at the limits of language''. This observation is related to one of the central ideas of his work: that the meaning of a word does not need to rely on some kind of definition; what really matters is the way we use this word in our everyday communications. In ``Philosophical Investigations'' (\cite{wittgenstein1963}), Wittgenstein presents a thought experiment:

\begin{quote}
Now think of the following use of language: I send someone shopping. I give him a slip marked `five red apples'. He takes the slip to the shopkeeper, who opens the drawer marked `apples'; then he looks up the word `red' in a table and finds a colour sample opposite it; then he says the series of cardinal numbers---I assume that he knows
them by heart---up to the word `five' and for each number he takes an apple of the same colour as the sample out of the drawer. It is in this and similar ways that one operates with words.
\end{quote}

For the shopkeeper, the meaning of words `red' and `apples' was given by ostensive definitions (provided by the colour table and the drawer label). But what was the meaning of word `five'? Wittgenstein is very direct on this:

\begin{quote}
No such thing was in question here, only how the word `five' is used.
\end{quote}

Note that this operational view of the ``meaning is use'' idea is quite different from the distributional perspective we discussed in the previous sections. It simply implies that language is not expressive enough to describe certain fundamental concepts of our world which, in turn, means that the application of distributional models is by definition limited to a subset of the vocabulary for any language. While this limitation is disappointing, it certainly does not invalidate the established status of this technology as a very useful tool in natural language processing; however, knowing the fundamental weaknesses of our models is imperative since it helps us to set realistic expectations and opens opportunities for further research.


\index{meaning|)}
\index{Wittgenstein, Ludwig|)}

\section{Unifying the two semantic paradigms}
\label{sec:compdistr}

Distributional models of meaning have been widely studied and successfully applied to a variety of language tasks, especially during the last decade with the availability of large-scale corpora, like Gigaword \cite{gigaword} and ukWaC \cite{ukwac}, which provide a reliable resource for training the vector spaces. For example, Landauer and Dumais \cite{Landauer} use vector space models in order to reason about human learning rates in language; Sch\"utze \cite{Schutze} performs word sense induction and disambiguation; Curran \cite{Curran} shows how distributional models can be applied to automatic thesaurus extraction; Manning et al. \cite{Manning} discuss possible applications in the context of information retrieval.

Despite the size of the various text corpora, though, the creation of vectors representing the meaning of multi-word sentences by statistical means is still impossible. Therefore, the provision of distributional models with compositional abilities\index{CDMs} similar to what was described in \S \ref{sec:compsem} seems a very appealing solution that could offer the best of both worlds in a unified manner. The goal of such a system would be to combine the distributional vectors of words into vectors of larger and larger text constituents, up to the level of a sentence. A sentence vector, then, could be compared with other sentence vectors, providing a way for assessing the semantic similarity between sentences as if they were words. The benefits of such a feature are obvious for many natural language processing tasks, such as paraphrase detection, machine translation, information retrieval, and so on, and in the following sections I will review all the important approaches and current research towards this challenging goal.

\subsection{Vector mixture models}
\label{sec:vecmix}

\index{CDMs!vector mixtures|(}
\index{vector mixtures|(}

The transition from word vectors to sentence vectors implies the existence of a composition operation that can be applied between text constituents: the composition of `red' and `car' into the adjective-noun compound `red car', for example, should produce a new vector derived from the composition of the distributional vectors for `red' and `car'. Since we work with vector spaces, the candidates that first come to mind are vector addition and vector (point-wise) multiplication. Indeed, Mitchell and Lapata \cite{Lapata} present and test various models, where the composition of vectors is based on these two simple operations. Given two word vectors $\ov{w_1}$ and $\ov{w_2}$ and assuming an orthonormal basis $\{\ov{n_i}\}_i$, the multiplicative model\index{multiplicative model} computes the meaning vector of the new compound as follows:

\begin{equation}
\label{equ:mult}
   \ov{w_1w_2} = \ov{w_1} \odot \ov{w_2} = \sum\limits_i c_i^{w_1}c_i^{w_2}\ov{n_i} 
\end{equation}

\noindent
Similarly, the meaning according to the additive model\index{additive model} is:

\begin{equation}
\label{equ:add}
  \ov{w_1w_2} = \alpha \ov{w_1} + \beta \ov{w_2} = \sum\limits_i (\alpha c_i^{w_1}+\beta c_i^{w_2})\ov{n_i}
\end{equation}

\noindent where $\alpha$ and $\beta$ are optional weights denoting the relative importance of each word. The main characteristic of these models is that all word vectors live in the same space, which means that in general there is no way to distinguish between the type-logical identities of the different words. This fact, in conjunction with the element-wise nature of the operators, makes the output vector a kind of \textit{mixture} of the input vectors, all of which contribute (in principle) \textit{equally} to the final result;
thus, each element of the output vector can be seen as an ``average'' of the two corresponding elements in the input vectors. In the additive case, the components of the result are simply the cumulative scores of the input components. So in a sense the output element embraces both input elements, resembling a union of the input features. On the other hand, the multiplicative version is closer to intersection: a zero element in one of the input vectors will eliminate the corresponding feature in the output, no matter how high the other component was.


Vector mixture models constitute the simplest compositional method in distributional semantics. Despite their simplicity, though, (or because of it) these approaches have been proved very popular and useful in many NLP tasks, and they are considered hard-to-beat baselines for many of the more sophisticated models we are going to discuss next. In fact, the comparative study of Blacoe and Lapata \cite{blacoe2012} suggests something really surprising: that, for certain tasks, additive and multiplicative models can be almost as  effective as state-of-the-art deep learning models, which will be the subject of \S \ref{sec:deeplearning}. 

\index{vector mixtures|)}
\index{CDMs!vector mixtures|)}

\subsection{Tensor product and circular convolution}
\label{sec:tensorproduct}

The low complexity of vector mixtures comes with a price, since the produced composite representations disregard grammar in many different ways. For example, an obvious problem with these approaches is the commutativity of the operators: the models treat a sentence as a ``bag of words'' where the word order does not matter, equating for example the meaning of the sentence `dog bites man' with that of `man bites dog'. This fact motivated researchers to seek solutions on non-commutative operations, such as the tensor product\index{tensor product!as a non-commutative operation} between vector spaces. Following this suggestion, which was originated by Smolensky \cite{Smolensky}, the composition of two words is achieved by a structural mixing of the basis vectors that results in an increase in dimensionality:

\begin{equation}
  \ov{w_1} \otimes \ov{w_2} = \sum\limits_{i,j} c^{w_1}_ic^{w_2}_j(\ov{n_i} \otimes \ov{n_j})
\end{equation}

Clark and Pulman \cite{ClarkPulman} take this original idea further and propose a concrete model in which the meaning of a word is represented as the tensor product of the word's distributional vector with another vector that denotes the grammatical role of the word and comes from a different abstract vector space of grammatical relationships. As an example, the meaning of the sentence `dog bites man' is given as:

\begin{equation}
  \ov{dog~bites~man} = 
  (\ov{dog} \otimes \ov{subj}) \otimes \ov{bites} \otimes
  (\ov{man} \otimes \ov{obj})
  \label{equ:clarkpulman}
\end{equation}

Although tensor product models solve the bag-of-words problem, unfortunately they introduce a new very important issue: given that the cardinality of the vector space is $d$, the space complexity grows exponentially as more constituents are composed together. With $d=300$, and assuming a typical floating-point machine representation (8 bytes per number), the vector of Equation \ref{equ:clarkpulman} would require $300^5 \times 8 = 1.944 \times 10^{13}$ bytes ($\approx$ 19.5 terabytes). Even more importantly, the use of tensor product as above only allows the comparison of sentences that share the same structure, i.e. there is no way for example to compare a transitive sentence with an intransitive one, a fact that severely limits the applicability of such models.

Using a concept from signal processing, Plate \cite{Plate} suggests the replacement of tensor product by circular convolution.\index{circular convolution} This operation carries the appealing property that its application on two vectors results in a vector of the same dimensions as the operands. Let $\ov{v}$ and $\ov{u}$ be vectors of $d$ elements, the circular convolution of them will result in a vector $\ov{c}$ of the following form:

\begin{equation}
  \ov{c} = \ov{v} \circledast \ov{u} = \sum\limits_{i=0}^{d-1} \left(\sum\limits_{j=0}^{d-1} v_ju_{i-j}\right) \ov{n}_{i+1}
\end{equation}

\noindent where $\ov{n_i}$ represents a basis vector, and the subscripts are modulo-$d$, giving to the operation its circular nature. This can be seen as a compressed outer product of the two vectors. However, a successful application of this technique poses some restrictions. For example, it requires a different interpretation of the underlying vector space, in which ``micro-features'', as PMI weights for each context word, are replaced by what Plate calls ``macro-features'', i.e. features that are expressed as whole vectors drawn from a normal distribution. Furthermore, circular convolution is commutative, re-introducing the bag-of-words problem.\footnote{Actually, Plate proposes a workaround for the commutativity problem; however this is not specific to his model, since it can be used with any other commutative operation such as vector addition or point-wise multiplication.}

\subsection{Tensor-based models}
\label{sec:tensorbased}

\index{tensor-based models|(}
\index{CDMs!tensor-based models|(}

A weakness of vector mixture models and the generic tensor product approach is the symmetric way in which they treat all words, ignoring their special roles in the sentence. An adjective, for example, lives in the same space with the noun it modifies, and both will contribute equally to the output vector representing the adjective-noun compound. However, a treatment like this seems unintuitive; we tend to see relational words, such as verbs or adjectives, as functions acting on a number of arguments rather than entities of the same order as them.\index{words!as functions} Following this idea, a recent line of research represents words with special meaning as multi-linear maps (tensors\footnote{Here, the word \textit{tensor} refers to a geometric object that can be seen as a generalization of a vector in higher dimensions. A matrix, for example, is an order-2 tensor.} of higher order) that apply on one or more arguments (vectors or tensors of lower order).\index{tensors} An adjective, for example, is not any more a simple vector but a matrix (a tensor of order 2) that, when matrix-multiplied with the vector of a noun, will return a modified version of it. That is, for an adjective $\ol{adj}=\sum_{ij} c^{adj}_{ij} (\ov{n_i}\ten\ov{n_j})$ and a noun $\ov{noun} = \sum_j c^{noun}_j \ov{n_j}$ we have:

\begin{equation}
  \ov{adj~noun} = \sum_{ij} c^{adj}_{ij} c^{noun}_j \ov{n_j}
\end{equation}



This approach is based on the principle of \textit{map-state duality}: every linear map from $V$ to $W$ (where $V$ and $W$ are finite-dimensional Hilbert spaces) stands in bijective correspondence to a tensor living in the tensor product space $V \otimes W$.\index{map-state duality} For the case of a multi-linear map (a function with more than one argument), this can be generalized to the following:

\begin{equation}
  f: V_1 \to \dots \to V_j \to V_k \cong V_1 \otimes \dots \otimes V_j \otimes V_k
\end{equation}

In general, the order of the tensor is equal to the number of arguments plus one order for carrying the result; so a unary function (such as an adjective) is a tensor of order 2, while a binary function (e.g. a transitive verb) is a tensor of order 3. The composition operation is based on the inner product and is nothing more than a generalization of matrix multiplication in higher dimensions, a process known as \textit{tensor contraction}.\index{tensor contraction} Given two tensors of orders $m$ and $n$, the tensor contraction operation always produces a new tensor of order $n+m-2$. Under this setting, the meaning of a simple transitive sentence can be calculated as follows:

\begin{equation}
  \label{equ:trans}
  \overline{subj~verb~obj} = \ov{subj}^{\mathsf{T}} \times \overline{verb} \times \ov{obj}
\end{equation}

\noindent where the symbol $\times$ denotes tensor contraction. Given that $\ov{subj}$ and $\ov{obj}$ live in $N$ and $\overline{verb}$ lives in $N \otimes S \otimes N$, the above operation will result in a tensor in $S$, which represents the sentence space of our choice. 

Tensor-based models provide an elegant solution to the problems of vector mixtures: they are not bag-of-words approaches and they respect the type-logical identities of special words, following an approach very much aligned with the formal semantics perspective. Furthermore, they do not suffer from the space complexity problems of models based on raw tensor product operations (\S \ref{sec:tensorproduct}), since the tensor contraction process guarantees that every sentence will eventually live in our sentence space $S$. On the other hand, the highly linguistic perspective they adopt has also a downside: in order for a tensor-based model to be fully effective, an appropriate linear map to vector spaces must be concretely devised for every functional word, such as prepositions, relative pronouns or logical connectives. As we will see in Chapter \ref{ch:extend}, this problem is far from trivial; actually it constitutes one of the most important open issues, and at the moment restricts the application of these models on well-defined text structures (for example, simple transitive sentences of the form `subject-verb-object' or adjective-noun compounds).

The notion of a framework where relational words act as multi-linear maps on noun vectors has been formalized by Coecke et al. \cite{Coeckeetal} in the abstract setting of category theory and compact closed categories, a topic we are going to discuss in more detail in Chapter \ref{ch:framework}. Baroni and Zamparelli's composition method for adjectives and nouns also follows the very same principle \cite{Baroni}.

\index{CDMs!tensor-based models|)}
\index{tensor-based models|)}

\subsection{Deep learning models}
\label{sec:deeplearning}

\index{deep-learning models|(}
\index{CDMs!deep-learning models|(}

\begin{figure}[h]
  \centering
  \includegraphics[scale=0.7]{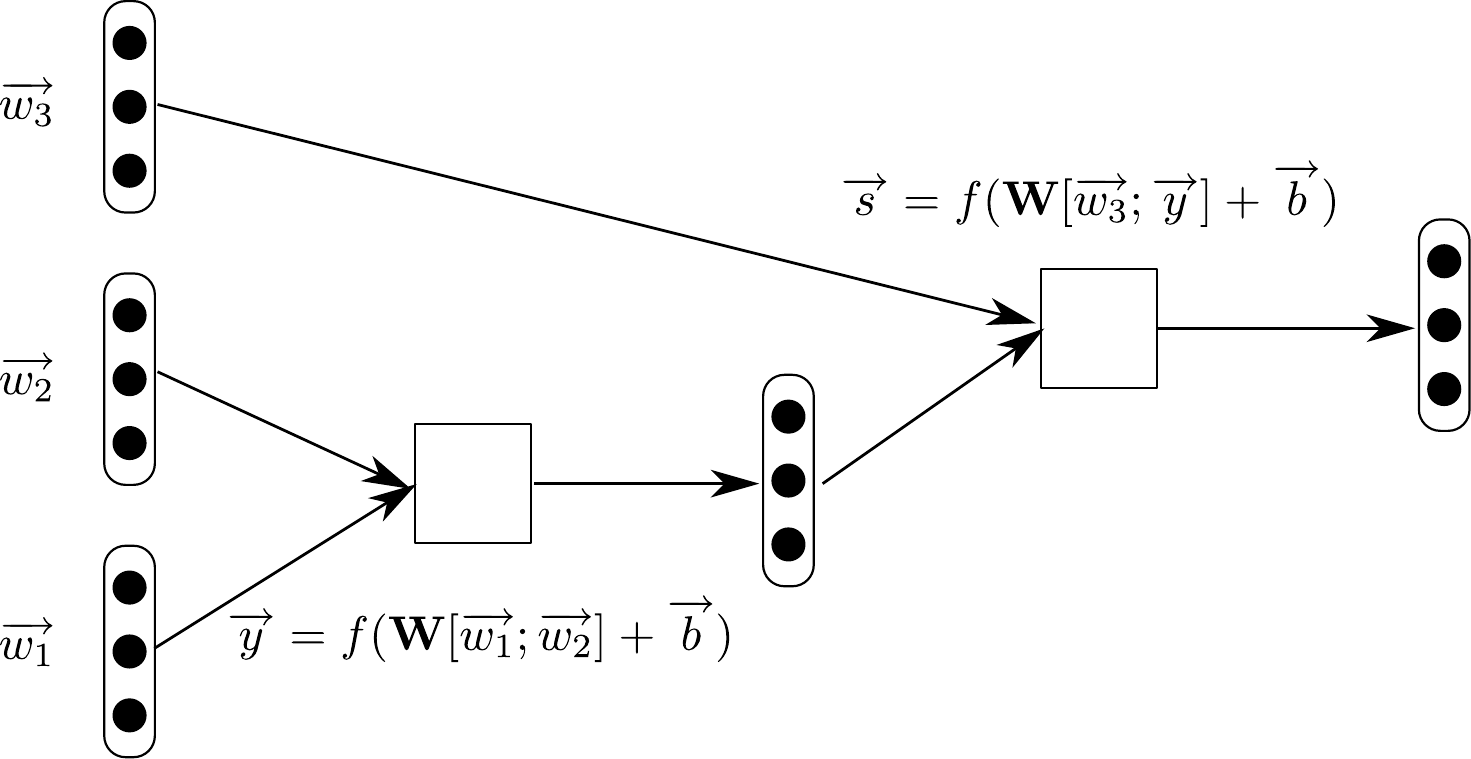}
  \caption{A recursive neural network with a single layer for providing compositionality in distributional models.}
  \label{fig:nn}
\end{figure}

A recent trend in compositionality of distributional models is based on deep learning techniques, a class of machine learning algorithms (usually neural networks) that approach models as multiple layers of representations, where the higher-level concepts are induced from the lower-level ones. For example, following ideas from the seminal work of Pollack \cite{pollack1990} conducted 25 years ago, Socher and colleagues \cite{socher2011,socher2012,socher2010} use recursive neural networks in order to produce compositional vector representations for sentences, with promising results in a number of tasks. In its most general form, a neural network like this takes as input a pair of word vectors $\ov{w_1},\ov{w_2}$ and returns a new composite vector $\ov{y}$ according to the following equation:

\begin{equation}
  \ov{y} = g(\textbf{W}[\ov{w}_1;\ov{w}_2] + \ov{b})
\end{equation}

\noindent where $[\ov{w}_1;\ov{w}_2]$ denotes the concatenation of the two child vectors, $\textbf{W}$ and $\ov{b}$ are the parameters of the model, and $g$ is a element-wise non-linear function such as $\tanh$. This output, then, will be used in a recursive fashion again as input to the network for computing the vector representations of larger constituents. This class of models is known as \textit{recursive neural networks} (RecNNs),\index{recursive neural networks} and adheres to the general architecture of Fig. \ref{fig:nn}.

In a variation of the above structure, each intermediate composition is performed via an auto-encoder instead of a feed-forward network. A \textit{recursive auto-encoder}\index{recursive auto-encoders} (RAE) acts as an identity function, learning to reconstruct the input, encoded via a hidden layer, as faithfully as possible. The state of the hidden layer of the network is then used as a compressed representation of the two original inputs. Since the optimization is solely based on the reconstruction error, a RAE is considered as a form of unsupervised learning \cite{socher2011}. Finally, neural networks can vary in design and topology. Kalchbrenner et al. \cite{kalchbrenner:ACL:2014}, for example, model sentential compositionality using a \textit{convolutional} neural network,\index{convolutional neural networks} the main characteristic of which is that it acts on small overlapping parts of the input vectors.


In contrast to the previously discussed compositional approaches, deep learning methods are based on a large amount of training: the parameters $\textbf{W}$ and $\ov{b}$ in the network of Fig. \ref{fig:nn} must be learned through an iterative algorithm known as \textit{backpropagation}, a process that can be very time-consuming and in general cannot guarantee optimality. However, the non-linearity in combination with the layered approach in which neural networks are based provide these models with great power, allowing them to simulate the behaviour of a range of functions much wider than the multi-linear maps of tensor-based approaches. Indeed, deep learning compositional models have been applied in various paraphrase detection and sentiment analysis tasks \cite{socher2011,socher2012,socher2010} delivering results that fall into state-of-the-art range.

\index{CDMs!deep-learning models|)}
\index{deep-learning models|)}

\subsection{Intuition behind various compositional approaches}
\label{sec:intuition}

\index{CDMs!intuition|(}

At this point, it is instructive to pause for a moment and consider what exactly the term \textit{composition} would mean for each one of the approaches reviewed in this section so far. Remember that in the case of vector mixtures every vector lives in the same base space, which means that a sentence vector has to share the same features and length with the word vectors. What a distributional vector for a word actually shows us is to what extent all other words in the vocabulary are related to this specific target word. If our target word is a verb, then the components of its vector can be seen as related to the \textit{action} described by the verb: a vector for  the verb `run' reflects the degree to which a `dog' can run, a `car' can run, a `table' can run and so on. The element-wise mixing of vectors $\ov{dog}$ and $\ov{run}$ then in order to produce a compositional representation for the meaning of the simple intransitive sentence `dogs run', finds an intuitive interpretation: the output vector will reflect the extent to which things that are related to dogs can also run (and vice versa); in other words, it shows how \textit{compatible} the verb is with the specific subject.\index{vector mixtures!combatibility check}\footnote{Admittedly, this rather idealistic behaviour does not apply to the same extent on every pair of composed words; however the provided basic intuition is in principle valid.}

\index{words!as functions|(}

A tensor-based model, on the other hand, goes beyond a simple compatibility check between the relational word and its arguments; its purpose is to \textit{transform} the noun into a sentence. Furthermore, the size and the form of the sentence space become tunable parameters of the model, which can depend on the specific task in hand. Let us assume that in our model we select sentence and noun spaces such that $S \in \mathbb{R}^s$ and $N \in \mathbb{R}^n$, respectively; here, $s$ refers to the number of distinct features that we consider appropriate for representing the meaning of a sentence in our model, while $n$ is the corresponding number for nouns. An intransitive verb, then, like `play' in `kids play', will live in $N \otimes S \in \mathbb{R}^{n \times s}$, and will be a map $f: N \to S$ built (somehow) in a way to take as input a noun and produce a sentence; similarly, a transitive verb will live in $N \otimes S \otimes N \in \mathbb{R}^{n \times s \times n}$ and will correspond to a map $f_{tr}: N\otimes N \to S$.

Let me demonstrate this idea using a concrete example, where the goal is to simulate the truth-theoretic nature of formal semantics view in the context of a tensor-based model (perhaps for the purposes of a textual entailment task). In that case, our sentence space will be nothing more than the following:

\small
\singlespace
\begin{equation}
S = 
\left\lbrace
\left( \begin{array}{c}
0 \\
1  
\end{array} \right)
,
\left( \begin{array}{c}
1 \\
0  
\end{array} \right)
\right\rbrace
\end{equation}
\onehalfspace
\normalsize

\noindent
with the two vectors representing $\top$ and $\bot$, respectively. Each individual in our universe will correspond to a basis vector of our noun space; with just three individuals (Bob, John, Mary), we get the following mapping:

\small
\singlespace
\begin{equation}
\ov{bob} = 
\left( \begin{array}{c}
0 \\
0 \\
1  
\end{array} \right)
~~~\ov{john}=
\left( \begin{array}{c}
0 \\
1 \\
0  
\end{array} \right)
~~~\ov{mary}=
\left( \begin{array}{c}
1 \\
0 \\
0  
\end{array} \right)
\end{equation}
\onehalfspace
\normalsize

In this setting, an intransitive verb will be a matrix formed as a series of truth values, each one of which is associated with some individual in our universe. Assuming for example that only John performs the action of sleeping, then the meaning of the sentence `John sleeps' is given by the following computation:\footnote{Clark \cite{clark2013} details an extension of this simple model from truth values to probabilities that show how plausible is the statement expressed by the composite vector.}

\small
\singlespace
\begin{equation}
\ov{john}^{\mathsf{T}} \times \overline{sleep} =
\left( \begin{array}{cccc}
0 & 1 & 0 
\end{array} \right)
\times
\left( \begin{array}{cccc}
1 & 0 \\
0 & 1 \\
1 & 0 \\
\end{array} \right)
=
\left( \begin{array}{c}
0 \\
1  
\end{array} \right)
=
\top
\end{equation}
\onehalfspace
\normalsize

\index{words!as functions|)}

\begin{figure}[h!]
  \centering
  \includegraphics[scale=0.5]{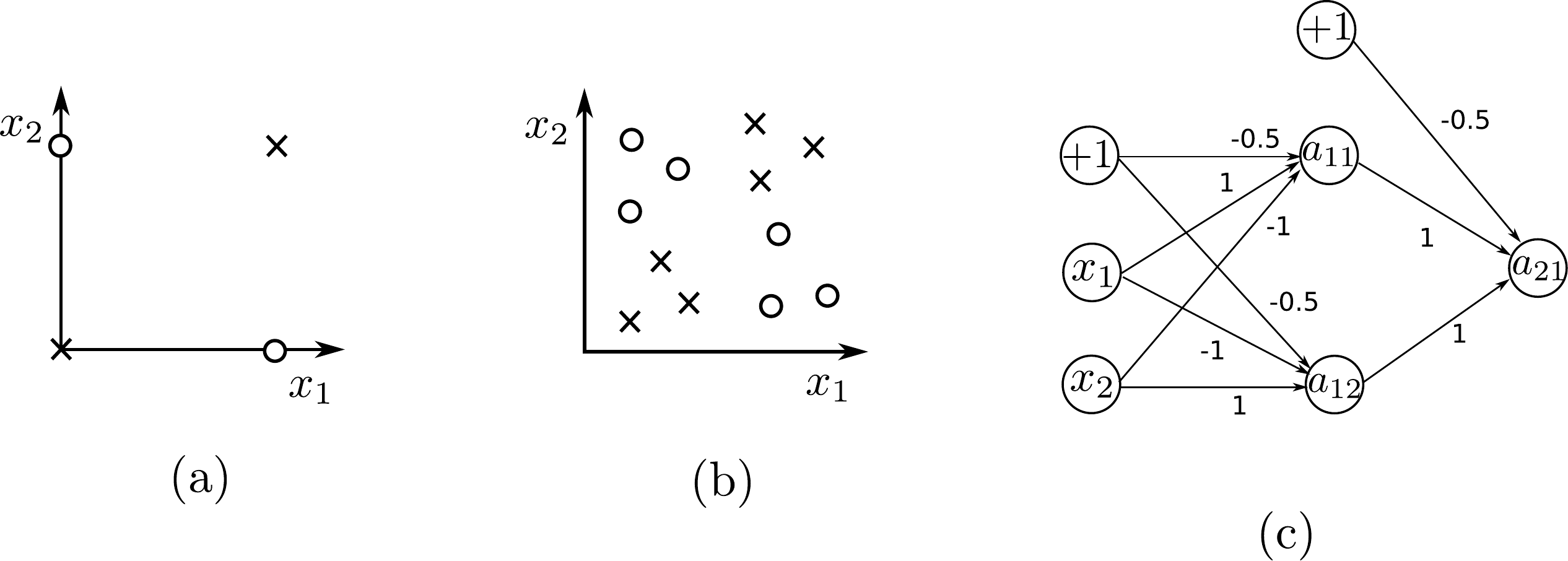}
  \caption{Intuition behind neural compositional models.}
  \label{fig:nn-intuition}
\end{figure}

For the case of tensor-based models, the nouns-to-sentence transformation has to be linear. The premise of deep-learning compositional models is to make this process much more effective by applying consecutive non-linear layers of transformation. Indeed, a neural network is not only able to project an input vector into a sentence space, but it can also \textit{transform the geometry} of the space itself in order to make it reflect better the relationships between the points (sentences) in it.\index{deep-learning models!and sentence space}\index{sentence space!and deep-learning models} This quality is better demonstrated with a simple example: although there is no linear map which sends an input $x \in \{0,1\}$ to the correct XOR value (as it is evident from part (a) of Fig. \ref{fig:nn-intuition}), the function can be successfully approximated by a simple feed-forward neural network with one hidden layer, similar to that in part (c) of the figure. Part (b) shows a generalization to an arbitrary number of points, which for our purposes can be thought as representing two semantically distinct groups of sentences.

\index{CDMs!intuition|)}

\section{A taxonomy of CDMs}
\label{sec:taxonomy}

\index{CDMs!taxonomy|(}
\index{taxonomy of CDMs|(}

\begin{figure}[h!]
  \includegraphics[scale=0.89]{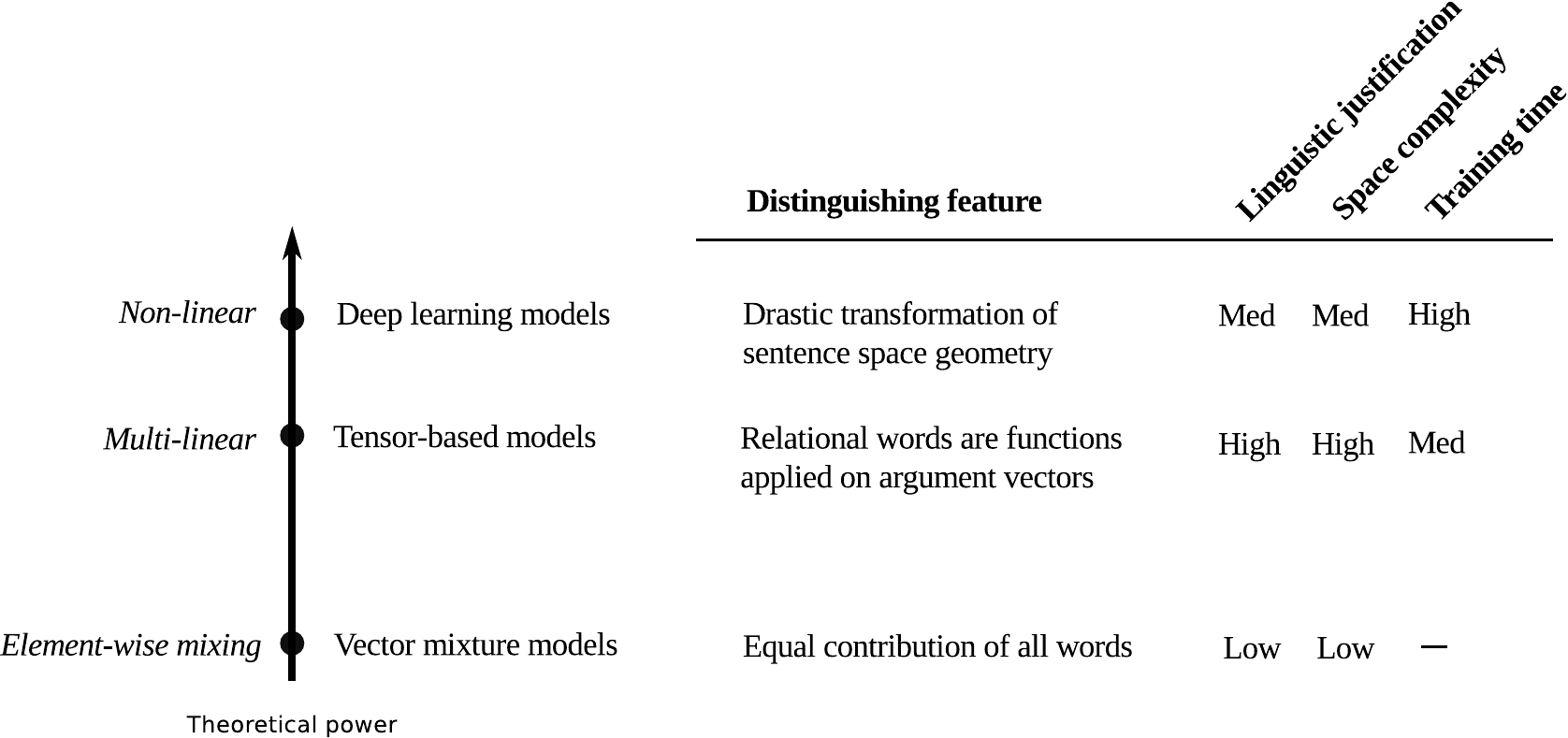}
  \caption{A taxonomy of CDMs.}
  \label{fig:taxonomy}
\end{figure}

I would like to conclude this chapter by summarizing the presentation of the various CDM classes in the form of a concise diagram
(Fig. \ref{fig:taxonomy}). As we will see in Part \ref{prt:theory}, this hierarchy is not complete; specifically, the work I present in this thesis introduces a new class of CDMs, which is flexible enough to include features from both tensor-based and vector mixture models. This is a direct consequence of applying Frobenius algebras on the creation of relational tensors, a topic that we will study in detail in Chapter \ref{ch:frobverbs}.

\index{taxonomy of CDMs|)}
\index{CDMs!taxonomy|)}

\chapter{A Categorical Framework for Natural Language}
\label{ch:framework}

\begin{chabstract}
We now proceed to our main subject of study, the categorical framework of Coecke et al. \cite{Coeckeetal}. The current chapter provides an introduction to the important notions of compact closed categories and pregroup grammars, and describes in detail how the passage from grammar to vector spaces is achieved. Furthermore, the reader is introduced to the concept of Frobenius algebras (another main topic of this work), as well as to a convenient diagrammatic notation that will be extensively used in the rest of this thesis. Material is based on \cite{kartsaklis2014,KartsaklisSpringer}. 
\end{chabstract}

\noindent
Tensor-based models stand in between the two extremes of vector mixtures and deep learning methods, offering an appealing alternative that can be powerful enough and at the same time fully aligned with the formal semantics view of natural language. Actually, it has been shown that the linear-algebraic formulas for the composite meanings produced by a tensor-based model emerge as the natural consequence of a structural similarity between a grammar and finite-dimensional vector spaces. Exploiting the abstract mathematical framework of category theory, Coecke, Sadrzadeh and Clark \cite{Coeckeetal} managed to equip the distributional models of meaning with compositionality in a way that every grammatical reduction corresponds to a linear map defining mathematical manipulations between vector spaces. This result builds upon the fact that the base type-logic of the framework, a pregroup grammar \cite{Lambek}, shares the same abstract structure with finite-dimensional vector spaces, that of a compact closed category. Mathematically, the transition from grammar types to vector spaces has the form of a strongly monoidal functor, that is, of a map that preserves the basic structure of compact closed categories. The following section provides a short introduction to the category theoretic notions above.

\section{Introduction to categorical concepts}
\label{sec:ctintro}

\textit{Category theory}\index{category theory} is an abstract area of mathematics, the aim of which is to study and identify universal properties of mathematical concepts that often remain hidden by traditional approaches. A \textit{category}\index{category} is a collection of objects and morphisms that hold between these objects, with composition of morphisms as the main operation. That is, for two morphisms $A\xrightarrow{f}B\xrightarrow{g}C$, we have $g \circ f: A\xrightarrow{}C$. Morphism composition is associative, so that $(h \circ g) \circ f = h \circ (g \circ f)$. Furthermore, every object $A$ has an identity morphism $1_A:A \to A$; for $f:A \to B$ we moreover require that:

\begin{equation}
 f \circ 1_A = f~~~\text{and}~~~ 1_B \circ f = f
\end{equation}

In category theory, the relationships between morphisms are often depicted through the use of \textit{commutative diagrams}.\index{commutative diagrams} We say that a diagram \textit{commutes} if all paths starting and ending at the same points lead to the same result by the application of composition. The diagram below, for example, shows the associativity of composition:

\begin{equation}
\label{equ:commutative}
\begin{tikzpicture}[scale=1.5,baseline=30pt]
font=\small
\node (A) at (0,2) {$A$};
\node (B) at (2,2) {$B$};
\node (C) at (2,0) {$C$};
\node (D) at (4,0) {$D$};
\path[->,font=\footnotesize]
(A) edge node[above]{$f$} (B)
(A) edge node[left]{$g\circ f~~~$} (C)
(B) edge node[left]{$g$} (C)
(B) edge node[right]{$h\circ g$} (D)
(C) edge node[below]{$h$} (D);
\end{tikzpicture}
\end{equation}

We now proceed to the definition of a \textit{monoidal category},\index{monoidal category}\index{category!monoidal} which is a special type of category equipped with another associative operation, a monoidal tensor $\otimes: \mathcal{C} \times \mathcal{C} \to \mathcal{C}$ (where $\mathcal{C}$ is our basic category).\index{tensor product!in monoidal categories} Specifically, for each pair of objects $(A,B)$ there exists a composite object $A\otimes B$, and for every pair of morphisms $(f:A \to C$, $g:B \to D)$ a parallel composite $f\otimes g: A \otimes B \to C \otimes D$. For a \textit{symmetric monoidal category},\index{monoidal category!symmetric} it is also the case that $A \otimes B \cong B \otimes A$. Furthermore, there is a unit object $I$ which satisfies the following natural isomorphisms:

\begin{equation}
\label{equ:moniso}
 A\otimes I \cong A \cong I\otimes A
\end{equation}

The natural isomorphisms in Eq. $\ref{equ:moniso}$ above, together with the natural isomorphism $A\ten (B\ten C) \xrightarrow{\alpha_{A,B,C}} (A\ten B) \ten C$ need to satisfy certain coherence conditions \cite{maclane}, which basically ensure the commutativity of all relevant diagrams. 

Finally, a monoidal category is \textit{compact closed}\index{compact closed category}\index{category!compact closed}\index{monoidal category!compact closed} if every object $A$ has a left and right adjoint,\index{adjoints} denoted as $A^l,A^r$ respectively, for which the following special morphisms exist:

\begin{equation}
\label{equ:eta}
  \eta^l: I \to A \otimes A^l~~~~
  \eta^r:I \to A^r \otimes A
\end{equation}
\vspace{-0.5cm}
\begin{equation}
\label{equ:epsilon}
  \epsilon^l: A^l \otimes A \to I~~~~
  \epsilon^r: A \otimes A^r \to I
\end{equation}

These adjoints are unique up to isomorphism. Furthermore, the above maps need to satisfy the following four special conditions (the so-called \textit{yanking conditions}),\index{yanking conditions}\index{compact closed category!yanking conditions} which again ensure that all relevant diagrams commute:

\begin{gather}
\begin{split}
  (1_A\ten\epsilon^l_A)\circ(\eta^l_A\ten 1_A) = 1_A~~~~~(\epsilon^r_A\ten 1_A)\circ (1_A\ten \eta^r_A)=1_A ~~~~ \\
  (\epsilon^l_A\ten 1_{A^l})\circ (1_{A^l}\ten \eta^l_A) = 1_{A^l}~~~~~~(1_{A^r}\ten\epsilon^r_A)\circ (\eta^r_A\ten 1_{A^r}) = 1_{A^r}
\end{split}  
\end{gather}

For the case of a \textit{symmetric compact closed category},\index{compact closed category!symmetric} the left and right adjoints collapse into one, so that $A^* := A^l = A^r$. As we will see, both a pregroup grammar and the category of finite-dimensional vector spaces over a base field conform to this abstract definition of a compact closed category. Before we proceed to this, though, we need to introduce another important concept, that of a strongly monoidal functor. 

In general, a \textit{functor}\index{functor} is a morphism between categories that preserves identities and composition. That is, a functor $\mathcal{F}$ between two categories $\mathcal{C}$ and $\mathcal{D}$ maps every object $A$ in $\mathcal{C}$ to some object $\mathcal{F}(A)$ in $\mathcal{D}$, and every morphism $f:A\to B$ in $\mathcal{C}$ to a morphism $\mathcal{F}(f):\mathcal{F}(A)\to \mathcal{F}(B)$ in $\mathcal{D}$, such that $\mathcal{F}(1_A)=1_{\mathcal{F}(A)}$ for all objects in $\mathcal{C}$ and $\mathcal{F}(g\circ f)=\mathcal{F}(g)\circ \mathcal{F}(f)$ for all morphisms $f:A\to B,g:B\to C$. Note that in the case of monoidal categories, a functor must also preserve the monoidal structure. Taking any two monoidal categories $(\mathcal{C},\ten,I_{\mathcal{C}})$ and $(\mathcal{D},\bullet,I_{\mathcal{D}})$, a \textit{monoidal functor}\index{monoidal functor}\index{functor!monoidal} is a functor that comes with a natural transformation (a structure-preserving morphism between functors) $\phi_{A,B}: \mathcal{F}(A) \bullet \mathcal{F}(B) \to \mathcal{F}(A\ten B)$ and a morphism $\phi:I_{\mathcal{D}} \to \mathcal{F}(I_{\mathcal{C}})$, which as usual satisfy certain coherence conditions. Finally, preserving the compact structure between two categories so that $\mathcal{F}(A^l) = \mathcal{F}(A)^l$ and $\mathcal{F}(A^r) = \mathcal{F}(A)^r$ requires a special form of a monoidal functor, one whose $\phi_{A,B}$ and $\phi$ morphisms are invertible. Note that in this case, we get:

\begin{eqnarray*}
   \M{F}(A^l) \bullet \M{F}(A) \xrightarrow{\phi_{A^l,A}} \M{F}(A^l\ten A) \xrightarrow{\M{F}(\epsilon^l_{\M{C}})} \M{F}(I_{\M{C}}) \xrightarrow{\phi^{-1}} I_{\M{D}} \\
   I_{\M{D}} \xrightarrow{\phi} \M{F}(I_{\M{C}}) \xrightarrow{\M{F}(\eta^l_{\M{C}})} \M{F}(A\ten A^l) \xrightarrow{\phi_{A,A^l}^{-1}} \M{F}(A)\bullet \M{F}(A^l) 
\end{eqnarray*}

Since adjoints are unique, the above means that $\M{F}(A^l)$ must be the left adjoint of $\M{F}(A)$; the same reasoning applies for the case of the right adjoint. We will use such a \textit{strongly monoidal functor}\index{strongly monoidal functor}\index{functor!strongly monoidal} in \S\ref{sec:functor} as a means for translating a grammatical derivation to a linear-algebraic formula between vectors and tensors. 

\section{Pregroup grammars}
\label{sec:pregroup}

\index{pregroup grammars|(}

A \textit{pregroup grammar} \cite{Lambek} is a type-logical grammar built on the rigorous mathematical basis of a pregroup algebra, i.e. a partially ordered monoid with unit $1$, whose each element $p$ has a left adjoint $p^l$ and a right adjoint $p^r$.\index{adjoints} In the context of a pregroup algebra, Eqs \ref{equ:eta} and \ref{equ:epsilon} are transformed to the following:

\begin{equation}
\label{equ:pregroups}
  p^l\cdot p \leq 1 \leq p\cdot  p^l~~~~\text{and}~~~~p\cdot  p^r \leq 1 \leq p^r\cdot  p
\end{equation}

\noindent where the dot denotes the monoid multiplication. It also holds that $1\cdot p = p = p\cdot 1$. A pregroup grammar is the pregroup freely generated over a set of atomic types, for example $n$ for well-formed noun phrases and $s$ for well-formed sentences. Atomic types and their adjoints can be combined to form compound types, e.g. $n^r\cdot s\cdot n^l$ for a transitive verb. This type reflects the fact that a transitive verb is an entity expecting a noun to its right (the object) and a noun to its left (the subject) in order to return a sentence. The rules of the grammar are prescribed by the mathematical properties of pregroups, and specifically by the inequalities in (\ref{equ:pregroups}) above. We say that a sequence of words $w_1w_2\hdots w_n$ forms a grammatical sentence whenever we have:

\begin{equation}
  t_1 \cdot t_2 \cdot \hdots \cdot t_n \leq s
\end{equation}

\noindent where $t_i$ is the type of the $i$th word in the sentence. As an example, consider the following type dictionary:

\small
\begin{center}
\begin{tabular}{c|c|c|c|c|c}
  John & saw & Mary & reading & a & book \\
  \hline
  $n$ & $n^r\cdot s \cdot n^l$ & $n$ & $n^r \cdot n \cdot n^l$ & $n\cdot n^l$ & $n$
\end{tabular}
\end{center}
\normalsize

The juxtaposition of these words in the given order would form a grammatical sentence, as the derivation below shows:

\begin{eqnarray}
 \label{equ:preg-der}
  n\cdot (n^r\cdot s \cdot n^l) \cdot n \cdot (n^r \cdot n \cdot n^l) \cdot (n \cdot n^l) \cdot n & = & \nonumber\\
  (n\cdot n^r) \cdot s \cdot n^l \cdot (n \cdot n^r) \cdot n \cdot (n^l \cdot n) \cdot (n^l \cdot n) & \leq  & ~~~~~\text{(associativity)}\nonumber\\
  1 \cdot s \cdot n^l \cdot 1 \cdot n \cdot 1 \cdot 1 & = & ~~~~~\text{(partial order)} \\
  s \cdot (n^l\cdot n) & \leq & ~~~~~\text{(unitors)}\nonumber \\
  s & & ~~~~~\text{(partial order)} \nonumber
\end{eqnarray}

Derivations in a pregroup grammar are usually shown as reduction diagrams,\index{reduction diagrams} where contractions to the unit object due to inequalities $p^l\cdot p \leq 1$ and $p\cdot p^r \leq 1$ are depicted by lines connecting the corresponding elements:

\begin{equation}
\label{equ:john}
\small

\InputIfFileExists{./tikz/john.tikz}{}{\input{./tikz/john.tikz}}

\normalsize
\end{equation}

A pregroup grammar forms a compact closed category, to which I will refer as \textbf{Preg_F}.\index{pregf@\textbf{Preg_F},category}\index{category!pregf@\textbf{Preg_F}} The objects of this category are the basic types, their adjoints, and all tensor products of them, while the morphisms are given by the partial ordering, i.e. we take $p\leq q$ to denote the morphism $p\to q$. Note that this means that \textbf{Preg_F} is a \textit{posetal} category;\index{posetal category} that is, a category between any two objects of which there exists \textit{at most} one morphism. The tensor product is the monoid multiplication: for two objects $p$ and $q$ we have $p\cdot q$, and for two morphisms $p\leq q$ and $r \leq t$ we get $p\cdot r \leq q \cdot t$ by monotonicity of the monoid multiplication. Furthermore, from Eq. \ref{equ:pregroups} we derive:

\begin{eqnarray}
\begin{split}
\epsilon^l: p^l\cdot p \leq 1 & \quad & \epsilon^r: p\cdot p^r \leq 1 \\
\eta^l: 1 \leq p\cdot p^l & \quad & \eta^r: 1 \leq p^r\cdot p 
\end{split}
\end{eqnarray} 

The category \textbf{Preg_F} is an obvious choice for accommodating our grammar, but it imposes a simplification: its posetal nature means that all possible derivations between two grammatical types would be identified, which is clearly an unwelcome property. Consider, for example the following case:

\begin{equation}
 \small
 
\InputIfFileExists{./tikz/rotten.tikz}{}{\input{./tikz/rotten.tikz}}

 \normalsize
 \label{equ:rotten}
\end{equation}

We obviously need a way to distinguish the two cases, since they lead to semantically different results. The solution is to replace the free pregroup with the \textit{free compact-closed category}\index{compact closed category!free} over a partially ordered set of basic grammar types, as described by Preller and Lambek \cite{preller2007free}, abandoning the assumption that there is only one morphism for each domain and co-domain pair. This category, which is referred to by \textbf{C_F} in this work, has form identical to \textbf{Preg_F} with the important difference that two objects might be connected with more than one morphism; for the example above, the \textit{hom-set}\footnote{The set of all morphisms between the two objects.} between $n\cdot n^l\cdot n \cdot n^r\cdot n\cdot n^l\cdot n$ and $n$ becomes:

\begin{equation*}
  \left\{(1_n\cdot \epsilon_n^l) \circ (1_{n\cdot n^l} \cdot \epsilon_n^l\cdot 1_n \cdot \epsilon_n^r),
  (\epsilon_n^r\cdot 1_n \cdot \epsilon_n^l)\circ(1_n\cdot \epsilon_n^l\cdot 1_{n^r \cdot n \cdot n^l \cdot n}) \right\}
\end{equation*}

\noindent in which the first element corresponds to the left-hand derivation in (\ref{equ:rotten}), and the second element to the right-hand derivation. In the next section we will use \textbf{C_F} as the domain of our syntax-to-semantics functor. 

\index{pregroup grammars|)}

\section{Quantizing the grammar}
\label{sec:functor}

Finite-dimensional vector spaces over a field also form a compact closed category, where the vector spaces are the objects, linear maps are the morphisms, the main operation is the tensor product between vector spaces, and the field over which the spaces are formed, in our case $\mathbb{R}$, is the unit. Following a category-theoretic convention, in this setting vectors $\ov{v} \in V$, $\ol{u} \in V\ten W$ etc. are considered specific \textit{states}\footnote{This view of vectors as states will take a more concrete form in Chapter \ref{ch:ambiguity}, where I resort to quantum mechanics for modelling ambiguity in language.} of their corresponding vector spaces,\index{vectors, as states}\index{quantum states} and are represented as morphisms from the unit object:

\begin{equation}
  \ov{v}: I \to V~~~~~~~~~~\ol{u}: I \to V\ten W
\end{equation}

In contrast to \textbf{Preg_F} and \textbf{C_F}, the category of finite-dimensional vector spaces and linear maps is symmetric. The adjoint\index{adjoints} of each vector space is its dual which, in the presence of a fixed basis, is isomorphic to the space itself. Note that any vector space with fixed basis (as it is always the case in distributional models of meaning) comes with an inner product; this provides a canonical way to model the behaviour of the $\epsilon$-maps, which become the inner product between the involved vectors:

\begin{equation}
\label{equ:innerprod}
  \epsilon^l = \epsilon^r: W \otimes W \to \mathbb{R}:: \sum\limits_{ij} c_{ij}(\ov{w_i} \otimes \ov{w_j}) \mapsto \sum\limits_{ij} c_{ij}\langle \ov{w_i}|\ov{w_j}\rangle
\end{equation}

On the other hand, $\eta$-maps define identity matrices, which by linearity generalize to scalar products of these:

\vspace{-0.2cm}
\begin{equation}
\label{equ:eta-vector}
  \eta^l = \eta^r: \mathbb{R} \to W \otimes W :: 1 \mapsto \sum\limits_i \ov{w_i} \otimes \ov{w_i}
\end{equation}

Let us refer to the category of finite-dimensional vector spaces and linear maps over a field as \textbf{FVect}\index{fvect@\textbf{FVect}, category}\index{category!fvect@\textbf{FVect}}. In the original paper \cite{Coeckeetal}, the unification of grammar and vector spaces was achieved by working on the product category $\textbf{FPreg_F} \times \textbf{FVect}$; for a type reduction $p_1\cdot p_2\cdot \hdots \cdot p_n \leq x$, the transition to vector spaces had the form of a map $(\leq,f)$ such that:

\begin{equation}
  (\leq,f): (p_1\cdot p_2 \cdot \hdots \cdot p_n, W_1\ten \hdots \ten W_n) \to (x,W)
\end{equation}

\noindent with $f$ representing the syntax-to-semantics morphism. Later, this was recast into a functorial passage from a pregroup dictionary to \textbf{FVect} from Preller and Sadrzadeh \cite{Preller2010} and Kartsaklis, Sadrzadeh, Pulman and Coecke \cite{kartsaklis2014}, while in \cite{Sadrzadeh2013} Coecke, Grefenstette and Sadrzadeh define a functor from Lambek monoids. The main difference of this latter work is that the grammar is expressed as a \textit{monoidal bi-closed category},\index{category!monoidal bi-closed}\index{monoidal category!bi-closed}\index{bi-closed monoidal category}
a structure that is closer than a pregroup to the traditional form of a categorial grammar\index{categorial grammar} as this was formalized in the context of Lambek calculus \cite{lambek1958} and the work of Ajdukiewicz \cite{ajdukiewicz1935} and Bar-Hillel \cite{bar1953}. For the purposes of this thesis I will not consider this approach, and I will define the transition from grammar type reductions to vector spaces along the lines of \cite{Preller2010} and \cite{kartsaklis2014}. 


Assuming the canonical choices of Eq. \ref{equ:innerprod} and Eq. \ref{equ:eta-vector} for the structural $\epsilon$ and $\eta$ morphisms, respectively, defining a strongly monoidal functor:

\begin{equation}
 \label{equ:functor}
 \mathcal{F}: \textbf{C_F} \to \textbf{FVect}
\end{equation}

\noindent culminates in choosing an appropriate vector space for each atomic type $x \in \M{T}$ (where $\M{T}$ is the partially ordered set of the atomic pregroup types) so that $\M{F}(x) = X$. Note that since the set $\op{Ob}(\textbf{C_F})$\footnote{The set $\op{Ob}(\M{C})$ is the set of objects in category $\M{C}$.} is freely generated over $\M{T}$ means that we can think of the objects in functor's co-domain as also being freely generated over a set $\{\M{F}(t)|t \in \M{T}\}$---which provides vector space counterparts for all complex types in our grammar. I proceed to make the above precise and give the basic properties of $\M{F}$. 

Firstly, it maps the pregroup unit to the field of our vector spaces:

\begin{equation}
  \M{F}(1) = \mathbb{R}
\end{equation}

As discussed above, for each atomic grammar type $x$, the functor assigns a vector space $X$ so that $\M{F}(x) = X$. For example:

\vspace{-0.3cm}
\begin{equation}
  \label{equ:map}
  \M{F}(n) = N~~~~~~~~~\M{F}(s) = S
\end{equation}

Note that in principle the assigned vector space does not have to be different for each grammatical type. Later, in Chapter \ref{ch:frobverbs}, we will see that using the same vector space for nouns and sentences leads to interesting consequences. 

Regarding the mappings of the adjoints, we first notice that since \textbf{FVect} is symmetric, it is the case that:

\vspace{-0.3cm}
\begin{equation}
   \label{equ:map-adjoints1}
   \mathcal{F}(x)^l = \mathcal{F}(x)^r = \mathcal{F}(x)^*
\end{equation}

Furthermore, the fact that we work with fixed basis means that $\M{F}(x)^* \cong \M{F}(x)$. As a result, we have:

\vspace{-0.3cm}
\begin{equation}
   \label{equ:map-adjoints}
   \mathcal{F}(x^l) = \mathcal{F}(x)^l = \mathcal{F}(x)^* = \mathcal{F}(x)
\end{equation}

\noindent and similarly for the right adjoint.\index{adjoints} From strong monoidality, it follows that the complex types are mapped to tensor products of vector spaces:

\vspace{-0.3cm}
\begin{equation}
  \M{F}(x\cdot y) = \M{F}(x)\ten \M{F}(y)
\end{equation}

Putting together all the above, we get type mappings of the following form:

\vspace{-0.3cm}
\begin{equation}
 \label{equ:complmap}
 \mathcal{F}(n\cdot n^l) = \mathcal{F}(n^r\cdot n) = N \otimes N~~~~~~~
 \mathcal{F}(n^r\cdot s\cdot n^l) = N \otimes S \otimes N
\end{equation}

The structural morphisms $\epsilon$ and $\eta$ are translated as follows:

\vspace{-0.3cm}
\begin{equation}
  \M{F}(\epsilon_{\textbf{C_F}}^l) = \M{F}(\epsilon_{\textbf{C_F}}^r) = \epsilon_{\textbf{FVect}}~~~~~~~~~
  \M{F}(\eta_{\textbf{C_F}}^l) = \M{F}(\eta_{\textbf{C_F}}^r) = \eta_{\textbf{FVect}}
\end{equation}

\noindent with $\epsilon_{\textbf{FVect}}$ and $\eta_{\textbf{FVect}}$ defined as in Eqs \ref{equ:innerprod} and  \ref{equ:eta-vector}, respectively. Finally, each other morphism in \textbf{C}$_F$ is mapped to a linear map in \textbf{FVect}. I will use the case of a transitive sentence as an example. Here, the subject and the object have the type $n$, whereas the type of the verb is $n^r\cdot s\cdot n^l$, as described above. The derivation in pregroups form proceeds as follows:

\vspace{-0.3cm}
\begin{equation}
  n\cdot (n^r\cdot s\cdot n^l)\cdot n = (n\cdot n^r)\cdot s \cdot (n^l\cdot n) \leq 1\cdot s\cdot 1 = s
\end{equation}

\noindent which corresponds to the following morphism in $\textbf{C_F}$:

\begin{equation}
 \epsilon^r_n\cdot 1_s \cdot \epsilon^l_n: n\cdot n^r\cdot s\cdot n^l\cdot n \to s
\end{equation}

Applying our functor $\cal{F}$ to this will give:

\begin{align}
  \mathcal{F}(\epsilon_n^r \cdot 1_s \cdot \epsilon_n^l) & = \mathcal{F}(\epsilon^r_n)\ten \mathcal{F}(1_s)\ten \mathcal{F}(\epsilon^l_n) \nonumber \\ 
  & = \epsilon_{\mathcal{F}(n)} \ten 1_{\mathcal{F}(s)} \ten \epsilon_{\mathcal{F}(n)} \\
  &= \epsilon_N \otimes 1_S \otimes \epsilon_N: N \otimes N\otimes S \otimes N \otimes N \to S \nonumber
\end{align}

\index{sentence meaning!categorically|(}

The function $f$ in Eq. \ref{equ:main}, suggested as a way of calculating the meaning of a sentence, now takes a concrete form strictly based on the grammar rules that connect the individual words. I will now proceed and formally define the meaning of a sentence as follows:

\vspace{0.8cm}
\begin{definition}
  Take $\ov{w_i}$ to be a vector $I\to \mathcal{F}(p_i)$ corresponding to word $w_i$ with pregroup type $p_i$ in a sentence $w_1w_2 \hdots w_n$. Given a type-reduction $\alpha: p_1\cdot p_2 \cdot \hdots \cdot p_n \to s$, the meaning of the sentence is defined as follows:
   \begin{equation}
     \mathcal{F}(\alpha)(\ov{w_1}\ten \ov{w_2} \ten \hdots \ten \ov{w_n})
   \end{equation}
 \label{def:categorical}  
\end{definition}

\index{sentence meaning!categorically|)}

Let us see what Def. \ref{def:categorical} means from a linear-algebraic perspective. Working on the transitive case a little further, we observe that the map of pregroup types to tensors prescribes that the subject and object are vectors (Eq. \ref{equ:map}) while the verb is a tensor of order 3 (Eq. \ref{equ:complmap}). Hence, for a simple sentence such as `dogs chase cats' we have the following geometric entities:

\[
\ov{dogs} = \sum_i c^{dogs}_i \ov{n_i}~~~~~~~\overline{chase} = \sum_{ijk}c^{chase}_{ijk}(\ov{n_i} \otimes \ov{s_j} \otimes \ov{n_k})~~~~~~~\ov{cats} = \sum_k c^{cats}_k \ov{n_k}
\]

\noindent where $c^v_i$ denotes the $i$th component in vector $\ov{v}$, $\ov{n_i}$ is a basis vector of $N$ and $\ov{s_i}$ a basis vector of $S$. Applying Def. \ref{def:categorical} on this sentence will give:

\begin{equation}
\label{deriv}
   \begin{split}
   \mathcal{F}(\epsilon^r_n \cdot 1_s \cdot \epsilon^l_n)(\ov{dogs} \otimes \overline{chase} \otimes \ov{cats}) & = \\
   (\epsilon_N \otimes 1_S \otimes \epsilon_N)(\ov{dogs} \otimes \overline{chase} \otimes \ov{cats}) & = \\
   \sum\limits_{ijk} c_{ijk}^{chase} \langle \ov{dogs}|\ov{n_i} \rangle \ov{s_j} \langle \ov{n_k}|\ov{cats}\rangle & = \\
   \sum\limits_{ijk} c_{ijk}^{chase} c_i^{dogs} c_k^{cats} \ov{s_j} & = \\
   \ov{dogs}^{\mathsf{T}} \times \overline{chase} \times \ov{cats}
   \end{split}
\end{equation}

\noindent where the symbol $\times$ denotes tensor contraction, and the result lives in $S$. Thus we have arrived at Eq. \ref{equ:trans}, presented in \S \ref{sec:tensorbased} as a means for calculating the vector of a transitive sentence in the context of a tensor-based model. 

The significance of the categorical framework lies exactly in this fact, that it provides an elegant mathematical counterpart of the formal semantics perspective as expressed by Montague \cite{Mon1}, where words are represented and interact with each other according to their type-logical identities. Furthermore, it seems to imply that approaching the problem of compositionality in a tensor-based setting is a step in the right direction, since the linear-algebraic manipulations are a direct consequence of the grammatical derivation. 

\section{Pictorial calculus}
\label{sec:pictorial}

\index{graphical language|(}

\tikzstyle{every picture}=[scale=0.35]

Compact closed categories are a special case of monoidal categories, which are \textit{complete} with regard to a convenient pictorial calculus \cite{kelly1972many,selinger2011survey}. This means that an equation between morphisms follows from the axioms of monoidal categories if and only if it holds, up to \textit{planar isotopy},\index{planar isotopy} in the pictorial calculus. Selinger \cite{selinger2011survey} explains the ``planar isotopy'' requirement as follows:

\begin{quote}
[...] two diagrams drawn in a rectangle in the plane [...] are equivalent if it is possible to transform one to the other by continuously moving around boxes in the rectangle, without allowing boxes or wires to cross each other or to be detached from the boundary of the rectangle during the moving.
\end{quote}

The diagrammatic language visualizes the derivations in an intuitive way, focusing on the interactions between the various objects, and I will use it extensively for demonstrating the main points of this thesis. In this section I introduce the fragment of the calculus that is relevant to the current work. A morphism $f:A \to B$ is depicted as a box with incoming and outgoing wires representing the objects; the identity morphism $1_A: A \to A$ is just a straight line.

\small
\ctikzfig{morphism1}
\normalsize

Recall that a vector is represented as a morphism from the unit object. In our graphical calculus, the unit object is depicted as a triangle, while the number of wires emanating from it denotes the order of the corresponding tensor:

\small
\ctikzfig{samples}
\normalsize

Tensor products of objects and morphisms are depicted by juxtaposing the corresponding diagrams side by side. Composition, on the other hand, is represented as a vertical superposition.  For example, from left to right,  here are the pictorial representations of the tensor of  a vector in $A$ with a vector in $B$, a tensor of morphisms $f \otimes g:A \otimes C \to B \otimes D$, and a composition of morphisms $h \circ f$ for $f: A \to B$ and $h: B \to C$.

\small
\ctikzfig{tensor}
\normalsize

The $\epsilon$-maps are represented as cups ($\cup$) and the $\eta$-maps as caps ($\cap$). Equations such as $(\epsilon^l_A \otimes 1_{A^l}) \circ (1_{A^l} \otimes \eta^l_A) = 1_A$ (one of the four yanking conditions of compact closed categories) now get an intuitive visual justification:

\small
\ctikzfig{morphisms}
\normalsize



Let us now see how the grammatical derivation in (\ref{equ:john}) is translated to \textbf{FVect}:\footnote{Despite the fact that our functor sends atomic types and their adjoints to the same base vector space (Eq. \ref{equ:map-adjoints}), in my diagrams I will occasionally take the liberty to refer to the tensored vector spaces with names denoting their original grammatical type for enhanced readability.}

\begin{equation}
 \footnotesize
 
\InputIfFileExists{./tikz/john-fvect.tikz}{}{\input{./tikz/john-fvect.tikz}}

 \normalsize
 \label{equ:fvect-der} 
\end{equation}

Categorically, the derivation of a general transitive sentence of the form \textit{subject-verb-object} can be depicted by the following commutative diagram:

\begin{equation}
\label{equ:trans-comm}
\begin{tikzpicture}[scale=1.5,baseline=30pt]
font=\small
\node (A) at (0,4) {$I\ten I \ten I$};
\node (B) at (0,0) {$I$};
\node (C) at (13,4) {$N\ten N^r\ten S\ten N^l\ten N$};
\node (D) at (13,0) {$S$};
\path[->,font=\footnotesize]
(A) edge node[above]{$\ov{subj}\ten\ol{verb}\ten\ov{obj}$} (C)
(C) edge node[right]{$\epsilon^r_N\ten 1_S \ten \epsilon^l_N$} (D)
(B) edge node[below]{$(\epsilon^r_N\ten 1_S \ten \epsilon^l_N)\circ(\ov{subj}\ten\ol{verb}\ten\ov{obj})$} (D)
(A) edge[-] node[above,rotate=90]{$\cong$} (B);
\end{tikzpicture}
\end{equation}

\noindent which demonstrates the fact that the composition morphism is a state vector in the sentence space.

\index{graphical language|)}

\section{Frobenius algebras}
\label{sec:frobenius}

\index{Frobenius algebras|(}

I will close this chapter by introducing the notion of a Frobenius algebra, which is going to play an important role for the rest of this thesis. Frobenius algebras were named after the German mathematician Ferdinand Georg Frobenius, and have been studied extensively from the 1930s onwards. The basis of a solid duality theory was set by Nakayama in 1939 \cite{nakayama1939} and 1941 \cite{nakayama1941}. Since then, they have found applications in various fields of mathematics and physics, e.g. in topological quantum field theories \cite{Kock} and in categorical quantum mechanics \cite{coecke2006quantum, CoeckeVic}. The  general categorical definitions recalled below are due to Carboni and Walters \cite{CarboniWal}, whereas their concrete instantiations to algebras over vector spaces were developed by Coecke and Pavlovic \cite{coecke2006quantum}. 

A Frobenius algebra over a symmetric monoidal category $({\cal C}, \otimes, I)$ is a tuple $(F, \Delta, \iota, \mu, \zeta)$, where for an $F$ object of ${\cal C}$ the triple $(F, \Delta, \iota)$  is an associative coalgebra; that is, the following are morphisms of ${\cal C}$, satisfying the coalgebraic associativity and unit conditions:

\begin{align}
\Delta \colon F \to F \otimes F&~& \iota \colon F \to I
\end{align}

The triple $(F, \mu, \zeta)$ is an associative algebra, which means that the following are morphisms of ${\cal C}$, satisfying the algebraic associativity and unit conditions:

\begin{align}
\mu \colon  F \otimes F \to F  &~& \zeta \colon I \to F 
\end{align}

Moreover, the above $\Delta$ and $\mu$ morphisms satisfy the following \emph{Frobenius} condition:\index{Frobenius condition}

\begin{align}
(\mu \otimes 1_F) \circ (1_F \otimes \Delta) \ = \  \Delta \circ \mu  \ = \  (1_F \otimes \mu) \circ (\Delta \otimes 1_F) 
\end{align}

A Frobenius algebra is \emph{commutative} if it satisfies the following two conditions for $\sigma: A \otimes B \to B \otimes A$, the symmetry morphism of $({\cal C}, \otimes, I)$:

\begin{equation}
\sigma \circ \Delta = \Delta \hspace{2cm} \mu \circ \sigma = \mu
\end{equation}

Finally, a Frobenius algebra is \emph{isometric} or \emph{special} if it satisfies the following condition:

\begin{equation}
\mu \circ \Delta = 1
\end{equation}

In the category \textbf{FVect}, any vector space $V$ with a fixed basis $\{\ov{v_i}\}_i$ has a commutative special Frobenius algebra over it, explicitly given as follows:

\begin{align}
\label{equ:frob}
  \Delta ::\ov{v_i} \mapsto \ov{v_i} \otimes \ov{v_i} & & \iota::\ov{v_i} \mapsto 1 \nonumber \\ 
 \mu:: \ov{v_i} \otimes \ov{v_j} \mapsto \delta_{ij}\ov{v_i} :=  \left\{\begin{array}{c l}
   \ov{v_i} & i=j \\ 
   \ov{0} & i \neq j
\end{array} \right. & &
  \zeta::1 \mapsto \sum_i \ov{v_i}
\end{align}


In such Frobenius algebras, therefore, the operation $\Delta$  corresponds to \textit{copying}\index{copying of basis}\index{semantic spaces!copying of basis} of the basis and its unit $\iota$ corresponds to deleting of the basis.\index{deleting of basis}\index{semantic spaces!deleting of basis} They enable one to faithfully encode vectors of $V$ into spaces with higher tensor orders, such as $V \otimes V, V \otimes V \otimes V, \cdots$. I will use this fact extensively in Chapter \ref{ch:frobverbs} in order to encode relational tensors of lower order into higher order spaces. In linear algebraic terms, for $\ov{v} \in V$, we have that $\Delta(\ov{v})$ is a diagonal matrix whose diagonal elements are the weights of $\ov{v}$. As an example, consider $V$ to be a 2-dimensional space with basis $\{\ov{v_1},\ov{v_2}\}$. A vector $\ov{v} = a\ov{v_1}+b\ov{v_2}$ then can be encoded into $V\ten V$ as follows:

\begin{equation}
  \Delta(\ov{v}) = \Delta \left( \begin{array}{c}
    a\\b\end{array} \right) =
    \left( \begin{array}{cc}
    a & 0 \\ 0 & b \\ \end{array} \right) =
    a \ov{v_1} \ten \ov{v_1} + b \ov{v_2} \ten \ov{v_2}
\end{equation}

The operation $\mu$ is referred to as \textit{uncopying} of the basis.\index{uncopying of basis}\index{semantic spaces!uncopying of basis} In linear algebraic terms, for $\ol{z} \in V\otimes V$, we have that $\mu(\ol{z})$ is a vector consisting only of the diagonal elements of $\ol{z}$, hence losing the information encoded in the non-diagonal part. Concretely, if $\ol{z} = a\ov{v_1} \ten \ov{v_1} + b\ov{v_1} \ten \ov{v_2} + c \ov{v_2} \ten \ov{v_1} + d \ov{v_2} \ten \ov{v_2}$ in $V \ten V$, this gets the following form:

\begin{equation}
  \mu(\ol{z}) = \mu \left( \begin{array}{cc}a&b\\c&d\end{array}\right) =
  \left(\begin{array}{c}a\\d\end{array}\right) = a \ov{v_1} + d \ov{v_2}
\end{equation}

\tikzstyle{every picture}=[scale=0.5]

Computations with Frobenius algebras can  also be represented within the more general diagrammatic calculus introduced in \S \ref{sec:pictorial}. Specifically, the  linear maps of the coalgebra and algebra are depicted by:

\begin{equation}

\InputIfFileExists{./tikz/frob1.tikz}{}{\input{./tikz/frob1.tikz}}

\end{equation}

\noindent which means that the Frobenius condition takes this form:

\begin{equation}

\InputIfFileExists{./tikz/equation.tikz}{}{\input{./tikz/equation.tikz}}

\end{equation}

%
%
%

Finally, the Frobenius condition guarantees that any diagram depicting a Frobenius algebraic computation can be reduced to a normal form that only depends on the number of input and output wires of the nodes, provided that the diagram of computation is connected. This property (a high level abstract account of which is due to Lack \cite{Lack}) is referred to as the \textit{spider equality}\index{spider equality} and is depicted as follows:

\begin{equation}

\InputIfFileExists{./tikz/spider.tikz}{}{\input{./tikz/spider.tikz}}

\label{equ:spider}
\end{equation}
\vspace{0.5cm}

\index{Frobenius algebras|)}

This completes the presentation of the compositional model of Coecke et al. and the related categorical notions. At this point I should note, though, that as this thesis progresses new concepts will be introduced along the way. In Chapter \ref{ch:ambiguity}, for example, the idea of an inner product space, which currently is only implicitly defined by the presence of fixed basis in our vector spaces, will be formalized in the context of \textit{dagger compact closed categories} and specifically the category of finite-dimensional Hilbert spaces and linear maps (\textbf{FHilb}). Dagger compact closed categories (\S\ref{sec:operators}), as well as \textit{dagger Frobenius algebras} (\S\ref{sec:frob-density}), constitute the fundamental structures of categorical quantum mechanics as set by Abramsky and Coecke \cite{abramsky2004}, to which I will resort later in order to provide an extension of the original model capable of handling lexical ambiguity.

We are now ready to proceed to Part \ref{prt:theory}, which details the theory behind the contributions of this thesis. Towards this purpose, the diagrammatic calculus presented in \S\ref{sec:pictorial} will be one of my main tools for demonstrating the various concepts and ideas. 
When I find appropriate, I will also resort to the use of standard categorical commutative diagrams, similar to (\ref{equ:trans-comm}), for additional clarity. Before closing this chapter I would like to refer the reader who is interested in a more complete introduction to monoidal categories, Frobenius algebras, and their diagrammatic calculi, to the work of Coecke and Paquette \cite{CoeckePaq}.

\mbox{}\newpage

\part{Theory}
\label{prt:theory}
\mbox{}\newpage
\chapter{Building Verb Tensors using Frobenius Algebras}
\label{ch:frobverbs}

\begin{chabstract}
I begin my analysis by providing an intuition of tensors as a means for modelling relations. Then I proceed and present an implementation framework for tensors of relational words based on Frobenius algebras that extends and improves previous proposals in a number of ways. After I provide a linguistic intuition of the suggested framework, I conclude the chapter by presenting a new CDM hierarchy that incorporates the newly devised class of models. Material is based on \cite{kartsaklis2014,kartsadrqpl2014,kartsaklis2012}. 
\end{chabstract}


\noindent
To a great extent, the essence of any tensor-based model lies in the functions that produce the sentences, i.e. the verbs.\index{words!as functions} As we saw in \S \ref{sec:intuition}, the purpose of such a function is to take a number of inputs (vectors representing nouns or noun phrases) and transform them to a sentence. The important question of how a function of this form can be constructed can obviously find more than one answer, many of which might depend on the underlying task at hand or the available resources. A goal of this thesis, though, is to not be dependent on factors of this sort; I am interested in reasoning about natural language meaning at a generic level, with the hope that the findings would have something more to say than how effective a specific CDM can be on some specific task. So let me start ``by the book'' and adopt a formal semantics perspective, where the meaning of a verb can be seen as the set of all elements in the universe of discourse who perform the action that the verb describes. As will become evident, this approach leads naturally to the tensor construction method of Grefenstette and Sadrzadeh \cite{GrefenSadr1}, which will serve as a starting point for the models presented in this thesis.

\section{Verb tensors as relations}
\label{sec:tensor-rel}

\index{tensors!as relations|(}
\index{relations, as tensors|(}

Recall from the discussion in \S \ref{sec:compsem} that under an extensional perspective of semantics\index{extensional semantics} the meaning of an intransitive verb is the set of individuals who perform the action denoted by the verb:

\begin{equation}
  \sem{sleep} = \{x|x \in \mathcal{U} \wedge x~\text{sleeps}\}
  \label{equ:sleep-set}
\end{equation}

\noindent where $\mathcal{U}$ denotes the universe of discourse. I will now start to transfer this setting to a vector space environment, adopting a truth-theoretic instantiation similar to the example in \S\ref{sec:intuition}. As in that case, the noun space $N$ is spanned by a set of mutually orthogonal vectors $\{\ov{n_i}\}_i$ denoting individuals (John is represented by $\ov{n_1}$, Mary by $\ov{n_2}$ and so on). I adopt the generic convention that maps individual $n_i$ to $\ov{n_i}$.\footnote{In other words, $n_i$ refers to an actual person such as John or Mary, while $\ov{n_i}$ is the model-specific representation of that person.} So the entirety of the universe of discourse is given by a vector of 1s $\ov{U}=\sum_{n \in \mathcal{U}} \ov{n}$. The vectorial representation of a set of individuals $A\subseteq \mathcal{U}$ is just the sum of the corresponding noun vectors; naturally, the $i$th element of such a vector will be 1 if the individual who is denoted by $\ov{n_i}$ is included in set $A$, and 0 otherwise.  Continuing on the intransitive verb example, let $A$ be the set of all individuals whose denotations occur at least once as a subject of the specific verb in some corpus. Then, according to the definition above, the vectorial representation of this set is the following:\index{sets, as vectors} 

\begin{equation}
   \label{equ:intr-formal}
   \ov{verb}_{IN} = \sum_{n\in A} \ov{n}
\end{equation}

We have now an intuitive way to represent the meaning of a relational word, which in our vector space semantics takes the form of a summation over the vectors denoting the arguments of that word. Inevitably, though, this leads to a result living in the noun space $N$, seemingly contradicting the theoretical basis laid out in Chapter \ref{ch:framework}, which clearly states that words of type $n^r\cdot s$ (such as an intransitive verb) should live in the tensor product space $N\ten S$. The functorial relation between grammatical types and tensors can be restored if we observe that any set of individuals is in one-to-one correspondence with a \textit{characteristic function}\index{characteristic function} that maps elements of the domain $\mathcal{U}$ to truth values; in other words, Eq. \ref{equ:sleep-set} can also be given as:

\begin{equation}
  \sem{sleep} = \{(x,t)|x \in \mathcal{U}, t=\top~\text{if}~x~\text{sleeps and}~\bot~\text{o.w.} \}
\end{equation}

Take $\chi_A$ to be the characteristic function of set $A$; furthermore, let the sentence space $S$ be the one-dimensional space spanned by the single vector $\ov{1}$, which represents the truth value $\top$, whereas value $\bot$ is given by vector $\ov{0}$. Then, by mapping each pair $(x,t)$ of $\chi_A$ to the tensor product of the vectors denoting $x$ and $t$,\index{tensor product!for representing elements in a relation} we can re-define the meaning representation of the verb as below:

\begin{equation}
   \label{equ:intr1-formal}
   \ol{verb}_{IN} = \sum_{(n_i,\sem{\ov{s_i}}) \in \chi_A} \ov{n_i}\ten \ov{s_i}
\end{equation}

\noindent with $\ov{s_i}$ to be $\ov{1}$ if the individual $n_i$ occurs as a subject of the intransitive verb and $\ov{0}$ otherwise, and $\sem{\ov{s_i}}$ the corresponding truth value. The resulting tensor lives in $N\ten S$ as dictated by the grammar; furthermore, and due to the form of $S$, it is isomorphic to the vector of Eq. \ref{equ:intr-formal}. Assuming now that individual Mary is denoted by $\ov{n_2}$, we can compute the meaning of the sentence ``Mary sleeps'' as follows:

\small
\vspace{-0.5cm}
\begin{eqnarray}
  (\epsilon^r_N \ten 1_S) \left(\ov{n_2}\ten\sum_i(\ov{n_i}\ten\ov{s_i})\right) & = &
  (\epsilon^r_N \ten 1_S) \left(\sum_i \ov{n_2}\ten\ov{n_i}\ten\ov{s_i}\right) \nonumber \\
  & = & \sum_i \langle\ov{n_2}|\ov{n_i}\rangle\ov{s_i} =
  \sum_i \delta_{2i}\ov{s_i} \\
  & = & \ov{s_2} \left\{ \begin{array}{lr}\ov{1} & \text{if}~n_2\in A \\ \ov{0} & o.w. \end{array} \right. \nonumber
\end{eqnarray}
\normalsize

Note that in practice this is just the inner product of the vector of Mary with the vector representing the subset of individuals who sleep, given by Eq. \ref{equ:intr-formal}. Let us now examine the case of a word with two arguments, such as a transitive verb. From an extensional perspective, this will be a relation of the following form:

\vspace{-0.3cm}
\begin{equation}
  \label{equ:trans-formal}
  \sem{likes} = \{(x,y)|x,y\in\mathcal{U} \wedge x~\text{likes}~y\}
\end{equation}

Let $\ov{n_i},\ov{n_j}$ be the vectors denoting the two individuals in a pair of that relation (so that $n_i$ likes $n_j$), and take $\ov{n_i}\ten\ov{n_j}$ to be the semantic representation of that pair.\index{tensor product!for representing elements in a relation} This will be a matrix of size $|\mathcal{U}|\times|\mathcal{U}|$ with all zeros except a single 1 at position $(i,j)$. It is now straightforward to model the relation of Eq. \ref{equ:trans-formal} as the sum of all its pairs, in the following manner:

\vspace{-0.3cm}
\begin{equation}
  \label{equ:trans1-formal}
  \ol{verb}_{TR} = \sum_{(\ov{n_s},\ov{n_o}) \in R} \ov{n_s}\ten \ov{n_o}
\end{equation}

\noindent where $R$ is the underlying relation and $\ov{n_s}$, $\ov{n_o}$ denote individuals occurring as subject and object, respectively, of the specific verb in the training corpus. The result will be a matrix with an element in position $(i,j)$ to be 1 if $(n_i,n_j)$ has occurred as context of the verb (i.e. with $n_i$ to be the subject of the verb and $n_j$ its object), and 0 otherwise. As in the one-argument case, the functorial relation with the grammatical type of the verb ($n^r\cdot s \cdot n^l$) can be retained by replacing the matrix of Eq. \ref{equ:trans1-formal} with the following isomorphic order-3 tensor:

\begin{equation}
   \label{equ:tran2-formal}
   \ol{verb}_{TR} = \sum_{((n_i,n_j),\sem{\ov{s_{ij}}}) \in \chi_R} \ov{n_i}\ten \ov{s_{ij}} \ten \ov{n_j}
\end{equation}

\noindent where this time $\chi_R$ is the characteristic function of the relation defining the contexts of the transitive verb, and $\ov{s_{ij}}$ is a one-dimensional vector denoting the inclusion or the absence of the corresponding pair of individuals in that relation. It is trivial to show that computing the meaning of a sentence such as ``$n_i$ likes $n_j$'' will always result in the corresponding $\ov{s_{ij}}$ vector, i.e. the $(i,j)$th element of the matrix created by Eq. \ref{equ:trans1-formal}. 

\index{tensors!created by argument summing|(}

The discussion so far suggests that the ``argument tensoring-and-summing'' process\index{argument summing} described above is a valid way for one to model the truth-theoretic nature of formal approaches in natural language semantics, when using a vector space setting. In the first concrete implementation of the categorical framework from Grefenstette and Sadrzadeh \cite{GrefenSadr1}, the authors suggest a generalization of this method to high dimensional real-valued vectors. Specifically, for the case of a transitive verb they propose the creation of a matrix as follows:

\begin{equation}
\label{equ:weightrel}
 \ol{verb} = \sum\limits_{i}(\ov{subj_i}\otimes \ov{obj_i})
\end{equation}

\noindent where $\ov{subj_i}$ and $\ov{obj_i}$ are the distributional vectors of subject and object, respectively, created as described in \S\ref{sec:dissem}, and $i$ iterates over all contexts of the specific verb in the training corpus. As explained in the doctoral thesis of Grefenstette \cite{GrefenstetteThesis2013}, this process results in a structural mixing of the argument vectors, which reflects the extent to which each component of the subject vector is compatible with all the components of the object vector, and vice versa. 

\index{tensors!created by argument summing|)}

For the purposes of this thesis, the argument summing procedure of \cite{GrefenSadr1} has many important advantages. As demonstrated above, it is obviously aligned with the formal semantics perspective, which makes perfect sense for a model that can be seen as a mathematical counterpart of this traditional view of meaning in language. Furthermore, it is generic enough not to be tied to any particular task or methodology, while at the same time its implementation and testing remain straightforward. Last, but not least, it creates \textit{reduced representations} for relational words, since in all cases the order of the produced tensor is lower by one than the order dictated by the grammatical type. This is a quite important achievement, since for tensor-based models space complexity is always an issue. If, for example, we assume noun and sentence vectors with 300 dimensions, a transitive verb (a tensor of order 3) will require 27 million parameters, while for the case of a ditransitive verb this number is increased to 8.1 billion. Overall, I consider the argument summing procedure an interesting test-bed and an appropriate tool for demonstrating the ideas presented in this thesis.

\index{relations, as tensors|)}
\index{tensors!as relations|)}

\section{Introducing Frobenius algebras in language}
\label{sec:recast}

\index{Frobenius algebras!in language|(}

\tikzstyle{every picture}=[scale=0.35,baseline=0pt]

Despite the many benefits of the reduced tensors presented in \S\ref{sec:tensor-rel}, there is still an important problem we have to solve in order to make this method viable for real-world tasks. Recall that in its current form, our sentence space is the unit (the field). The transition from the truth-theoretic setting to real-valued vector spaces means that the single element that will result from a composition would not be restricted to 0 or 1, but it would be some number that, under proper normalization conditions, can show how  \textit{probable} is the specific subject-object pair for the verb at hand. Our composition function in this case is nothing more than an endomorphism of the field, as shown below for a transitive verb matrix created by Eq. \ref{equ:weightrel}:

\begin{equation}
  I \cong I\ten I \ten I \xrightarrow{\ov{s}\ten\ol{v}\ten\ov{o}}N\ten N \ten N \ten N \xrightarrow{\epsilon^r_N \ten \epsilon^l_N} I\ten I \cong I
\end{equation}

For any practical application, this is of course inadequate. We need a way to expand the verb matrix into a tensor of higher order, thus providing somehow the sentence dimension that is now missing. Recall from our discussion in \S \ref{sec:frobenius} that every vector space with a fixed basis comes with a Frobenius algebra over it, which provides canonical ways for \textit{copying} or \textit{deleting} elements of that basis.\index{semantic spaces!copying of basis}\index{semantic spaces!deleting of basis}\index{copying of basis}\index{deleting of basis} Since in this case we are interested in copying, our tool will be a Frobenius map $\Delta: N \to N\ten N$, applied on one or more of the original dimensions of our matrix, which currently lives in $N\ten N$. 

As it turns out, the decision of which (and how many) dimensions of the verb matrix to copy has some important consequences. We will start by examining the case of copying both dimensions, in which our inflated tensor takes the following form:

\begin{equation}
  I \xrightarrow{\ol{verb}} N\ten N \xrightarrow{\Delta_N \ten \Delta_N} N\ten N \ten N \ten N
\end{equation}

How does this tensor form fit to the required grammatical type $n^r\cdot s \cdot n^l$? What we implicitly assume here is that $S=N\ten N$, which means that the composition of a transitive verb with a specific subject-object pair will result in a matrix. In fact, this is exactly the composition model proposed in \cite{GrefenSadr1} and \cite{GrefenstetteThesis2013}, where the meaning of a sentence is represented as an order-$n$ tensor, with $n$ to be the number of arguments for the head word of the sentence. In other words, when following this approach an intransitive sentence lives in a space $S=N$, a transitive one in $S=N\otimes N$ and so on. Let us show this graphically; in our diagrammatic calculus, the inflated tensor is depicted by the left-hand diagram below. When substituting the verb in Def. \ref{def:categorical}, the composition proceeds as in the right-hand diagram:

\begin{equation}
\footnotesize

\InputIfFileExists{./tikz/rel.tikz}{}{\input{./tikz/rel.tikz}}
 
\normalsize
\label{fig:frobrel-ed}
\end{equation}
\vspace{0.1cm}

It is immediately obvious that the involvement of $\Delta$-maps in the creation of the verb tensor imposes a restriction: since now our sentence space is produced by copying basis elements of the noun space, our functor $\mathcal{F}$ cannot any more apply different mappings on the two atomic pregroup types $s$ and $n$; both of these should be mapped onto the same basic vector space, bringing the model closer to vector mixtures than originally aimed at by the theory. 
Indeed, it is easy to show that in the extreme case that one copies \textit{all} the dimensions of a relational tensor, as we do here, the model reduces itself to a simple point-wise multiplication of the arguments with the verb tensor itself. This is exactly what diagram (\ref{fig:frobrel-ed}) reflects, and it is also obvious from the closed-form formula produced by the composition: Let $\ov{subj} = \sum_i s_i \ov{n_i}$, $\ov{obj} = \sum_j o_j \ov{n_j}$ be the vectors for the context of a transitive verb, the initial matrix of which is given by $\ol{verb} = \sum_{ij} v_{ij} \ov{n_i} \ten \ov{n_j}$. Taking $\ol{verb}~' = (\Delta_N\ten \Delta_N)(\ol{verb})=\sum_{ij} v_{ij} \ov{n_i} \ten \ov{n_i} \ten \ov{n_j} \ten \ov{n_j}$ leads to the following composition:

\vspace{-0.7cm}
   \begin{eqnarray}
     (\epsilon^r_N \ten 1_{N\ten N} \ten \epsilon^l_N)
     (\ov{subj} \ten \ol{verb}~' \ten \ov{obj}) & = \nonumber \\
     \sum\limits_{ij} v_{ij} \langle \ov{subj}|\ov{n_i} \rangle \langle \ov{n_j}|\ov{obj} \rangle \ov{n_i} \ten \ov{n_j} & = \\
     \sum\limits_{ij} v_{ij} s_i o_{j} \ov{n_i} \ten \ov{n_j} = \ol{verb} \odot (\ov{subj} \ten \ov{obj}) \nonumber
   \end{eqnarray}

Hence, what we actually get is a sophisticated version of a vector mixture model (although not exactly in the word order forgetting sense of it), where the representation of a sentence is always the tensor product of the arguments point-wise multiplied by the tensor of the head verb. Which now makes very clear the reason behind the other unwelcome property of this formulation, that every sentence has to live in a space of a different tensor power, equal to that of the head verb. Especially this latter side-effect is in direct conflict with the premises of the categorical framework, since there is no way for one to compare sentences of different structures, say an intransitive one with a transitive one. Even more importantly, it means that the model is unable to assign a meaning to sentences or phrases with nested grammatical structures, such as in the following case:

\begin{equation}
\small

\InputIfFileExists{./tikz/mary.tikz}{}{\input{./tikz/mary.tikz}}

\normalsize
\end{equation}

Due to the mismatch between the logical and the concrete types, the translation of the above derivation to vector spaces is not possible; the multi-linear map for `read' expects a vector in $N$ as a semantic representation for its object, but what it actually receives is a tensor of order 2 (a matrix), since preposition `about' is a function of two arguments. The next section deals with all the above issues, building on and extending my previously published work with Sadrzadeh, Coecke and Pulman \cite{kartsaklis2012,kartsaklis2014} on this topic. 

\index{Frobenius algebras!in language|)}

\section{Unifying the sentence space}
\label{sec:sentencespace}

\index{sentence space!unification|(}
\index{Frobenius algebras!unification of sentence space|(}

The remedy for the problems raised with the approach of \cite{GrefenSadr1} is simple and stems directly from the two main limitations we have to take into account: (a) $S$ must be equal to $N$; (b) a tensor with $n$ arguments will always return a vector in $N$ iff its order is $n+1$. Let me enforce the first condition by modifying the syntax-to-semantics functor so that the assignments of the basic grammar types in Eq. \ref{equ:map} are replaced with the following:

\begin{equation}
  \label{equ:mapmod}
  \M{F}(n) = N~~~~~~~~~\M{F}(s) = N
\end{equation}

Since we start from reduced tensors, the order of which is equal to the number of the verb arguments, we can conform to the second condition and properly restore the functorial passage from grammar to vector spaces by restricting ourselves in copying only \textit{one} of the tensor dimensions. As we will see later, this leads to interesting consequences that allow intuitive linguistic interpretations of the resulting framework. 

\index{transitive verbs, and Frobenius algebras|(}

I will continue using the case of a transitive verb as a running example. We start from a matrix created as dictated by Eq. \ref{equ:weightrel}; this is a tensor living in $N \ten N$ that has to be encoded in $N \ten N \ten N$. A transitive verb is a function of two arguments, so there are two different ways to apply the Frobenius $\Delta$-map:

\index{Copy-Subject model|(}
\index{Frobenius algebras!Copy-Subject model|(}

\paragraph{Copy-Subject} The first option is to copy the ``row'' dimension of the matrix corresponding to the verb. This dimension is the one that interacts with the subject noun during the composition. In diagrammatic form, the tensor and the composition are as follows:

\begin{equation}
\footnotesize

\InputIfFileExists{./tikz/copysbj.tikz}{}{\input{./tikz/copysbj.tikz}}
 
\label{fig:copysbj}
\end{equation}

\vspace{0.5cm}
The compositional morphism in this case is the following:

\begin{eqnarray}
  (\epsilon^r_N \ten 1_N \ten \epsilon^l_N) \circ 
  (1_N \ten \Delta_N \ten 1_N \ten 1_N)
  (\ov{subj} \ten \ol{verb} \ten \ov{obj}) & = \\
  (\mu_N\ten\epsilon^l_N)(\ov{subj} \ten \ol{verb} \ten \ov{obj}) 
  \nonumber
 \label{equ:csbj-cat}
\end{eqnarray}

\noindent as the normal form in (\ref{fig:copysbj}) and the commutative diagram below makes clear:

\begin{equation}
\begin{tikzpicture}[scale=1.5,baseline=30pt]
font=\footnotesize
\node (A) at (0,4) {$I\ten I\ten I$};
\node (B) at (8,4) {$N\ten N\ten N\ten N$};
\node (C) at (20,4) {$N\ten N\ten N\ten N\ten N$};
\node (D) at (20,0) {$N$};
\path[->,font=\scriptsize]
(A) edge node[above]{$\ov{s}\ten\ol{v}\ten\ov{o}$} (B)
(B) edge node[above]{$1_N\ten \Delta_N \ten 1_N\ten 1_N$} (C)
(C) edge node[left]{$\epsilon^r_N\ten 1_N \ten \epsilon^l_N$} (D)
(B) edge node[left]{$\mu_N \ten \epsilon^l_N~~~~$} (D);
\end{tikzpicture}
\end{equation}

Linear-algebraically (and using the subject, verb, and object vectors of our previous example), the computation proceeds as follows:

\begin{eqnarray}
(\mu_N\ten\epsilon_N^l)(\ov{subj}\ten\ol{verb}\ten\ov{obj}) =
\sum\limits_{ij}v_{ij}s_io_j \mu_N(\ov{n_i}\ten\ov{n_i}) \epsilon^l_N(\ov{n_j}\ten \ov{n_j}) & = \nonumber \\
\sum\limits_{ij}v_{ij}s_io_j \ov{n_i} \langle \ov{n_j}|\ov{n_j} \rangle = \sum\limits_{ij}v_{ij}s_io_j \ov{n_i} & = \\
\ov{subj} \odot (\ol{verb} \times \ov{obj}) \nonumber
\end{eqnarray}

\index{Frobenius algebras!Copy-Subject model|)}
\index{Copy-Subject model|)}

\index{Copy-Object model|(}
\index{Frobenius algebras!Copy-Object model|(}

\paragraph{Copy-Object} Our other option is to copy the ``column'' dimension of the verb matrix, the one that interacts with objects:

\begin{equation}
\footnotesize

\InputIfFileExists{./tikz/copyobj.tikz}{}{\input{./tikz/copyobj.tikz}}
 
\normalsize
\label{fig:copyobj}
\end{equation}

\vspace{0.5cm}
Eq. \ref{equ:cobj-cat} provides the categorical morphism and the linear-algebraic form (which I am not going to derive explicitly this time):

\begin{eqnarray}
  \label{equ:cobj-cat}
  (\epsilon^r_N \ten 1_N \ten \epsilon^l_N)\circ
  (1_N\ten 1_N\ten \Delta_N \ten 1_N) 
  (\ov{subj} \ten \ol{verb} \ten \ov{obj}) & = \nonumber \\
  (\epsilon^r_N \ten \mu_N)
  (\ov{subj} \ten \ol{verb} \ten \ov{obj}) & = \\
  \ov{obj} \odot (\ol{verb}^{\mathsf{T}} \times \ov{subj}) \nonumber
\end{eqnarray}

\index{Frobenius algebras!Copy-Object model|)}
\index{Copy-Object model|)}

\begin{figure}[h!]
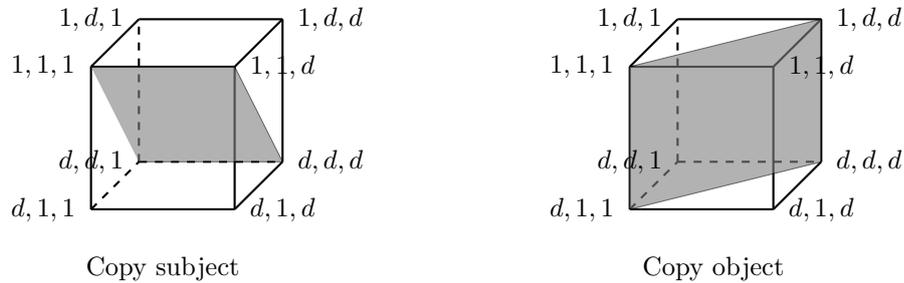

  \footnotesize
  \ctikzfig{cubes}
  \normalsize
  \caption{Geometric interpretation of the Frobenius models.}
  \label{fig:cubes}
\end{figure}

From a geometric perspective, the two Frobenius models correspond to different ways of ``diagonally'' placing a plane into a cube. This is shown in Fig. \ref{fig:cubes}.

\index{transitive verbs, and Frobenius algebras|)}

\index{Copy-Subject model!interpretation|(}
\index{Copy-Object model!interpretation|(}
\index{Frobenius algebras!interpretation of applying on language|(}

In contrast to the original model of \cite{GrefenSadr1}, the interpretation of the Copy-Subject and Copy-Object models is quite different. Although there is still an element of point-wise multiplication, now part of the sentence is computed by tensor contraction. For the case of the Copy-Subject model, the composite part corresponds to the verb phrase (verb-object) and is produced by matrix-multiplying the verb matrix with the object vector. In other words, what actually is achieved is a merging of the two parts of the sentence (subject and verb phrase), by point-wise multiplying the vector of the subject with the composite vector of the verb phrase.

\begin{equation}
 \footnotesize
 
\InputIfFileExists{./tikz/copysbj-normal.tikz}{}{\input{./tikz/copysbj-normal.tikz}}

 \normalsize
\end{equation}

\vspace{0.5cm}
The situation is very similar for the Copy-Object model, with the difference that now a vector is ``categorically'' composed for the intransitive part of the sentence (subject-verb), and then this vector is merged with the vector of the object:

\begin{equation}
 \footnotesize
 
\InputIfFileExists{./tikz/copyobj-normal.tikz}{}{\input{./tikz/copyobj-normal.tikz}}

 \normalsize
\end{equation}

\vspace{0.5cm}
The particular perspective of breaking the sentence into two parts with the verb as a common joint provides another important advantage: it actually \textit{justifies} the somewhat ad-hoc decision to use a matrix in the compositional equations, since in both the cases of a verb phrase and an intransitive sentence the verb is a function of a single argument---that is, it can be canonically represented as a tensor of order 2. As we will see, my analysis suggests that the matrix in the Copy-Subject model must be different from the matrix of the Copy-Object model, since these two tensors represent different functions. I will return to this topic in \S \ref{sec:frob-regression}, when I will discuss statistical techniques for building the relational tensors. Furthermore, in \S\ref{sec:intonation} we will see that the proposed model finds an appealing interpretation related to the linguistic phenomenon known as intonation. For the moment, though, I will show how the Frobenius setting can be extended to tensors of any arity.

\index{Frobenius algebras!interpretation of applying on language|)}
\index{Copy-Subject model!interpretation|)}
\index{Copy-Object model!interpretation|)}

\index{Frobenius algebras!unification of sentence space|)}
\index{sentence space!unification|)}

\section{Extending the setting to any tensor}
\label{sec:ditransitive}

\index{Frobenius algebras!on tensors of any arity|(}

The application of Frobenius operators extends to any relational word, providing a generic treatment for all aspects of language.   According to the ``argument summing'' procedure of \cite{GrefenSadr1},\index{argument summing} for example, the linear maps of adjectives and intransitive verbs (both functions of a  single argument) will be elements of $N$ created as follows:

\begin{equation}
   \ov{adj} = \sum\limits_i \ov{noun_i}\quad\quad\quad
   \ov{verb}_{IN} = \sum\limits_i \ov{subj_i}
\end{equation}

\noindent with $i$ iterating through all relevant contexts. In both cases the application of a $\Delta$ operation will create a diagonal matrix, which when substituted in Def. \ref{def:categorical} will produce the result shown below. 

\begin{equation}
\footnotesize

\InputIfFileExists{./tikz/adj-intr.tikz}{}{\input{./tikz/adj-intr.tikz}}

\normalsize
\label{fig:single-arg}
\end{equation}

Note that this is nothing more than another case of copying ``all'' dimensions of a tensor (here just one), that as we discussed in \S \ref{sec:sentencespace} results in a degeneration to an element-wise multiplicative model. Specifically, the meaning of adjective-noun compounds and intransitive sentences reduces to the following:

\vspace{-0.5cm}
\begin{gather}
  \ov{adj~noun} = \mu(\ov{adj}\ten\ov{noun}) = \ov{adj} \odot \ov{noun} \\
  \ov{s_{IN}} = \mu(\ov{subj}\ten\ov{verb}) = \ov{subj} \odot \ov{verb} 
\end{gather}

\index{ditransitive verbs, and Frobenius algebras|(}

I will now proceed to the more interesting case of a ditransitive sentence, the derivation of which is depicted here: 

\begin{equation}
\begin{minipage}{0.35\linewidth}
  \footnotesize
  \ctikzfig{dtrans-preg}
\end{minipage}
\begin{minipage}{0.35\linewidth}
   \footnotesize
   \ctikzfig{dtrans}
\end{minipage}
\label{fig:dtrans}
\end{equation}

A ditransitive verb can be seen as a function of 3 arguments; hence, we can start by creating a tensor of order 3 that represents the verb, as follows:

\begin{equation}
  \ol{verb}_{DT} = \sum\limits_i \ov{subj_i} \ten \ov{iobj_i} \ten \ov{dobj_i}
\end{equation}

\noindent
where $\ov{iobj}$ and $\ov{dobj}$ refer to the vectors of the indirect and direct object, respectively. In this case, the Frobenius operators offer to us three alternatives, all of which are shown below:

\begin{eqnarray}
  \begin{minipage}{0.35\linewidth}
    \footnotesize
    \ctikzfig{dt-copysbj}
    \centering
    Copy subject
  \end{minipage}
  \begin{minipage}{0.35\linewidth}
    \footnotesize
    \ctikzfig{dt-copydobj}
    \centering
    Copy direct object
    \label{fig:frob-ditrans}
  \end{minipage} \\\nonumber
  ~\\\nonumber
  \begin{minipage}{0.70\linewidth}
    \footnotesize
    \ctikzfig{dt-copyiobj}
    \centering
    Copy indirect object
  \end{minipage}
\end{eqnarray}

\vspace{0.2cm}
Note that the third case (copy indirect object) makes use of the symmetry of the category. In all cases the result is a vector, computed as below:

\begin{equation}
  \text{Copy subject:}~\ov{s_{DT}} = \ov{subj} \odot \left((\ol{verb} \times \ov{iobj})\times \ov{dobj} \right)
  \label{equ:copysbj-dit}
\end{equation}

\vspace{-0.6cm}
\begin{equation}
  \text{Copy direct object:}~\ov{s_{DT}} = \ov{dobj} \odot \left((\ol{verb}\times \ov{iobj})^{\mathsf{T}} \times \ov{subj} \right)
  \label{equ:copydobj-dit}
\end{equation}

\vspace{-0.6cm}
\begin{equation}
  \text{Copy indirect object:}~\ov{s_{DT}} = \ov{iobj} \odot \left((\ol{verb}\times \ov{dobj})^{\mathsf{T}} \times \ov{subj} \right)
  \label{equ:copyiobj-dit}
\end{equation}

\index{ditransitive verbs, and Frobenius algebras|)}
\index{Frobenius algebras!on tensors of any arity|)}

As in the transitive case, the sentence is broken into two parts: one part for the copied argument and one part for the rest of the sentence. In all cases, the argument vector is merged by point-wise multiplication with the composite vector of the context. Let me now formally define the concept of a \textit{Frobenius  vector}:\index{Frobenius vector}

\begin{definition}
\label{def:frobvector}
  Let a text constituent with head word $w$ of arity $n$ and grammatical derivation $\alpha$; the $i$th Frobenius vector of this constituent is defined as follows:
  
  \begin{equation}
    \ov{v}_i^{\mathsf{F}} = \mathcal{F}(\alpha) (\ov{arg_1}\ten \hdots \ten \Delta_i(\ol{w}) \ten \hdots \ten \ov{arg_n})
  \end{equation}
  
\noindent where $\Delta_i(\ol{w})$ represents the application of a Frobenius $\Delta$-map on the $i$th dimension of the reduced tensor $\ol{w}$ for that word.

\end{definition}

%

If $w$ is a verb, then the Frobenius vectors $\ov{v}^{\mathsf{F}}_i$ will represent different aspects of the meaning of a sentence. In the next section I will provide an intuitive interpretation of these different aspects in the context of intonation.

\section{Modelling intonation}
\label{sec:intonation}

\index{Frobenius algebras!and intonation|(}

The term \textit{intonation}\index{intonation} refers to variations of spoken pitch, the purpose of which is to emphasize parts of the utterance that might be important for the conveyed message. Consider again the ditransitive sentence of \S \ref{sec:ditransitive}:

\begin{exe}
\ex\label{ex:john} John gave Mary a flower
\end{exe}

This statement might have been used in response to a number of questions, each one of which requires the use of a different intonational emphasis from the speaker. For example:

\begin{exe}
  \ex
  \begin{xlist}
  \ex \textit{Who gave Mary a flower?}\\
      \textbf{\underline{John}} gave Mary a flower
  \ex \textit{What did John give to Mary?}\\
      John gave Mary \textbf{\underline{a flower}}
  \ex \textit{To whom did John give a flower?} \\
      John gave \textbf{\underline{Mary}} a flower    
  \end{xlist}    
\end{exe}

The bold-faced part above is what Steedman \cite{steedman2000information} calls \textit{rheme}\index{rheme}---the information that the speaker wishes to make common ground for the listener. The rest of the sentence, i.e. what the listener already knows, is called \textit{theme}.\index{theme} In the same paper, Steedman argues that the different syntactic derivations that one can get even for very simple sentences when using Combinatory Categorial Grammar (CCG)\index{combinatory categorial grammar (CCG)} \cite{steedman} (referred to with the somewhat belittling term ``spurious readings''),\index{spurious readings, in CCG} actually serve to reflect changes in the intonational patterns. In CCG, our example sentence in (\ref{ex:john}) has a number of different syntactic derivations, two of them are the following:

\begin{equation}
\begin{minipage}{0.80\linewidth}
\begin{center}  
\deriv{4}{
{\rm John} & {\rm gave} & {\rm Mary} & {\rm a~flower} \\
\uline{1} & \uline{1} &\uline{1} & \uline{1} \\
\textsc{np} & \textsc{((s\bs np)/np)/np} & \textsc{np} & \textsc{np} \\
   & \fapply{2} & \\
   & \cmc{2}{\textsc{(s\bs np)/np}} & \\
   & \fapply{3} \\
   & \cmc{3}{\textsc{s\bs np}} \\
\bapply{4} \\
\cmc{4}{\textsc{s}} \\
}  
\end{center}
\end{minipage}
\label{fig:spur1}
\end{equation}

\begin{equation}
\begin{minipage}{0.80\linewidth}
\begin{center}  
\deriv{4}{
{\rm John} & {\rm gave} & {\rm Mary} & {\rm a~flower} \\
\uline{1} & \uline{1} &\uline{1} & \uline{1} \\
\textsc{np} & \textsc{((s\bs np)/np)/np} & \textsc{np} & \textsc{np} \\
\ftype{1} & \fapply{2} & \\
{\textsc{s/(s\bs np)}} & \cmc{2}{\textsc{(s\bs np)/np}} & \\
\fcomp{3} & \\
\cmc{3}{\textsc{s/np}} & \\
\fapply{4} \\
\cmc{4}{\textsc{s}} \\
}  
\end{center}
\end{minipage}
\label{fig:spur2}
\end{equation}

Note that (\ref{fig:spur1}) proceeds by first composing the part of the verb phrase (``gave Mary a flower''); later, in the final step, the verb phrase is composed with the subject `John'. The situation is reversed for (\ref{fig:spur2}), where the use of type-raising and composition rules of CCG allow the construction of the fragment ``John gave Mary'' as valid grammatical text constituent, which is later combined with the direct object of the sentence (`a flower'). According to Steedman, each one of these derivations expresses a different intonational pattern, distinguishing the rheme from the theme when the sentence is used for answering different questions:  (\ref{fig:spur1}) answers to ``Who gave Mary a flower?'', whereas (\ref{fig:spur2}) to ``What did John give to Mary?''. 

It is interesting to note that in our grammatical formalism, pregroup grammars, variations in a grammatical derivation similar to above are only implicitly assumed, since the order of composition between a relational word and its argument remains unspecified. This fact is apparent in the pregroup derivation of our sentence in (\ref{fig:dtrans}), and is directly reflected in our semantic space through the functorial passage:

\begin{eqnarray}
  \nonumber
 \ov{john}^{\mathsf{T}} \times  \left( (\ol{gave} \times \ov{mary}) \times \ov{flower} \right) & = \\
  \left( (\ov{john}^{\mathsf{T}} \times \ol{gave}) \times \ov{mary} \right) \times \ov{flower} & = \label{equ:transparency}\\
  \left( (\ov{john}^{\mathsf{T}} \times \ol{gave})^{\mathsf{T}} \times \ov{flower} \right) \times \ov{mary} \nonumber
\end{eqnarray}

Eq. \ref{equ:transparency} constitutes a natural manifestation of the \textit{principle of combinatory transparency} \cite{steedman}:\index{combinatory transparency, principle} no matter in what order the various text constituents are combined, the semantic representation assigned to the sentence is always the same. However, if the claim that different grammatical derivations (thus different ways of combining constituents) subsume different intonational patterns is true, then this must also be reflected somehow in the semantic form of the sentence; in our case, the sentence vector. In other words, the claim here is that the meaning of sentence ``\textbf{\underline{Mary}} wrote a book about bats'' is slightly different of that of sentence ``Mary wrote a book about \textbf{\underline{bats}}''. Indeed, if one uses the former form instead of the latter to answer the question ``I \textit{know} that Mary wrote a book, but what was this book about?'', the person who posed the question would require a moment of mental processing to translate the semantics of the response in her mind.

Let me try making this idea more concrete by extending our pregroup grammar\index{intonation!and pregroup grammars}\index{pregroup grammars!and intonation} in order to reflect intonational information; specifically, I will introduce a new atomic type $\rho \leq n$ which will mark the part of a sentence that corresponds to rheme. The different intonational patterns in our example sentence now get the following distinct derivations:

\begin{equation}
  \footnotesize
  
\InputIfFileExists{./tikz/intonation1.tikz}{}{\input{./tikz/intonation1.tikz}}

  \normalsize
  \label{equ:intonation}
\end{equation}
\vspace{0.3cm}

Each intonational pattern requires a variation of the basic $n^r\cdot s\cdot n^l\cdot n^l$ grammatical type which, when composed with the rest of the context, leads to slightly different versions of the final result. The Frobenius setting detailed in this chapter provides a natural way to incorporate this kind of variation in the meaning of the relational tensors, and in turn in the meaning of sentences. Recall that each one of the alternative options for copying a dimension of a tensor results in (a) splitting the sentence in two parts, one of which consists of a single argument (the one corresponding to the copied dimension) and one for the rest of the sentence; (b) composing a vector for the rest of the sentence; and (c) point-wise multiplying the vector of the argument with the composite vector. I argue that the isolated argument corresponds to the rheme, while the composite part serves as the theme. In other words, each one of the types of `gave' in (\ref{equ:intonation}) must be assigned to a tensor produced by copying a different argument, as below:

\begin{equation}
  \small
  
\InputIfFileExists{./tikz/intonation3.tikz}{}{\input{./tikz/intonation3.tikz}}

  \normalsize
  \label{equ:intonation2}
\end{equation}
\vspace{0.3cm}

The point-wise multiplication of the theme with the rheme provides a unified meaning for the sentence, in which however the rheme will play a very important role: the nature of the compositional operator (a vector mixture) guarantees that the resulting vector will be affected \textit{equally} from both arguments, rheme and theme. This puts the necessary focus on the appropriate part of the sentence, reflecting the variation in semantics intended by the intonational pattern (see also Diagram \ref{fig:frob-ditrans}):

\begin{exe}
\ex
\begin{xlist}
 \ex Who gave Mary a flower?\\
     $\ov{john} \odot \ov{gave~mary~a~flower}$
 \ex What did John give to Mary?\\
     $\ov{john~gave~mary} \odot \ov{a~flower}$
 \ex To whom did John give a flower?\\
     $\ov{john~gave~a~flower} \odot \ov{(to)~mary}$
\end{xlist} 
\end{exe}

The following definition makes this idea precise:\index{Frobenius vector!as means for modelling intonation}

\begin{definition}
  The meaning of a text constituent with head word $w$ and arity $n$ carrying intonational information on the $i$th argument is given by its $i$th Frobenius vector.
\end{definition}

There is a different kind of question that has not been addressed yet:

\begin{exe}
\ex What just happened?\\
    \textbf{\underline{John gave Mary a flower}}
\end{exe}

For simple declarative cases such that the above, where the theme is unmarked,\index{unmarked theme} I will define the meaning of a sentence as the sum of all intonational patterns. Since each intonational pattern corresponds to a different Frobenius vector, as these defined in Def. \ref{def:frobvector}, we conclude to the following:\index{Frobenius additive model}

\begin{definition}
\label{def:frobadd}
  The meaning of a sentence with an unmarked theme is defined as the sum of all its Frobenius vectors:
  
  \begin{equation}
     \ov{s_U} = \ov{v}^{\mathsf{F}}_1 + \ov{v}^{\mathsf{F}}_2 \hdots + \ov{v}^{\mathsf{F}}_n
  \end{equation}
\end{definition}

The material presented in this section is admittedly only the first step towards a complete treatment of intonation in the context of CDMs.\footnote{A more extended version of this study can be found in \cite{intonation} (joint work with M. Sadrzadeh).} An interesting point that requires further study, for example, is how the above setting behaves in the presence of focus-sensitive elements, such as adverbs like `only' and `even'. Dealing with questions like these, together with a practical evaluation explicitly focused on the topic of intonation, is deferred for future research. For my current purposes, the model of Def. \ref{def:frobadd} (to which I will refer as \textbf{Frobenius additive}) and the other Frobenius models presented in this chapter will be evaluated on a number of generic tasks in Chapter \ref{ch:frobexp}. Furthermore, in the next section I provide a theoretical example of the Frobenius setting that hopefully will make the concept better understood to the reader.

\index{Frobenius algebras!and intonation|)}

\section{A truth-theoretic instantiation}
\label{sec:truth-theoretic}

\index{Frobenius algebras!truth-theoretic example|(}

The purpose of this section is to provide some intuition for the Frobenius framework by showing how it behaves in a simple truth-theoretic setting similar to that of \S\ref{sec:tensor-rel}. Recall that a relation corresponding to the meaning of a transitive verb, such as `likes', can be given by a matrix $\ol{likes}=\sum likes_{ij} \ov{n_i}\ten\ov{n_j}$ where $likes_{ij}$ is the scalar 1 if individual $n_i$ likes individual $n_j$ and 0 otherwise. For individuals $n_1$ and $n_3$, copying the subject dimension of this matrix proceeds as follows:

\vspace{-0.5cm}
\small
\begin{eqnarray}
\label{equ:likes-frob}
(\mu_N \ten \epsilon^l_N) \left( \ov{n_1} \ten \left( \sum\limits_{ij} likes_{ij} \ov{n_i} \ten \ov{n_j}\right) \ten \ov{n_3} \right) & = \nonumber \\
\sum\limits_{ij} likes_{ij} \mu(\ov{n_1}\ten \ov{n_i}) \langle \ov{n_j}|\ov{n_3} \rangle = 
\sum\limits_{ij} likes_{ij} \delta_{1i} \delta_{j3} \ov{n_i} & = \\ likes_{13} \ov{n_1} = 
\left\{ \begin{array}{lr} \ov{n_1} & \text{if~}likes_{13}=1 \\ \ov{0} & o.w. \end{array} \right. \nonumber
\end{eqnarray}
\normalsize

The above example illustrates clearly the necessity of a shared space between sentences and words when using Frobenius operators, since what we get back is a vector in $N$; in case that the sentence is true, this vector corresponds to the subject of the verb, otherwise it is the zero vector. However, how can one interpret the result of (\ref{equ:likes-frob}) in a way consistent with the model? Let us repeat the derivation step-wise, starting from the composition of the verb with the object:

\vspace{-0.5cm}
\small
\begin{eqnarray}
  (1_N \ten \epsilon^l_N) \left( \left( \sum\limits_{ij} likes_{ij} \ov{n_i}\ten \ov{n_j} \right) \ten \ov{n_3} \right) =
  \sum\limits_{ij} likes_{ij} \ov{n_i} \langle \ov{n_j}|\ov{n_3}\rangle & = \\
  \sum\limits_{ij} likes_{ij} \ov{n_i} \delta_{j3} = \sum\limits_{i} likes_{i3} \ov{n_i} \nonumber
\end{eqnarray}
\normalsize

The vectorial representation of the verb phrase ``likes $n_3$'' gets a very intuitive explanation: it is the sum of all individuals who like the person denoted by vector $\ov{n_3}$. Since in our setting individuals form an orthonormal basis, the $i$th element of this vector corresponds to the person $n_i$, and has value 1 if this person likes $n_3$ or 0 otherwise. Indeed, as we discussed in \S\ref{sec:tensor-rel} this sum can be seen as a \textit{subset} of elements in our universe of discourse, the entirety of which is represented by the vector $\sum_i \ov{n_i}$. 

What differentiates the Frobenius setting from the standard categorical composition is the second step, in our case that of composing the meaning of the verb phrase with the subject. Recall that this is just the point-wise multiplication of the vector of the verb phrase with the vector of the subject, which in our setting can be interpreted as the \textit{intersection} of the singleton set formed by the subject with the set of people who like individual $n_3$. 

\small
\singlespace
\begin{equation}
  \ov{n_1} \odot \ov{likes~n_3} = 
  \left( \begin{array}{c} 0\\1\\0\\0 \end{array} \right) \odot
  \left( \begin{array}{c} 1\\1\\0\\1 \end{array} \right) = 
  \left( \begin{array}{c} 0\\1\\0\\0 \end{array} \right)
\end{equation}
\onehalfspace
\normalsize
\vspace{0.1cm}

This is just a test of set membership: if the result is the subject vector, this means that the corresponding individual is a member of the set of people who like $n_3$; in the opposite case, the result will be the zero vector (the empty set). Let me summarize these observations using as an example the concrete sentence ``John likes Mary''. In the traditional categorical composition, the meaning of this sentence is given by:

\vspace{-0.4cm}
\begin{equation}
  \text{John}~\in~\{x|x~\text{likes Mary}\}
\end{equation}

\noindent and the result can be one of the two values in our sentence space, i.e. $\top$ or $\bot$. In the Frobenius setting, the meaning of the sentence changes to:

\vspace{-0.4cm}
\begin{equation}
   \{\text{John}\} \cap \{x|x~\text{likes Mary}\}
\end{equation}

\noindent and the result is $\{\text{John}\}$ if John is among the people who like Mary, or the empty set otherwise. Hence, while the goal is still the same (to state the truth or falseness of the sentence), now we have a different way to express the result.

\index{intonation!truth-theoretic example|(}

Even more interestingly, the above analysis provides a direct justification of why the Frobenius models constitute an appropriate way of modelling intonation: when answering the question ``Who likes Mary?'', the correct answer is ``John'' and not ``true'' or ``false''. Furthermore, it is not difficult to show that when the object dimension of the verb matrix is copied, the evaluation of the sentence takes the following form:

\begin{equation}
   \{\text{Mary}\} \cap \{x|x~\text{is liked by John}\}
\end{equation}

A positive result ($\{\text{Mary}\}$) answers directly to the question ``Whom John likes?'', setting the focus on the appropriate part of the sentence. 

%
%
%
%
%

\index{intonation!truth-theoretic example|)}
\index{Frobenius algebras!truth-theoretic example|)}

\section{Frobenius operators and entanglement}
\label{sec:entanglement}

\index{Frobenius algebras!and entanglement|(}
\index{entanglement|(}
\index{tensor-based models!and entanglement|(}

In this section I will shortly demonstrate the benefits of using Frobenius operators for constructing relational tensors with regard to \textit{entanglement}. As in quantum mechanics, entanglement is a necessary requirement for tensor-based models to allow the unhindered flow of information between the different parts of the sentence. Recall that a word tensor living in vector space $W$ is seen as a state of $W$. This state is \textit{separable} if it can be expressed as the tensor product of two or more vectors. In our graphical calculus, these objects  are depicted  by  the juxtaposition of two or more triangles:

\begin{equation}
\footnotesize

\InputIfFileExists{./tikz/separ.tikz}{}{\input{./tikz/separ.tikz}}

\normalsize
\end{equation}

In general, a state is not separable if it is a linear combination of many separable states. The number of separable states needed to express the original tensor is equal to  the \textit{tensor rank}. Graphically, a tensor of this form is shown as a single triangle with two or more legs: 

\begin{equation}
\footnotesize

\InputIfFileExists{./tikz/entang.tikz}{}{\input{./tikz/entang.tikz}}

\normalsize
\end{equation}

\index{flow of information|(}
\index{separability|(}
\index{tensors!separability|(}

In categorical quantum mechanics terms, entangled states are necessary to allow the flow of information between the different subsystems; this is exactly the case for linguistics as well. Consider the diagram below, in which all relational words are represented by separable tensors (in other words, no entanglement is present). 

\begin{equation}
\label{equ:degrade}
\footnotesize

\InputIfFileExists{./tikz/sentencesep.tikz}{}{\input{./tikz/sentencesep.tikz}}

\normalsize
\end{equation}

In this version, the  $\epsilon$-maps are  completely detached from the components of the relational tensors that carry the results (left-hand wire of the adjective and middle wire of the verb); as a consequence, flow of information is obstructed, all compositional interactions have been eliminated, and the meaning of the sentence is reduced to the middle component of the verb (shaded vector) multiplied by a scalar, as follows (superscripts denote the left-hand, middle, and right-hand components of separable tensors):

\begin{equation*}
  \langle \ov{happy}^{(r)}|\ov{kids}\rangle \langle \ov{happy}^{(l)}|\ov{play}^{(l)} \rangle 
  \langle \ov{play}^{(r)}|\ov{games}\rangle \ov{play}^{(m)} 
\end{equation*}

\index{tensors!separability|)}
\index{separability|)}
\index{flow of information|)}

Depending on how one measures the distance between two sentences, this is a very unwelcome effect, to say the least. When using cosine distance, the meaning of all sentences with `play' as the verb will be exactly the same and equal to the middle component of the  `play' tensor. For example, the sentence  ``trembling  shadows play hide-and-seek'' will have the same meaning  as our example sentence. Similarly, the comparison of two arbitrary transitive sentences will be reduced to comparing just the middle components of their verb tensors, completely ignoring any surrounding context. The use of Euclidean distance\index{Euclidean distance} instead of cosine would slightly improve things, since now we would be at least able to also detect differences in the magnitude between the two middle components. Unfortunately, this metric has been proved not very appropriate for distributional models of meaning, since in the vastness of a highly dimensional space every point ends up to be almost equidistant from all the others. As a result, most implementations of distributional models prefer the more relaxed metric of cosine distance which is length-invariant. Table \ref{tbl:cons} presents the consequences of separability in a number of grammatical constructs.

\renewcommand{\arraystretch}{2.0}
\begin{table}[h!]
  \begin{center}
  \scriptsize
  \begin{tabular}{l|l|c}
     \hline
     \textbf{Structure} & \textbf{Simplification} & \textbf{Cos-measured} \\
     \hline\hline
      adjective-noun & $\overline{adj} \times \ov{noun} = (\ov{adj}^{(l)} \otimes \ov{adj}^{(r)}) \times \ov{noun} = \langle \ov{adj}^{(r)}|\ov{noun}\rangle \cdot \ov{adj}^{(l)}$ & $\ov{adj}^{(l)}$  \\
      \hline
     intrans. sentence & $\ov{subj} \times \overline{verb} = \ov{subj} \times (\ov{verb}^{(l)} \otimes \ov{verb}^{(r)}) = \langle \ov{subj}|\ov{verb}^{(l)}\rangle \cdot \ov{verb}^{(r)}$ & $\ov{verb}^{(r)}$  \\
     \hline
     verb-object & $\overline{verb} \times \ov{obj} = (\ov{verb}^{(l)} \otimes \ov{verb}^{(r)}) \times \ov{obj} = \langle \ov{verb}^{(r)}|\ov{obj}\rangle \cdot \ov{verb}^{(l)}$ &  $\ov{verb}^{(l)}$ \\
     \hline
     transitive sentence & $\begin{array}{r l} \ov{subj} \times \overline{verb} \times \ov{obj} =  
            \ov{subj} \times (\ov{verb}^{(l)} \otimes \ov{verb}^{(m)} \otimes \ov{verb}^{(r)}) \times \ov{obj} & = \\ 
            \langle \ov{subj}|\ov{verb}^{(l)}\rangle \cdot \langle \ov{verb}^{(r)}|\ov{obj}\rangle \cdot \ov{verb}^{(m)} \end{array}$ &  $\ov{verb}^{(m)}$\\
     \hline
  \end{tabular}
  \normalsize
  \end{center}

  \caption[Consequences of separability in various grammatical structures.]{Consequences of separability in various grammatical structures. Superscripts $(l)$, $(m)$ and $(r)$ refer to left-hand, middle, and right-hand component of a separable tensor.}
  \label{tbl:cons}
\end{table}
  \renewcommand{\arraystretch}{1.0}
  
This aspect of tensor-based models of meaning is quite important and, surprisingly, almost completely neglected by the current research.  In fact, the only relevant work I am aware of comes from personal research (joint with Sadrzadeh) \cite{kartsadrqpl2014}. In the context of current discussion, the interesting point is that no matter the actual level of entanglement in the relational tensors, the Frobenius framework detailed in this chapter provides an additional layer that prevents the compositional process from degrading to the invariant vectors showed in (\ref{equ:degrade}). The reason behind this is that, given a vector $\ov{v}$, the diagonal matrix that results from $\Delta(\ov{v})$ cannot be expressed as a product state:\index{product state}

\begin{equation}
  
\InputIfFileExists{./tikz/entangl1.tikz}{}{\input{./tikz/entangl1.tikz}}

\end{equation}

A tensor like this is essentially constructed from a \textit{single} wire, providing an analogy with a \textit{maximally entangled state} in quantum mechanics. I give an example of this inseparability for a two dimensional space, where we have:


\vspace{-0.5cm}
\small
\begin{equation}
  \Delta\left(\begin{array}{c}a \\ b\end{array}\right) = 
  \left(\begin{array}{cc}a & 0 \\ 0 & b\end{array}\right)~~~~\neq~~~~
  \left(\begin{array}{c}v_1 \\ v_2\end{array}\right) \ten
  \left(\begin{array}{c}w_1 \\ w_2\end{array}\right) =
  \left(\begin{array}{cc}v_1w_1 & v_1w_2 \\ v_2w_1 & v_2w_2\end{array}\right) 
\end{equation}
\normalsize

\noindent since the only case in which the above would be trivially satisfied is when $a=b=0$. Let us now see how this property affects the various models presented in this chapter in the case of a separable verb matrix. When one copies both dimensions of the original verb matrix the result is the following:

\begin{equation}

\InputIfFileExists{./tikz/relsep.tikz}{}{\input{./tikz/relsep.tikz}}

\end{equation}

\noindent which means that, linear-algebraically, the meaning of a transitive sentence becomes:

\vspace{-0.5cm}
\begin{equation}
 \ol{subj~verb~obj} = (\ov{subj} \odot \ov{verb}^{(l)}) \otimes (\ov{verb}^{(r)} \odot \ov{obj})
\end{equation} 

\index{Copy-Subject model!and entanglement|(}
\index{Copy-Object model!and entanglement|(}
\index{Frobenius additive model!and entanglement|(}

Furthermore, the Copy-Subject and Copy-Object models simplify to the following:

\vspace{-0.5cm}
\begin{equation}
\begin{tabular}{ccc}
 
\InputIfFileExists{./tikz/copysbjsep.tikz}{}{\input{./tikz/copysbjsep.tikz}}
 & & 
\InputIfFileExists{./tikz/copyobjsep.tikz}{}{\input{./tikz/copyobjsep.tikz}}

\end{tabular}
\end{equation}

Therefore, the actual equation behind the Frobenius additive model becomes:

\vspace{-0.2cm}
\begin{equation}
\ov{subj~verb~obj} = (\ov{subj} \odot \ov{verb}^{(l)}) + (\ov{verb}^{(r)} \odot \ov{obj})
  \label{equ:frobadd}
\end{equation}

\index{Frobenius additive model!and entanglement|)}
\index{Copy-Object model!and entanglement|)}
\index{Copy-Subject model!and entanglement|)}

Despite the simplifications presented above, note that none of these models degenerates to the level of producing ``constant'' vectors or matrices. The reason behind this lies in the use of Frobenius $\Delta$ operators for copying the original dimensions of the verb matrix, a computation that equips the fragmented system with flow, although not in the originally intended sense. Actually, the results of our aforementioned study in \cite{kartsadrqpl2014} suggest that this might be very important when one constructs verb matrices by using the argument summing procedure of \cite{GrefenSadr1}. In that work we empirically evaluate the level of entanglement in verb matrices created according to Eq. \ref{equ:weightrel} by comparing them with two separable versions. In the first case, the verbs matrices were created by a straightforward variation of the original version that results in a product state:\index{argument summing!as product state}

\vspace{-0.2cm}
\small
\begin{equation}
  \ol{verb} = \left(\sum_i \ov{subj_i}\right) \otimes \left(\sum_i \ov{obj_i}\right)
  \label{equ:sep-model}
\end{equation}
\normalsize

\noindent with $i$ iterating as usual over all instances of the verb in the corpus. Furthermore, each verb matrix was compared with its rank-1 approximation, which is again a product state constructed by using only the highest eigenvalue and the related left and right singular vectors of the original matrix, as produced by applying singular value decomposition. In both cases, our findings suggest a very high \textit{cosine} similarity (up to 0.97) between the original matrix and the separable version, probably due to a high level of linear dependence between our word vectors. Although such a result depends heavily on the form of vector space used for constructing the verb tensors, our discussion above shows clearly that the Frobenius framework offers a safety net that can help us avoid these kinds of pitfalls. I refer the interested reader to \cite{kartsadrqpl2014} for more details about this topic.


\index{tensor-based models!and entanglement|)}
\index{entanglement|)}
\index{Frobenius algebras!and entanglement|)}

\section{From argument summing to linear regression}
\label{sec:frob-regression}

\index{tensors!created by linear regression|(}
\index{tensor-based models!and linear regression|(}
\index{Copy-Subject model!learning with linear regression|(}
\index{Copy-Object model!learning with linear regression|(}

Most of my analysis in this chapter assumes as a starting point the reduced verb representations produced by applying Eq. \ref{equ:weightrel}. I will now explore a different approach based on statistical learning that, as we are going to see, fits quite nicely to the Frobenius setting, and even extends its power. Recall that the Frobenius model essentially acts on two steps: (a) it prepares a composite vector for the theme by applying tensor contraction; and (b) point-wise multiplies the result with the rheme. When the rheme is the subject of a transitive verb, this is translated to the following:

\vspace{-0.3cm}
\begin{equation}
 \label{equ:reg-copysbj}
 \ov{subj}\odot(\ol{verb}\times\ov{obj})
\end{equation}

\noindent with the meaning of the verb-object part to be computed ``categorically'', as the matrix multiplication of the verb with the object vector. 

Let me now attempt for a moment a twist of perspective, and forget any assumptions about using Frobenius operators. How can one describe the compositional model depicted in Eq. \ref{equ:reg-copysbj} in that case? This is obviously a vector mixture model, where the operands are always some rheme and its corresponding theme; furthermore, the vector of the theme has been prepared ``categorically''. However, there is an important difference between this new view and the Frobenius one we were discussing so far: note that now the notion of a well-defined sentence does not exist any more---actually, it has been replaced by the notion of a \textit{well-defined theme}.\index{theme!replacing the notion of sentence} Let me formalize this observation, by replacing type $s$ with a new atomic type, call it $\theta$. The pregroup derivation for the theme part of Eq. \ref{equ:reg-copysbj} becomes:

\begin{equation}
  (\theta\cdot n^l) \cdot n = \theta\cdot (n^l \cdot n)  \leq \theta
  \label{equ:preg-theme}
\end{equation}

The new view reveals two important things. First, it is now evident why the representation of the verb as a tensor of order 2 (a matrix) makes sense in this setting: it is just a function that inputs a noun and outputs a well-defined theme, that is, $\ol{verb}:N \to \Theta$, which can canonically be expressed as a matrix in $N\ten \Theta$. More importantly, computing the meaning of the theme that results in by copying the object dimension, in $(\ov{subj}^{\mathsf{T}} \times \ol{verb}) \odot \ov{obj}$, will naturally require a different version of $\ol{verb}$, since this time the output of our function is a syntactically different construct (subject-verb).

Assuming now we have a means to prepare these two different aspects of our verb tensor, we can think of a transitive verb as being represented by a tuple $(\ol{verb}_s,\ol{verb}_o)$. This form has the advantage that, while it retains the benefits of our previous reduced representation (we still do not need to create a verb tensor of order 3), it is more expressive since the interaction of the verb with each argument is handled by a specialized function. Using this perspective in our Frobenius additive model\index{Frobenius additive model} results in the following equation:

\begin{equation}
  \ov{subj~verb~obj} = \ov{subj}\odot (\ol{verb}_o\times \ov{obj}) + (\ov{subj}^{\mathsf{T}}\times \ol{verb}_s)\odot \ov{obj}
  \label{equ:fadd-regr}
\end{equation}

At this stage, the separation of a transitive sentence into two distinct themes by the Frobenius operators makes it easier for us to apply more sophisticated statistical methods in order to create the corresponding verb tensors. Imagine for example the case of the left-hand theme (subject-verb); what we want is a linear map $\ol{verb}_s$ that, given an input noun, will approximate some ideal distributional behaviour of the subject-verb construct in a large corpus. Assuming that, for the $i$th occurrence of the verb in the corpus, this ``ideal behaviour'' is denoted by a vector $\ov{subj_i~verb}$, with $subj_i$ referring to the corresponding subject noun, the matrix we seek is the following:

\begin{equation}
   \hat{\ol{verb}_s} = \underset{\textbf{W}}{\arg\min} \sum\limits_i \left( \textbf{W} \times \ov{subj_i} - \ov{subj_i~verb} \right)^2
   \label{equ:linregr}
\end{equation}

Learning the matrix $\textbf{W}$ that minimizes the sum in Eq. \ref{equ:linregr} is a \textit{multi-linear regression} problem,\index{linear regression, as composition method}\index{CDMs!and linear regression} a technique that obviously constitutes the most appropriate way for training a model based on linear and multi-linear maps. What remains to be answered is how one should create the vectors that will be used as a ``gold reference'' in the optimization process. In one of the first applications of linear regression for CDMs, Baroni and Zamparelli \cite{Baroni} propose the creation of a distributional vector for every two-word construct using the same techniques as for single words (i.e. by collecting word co-occurrences from the context).\index{distributional hypothesis!on multi-word expressions} The goal of the model, then, would be to create a matrix that when multiplied with a subject vector will approximate the corresponding co-occurrence vector of the subject-verb construct. Of course, exactly the same method can be applied on the right-hand theme of a transitive sentence, learning a verb matrix that produces as output a verb-object construct.

Interestingly, if we discard the point-wise multiplications in Eq. \ref{equ:fadd-regr}, what we get back is very close to a recent trend in constructing tensor-based compositional models where each argument of a multi-linear map is addressed separately by learning a specialized matrix \cite{paperno2014,polajnar2014}. These models follow a methodology very similar to what described in this section, in the sense that the composition function does not aim to produce a sentence, but a specific part of it; the individual parts are somehow combined (usually by summing their vectors) in a later stage to produce a vectorial representation for the whole sentence.

Despite the convenience that comes with such methods, the decoupling of the arguments in the above way hides an important caveat: although the objective function of Eq. \ref{equ:fadd-regr} guarantees that the result will be a reliable (to the extent that the training corpus allows) map which, given an input, will produce an appropriate meaning vector for the left-hand (or right-hand) part of the sentence, there is no guarantee whatsoever that the addition of these two vectors reflects appropriately the meaning of the  sentence in its entirety. The solution to this problem is to restore our initial perspective and start again working with sentences instead of themes. From a statistical point of view, the proper way to achieve that is to \textit{jointly} learn the two matrices of a transitive verb; in the case of the Frobenius additive model, this could be done by training a regression model that directly minimizes the following quantity:

\begin{equation}
   \sum\limits_i \left( (\textbf{W_s} \times \ov{subj_i}) \odot \ov{obj_i} + (\textbf{W_o} \times \ov{obj_i}) \odot \ov{subj_i} - \ov{subj_i~verb~obj_i} \right)^2
   \label{equ:lr-fradd}
\end{equation}

\noindent or, for the simpler version without the point-wise interaction:

\begin{equation}
   \sum\limits_i \left( \textbf{W_s} \times \ov{subj_i} + \textbf{W_o} \times \ov{obj_i} - \ov{subj_i~verb~obj_i} \right)^2
   \label{equ:lr-fradd-sim}
\end{equation}

In the above equations, $\ov{subj~verb~obj}$ refers to a distributional vector created from the contexts of the specific triplet across the corpus. Inevitably, this might lead to data sparsity problems when creating the vectors for the triplets, and certainly is not applicable for training argument matrices in cases of relational words of higher arity (e.g. ditranstive verbs). For these cases, we have to adopt the less generic solution to optimize the objective function directly on the goal set by the underlying task---for example, on some form of classification as in the case of sentiment analysis. 

The application of the linear regression method can be seen in practice in \S\ref{sec:wsdfull}, where I use Eq. \ref{equ:linregr} in order to train matrices for verbs taking one argument.

\index{Copy-Object model!learning with linear regression|)}
\index{Copy-Subject model!learning with linear regression|)}
\index{tensor-based models!and linear regression|)}
\index{tensors!created by linear regression|)}

\section{A revised taxonomy of CDMs}
\label{sec:rev-taxonomy}

\index{CDMs!taxonomy, revised|(}
\index{taxonomy of CDMs|(}
\index{partial tensor-based models|(}
\index{CDMs!partial tensor-based models|(}
\index{tensor-based models!partial|(}

The Frobenius framework presented in this chapter has the unique characteristic that it allows the combination of two quite different compositional models: a vector mixture model based on point-wise multiplication, where both operands have equal contribution to the final composite result; and a tensor-based model in which relational words are tensors of higher order acting on vectors. As we saw in the previous pages, this flexibility can be used to model linguistic phenomena, such as intonation, or to provide solutions that lead to models more robust against physical properties of the verb tensors, as in the case of separability. In Chapter \ref{ch:extend} I move one step further, showing that appropriate use of Frobenius operators can help us to reduce the space complexity imposed by the higher order tensors of functional words such as prepositions and conjunctions in a way that makes linguistic sense. 

In general, the application of Frobenius algebras to language constitutes an interesting novel compositional methodology, resulting in a class of CDMs that deserves its own place in the taxonomy we put together in Fig. \ref{fig:taxonomy}. I will now call this class \textit{partial tensor-based models}\footnote{Thanks to a suggestion by Mehrnoosh Sadrzadeh.}, and I will introduce it in Fig. \ref{fig:rev-taxonomy} below. The practical evaluation of the new class of models is deferred for Chapter \ref{ch:frobexp}, where a series of experiments on various tasks will reveal steady and robust performance against previous implementations and baselines.

\begin{figure}[h!]
  \includegraphics[scale=0.89]{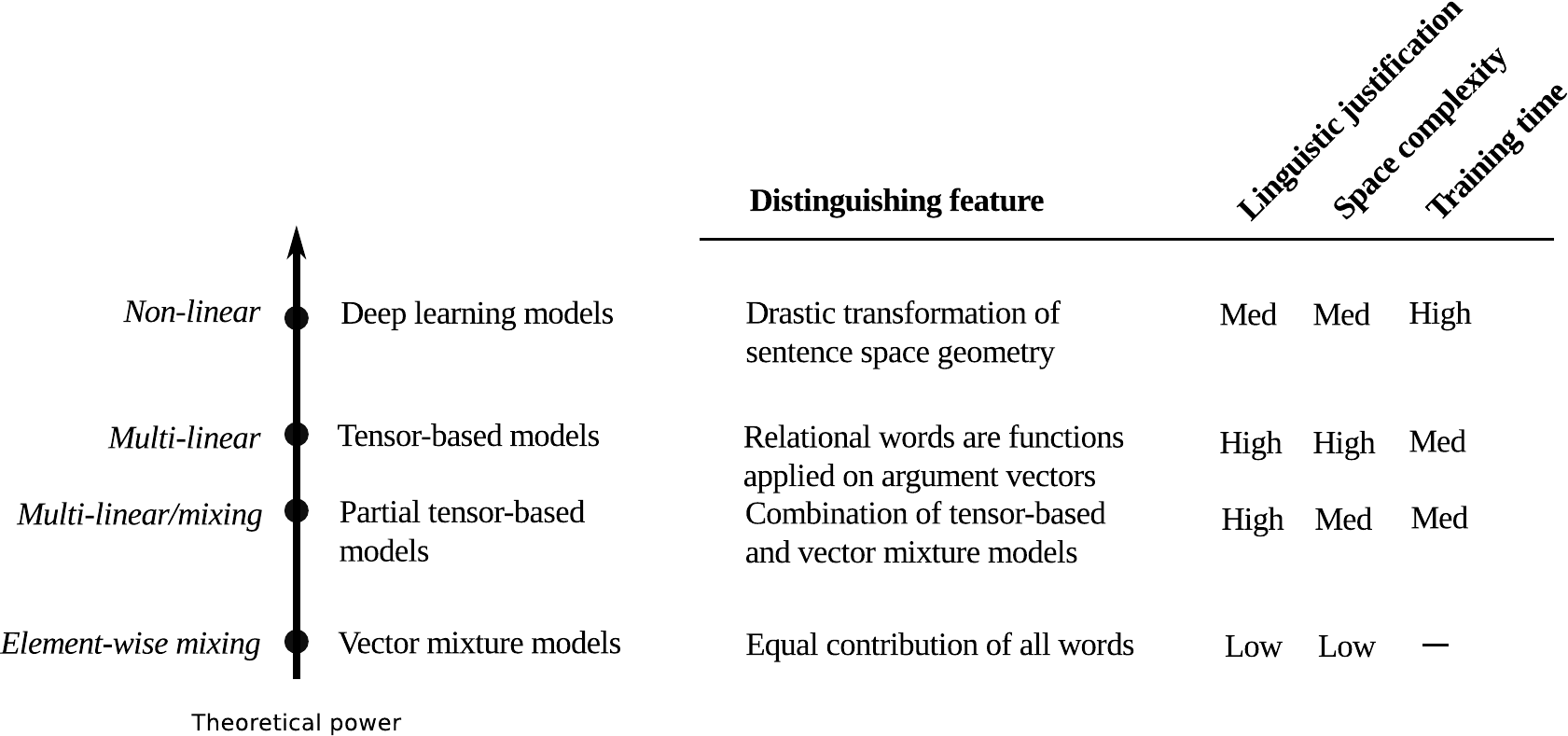}
  \caption{A revised taxonomy of CDMs.}
  \label{fig:rev-taxonomy}
\end{figure}

\index{tensor-based models!partial|)}
\index{CDMs!partial tensor-based models|)}
\index{partial tensor-based models|)}
\index{taxonomy of CDMs|)}
\index{CDMs!taxonomy, revised|)}

\chapter{Covering Larger Fragments of Language}
\label{ch:extend}

\begin{chabstract}
In contrast to other classes of CDMs, a tensor-based model has to assign a form of meaning in every kind of functional word, such as prepositions and relative pronouns. In this chapter I use Frobenius operators in conjunction with unit maps in \textbf{FVect} in order to provide intuitive linear-algebraic interpretations for a number of functional words, extending the applicability of the categorical framework of \cite{Coeckeetal} to larger fragments of language. 
\end{chabstract}

\noindent
Functional words,\index{words!with functional roles} such as prepositions, determiners, or relative pronouns, form a special category of lexical items that pose an important problem for any CDM. In the context of a distributional model, the problem arises from the ubiquitous nature of these words, which means that creating a context vector for preposition `in', for example, would not be especially useful, since this word can be encountered in almost every possible context. Vector mixture models ``address'' this problem by considering functional words as semantically vacuous and ignoring them for any practical purpose. This solution seems questionable since it essentially discards specific relations between the various text constituents. At the other end of the scale, deep learning techniques usually rely on their brute force and make no distinction between functional words and content words, hoping that the learning process will eventually capture at least some of the correct semantics. 

For tensor-based compositional models,\index{tensor-based models!and functional words} the problem is even more complicated. Models like these, being based on deep semantic analysis, do not allow us to arbitrarily drop a number of words from a sentence based on subjective criteria regarding their meaning, and for a good reason: there is simply no linguistic justification behind such decisions. Even in cases where dropping a word might seem justified (e.g. for a determiner), we need to proceed in a canonical way that does not violate the underlying theory. Even more importantly, the grammatical types of some functional words usually translate to tensors of very high order (for example, of order 6 for the case of verb-phrase coordinating conjunctions), a fact that makes the efficient treatment of the relevant constructs impossible. In this chapter I build on and extend previous work of Preller, Sadrzadeh and others \cite{Preller2010,relpronouns1,relpronouns2} by exploiting the compact closed structure of the semantic space and the convenient tool of Frobenius algebras in order to address the above issues. Specifically, I provide canonical ways to create compact representations for various functional words, which result in compositions that reflect the essence hidden behind the complex grammatical interactions. As we will see, a valuable tool for achieving this will be a morphism of a compact closed category that we have not used until now: the $\eta$-map.

\section{Omitting words}
\label{sec:omitting}

\index{words!canonically omitting|(}
\index{omitting words|(}

\tikzstyle{every picture}=[scale=0.41,baseline=0.pt]

In English language there exist certain cases where one can omit a word without altering the meaning of the overall sentence. Imagine, for example, a typical case of a \textit{sentential complement}:\index{sentential complements}

\begin{equation}
  \footnotesize
  
\InputIfFileExists{./tikz/sentcompl1.tikz}{}{\input{./tikz/sentcompl1.tikz}}

  \normalsize
  \label{equ:complement1}
\end{equation}

To comply with the English grammar, we would like the vector of the above sentence to be equal to that of ``Mary knew she loved John'', where the word `that' has been omitted. As first showed by Preller and Sadrzadeh \cite{Preller2010}, compact closed categories provide a means to achieve this through the unit morphism $\eta$ (Eq. \ref{equ:eta}). In the specific case, we can translate the $s\cdot s^l$ type of the complementizer\index{complementizer} to \textbf{FVect} by using a morphism $\eta^l_S:I \to S\ten S^l$ in the following manner:\index{compact closed category!unit morphism}

\begin{equation}
\footnotesize

\InputIfFileExists{./tikz/sentcompl2.tikz}{}{\input{./tikz/sentcompl2.tikz}}

\normalsize
\end{equation}

The interaction depicted above takes an intuitive form: the matrix of `that' allows the information from the right-hand sentence to \textit{flow} through it, acting as an identity function:

\vspace{-0.5cm}
\begin{equation}
\footnotesize

\InputIfFileExists{./tikz/sentcompl3.tikz}{}{\input{./tikz/sentcompl3.tikz}}

\normalsize
\end{equation}

Indeed, in categorical terms the composition of `that' with sentence ``she loved John'' becomes an instance of a yanking condition,\index{yanking conditions} and takes place as below:

\vspace{-0.3cm}
\begin{equation}
  (1_S\ten \epsilon^l_S) \circ (\eta^l_S\ten 1_S)(\ov{she~loved~john}) = 1_S(\ov{she~loved~john})
\end{equation}

We have now a canonical way to ``ignore'' a word\footnote{Note that technically we do not ignore anything; we have just decided to assign a particular meaning to the complementizer, and we construct its linear map in a way to reflect this meaning.}---or, much more interestingly, parts of the internal structure of words that correspond to redundant interactions, as we will see later in this chapter. For the moment, I would like to note that for certain functional words that have little meaning of their own, such as copulas, auxiliaries and determiners, the decision to ignore them is an acceptable compromise that can save us from trying to provide answers to perhaps unanswerable questions (e.g. what is the distributional meaning of `some'?). Applying the above method to the case of a determiner,\index{determiners} for example, we get the following:

\vspace{-0.2cm}
\begin{equation}
\footnotesize

\InputIfFileExists{./tikz/determiner.tikz}{}{\input{./tikz/determiner.tikz}}

\normalsize
\end{equation}
\vspace{0.5cm}
\vspace{-0.5cm}

\noindent
while for the slightly more complicated case of an auxiliary verb,\index{auxiliary verbs} we have:

\vspace{-0.1cm}
\begin{equation}
\footnotesize

\InputIfFileExists{./tikz/auxiliary.tikz}{}{\input{./tikz/auxiliary.tikz}}

\normalsize
\end{equation}

This time we create the tensor by nesting an $\eta$-map within another, getting the following composition (better depicted by a commutative diagram):


\vspace{-0.5cm}
\begin{equation}
\begin{tikzpicture}[scale=1.4,baseline=40pt]
font=\footnotesize
\node (A) at (0,4.5) {$N\ten N^r\ten N\ten N^r\ten S$};
\node (A1)at (0,6) {$N\ten N^r\ten I\ten N\ten N^r\ten S$};
\node (B) at (15,6) {$N\ten N^r\ten S\ten S^l\ten N\ten N^r\ten S$};
\node (C1)at (0,1.5) {$N\ten I\ten N^r\ten S$};
\node (C) at (0,0) {$N\ten N^r\ten S$};
\node (D) at (15,0) {$S\ten S^l \ten S$};
\node (E) at (8.5,0) {$S$};
\node[rotate=90] (l1) at (0,0.7) {$\cong$};
\node[rotate=90] (l2) at (0,5.2) {$\cong$};
\path[->,font=\scriptsize]
(A1) edge node[above]{$1_{N\ten N^r} \ten \eta^l_S \ten 1_{N \ten N^r \ten S}$} (B)
(C1) edge node[right]{$1_N\ten \eta^r_N\ten 1_{N^r}\ten 1_S$} (A)
(B) edge node[left]{$\epsilon_N^r \ten 1_S\ten 1_{S^l}\ten \epsilon^r_N\ten 1_S$} (D)
(C) edge node[above]{$\epsilon^r_N\ten 1_S$} (E)
(D) edge node[above]{$1_S\ten \epsilon^l_S$} (E);
\end{tikzpicture}
\end{equation}
\vspace{0.2cm}
\vspace{-0.5cm}

Despite the usefulness of the above method in the aforementioned cases, it should be evident that it would not be adequate for words that occur in more complex grammatical interactions within a sentence. In what follows I show how to extend these ideas to various other functional words of the English language in a linguistically justified way, often resorting to using $\eta$-maps in combination with Frobenius operators. 

\index{omitting words|)}
\index{words!canonically omitting|)}

\section{Prepositions}
\label{sec:preposition}

\index{prepositions|(}

Not all functional words can be ignored in the sense described in \S \ref{sec:omitting}. Prepositions, for example, signify specific relations between text constructs that need to be semantically reflected: our model must be capable of expressing that the meaning of ``John put the bag \textit{on} the table'' is different from that of ``John put the bag \textit{under} the table''. At the same time, we need to find a way to deal with the fact that a verb-modifying (for example) preposition, is a tensor of order 5:

\vspace{-0.5cm}
\begin{equation}
  \label{fig:prep}
  \begin{minipage}{0.45\linewidth}
  \centering
  \footnotesize
  
\InputIfFileExists{./tikz/prep1.tikz}{}{\input{./tikz/prep1.tikz}}

  \normalsize
  \end{minipage}
  \begin{minipage}{0.45\linewidth}
  \centering
  \footnotesize
  
\InputIfFileExists{./tikz/prep2.tikz}{}{\input{./tikz/prep2.tikz}}

  \normalsize
  \end{minipage}  
\end{equation}

From a formal semantics perspective, a transitive preposition is seen as a restriction applied to the set of individuals performing a specific action. Consider, for example, the logical form for the above sentence:

\vspace{-0.4cm}
\begin{equation}
  \exists x [park(x) \wedge in(x)(walks)(john)]
\end{equation}

Taking the meaning of predicate \textit{walks} to be the set of individuals who generally walk, \textit{in} is a function that takes as input a place $x$ and applied to this set of individuals who walk, eventually returning a subset with all individuals who walk in $x$ \cite[p.~244]{dowty}. Thus, a transitive preposition can be seen as a function with three arguments: a subject (\textit{John}), an action (\textit{walks}), and an object (\textit{park}). This is directly reflected in the right-hand diagram of (\ref{fig:prep}), where all this information is fed to the tensor of the preposition for processing.

Again, the compact closed structure provides us some options. I will begin by noticing that the type $s^r\cdot s\cdot n^l$ is a more generic case
of the type $s^r\cdot n^{rr}\cdot n^r\cdot s\cdot n^l$.\footnote{Note that the atomic type $n^{rr}$ expresses an \textit{iterated adjoint}: the right adjoint of a right adjoint.} Indeed,

\begin{equation}
  \label{equ:prep-sim1}
  s^r\cdot s\cdot n^l = s^r\cdot 1\cdot s\cdot n^l \leq s^r\cdot n^{rr}\cdot n^r\cdot s\cdot n^l
\end{equation}

This type form denotes a function of two arguments which, in the case of verb-modifying prepositions, expects a sentence to the left (an intransitive verb \textit{already combined} with a subject noun phrase), and an object noun phrase at the right in order to return a sentence. The translation of Eq. \ref{equ:prep-sim1} to vector spaces is: 

\begin{equation}
   S^r \otimes S \otimes N^l \cong S^r \otimes I \otimes S \otimes N^l \xrightarrow{1_{S^r} \otimes \eta^r_{N^r} \otimes 1_S \otimes 1_{N^l}} S^r \otimes N^{rr} \otimes N^r \otimes S \otimes N^l
\end{equation}

\noindent
Using this tensor for the composition will produce the following normal form:

\begin{equation}
\footnotesize

\InputIfFileExists{./tikz/prep3.tikz}{}{\input{./tikz/prep3.tikz}}

\normalsize
\end{equation}

The representation of a verb-modifying preposition as a function of two arguments is a very convenient outcome, since we already know how to handle such a function from our work with transitive verbs in Chapter \ref{ch:frobverbs}. Indeed, now we can proceed one step further and adopt the argument summing process of \cite{GrefenSadr1}, by defining the meaning of a preposition as follows:\index{argument summing}

\begin{equation}
  \ol{pr}_{TR} = \sum\limits_i \ov{verb_i} \ten \ov{noun_i}
\end{equation}

\noindent where $\ov{verb_i}$ corresponds to the verb modified by the preposition, and $\ov{noun_i}$ the object noun from the right. The usual Frobenius manipulations apply here; copying the object dimension of $\ol{in}$, for example, will compute the meaning of the sentence as $(\ol{in}^{\mathsf{T}} \times (\ol{walks}^{\mathsf{T}}\times \ov{john})) \odot \ov{park}$, which brings the same intonational interpretation as discussed in \S \ref{sec:intonation} by answering the question ``where does John walk in?''.\index{intonation!and prepositions}

\index{PP-modification|(}

The above treatment scales naturally to cases where a verb is modified by more than one prepositional phrase, satisfying in this way one of the basic requirements of a PP-modification analysis. Extending further our previous example:

\begin{equation}
  \footnotesize
  
\InputIfFileExists{./tikz/prep-rec.tikz}{}{\input{./tikz/prep-rec.tikz}}

  \normalsize
\end{equation}

\noindent which produces the following normal form:

\begin{equation}
  \footnotesize
  
\InputIfFileExists{./tikz/prep-rec1.tikz}{}{\input{./tikz/prep-rec1.tikz}}

  \normalsize
\end{equation}

\index{PP-modification|)}

Before closing this section, I would like to note that in general noun-modifying prepositions, as in ``a man in uniform'', do not need any special treatment since they are already in a manageable 2-argument form, as it is evident by their type ($n^r\cdot n\cdot n^l$). Obviously, the matrix of a noun-modifying preposition must be different of that corresponding to the verb-modifying variant; with $noun^{(s)}$ and $noun^{(o)}$ denoting the nouns at the ``subject'' and ``object'' position, we get:

\begin{equation}
  \ol{pr}_{N} = \sum\limits_i \ov{noun}^{(s)}_i \ten \ov{noun}^{(o)}_i
\end{equation}

\index{prepositions|)}

\section{Phrasal verbs}
\label{sec:phrasal-verbs}

\index{phrasal verbs|(}

Interestingly, the type $s^r\cdot s\cdot n^l$ we derived in \S\ref{sec:preposition} for a verb-modifying preposition can be further reduced if we create the preposition tensor by using a second $\eta$-map:

\begin{eqnarray}
     N^l \cong I \otimes N^l \xrightarrow{\eta^r_S \otimes 1_{N^l}} S^r \otimes S \otimes N^l \cong~~~~~~~~~~~~~~~~~~~~~~~~~~~~~~~~~~~~~~~~~~~~\\ \nonumber
     ~~~~~~~~~~~~~~~~~~~~S^r \otimes I \otimes S \otimes N^l \xrightarrow{1_{S^r} \otimes \eta^r_{N^r} \otimes 1_S \otimes 1_{N^l}} S^r \otimes N^{rr} \otimes N^r \otimes S \otimes N^l
\end{eqnarray}

Doing this might seem not very helpful in the beginning, since as it is evident in (\ref{fig:prep-join}) below what we have ``achieved'' is to cancel out the interaction of the object (``the park'') with the rest of the sentence. However, the interesting point of this treatment is that it provides a justification for viewing a preposition from a different perspective: as \textit{part} of the preceding verb. Indeed, it looks like the sole purpose of the reduced preposition tensor is to provide another argument to the adjacent verb, creating a new \textit{transitive} version of it:

\vspace{-0.5cm}
\begin{equation}
\label{fig:prep-join}
\footnotesize

\InputIfFileExists{./tikz/prep4.tikz}{}{\input{./tikz/prep4.tikz}}

\normalsize
\end{equation}

\vspace{0.3cm}
This second model resembles a pre-Davidsonian way of working with prepositions, when the standard practice was to treat them as integral parts of the verbs. In his seminal paper, ``The Logical Form of Action Sentences'' \cite{Davidson:1967}, Davidson\index{Davidson, Donald} introduces the representation of prepositions as distinct predicates, explaining his reasoning as follows:

\begin{quote}
In general, we conceal logical structure when we treat prepositions as integral parts of verbs; it is a merit of the present proposal that it suggests a way of treating prepositions as contributing structure. 
\end{quote}

However, the model of (\ref{fig:prep-join}) still carries an important advantage, since it provides a canonical way to treat special cases where actually the contribution of prepositions to structure is not very strong, such as phrasal verbs (`look after', `run into', `stand by', `hand in', etc). Creating a verb matrix for a phrasal verb is a procedure slightly more complicated than in the case of a normal verb. Specifically, given a text corpus that has been \textit{dependency-parsed} into a set of relations $(w_D,w_H,R)$, where $w_H$ is the head word, $w_D$ the dependent, and $R$ the grammatical relation that holds between the two words, we can construct the reduced tensor of a phrasal verb $verb~prep$ by looking for chains of the following form:

\begin{equation}
  (\textit{noun}_1,\textit{verb},\text{SUBJ}) \to (\textit{verb},\textit{prep},\text{MOD}) \to (\textit{noun}_2,\textit{verb},\text{OBJ})
\end{equation}

Here, MOD denotes a modifier. Then, \textit{noun}$_1$ and \textit{noun}$_2$ are the subject and the object, respectively, of the ``phrasal'' version of the verb, produced by attaching the preposition to it; the application of Eq. \ref{equ:weightrel}, by tensoring the vectors of the two nouns and adding them to a running total, will create a reduced tensor as required.

\index{phrasal verbs|)}

\section{Coordination}
\label{sec:coordination}

\index{coordination|(}
\index{Frobenius algebras!and coordination|(}

Two coordinated phrases or sentences can be seen as contributing equally to the final outcome; we would expect, for example, the vector of the sentence ``John reads and Mary sleeps'' to reflect equally the vectors of the two coordinated sentences, ``John reads'' and ``Mary sleeps''. This is exactly one of the cases in which the necessity of a compositional model with the characteristics of our Frobenius framework (categorical composition in combination with point-wise multiplication) is very clear; the tensors of the coordinators must be created in a way to allow categorical composition for the conjuncts, and then merge the two composite vectors using point-wise multiplication. Motivated by the work of Sadrzadeh and colleagues on relative pronouns\footnote{Which is quite relevant to the models of Chapter \ref{ch:frobverbs} and will be presented in more detail in \S\ref{sec:relpronouns}.} \cite{relpronouns1,relpronouns2}, in this section I will show how this can be achieved by a combination of Frobenius operators and $\eta$-maps. 

For the analysis that follows I consider the usual ternary rule $X$ CONJ $X \to X$,\index{ternary rule for coordination} which states that coordination always take place between conjuncts of the same type and produces a result that again matches that specific type. In pregroups terms, this is achieved by assigning the type $x^r\cdot x\cdot x^l$ to the conjunction (where $x$ can be an atomic or a complex type), which leads to the following generic derivation:

\vspace{-0.5cm}
\begin{equation}
 x\cdot (x^r\cdot x\cdot x^l)\cdot x \leq 1\cdot x\cdot 1 = x
 \label{equ:ternary}
\end{equation}

\vspace{-0.5cm}
\subsection{Coordinating noun phrases}
\label{sec:coord-np}

\index{NP-coordination|(}

I will start with the simple case of a noun-phrase coordinating conjunction, the type of which is $n^r\cdot n\cdot n^l$.
In order to achieve the desired result, I will create the corresponding construction in \textbf{FVect} using the following morphism:

\begin{equation}
  I \xrightarrow{\eta^r_N \ten \eta^l_N} N^r \ten N \ten N \ten N^l \xrightarrow{1_{N^r}\ten \mu_N \ten 1_{N^l}} N^r \ten N \ten N^l
\end{equation} 

The composition now of a coordinated noun phrase takes this form:

\begin{equation}
\footnotesize

\InputIfFileExists{./tikz/coordination2.tikz}{}{\input{./tikz/coordination2.tikz}}

\normalsize
\end{equation}

As shown above, the closed-form formula of this construction is the morphism:

\begin{equation}
  \mu(\ov{apples} \ten \ov{oranges}) = \ov{apples} \odot \ov{oranges}
\end{equation}

\index{NP-coordination|)}

\subsection{Coordinating verb phrases}
\label{sec:coord-vp}

\index{VP-coordination|(}

The case of verb phrase coordination is more interesting and involved. Recall that the pregroup type of a verb phrase is $n^r\cdot s$; that is, something that expects a noun (a subject) from the left in order to return a sentence. Using the ternary rule in Eq. \ref{equ:ternary} and the properties of  pregroups,\footnote{It can be shown that $(x\cdot y)^l=y^l\cdot x^l$ and $(x\cdot y)^r=y^r\cdot x^r$; furthermore, it also holds that $x^{lr}=x=x^{rl}$.} we derive the following type for a coordinator between two verb phrases:

\vspace{-0.5cm}
\begin{equation}
  (n^r\cdot s)^r\cdot (n^r\cdot s)\cdot (n^r\cdot s)^l \leq
  s^r\cdot n^{rr} \cdot n^r \cdot s \cdot s^l \cdot n^{rl} \leq
  s^r\cdot n^{rr} \cdot n^r \cdot s \cdot s^l \cdot n
\end{equation}
\vspace{-0.5cm}

This requires a tensor of order 6, as in the following example:
 
\begin{equation}
\footnotesize

\InputIfFileExists{./tikz/coordination.tikz}{}{\input{./tikz/coordination.tikz}}

\normalsize
\end{equation}

One intuitive way to model this so that it imposes ``equal contribution'' of the two verbs in the final outcome is to define the inner structure of the coordinator as follows:

\vspace{-0.5cm}
\begin{equation}
\label{equ:vp-coord}
\footnotesize

\InputIfFileExists{./tikz/sent-coord.tikz}{}{\input{./tikz/sent-coord.tikz}}

\normalsize
\end{equation}

Note that in this case we use for the first time the Frobenius $\iota$-map,\index{deleting of basis} the purpose of which is to \textit{delete} a wire. The motivation behind this becomes apparent in the normal form that results after the composition of the coordinator with the rest of the context:

\begin{equation}
\label{equ:vp-coord1}
\footnotesize

\InputIfFileExists{./tikz/coordination1.tikz}{}{\input{./tikz/coordination1.tikz}}

\normalsize
\end{equation}

Deleting the noun dimension of the second verb helps us to create a coherent compact representation for the coordinated structure ``sleeps and snores'', that can indeed be seen as a new verb phrase. Note that the sentence dimension of this structure is produced by ``merging'' the sentence dimensions of the two conjuncts:

\begin{equation}
\footnotesize

\InputIfFileExists{./tikz/coord-verb.tikz}{}{\input{./tikz/coord-verb.tikz}}

\normalsize
\end{equation}

Categorically, the original tensor of the coordinator is created by the following morphism:

\vspace{-0.5cm}
\begin{eqnarray}
   I \xrightarrow{\eta^r_S \ten \eta^l_S} S^r\ten S\ten S\ten S^l
   \cong~~~~~~~~~~~~~~~~~~~~~~~~~~~~~~~~~~~~~~~~~~~~~~~~~~~~~ \nonumber \\
   ~~~~~~~~~~~~~~S^r\ten I \ten S \ten S \ten S^l
   \xrightarrow{1_{S^r}\ten \eta^r_{N^r}\ten \mu_S\ten 1^l_S} S^r\ten N^{rr} \ten N^r \ten S \ten S^l
\end{eqnarray}
\vspace{-0.5cm}

I will use the normal form of (\ref{equ:vp-coord1}) to compute the meaning of the coordinated structure as below:

\vspace{-0.5cm}
\begin{equation}
  (\epsilon^r_N \ten \mu_S)(\ov{john}\ten \ol{sleeps} \ten \ol{snores}) = (\ov{john}^{\mathsf{T}}\times \ol{sleeps}) \odot \ov{snores}
\end{equation}
\vspace{-0.5cm}

\noindent
where $\ov{snores}$ is the vector produced by deleting the row dimension of the original verb matrix for `snores'. Linear-algebraically (and when one uses full tensors, as opposed to the reduced Frobenius tensors presented in Chapter \ref{ch:frobverbs}), the application of $\iota$-map (Eq. \ref{equ:frob}) to a matrix has the following result:\index{deleting of basis}

\begin{equation}
  (\iota\ten 1_N) \left( \sum\limits_{ij} v_{ij} \ov{n_i}\ten\ov{n_j}\right) = \sum\limits_{ij} v_{ij} \ov{n_j}
\end{equation}

This is exactly the vector created by summing the rows of the verb matrix. It would be also instructive to see how this method fits into the Frobenius framework of Chapter \ref{ch:frobverbs}. Imagine the following more complicated case of verb phrase coordination, where the coordinated verbs occur with an object noun:

\begin{equation}
  \footnotesize
  
\InputIfFileExists{./tikz/coord-vp1-new.tikz}{}{\input{./tikz/coord-vp1-new.tikz}}

  \normalsize
\end{equation}

\index{intonation!and coordination|(}
\index{Copy-Subject model!and coordination|(}

Let me first create the tensors for verbs `like' and `play' as reduced matrices (\S\ref{sec:tensor-rel}). I will expand the tensor of `like' by copying its subject dimension; however, I will not apply any dimension copying on the `play' tensor, since one of its dimensions needs to be deleted anyway. The above derivation gets the following form:

\begin{equation}
  \footnotesize
  
\InputIfFileExists{./tikz/coord-vp2-new.tikz}{}{\input{./tikz/coord-vp2-new.tikz}}

  \normalsize
\end{equation}

\noindent which has the following normal form, by the application of the spider theorem:

\begin{equation}
  \footnotesize
  
\InputIfFileExists{./tikz/coord-vp3.tikz}{}{\input{./tikz/coord-vp3.tikz}}

  \normalsize
\end{equation}

In linear-algebraic terms, the result is once again highly intuitive: the vectors of the two verb phrases are merged together (as required by the conjunction), and the result is further merged with the subject of the sentence---which has to contribute equally to the final outcome, since according to the Copy-Subject model we applied it has the role of the rheme:

\begin{equation}
  \ov{men} \odot (\ol{like}\times \ov{sports}) \odot (\ol{play}\times \ov{football})
   \label{equ:frob-coord}
\end{equation}

\index{Copy-Subject model!and coordination|)}
\index{intonation!and coordination|)}
\index{VP-coordination|)}

\subsection{Coordinating sentences}
\label{sec:coord-s}

\index{sentence coordination|(}

The type of a sentence coordinator is $s^r\cdot s \cdot s^l$, leading to a situation very similar to that of the noun phrase case. We create the tensor as follows:

\begin{equation}
  I \xrightarrow{\eta^r_S \ten \eta^l_S} S^r \ten S \ten S \ten S^l \xrightarrow{1_{S^r}\ten \mu_S \ten 1_{S^l}} S^r \ten S \ten S^l
\end{equation} 

\noindent and get the following generic derivation:

\begin{equation}
  \footnotesize
  
\InputIfFileExists{./tikz/coord-sent-new.tikz}{}{\input{./tikz/coord-sent-new.tikz}}

  \normalsize
  \label{equ:generic-sent}
\end{equation}

Linear-algebraically:

\begin{equation}
   (\ov{men}^{\mathsf{T}}\times \ol{watch}\times\ov{football}) \odot (\ov{women}^{\mathsf{T}}\times \ol{knit})
\end{equation}

Note that in general the type of a sentence coordinator can occur in expanded form, as below:

\begin{equation}
  \footnotesize
  
\InputIfFileExists{./tikz/coord-sent-exp.tikz}{}{\input{./tikz/coord-sent-exp.tikz}}

  \normalsize
  \label{equ:exp-sent}
\end{equation}

As in the case of verb-modifying prepositions (\S\ref{sec:preposition}), it can be shown that all these cases reduce to the canonical form $s^r\cdot s \cdot s^l$. For the derivation in (\ref{equ:exp-sent}), we get:

\begin{equation}
  \footnotesize
  
\InputIfFileExists{./tikz/coord-sent-exp1.tikz}{}{\input{./tikz/coord-sent-exp1.tikz}}

  \normalsize
  \label{equ:exp-sent1}
\end{equation}

Applying the previous Frobenius method to the coordinator tensor will give:

\begin{equation}
  \footnotesize
  
\InputIfFileExists{./tikz/coord-sent-exp2.tikz}{}{\input{./tikz/coord-sent-exp2.tikz}}

  \normalsize
  \label{equ:exp-sent2}
\end{equation}

\noindent the normal form of which is exactly the same as the generic case depicted in (\ref{equ:generic-sent}).

\index{sentence coordination|)}

\subsection{Distributivity}

\index{distributivity in coordination|(}

The coordination treatment presented above fits nicely with the theoretical basis detailed in this thesis, but it has a drawback: the element-wise multiplications introduced by the Frobenius operators do not interact well with tensor contraction with regard to distributivity, which means that the proposed system cannot adhere to basic entailments of the following form:

\begin{exe}
  \ex\label{ex:np-dist} Mary studies philosophy and history $\models$ Mary studies philosophy and Mary studies history 
  \ex\label{ex:vp-dist} Men like sports and play football $\models$ Men like sports and men play football
\end{exe}

\noindent where symbol $\models$ denotes semantic entailment. In (\ref{ex:np-dist}) the subject-verb part distributes over noun phrases, while in (\ref{ex:vp-dist}) the subject distributes over verb phrases. In both cases the entailed part is a coordination of sentences. The obvious way to allow this kind of distributivity in a tensor-based setting is to replace the element-wise multiplication with vector addition, since for a tensor \textbf{W} and two vectors $\ov{v},\ov{w}$ it is always the case that $\textbf{W}(\ov{v}+\ov{w})=\textbf{W}\ov{v}+\textbf{W}\ov{w}$. For our examples we get the out-of-the-box derivations shown below:\index{additive model!for coordination distributivity}

\begin{gather*}
  \small
  \ov{mary}\times \ol{studies} \times (\ov{phil}+\ov{hist}) = 
  \ov{mary}\times \ol{studies} \times \ov{phil} +
  \ov{mary}\times \ol{studies} \times \ov{hist} \\
  \ov{men}\times (\ol{like}\times \ov{sp}+\ol{play} \times \ov{fb}) =
  \ov{men}\times\ol{like}\times\ov{sp} + \ov{men}\times\ol{play}\times\ov{fb}
  \normalsize
\end{gather*}

The result is a setting similar to the original, in which the multiplicative vector mixture model we initially used to represent coordination has been replaced by the additive variation. So why not prefer this method in the first place over the Frobenius alternative? There are a couple of reasons. 
First, from a linguistic perspective vector addition seems an operation much more appropriate for modelling \textit{disjunctive} cases of coordination (a topic we did not address in this thesis), as explained by Widdows \cite{widdows2003orthogonal} in the context of quantum logic.\index{intonation!and coordination|(}\index{quantum logic} There is a second very important reason: In the case of an intonational interpretation, as described in \S\ref{sec:intonation}, the forms of basic entailment in (\ref{ex:np-dist}) and (\ref{ex:vp-dist}) \textit{do not} actually hold. Take for example the following Copy-Subject variation of the left-hand case in (\ref{ex:vp-dist}), as computed in Eq. \ref{equ:frob-coord}:

\begin{exe}
  \ex\label{ex:question-conj} \textit{Who likes sports and plays football?} \\
      \textbf{\underline{Men}} like sports and play football: \\
      $\ov{men}\odot \ol{like}\times \ov{sports} \odot \ol{play}\times \ov{football}$
\end{exe}

Note, however, that the Copy-Subject version of the right-hand part in (\ref{ex:vp-dist}) answers to a slightly different question---or, more accurately, to a conjunction of two \textit{distinct} questions:

\begin{exe}
  \ex \textit{Who likes sports and who plays football?} \\
      \textbf{\underline{Men}} like sports and \textbf{\underline{men}} play football: \\
      $(\ov{men}\odot \ol{like}\times \ov{sports}) \odot (\ov{men}\odot \ol{play}\times \ov{football}) = $ \\ $\ov{men}^2 \odot \ol{like}\times \ov{sports} \odot \ol{play}\times \ov{football}$
\end{exe}

The two resulting vectors are quite similar, with the important and necessary difference that in the second case the contribution of word `men' is intensified, exactly as required by the question at hand.

\index{intonation!and coordination|)}

%
%
%
%

%


\index{distributivity in coordination|)}
\index{Frobenius algebras!and coordination|)}
\index{coordination|)}

\section{Adverbs}

\index{adverbs|(}

An adverb is something that usually modifies the meaning of a verb or an adjective. For temporal or spatial adverbs that occur \textit{after} the modified verb, I propose the following representation:

\begin{equation}
\footnotesize

\InputIfFileExists{./tikz/adverb-temp.tikz}{}{\input{./tikz/adverb-temp.tikz}}

\normalsize
\end{equation}

Intuitively, a verb modified by an adverb gets this new form:

\begin{equation}
\footnotesize

\InputIfFileExists{./tikz/adverb-temp1.tikz}{}{\input{./tikz/adverb-temp1.tikz}}

\normalsize
\end{equation}

It is even possible for one to apply the argument summing procedure and create the meaning of the adverb as the sum of all verbs that the specific adverb modifies in the training corpus. The case of an adverb that occurs before the verb, as below:

\begin{equation}
\footnotesize

\InputIfFileExists{./tikz/adverb.tikz}{}{\input{./tikz/adverb.tikz}}

\normalsize
\end{equation}

\noindent is more complicated, since the adverbial linear map must be applied on the sentence dimension of the verb. In order to achieve this we need to follow \cite{Preller2010} and define the inner $S\ten S^l$ part of the adverb tensor as an \textit{operator}\index{operators!for modelling adverbs}---that is, a matrix that corresponds to a linear map $f:S\to S$. In our graphical calculus, this is depicted as follows:\footnote{The concept of an operator, as well as the specific graphical notation I used to denote it, will be explained in more detail in \S\ref{sec:operators}.}

\begin{equation}
\footnotesize

\InputIfFileExists{./tikz/adverb1.tikz}{}{\input{./tikz/adverb1.tikz}}

\normalsize
\end{equation}


Again in this case, our linear map can be created as the sum of all verbs that the specific adverb modifies in the training corpus. Note that the use of an operator\index{operators!for negation} is also the canonical way to define negation\index{negation} in a simple declarative sentence, assuming one has a way to create a map that ``negates'' the meaning of a sentence:

\begin{equation}
\footnotesize

\InputIfFileExists{./tikz/negation.tikz}{}{\input{./tikz/negation.tikz}}

\normalsize
\end{equation}

In a truth-theoretic setting, for example, similar to that of \S\ref{sec:intuition}, $\ol{not}$ can simply be $\left(\begin{smallmatrix}0 & 1\\1 & 0 \end{smallmatrix}\right)$ as explained in \cite{Preller2010}. Unfortunately, it is not quite clear how one can successfully generalize this map to high dimensional real-valued vector spaces; the work of Widdows \cite{widdows2003orthogonal} provides valuable insights on this in the context of quantum logic.\index{quantum logic} 

\index{adverbs|)}

\section{Infinitive phrases}
\label{sec:infinitive}

\index{infinitive phrases|(}

An infinitive phrase in the English language is a verb phrase in which the verb occurs in its infinitive form, that is, following the particle \textit{to}.\index{infinitive particle} Perhaps the most typical use of an infinitive verb phrase is to \textit{complement} some other verb, as below:

\begin{equation}
\footnotesize

\InputIfFileExists{./tikz/infinitive1.tikz}{}{\input{./tikz/infinitive1.tikz}}

\normalsize
\end{equation}

For these cases, I propose the complete ``bridging'' of the infinitive particle in conjunction with the deletion of the noun dimension of the verb, as below:\index{deleting of basis}

\begin{equation}
\footnotesize

\InputIfFileExists{./tikz/infinitive2.tikz}{}{\input{./tikz/infinitive2.tikz}}

\normalsize
\end{equation}

Note that this approach assigns the type $n^r\cdot s \cdot s^l$ to the complemented verb `want', which is now a function of two arguments: a subject noun from the left, and the bare infinitive form of the verb at the right. Since in the new formulation there is no direct connection between the noun (`John') and the bare infinitive (`sleep'), the interaction of those two words is assumed to be handled by the head word of the sentence (`wants'). This suggests that an appropriate method of constructing the tensors of the complemented verb according to the argument summing procedure is the following:\index{argument summing}

\begin{equation}
  \ol{verb}_{C} = \sum_i \ov{subj}_i \ten \ov{verb}^{BI}_i
  \label{equ:inf-construct}
\end{equation}

\noindent where $\ov{verb}^{BI}_i$ denotes the distributional vector of a verb whose infinitive form complements $verb_C$ in some sentence.  

\index{infinitive phrases|)}

\section{Relative pronouns}
\label{sec:relpronouns}

\index{relative pronouns|(}
\index{Frobenius algebras!and relative pronouns|(}

Recently, the treatment of relative pronouns in a tensor-based setting received a lot of attention from Sadrzadeh, Clark and Coecke, who provide intuitive insights regarding the ``anatomy'' (i.e. internal structure) of tensors in a series of papers \cite{relpronouns1,relpronouns2}. Their method is based on the application of Frobenius algebras, providing connections and common points with the way we model transitive verbs in Chapter \ref{ch:frobverbs}. The work in \cite{relpronouns1,relpronouns2} is very relevant to the material presented in this thesis in general and an inspiration for some of the methods I used in this chapter in particular, and in this section I'm going to review its most important points and make the connection with the Frobenius framework of Chapter \ref{ch:frobverbs} explicit. 

\index{subject relative pronouns|(}
\index{object relative pronouns|(}

The purpose of a relative pronoun is to mark a relative clause, i.e. a text constituent that modifies the noun at the left (the \textit{antecedent}). In its most common form, the antecedent can be the subject or the object of the relative clause's head verb, a fact that leads to slightly different grammatical types and derivations for the two cases:

\begin{equation}
\begin{minipage}{0.45\linewidth}
\centering
\footnotesize

\InputIfFileExists{./tikz/relpron1.tikz}{}{\input{./tikz/relpron1.tikz}}

~\\
~\\
Subject case
\normalsize
\end{minipage}
\begin{minipage}{0.45\linewidth}
\centering
\footnotesize

\InputIfFileExists{./tikz/relpron2.tikz}{}{\input{./tikz/relpron2.tikz}}

Object case
\normalsize
\end{minipage}
\end{equation}

\index{object relative pronouns|)}
\index{subject relative pronouns|)}

In \cite{relpronouns1}, the copying abilities of Frobenius algebras are cleverly exploited to allow the passing of information of the antecedent from the left-hand part of the phrase to the right-hand part, in order to let it properly interact with the modifier part:

\begin{equation}
\label{fig:relpron-obj}
\footnotesize

\InputIfFileExists{./tikz/relpron3.tikz}{}{\input{./tikz/relpron3.tikz}}

\normalsize
\end{equation}

\index{Copy-Subject model!and relative pronouns|(}

Given a reduced verb tensor created according to Eq. \ref{equ:weightrel}, the normal form of (\ref{fig:relpron-obj}) above is identical to the Frobenius vector that results after copying the subject dimension of a transitive verb (Diagram \ref{fig:copysbj}). Recall from the discussion in \S \ref{sec:intonation} that this is exactly the vector that puts intonational emphasis on the subject, since this part of the sentence/phrase contributes equally with the verb phrase to the final outcome through point-wise multiplication. The interpretation in the context of a subject relative pronoun follows the same intuition: the subject still constitutes the most important part of the phrase, \textit{modified} by the part that follows the relative pronoun. The relative pronoun itself does not appear in the final calculation. 

\index{Copy-Subject model!and relative pronouns|)}

\index{Copy-Object model!and relative pronouns|(}
Following the same idea, we can model an object relative pronoun as below:

\begin{equation}
\footnotesize

\InputIfFileExists{./tikz/relpron31.tikz}{}{\input{./tikz/relpron31.tikz}}

\normalsize
\end{equation}

\noindent
which this time puts the emphasis on the object.\index{Copy-Object model!and relative pronouns|)} Finally, for the case of possessive relative pronouns (e.g. ``the man \textit{whose} friend likes Mary''), which will not be discussed here, the reader may refer to \cite{relpronouns2}.

When one tries to put together the Frobenius framework of Chapter \ref{ch:frobverbs} with the relative pronouns treatment presented in this section, one has to face the apparent inconsistency that the meaning of a noun phrase, say ``men who like football'', will be identified with the meaning of the sentence ``\textbf{\underline{men}} like football'' (used as an answer to the question ``Who likes football?''). It is interesting, though, to note that despite the obvious simplification that such an identification imposes, the underlying reasoning remains quite valid. First and foremost, we observe that the subject noun is the most important part in both constructs. What changes is the perspective from which we see the composition with the context: in the relative pronoun case, we consider the context as a modifier of the noun, whereas in the intonational case, we emphasize the equal contribution of the two parts to the final composite vector. Setting aside the unfortunate effect that they produce the same concrete vectorial representation, these solutions are indeed intuitive and linguistically justified.

\index{Frobenius algebras!and relative pronouns|)}
\index{relative pronouns|)}

\section{A step towards complete coverage of language}

Back in \S\ref{sec:tensorbased}, we identified as one of the most important challenges related to tensor-based models the fact that the designer has to devise maps for every word type of the language. In this chapter and in Chapter \ref{ch:frobverbs} we covered some of the necessary ground towards this purpose. The proposed methodology, as hopefully it is evident from the material presented in this thesis,  allows for a scalable framework that is open to creative solutions and linguistically motivated reasoning. Besides a challenge, then, I consider the glass-box approach\index{glass-box approach}\index{tensors!glass-box approach}\index{tensor-based models!glass-box approach} that this thesis adopts a very important advantage, one step further from running (for example) a deep net-based black box a number of times with different parameters until our numbers exceed some threshold. This is actually one of the most appealing characteristics of tensor-based models: they allow us to actively express our intuitions regarding both the compositional and distributional behaviour of the various relational and functional words.

The use of the Frobenius algebras in the way described in this chapter and Chapter \ref{ch:frobverbs} is a means of providing additional structure to the compact setting of the original model, which essentially comes with a rather limited set of tools consisting of a tensor operation and $\eta$ and $\epsilon$ maps. As we saw, language can be more complex than this. The introduction of Frobenius operators provides a second form of composition which co-exists with tensor product and can help modelling more complex linguistic phenomena such as intonation and coordination. From this perspective, the research presented here can be seen as laying the groundwork for a novel form of a grammar enriched with Frobenius structure. In such a grammar, the appropriate translation from syntax to semantics would be explicitly handled in the functorial passage itself, rather than by direct tensor assignments as in this thesis. The intuition in both cases is the same and aims at providing a compositional model better tailored to the nature of vector space semantics than the current attempts. The goal is twofold: (a) to optimize the compositional process by eliminating redundant interactions (especially in the case of functional words); and (b) to circumvent tensor product and impose equal contribution between words when this makes linguistic sense.

A number of proposals detailed in this chapter have been applied in practice with positive results in the context of an experiment we are going to see later in \S\ref{sec:term-definition}. The goal of the task is to classify simplified dictionary definitions to their corresponding terms based on the distance of their vectors. The experiment involves the creation of composite vectors for the definitions, a process that required linear maps for verb- and noun- modifying prepositions as well as for various types of conjunctions. However, a large scale evaluation of the categorical model on sentences of completely arbitrary structure remains still a long-standing goal to which hopefully the current work contributes positively.

\chapter{Dealing with Lexical Ambiguity}
\label{ch:ambiguity}

\begin{chabstract}
This chapter investigates the relation between disambiguation and composition in the context of CDMs. A new compositional methodology is proposed which is capable of efficiently handling the different levels of lexical ambiguity in language by introducing an explicit disambiguation step before the composition. Furthermore, the categorical model of \cite{Coeckeetal} is recast in a higher-level framework which incorporates ambiguity by following a quantum-theoretic perspective. 
\end{chabstract}

\noindent
Section \ref{sec:compdistr} provided a concise introduction to a number of CDMs, which were further categorized according to the method they utilize to represent relational words and the composition function they employ (\S \ref{sec:taxonomy}). Regardless of their level of sophistication, however, most of these models do share a common feature: they all rely on ambiguous vector representations, where all the meanings of a homonymous word, such as the noun `bank', are merged into the same vector or tensor. An interesting question, then, refers to the way in which CDMs affect (and get affected by) those ambiguous words; in other words, given a sentence such as ``a man was waiting by the bank'', we are interested to know to what extent a composite vector can appropriately reflect the intended use of word `bank' in that context, and how such a vector would differ, for example, from the vector of the sentence ``a fisherman was waiting by the bank''.

Interestingly, from a formal semantics perspective this question is irrelevant, since the meanings of words are represented by logical constants that have been explicitly set before the compositional process. In the case of CDMs, however, ambiguity is a very important factor.\index{CDMs!and ambiguity} When acting on an ambiguous vector space, a CDM seems to perform two tasks at the same time, composition \textit{and} disambiguation, leaving the resulting vector in a state that is hard to interpret. In the ideal case, the composition function should be able to perform both tasks in parallel, providing an equivalent of the notion of \textit{continuous meaning}, as this is discussed in \S \ref{sec:continuous}. In practice, though, as a series of experiments in Chapter \ref{ch:wsdexp} will reveal, this is hardly the case. I argue that the reason for this is that a CDM is not capable of treating all kinds of ambiguity in a uniform way. There should be a clear separation of cases of \textit{homonymy}, where the same word has more than one completely disjunctive meaning, from \textit{polysemous} cases, in which a word has multiple but still related senses.

The purpose of this chapter is to investigate these issues in the context of compositional models of meaning to an appropriate depth, and to provide insights that can lead to more efficient CDMs. I will start by discussing the different levels of lexical ambiguity that may occur in language and I will attempt a connection with two different notions of meaning originated in psycholinguistics, that of continuous and discretized meaning. Later, in \S\ref{sec:quantum}, we will see how these concepts can be incorporated into the categorical framework of Coecke et al. \cite{Coeckeetal} by exploiting the fact that both our model and quantum mechanics follow a vector space semantics.

%
%

\section{Understanding lexical ambiguity}
\label{sec:continuous}

\index{ambiguity!understanding the nature of|(}

In order to deal with lexical ambiguity, as the title of this chapter claims it is aiming to do, we firstly need to understand its nature. In other words, we are interested to study in what way an ambiguous word differs from an unambiguous one, and what is the defining quality that makes this distinction clear. On the surface, the answer to these questions seems straightforward: an ambiguous word is one with more than one definition in the dictionary. Inevitably, though, this leads to another question, for which, as Hoffman et al. note \cite{hoffman2013}, there exists no satisfactory answer:

\vspace{-0.3cm}
\begin{quote}
  \textit{What is the criterion of considering two word uses as separate senses?}
\end{quote}
\vspace{-0.3cm}

For \textit{homonymous} cases,\index{homonymy} in which due to some historical accident words that share exactly the same spelling and pronunciation refer to completely different concepts, the distinction is straightforward. One such case is the word `bank', which can mean financial institution, land alongside river, or a sequence of objects in a row. However, for \textit{polysemous}\index{polysemy} words things are not so simple. In contrast to homonymy, \textit{polysemy} relates to subtle deviations between the different senses of the same word. Such a case for example would be `bank' again as a financial institution and as the concrete building in which the financial institution is accommodated. And while it is not wrong for one to claim that the number of meanings\footnote{From this point on, I adopt the convention to refer to \textit{meanings} of a word when talking about homonymous cases, and to \textit{senses} of a word for polysemous cases.}
of a homonymous word can be considered almost fixed across different dictionaries, the same is not true for the small and overlapping variations of senses that might be listed under a word expressing a polysemous case. 

The approach of classifying a word as ambiguous or not based on the number of its discrete lexicographic entries in a dictionary is in line with most psycholinguistic studies of ambiguity (see, for example \cite{borowsky1996,rodd2004}). However, a different perspective that is progressively becoming more popular among researchers rejects the assumption that a word is restricted to a limited number of predefined senses. Instead of this, the meaning of every word is considered dynamic, varying continuously depending on the context in which the specific word appears  \cite{mcclelland1989,landauer2001,hoffman2013}. Under this perspective, two uses of the same word are never truly identical in meaning, as their exact connotations depend on the specific context in which they appear. Hence, the separation of a word's meaning to some predefined number of senses is nothing more than an artificial \textit{discretization}\index{discretized meaning}\index{meaning!discretized} of the underlying continuous spectrum of senses.

\index{ambiguity!understanding the nature of|)}

The concept of continuous meaning\index{continuous meaning}\index{meaning!continuous} is very intuitive and fits naturally with the distributional semantics perspective, where a word is expressed in terms of all other words in the vocabulary. However, I argue that it can be only applied on polysemous cases, where a word still expresses a coherent and self-contained concept. Recall the example of the polysemous use of `bank' as a financial institution and the building where the services of the institution are offered; when we use the sentence `I went to the bank' (with the financial meaning of the word in mind) we essentially refer to both of the polysemous meanings of `bank' at the same time---at a higher level, the word `bank' expresses an abstract but concise concept that encompasses all of the available polysemous meanings. On the other hand, the fact that the same name can be used to describe a completely different concept (such as a river bank or a number of objects in a row) is nothing more than an unfortunate coincidence expressing lack of specification. 

For cases like these, a combination of the notions of discretized and continuous meaning seems a better strategy. The lack of specification imposed by the homonymous meanings must be addressed by an initial discretization step, responsible for defining explicit semantic representations for every homonymous use of the word at hand; furthermore, the notion of continuous meaning will be still retained within the produced parts, which will now provide much more coherent and accurate semantic representations for the corresponding concepts.

\index{continuous meaning!relation with polysemy|(}
\index{discretized meaning!relation with homonymy|(}
\index{homonymy!and discretized meaning|(}
\index{polysemy!and continuous meaning|(}

We have now arrived at a central claim of this thesis: when the use of the same word to describe two concepts is nothing more than an unfortunate historical coincidence, discretized meaning should be favoured over continuous meaning. Let me make this idea precise:

\vspace{-0.5cm}
\begin{equation}
\label{equ:meaning}
\begin{minipage}{0.90\linewidth}
   \centering
   $\text{homonymy} = \text{discretized meaning}$ \\
   $\text{polysemy} = \text{continuous meaning}$ 
\end{minipage}   
\end{equation}

\index{polysemy!and continuous meaning|)}
\index{homonymy!and discretized meaning|)}
\index{discretized meaning!relation with homonymy|)}
\index{continuous meaning!relation with polysemy|)}

In the next section I outline a concrete proposal for incorporating the above relationships into a CDM.

\section{Prior word disambiguation and composition}
\label{sec:priordis}

\index{prior disambiguation|(}
\index{homonymy!and prior disambiguation|(}
\index{polysemy!and prior disambiguation|(}

\begin{figure}[b!]
  \centering
  \includegraphics[scale=0.43,clip=true,trim=1.0cm 0.4cm 0cm 0cm]{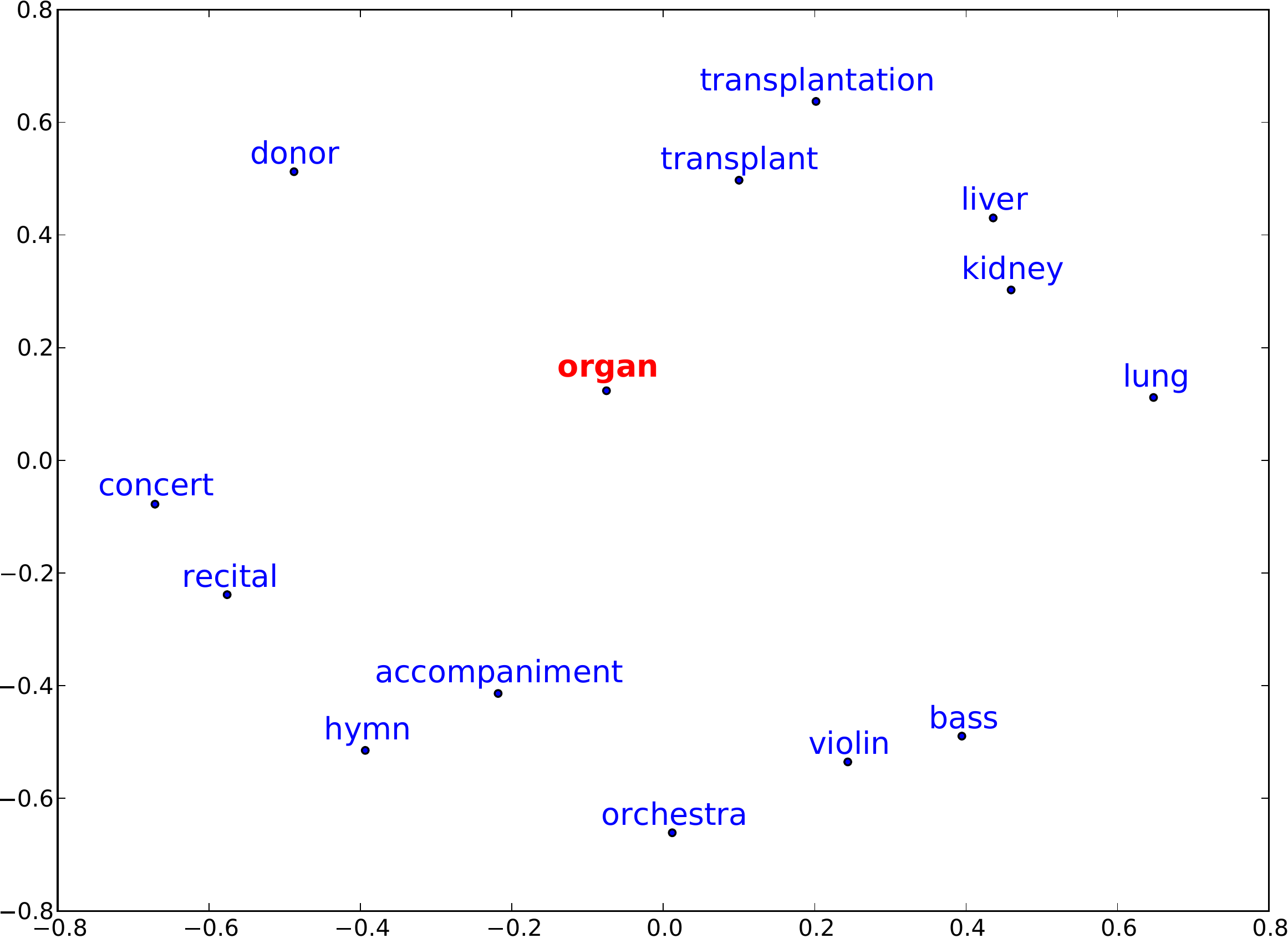}
  \caption[The vector of a homonymous word in a two-dimensional semantic space.]{The vector of a homonymous word in a two-dimensional semantic space (projection of real word vectors using multi-dimensional scaling).}
  \label{fig:organ-amb}
\end{figure}

Transferring the above discussion to high dimensional vector spaces, we start by observing that a vector space in which all the meanings of an ambiguous word are fused into the same vector representation is a setting far less than optimal. Indeed, having a single vector representing all meanings of a genuinely homonymous word does not make more sense than having a single vector representing the meanings of some other completely unrelated words, say `book' and `banana'. A vector like this would be the average of all senses, inadequate to reflect the meaning of any of them in a reliable way. This is demonstrated in Fig. \ref{fig:organ-amb} for the homonymous word `organ'.

In order to avoid using such uninformative inputs in our compositional function, every distinct \textit{meaning} of a word should be represented by a different vector. We can think of this requirement as the inevitable discretization related to the homonynous cases. On the other hand, we would expect that any composition function between word vectors would at least be able to adequately resolve the small deviations of senses due to polysemy, providing in this way an analogy of the continuous meaning concept. 


I will now proceed and present a way to achieve the above. Following Pulman \cite{pulman2013} and subsequent work I conducted with him and Sadrzadeh \cite{kartsaklis:2013:CoNLL}, I propose the introduction of an explicit disambiguation step that precedes the composition, which will be responsible for the homonymy-related discretization of the meaning. The disambiguated vectors will be used as inputs to the actual composition step, which will further refine the meaning of words to arbitrary levels of polysemy. The main idea is to replace Eq. \ref{equ:main}, aiming at providing a meaning for a sentence by composing the vectors for the words therein, with the following:

\begin{equation}
  \ov{s} = f\left(\phi(\ov{w_1},C_1),\phi(\ov{w_2},C_2),\hdots,\phi(\ov{w_n},C_n) \right)
  \label{equ:maindis}
\end{equation}

\noindent
where $f$ is our compositional function as before, and  $\phi(\ov{w_i},C_i)$ is a function that provides a disambiguated version of the vector (or tensor) for $i$th word given the context $C_i$ of this word (e.g. all other words in the same sentence). We can think of function $\phi$ as a means to handle all the homonymous cases, when the gap in meaning is quite big to be handled by the composition function alone.

A reasonable question that the above proposal poses is how one can  explicitly recognize and treat differently homonymous and polysemous cases of words, a problem that has been proved far from trivial in linguistics. The answer is that such an action is not necessary, given any rough discretization of meaning that at least captures the homonymous uses of the word to an adequate level. Assuming that a word $w$ is polysemous, and not homonymous, we can expect that all ``meaning'' vectors that might be produced by $\phi(\ov{w},C_w)$ will be very similar to each other as well as to the original ambiguous vector $\ov{w}$, reflecting the fact that there are only small deviations between the different senses of the word. Hence, the effect for the outer composition function $f$ should be minimal (see also discussion in \cite{baroni2014frege}, p. 94).

%
%

The effectiveness of the methodology expressed by Eq. \ref{equ:maindis} has been experimentally tested and verified in a series of experiments, which will be presented in Chapter \ref{ch:wsdexp} along with concrete methods for performing sense induction and disambiguation of word vectors and tensors. My goal for the rest of the present chapter is to show how the idea of treating each level of lexical ambiguity in a different way finds an intuitive manifestation in the context of the categorical model of \cite{Coeckeetal}. Specifically, I will exploit the framework of quantum mechanics in order to provide an ambiguity-aware extension of the model that is capable of handing the notions of homonymy and polysemy in a unified and natural manner.

\index{polysemy!and prior disambiguation|)}
\index{homonymy!and prior disambiguation|)}
\index{prior disambiguation|)}

\section{A quantum perspective in linguistics}
\label{sec:quantum}

Although seemingly unrelated, quantum mechanics and linguistics share a common link through the framework of compact closed categories, which has been used in the past to study and reason about phenomena related to both of these areas. The application of the categorical framework to linguistics is the subject of this thesis; even more interestingly, exactly the same machinery has been used in the past by Abramsky and Coecke \cite{abramsky2004} for providing structural proofs for a class of quantum protocols, essentially recasting the vector space semantics of quantum mechanics in a more abstract way. In this section I will try to make the connection between the two fields even more explicit, taking advantage of the fact that the ultimate purpose of quantum mechanics is to deal with uncertainty---and this is essentially what we need to achieve here in the context of language.


\subsection{From simple vectors to quantum states}
\label{sec:vec-states}

\index{quantum states|(}

As we discussed in \S \ref{sec:compsem}, the formal semantics perspective interprets the meanings of words as unique primitives. For a vocabulary $V$, let these primitives be given by a set of orthogonal vectors $\{\ket{n_i}\}_i$ with $0 \leq i < |V|$; the vector in which the $i$th element is 1 represents the $i$th word in $V$ in a classical sense. Now let us use the distributional hypothesis in order to start moving towards the quantum realm. Take a semantic space fixing a basis as above; a vector in that space will be represented as the sum:\footnote{Traditionally, vectors representing quantum states are given in \textit{Dirac notation}, and in this chapter I am going to follow this convention.}

\begin{equation}
\label{equ:vec}
  \ov{w} = a_1\ket{n_1} + a_2\ket{n_2} + \hdots + a_n\ket{n_n}
\end{equation}

Under this setting, we think of words as \textit{quantum systems}\index{words!as quantum systems} and bases $\ket{n_i}$ as describing a specific unambiguous \textit{state}: the state of word $w$ occurring in the vicinity (i.e. same context) of the $i$th word in the vocabulary. Then Eq. \ref{equ:vec} can be seen as a \textit{superposition}\index{superposition} of all these states, and the coefficients $a_i$ as denoting the strength of every state. In quantum mechanics terms, we have a \textit{pure state}\index{pure states} that expresses our knowledge regarding the words with which $w$ co-occurs in language (and how often), when $w$ is taken out of context. That is, when $w$ is seen alone, it exists in an abstract state where it co-occurs with \textit{all} the words of the vocabulary with strengths denoted by the various coefficients. 
%

Note that, in the ideal case, every disjoint meaning of a genuinely homonymous word must be represented by a different pure state. We expect the word \textit{bank} to have at least two different meanings, which will correspond to distinct pure states:

\begin{eqnarray}
\begin{split}
 \ov{bank}_{fin} = a_1\ket{n_1} + a_2\ket{n_2} + \hdots + a_n\ket{n_n} \\
 \ov{bank}_{riv} = b_1\ket{n_1} + b_2\ket{n_2} + \hdots + b_n\ket{n_n}
\end{split} 
\end{eqnarray}

This is a direct consequence of the fact that \textit{bank} as a land alongside a river is expected to be seen in drastically different contexts than as a financial institution, so that $\{b_i\}_i \neq \{a_i\}_i$---that is, it is essentially a different word. However, since the same ``label'' is used to express two different concepts, we need a way to express our genuine uncertainty (in a classical sense) regarding the state in which our system may be found (i.e. under which sense the specific word is used in a specific context). In quantum mechanics terms, this is achieved by the means of a \textit{mixed state},\index{mixed states} which is seen as a statistical ensemble of pure states. A mixed state is described by a \textit{density matrix}:\index{density matrices} a self-adjoint ($\rho=\rho^{\dagger}$) positive semi-definite ($\langle v|\rho|v \rangle \geq 0$ for all $\ket{v}$) matrix of trace one, created as below:\index{words!as density matrices}

\begin{equation}
   \rho(w) = \sum_i p_i \ket{s_i}\bra{s_i}
\end{equation}

\noindent where $\{ (p_i,\ket{s_i}) \}_i$ denotes an ensemble of pure states, each one of which can occur with probability $p_i$; we also have $\sum_i p_i = 1$. A density matrix can also be used for expressing a single pure state (that is, a word with a single  definition); in that case we simply have:

\begin{equation}
   \rho(w) = \ket{w}\bra{w}
\end{equation}

\index{homonymy!and mixed states|(}
\index{polysemy!and pure states|(}

Before proceeding further I would now like to restate ``Eq.'' \ref{equ:meaning}, incorporating in it the quantum perspective we have just discussed. Under this new view, quantum superposition (pure states) accounts for the continuous meaning, while the discretization of meaning is achieved through density matrices (mixed states). That is, we have:

\begin{equation}
\begin{minipage}{0.90\linewidth}
   \centering
   $\text{homonymy} = \text{discretized meaning} = \text{mixed state (density matrix)}$ \\
   $\text{polysemy} = \text{continuous meaning} = \text{pure state (superposition)}$
\end{minipage}   
\end{equation}

\index{polysemy!and pure states|)}
\index{homonymy!and mixed states|)}

The next section will hopefully make clear what this means in practice.

\index{quantum states|)}

\subsection{Quantum measurements}
\label{sec:qmeasure}

\index{quantum measurements|(}

Density matrices interact with other self-adjoint operators known as \textit{observables},\index{observables} to produce quantum measurements. Note that, by the spectral theorem, every self-adjoint matrix $A$ has an eigen-decomposition\index{eigen-decomposition} $A=\sum_i e_i \ket{e_i}\bra{e_i}$, where $e_i$ is an eigenvalue associated with eigenvector $\ket{e_i}$; furthermore, the eigenvectors of $A$ constitute an orthonormal basis. An observable describes some quantity, such as spin or momentum; its eigenvalues correspond to possible values of this quantity, while the associated eigenvectors (or \textit{eigenstates})\index{eigenstate} are the unambiguous states assigned to these values. Assuming a system in state $\ket{\psi}$, a measurement of $A$ with respect to that (pure) state will return the expectation value of $A$ in state $\psi$:

\begin{equation}
  \langle A \rangle_{\psi} = \bra{\psi} A \ket{\psi}
  \label{equ:qmeas-pure}
\end{equation}

Note that when $\ket{\psi}$ corresponds to one of the unambiguous eigenstates, the above calculation returns the associated eigenvalue with absolute certainty, as claimed above: 

\begin{equation}
\bra{e_i}A\ket{e_i}=\bra{e_i}e_i\ket{e_i}=e_i
\end{equation}

Now, in the case of an arbitrary state vector $\ket{\psi}$, we observe that any such vector can be expressed in terms of the orthonormal basis $\{\ket{e_i}\}_i$ as $\sum_i \langle e_i | \psi\rangle \ket{e_i}$. Substituting this form into Eq. \ref{equ:qmeas-pure} above will give:

\begin{equation}
 \langle A \rangle_{\psi} = \sum_i \langle\psi|e_i\rangle \bra{e_i}A\ket{e_i} \langle e_i|\psi\rangle =
 \sum_i |\langle e_i|\psi\rangle|^2 e_i
\end{equation}

According to the \textit{Born rule},\index{Born rule} the quantity $|\langle e_i|\psi\rangle|^2$ gives the probability of the measurement to return eigenvalue $e_i$; in other words, the result is the average of all potential values of the observable weighted by their probabilities with respect to $\ket{\psi}$. It can be shown that for a density matrix $\rho = \sum_j p_j \ket{\psi_j}\bra{\psi_j}$ expressing a mixed state, and an observable $A$ with eigen-decomposition $A=\sum_i e_i \ket{e_i}\bra{e_i}$, the measurement takes the following form:

\begin{equation}
  \langle A \rangle_{\rho} = \sum_j p_j \bra{\psi_j}A\ket{\psi_j} = 
  \sum_j \sum_i p_j e_i |\langle e_i|\psi_j\rangle|^2 =
   \op{Tr}(\rho A)
   \label{equ:qmeas-mixed}
\end{equation}

\noindent where $\op{Tr}$ denotes trace.\index{trace} From a statistical point of view, this is the average of the expectation values of $A$ for each one of the pure states $\psi_i$, weighted by the corresponding probabilities. 

Note that a density matrix, being a self-adjoint operator itself, can serve as an observable with respect to another density matrix describing the state of a different system. We can interpret such a measurement as the probability of the state expressed by the first matrix to be the configuration corresponding to the second one. On the linguistics side, taking the trace of $\rho_1\rho_2$ when those two operators describe words as detailed above, can be seen as a measurement of how probable it is for the two words to occur in the same context. In other words, we have a similarity notion\index{quantum measurements!as similarity notion between words} that will take the place of the usual inner product we have been using until now. Observe in Eq. \ref{equ:qmeas-mixed} that during this computation two levels of averaging occur in parallel: the inner sum (indexed by $i$) is responsible for the superposition part over the pure states, hence our means to handle polysemy cases; on the other hand, the outer sum (indexed by $j$) carries the statistical average related to the homonymous cases. 

There is one last concept to discuss before proceeding to formalize all these ideas in a more rigorous way; that of a \textit{partial trace}.\index{partial trace} Assuming a composite quantum system represented by a density matrix $\rho \in \mathcal{H}_A\ten \mathcal{H}_B$, we might want to obtain a measurement of an observable $M$ with respect to one of the subsystems, say $A$. For these cases we compute the partial trace as $\op{Tr}(\rho(M\ten 1_B))$, an operation that ``destroys'' system $A$ but leaves system $B$ intact. Recall that, in our categorical framework, composite systems such as the above correspond to words with complex types, such as adjectives and verbs. It turns out that in the quantum setting this notion of partial trace serves as the composition function between words,\index{quantum measurements!as composition function} replacing the $\epsilon$-maps of the ``classical'' setting. Intuitively, a ``measurement'' of a noun serving as the subject of some intransitive verb, with respect to the density matrix of that verb, produces a new density matrix for the compound system that reflects the probability of the two words occurring together. 

\index{quantum measurements|)}

\subsection{Operators and completely positive maps}
\label{sec:operators}

\index{operators!positive|(}

I will now start to adapt the categorical framework of Chapter \ref{ch:framework} in a way that adopts the quantum view of \S \ref{sec:vec-states} and \S\ref{sec:qmeasure}. First and foremost, this view requires us to be a little more precise regarding the form of our semantic space. We start from the fact that, traditionally, a pure quantum state is a vector in a Hilbert space over the field of complex numbers $\mathbb{C}$.\footnote{This is because quantum mechanics deals with \textit{wave functions}, the proper representation of which requires complex numbers. In practice, though, it should be noted that all semantic spaces for linguistic purposes are built over real numbers.} Furthermore, for our purpose this space will be finite-dimensional. We refer to the category of finite-dimensional Hilbert spaces and linear maps as \textbf{FHilb}.\index{fhilb@\textbf{FHilb}, category}\index{category!fhilb@\textbf{FHilb}} This category is \textit{dagger compact closed},\index{category!dagger compact closed}\index{dagger compact closed category}\index{compact closed category!dagger} that is, equipped with an involutive contravariant functor $\dagger:\textbf{FHilb} \to \textbf{FHilb}$ with identity on objects. Specifically, every morphism $f:A \to B$ has a dagger which is the unique map $f^\dagger: B \to A$ satisfying $\langle f(u)|v \rangle = \langle u|f^\dagger(v) \rangle$ for all $u \in A, v \in B$. Additionally, the following diagram commutes for all objects:

\begin{equation}
\begin{tikzpicture}[scale=1.5,baseline=30.0pt]
\node (A) at (0,2.5) {$I$};
\node (B) at (4,2.5) {$A\ten A^*$};
\node (C) at (4,0) {$A^*\ten A$};
\path[->,font=\small]
(A) edge node[above]{$\epsilon^\dagger_A$} (B)
(B) edge node[right]{$\sigma_{A,A^*}$} (C)
(A) edge node[left]{$\eta_A$}(C);
\end{tikzpicture}
\end{equation}

The existence of a dagger provides a categorical definition for the notion of inner product;\index{inner product, with daggers} for two state vectors $\ket{\psi}$ and $\ket{\phi}$ in a Hilbert space $\mathcal{H}$, we can now define $\bra{\psi}\phi\rangle$ as the following morphism:

\begin{equation}
  I \xrightarrow{~\psi~} \mathcal{H} \xrightarrow{~\phi^{\dagger}~} I
  \label{equ:cat-inproduct}
\end{equation}

In category \textbf{FHilb}, a pure state is a morphism $\psi: I \to A$ such that $\psi^\dagger \circ \psi = 1_I$, quite similar to a word vector of meaning as those introduced in \S \ref{sec:dissem}.  However, our decision to adopt the density matrix representation results in a \textit{doubling};\index{doubling of dimensions}\index{density matrices!doubling of dimensions} indeed, the density matrix associated with a pure state is the positive operator $\psi \circ \psi^\dagger:A \to A$ that sends states to states. Similarly, the density matrix of a mixed state is the operator $\rho:A \to A$. We can retrieve the original notation of representing vectors in some state space as morphisms from the unit by using the notion of \textit{name},\index{name of an operator}\index{operators!names} which, for a mixed-state operator is defined as follows:

\begin{equation}
  \name{\rho} = I \xrightarrow{\eta^r} A^*\ten A \xrightarrow{1_{A^*} \ten \rho} A^*\ten A
\end{equation}

Intuitively, we can think of a name as the concrete matrix corresponding to the linear map of the specified morphism. In our graphical calculus, $\name{\rho}$ has the following representation:

\begin{equation}
\footnotesize

\InputIfFileExists{./tikz/name.tikz}{}{\input{./tikz/name.tikz}}

\normalsize
\end{equation}

Note that an operator representing a pure state $\psi$ has the following name:

\begin{equation}
  I \xrightarrow{\eta^r} A^*\ten A \xrightarrow{1_{A^*}\ten \psi^\dagger}
  A^*\ten I \xrightarrow{1_{A^*}\ten \psi} A^*\ten A
\end{equation}

\noindent which, as is evident in the graphical language, collapses to tensor product of the vector of $\psi$ with itself.

\begin{equation}
\footnotesize

\InputIfFileExists{./tikz/purename.tikz}{}{\input{./tikz/purename.tikz}}

\normalsize
\end{equation}

Since we deal with positive operators (i.e. $f$ is positive if it satisfies $\langle f(v)|v\rangle\geq 0$ for all $v$), it would be helpful to define the notion categorically: In a dagger compact closed category, a morphism $f:A\to A$ is positive when there exists an object $B$ and a morphism $g:A\to B$ such that $f=g^\dagger \circ g$. Graphically:

\begin{equation}
 \footnotesize
 
\InputIfFileExists{./tikz/positive.tikz}{}{\input{./tikz/positive.tikz}}

 \normalsize
\end{equation}

\index{operators!positive|)}

We also need to formally define the notion of trace,\index{trace} since, as we saw in \S \ref{sec:vec-states}, this will be our means to perform measurements between states; for a morphism $f:A \to A$, this takes the following form:

\begin{equation}
  \op{Tr}(f) = I \xrightarrow{\eta_A} A^*\ten A \xrightarrow{\sigma_{A^*,A}} A\ten A^* \xrightarrow{f\ten 1_{A^*}} A\ten A^* \xrightarrow{\epsilon_A} I
\end{equation}

\noindent and, graphically:

\begin{equation}
\footnotesize

\InputIfFileExists{./tikz/trace.tikz}{}{\input{./tikz/trace.tikz}}

\normalsize
\end{equation}

Similarly, we define what a partial trace is.\index{partial trace} As informally discussed before and will be more evident in \S \ref{sec:quantum-recasting}, in the new framework we are introducing this takes the role of our composition function. For a morphism $A_1\ten \hdots \ten A_n \ten C \to B_1\ten \hdots \ten B_n \ten C$ it is defined as:

\begin{equation}
\footnotesize

\InputIfFileExists{./tikz/partrace.tikz}{}{\input{./tikz/partrace.tikz}}

\normalsize
\end{equation}

Recasting the categorical framework of \cite{Coeckeetal} to the density matrices formalism requires another important step: the morphisms in \textbf{FHilb} are simple linear maps between Hilbert spaces, but what we actually need are maps that send positive operators to positive operators, while at the same time respect the monoidal structure. It turns out that the notion we are looking for is that of \textit{completely positive maps}.\index{completely positive maps} Specifically, according to the \textit{Stinespring dilation theorem}\index{Stinespring dilation theorem} a morphism $f: A^*\ten A \to B^*\ten B$ is completely positive if 
there exists an object $C$ and a morphism $g:A\to C\ten B$ such that the following is true:

\begin{equation}
\label{fig:defcpm}
  \footnotesize
  
\InputIfFileExists{./tikz/spinespring.tikz}{}{\input{./tikz/spinespring.tikz}}

  \normalsize
\end{equation}
\vspace{0.3cm}

Thankfully, it has been shown that dagger compact closed categories stand in a certain correspondence with categories in which morphisms are completely positive maps, as required for our construction. This is the topic of the next section.

\subsection{The CPM construction}

\index{CMP construction|(}
\index{fhilb@\textbf{FHilb}, category!and CPM construction|(}

Selinger \cite{selinger2007dagger} has showed that any dagger compact closed category can be associated with a category in which the objects are the objects of the original category, but the maps are completely positive maps. Furthermore, he showed that the resulting category is again dagger compact closed. The \textit{CPM construction} is exactly what we need in order to impose complete positivity in our model, and in this section I am going to detail how it can be applied to \textbf{FHilb}. 

\vspace{0.8cm}
\begin{definition}
  \label{def:cpm}
  The category \textbf{CPM(FHilb)}:
  \begin{itemize}
    \item Has as objects finite-dimensional Hilbert spaces (exactly as \textbf{FHilb}),
    \item Each morphism $f:A \to B$ in \textbf{CPM(FHilb)} is a completely positive map $f:A^*\ten A \to B^*\ten B$ in \textbf{FHilb}; furthermore, composition of morphisms is as in \textbf{FHilb},
    \item Tensor product of objects is as in \textbf{FHilb}; the tensor product of two morphisms $f:A^*\ten A \to B^*\ten B$ and $g:C^*\ten C \to D^*\ten D$ is given as below:
    
    \index{tensor product!in CPM construction|(}
    
    \vspace{-0.8cm}
    \begin{eqnarray}
      C^* \ten A^* \ten A \ten C \xrightarrow{\cong} A^*\ten A \ten C^*\ten C \xrightarrow{f\ten g}~~~~~~~~~~~~~~~~~~~~~\nonumber \\
      B^* \ten B \ten D^* \ten D \xrightarrow{\cong} D^* \ten B^* \ten B \ten D
      \label{equ:tensor-cpm}
    \end{eqnarray}

    Graphically:    
    
    \begin{equation}
    \label{fig:tensorcpm}
      \footnotesize
      
\InputIfFileExists{./tikz/cpm-tensor.tikz}{}{\input{./tikz/cpm-tensor.tikz}}

      \normalsize
    \end{equation}
    
    \index{tensor product!in CPM construction|)}
    
  \end{itemize}
\end{definition}

Eq. \ref{equ:tensor-cpm} and Diagram (\ref{fig:tensorcpm}) above make use of the fact that \textbf{FHilb} is a symmetric dagger compact closed category, equipped with a natural isomorphism $\sigma_{A,B}: A\ten B \to B \ten A$. Note that the tensor product of morphisms is again a completely positive map according to Diagram (\ref{fig:defcpm}); indeed, if we express the two tensored morphisms in their factorized form we get:

\begin{equation}
  \footnotesize
  
\InputIfFileExists{./tikz/tensor-factorized.tikz}{}{\input{./tikz/tensor-factorized.tikz}}

  \normalsize
\end{equation}

The resulting category is again dagger compact closed (see Theorem 4.20 and related proof in \cite{selinger2007dagger}). Table \ref{tbl:cpm} shows how the CPM construction translates all the important morphisms we are going to need for our purposes (in general by doubling them).

\begin{table}[t]
\begin{center}
\footnotesize
\begin{tabular}{c}
\hline
~~~~~~\textbf{CPM(FHilb)}~~~\textbf{FHilb}~~~~~~~~~~~~~~~~~~~~~~~~~~\textbf{CPM(FHilb)}~~~~~~~~~\textbf{FHilb} \\
\hline
~~ \\
 
\InputIfFileExists{./tikz/table-cpm.tikz}{}{\input{./tikz/table-cpm.tikz}}
 \\
~~ \\ 
 \hline
\end{tabular}
\caption[The application of the CPM construction on \textbf{FHilb} using the graphical calculus.]{The application of the CPM construction on \textbf{FHilb} using the graphical calculus. Tensor product is shown in (\ref{fig:tensorcpm}).}
\label{tbl:cpm}
\end{center}
\end{table}

\index{fhilb@\textbf{FHilb}, category!and CPM construction|)}
\index{CMP construction|)}

\section{An ambiguity-aware model of language}
\label{sec:quantum-recasting}

We are now ready to put together all the concepts introduced above in the context of a compositional model of meaning. We deal first with the generic case of using full (as opposed to reduced) tensors for relational words; \S \ref{sec:frob-density} describes how one can apply the Frobenius framework of Chapter \ref{ch:frobverbs} to the new model. 

\begin{definition}
Let $\mathcal{L}: \textbf{FHilb} \to \textbf{CPM(FHilb)}$ be the canonical functor applying the CPM construction as in Def. \ref{def:cpm}. Then, the passage from a grammar to distributional meaning is defined according to the following composition:

\begin{equation}
  \textbf{C_F} \xrightarrow{\mathcal{F}} \textbf{FHilb} \xrightarrow{\mathcal{L}} \textbf{CPM(FHilb)}
\end{equation}

\noindent where $\mathcal{F}$ is a strongly monoidal functor that sends pregroup types from the free category generated over a partial set of atomic grammar types to finite-dimensional Hilbert spaces in \textbf{FHilb}.

\end{definition}

\index{sentence meaning!in the quantum model|(}

Let us now proceed and redefine what we would consider as ``meaning'' of a sentence in the new model.

\begin{definition}
Take $\rho(w_i)$ to be a state vector $I\to \mathcal{L}(\mathcal{F}(p_i))$ corresponding to operator $\mathcal{F}(p_i)\to \mathcal{F}({p_i})$ for word $w_i$ with pregroup type $p_i$ in a sentence $w_1w_2\hdots w_n$. Given a type reduction $\alpha:p_1\cdot p_2 \cdot \hdots \cdot p_n \to s$, the meaning of the sentence is defined as follows:

\begin{equation}
  \label{equ:meaning-new}
  \mathcal{L}(\mathcal{F}(a))\left(\rho(w_1)\ten \rho(w_2) \ten \hdots \ten \rho(w_n)\right)
\end{equation}

\noindent where $\ten$ here denotes tensor product in \textbf{CPM(FHilb)}. We further require that $\rho(w_i)$ has been prepared as a proper density matrix from a statistical ensemble of pure states associated with the $i$th word in the sequence.

\index{sentence meaning!in the quantum model|)}

\end{definition}

Eq. \ref{equ:meaning-new} is quite similar to our original equation of Def. \ref{def:categorical}; the major difference is that a sentence is not represented any more as the tensor product of vectors, but as the tensor product of the density matrices associated with them. It would be helpful to clarify at this point what this means exactly. \index{tensor product!in CPM construction|(} Recall that tensoring any two positive maps $A^*\ten A \xrightarrow{f} B^*\ten B$ and $C^*\ten C \xrightarrow{g} D^*\ten D$ by a simple juxtaposition of them as done until now is not enough, since the resulting morphism $A^*\ten A \ten C^*\ten C \xrightarrow{f\ten g} B^*\ten B \ten D^*\ten D$ will not be completely positive. Complete positivity requires from us to \textit{embed} one object into another starting from $C^*\ten A^* \ten A \ten C$, which, for two words with density matrices $\rho_1$ and $\rho_2$, leads to the following graphical representation:

\begin{equation}
\footnotesize

\InputIfFileExists{./tikz/cpm-juxtap.tikz}{}{\input{./tikz/cpm-juxtap.tikz}}

\normalsize
\end{equation}
\vspace{0.3cm}

\index{tensor product!in CPM construction|)}

Exactly as in the original model, measuring the semantic similarity between two words is an action performed by the application of an $\epsilon$-map, which in \textbf{CPM(FHilb)} is doubled as everything else. Note that for two words with operators $\rho_1$ and $\rho_2$, the result of this is nothing more than $\op{Tr}(\rho^\dagger_2 \rho_1)$:

\begin{equation}
\footnotesize

\InputIfFileExists{./tikz/cpm-sim.tikz}{}{\input{./tikz/cpm-sim.tikz}}

\normalsize
\label{equ:cpm-sim}
\end{equation}
\vspace{0.3cm}

I will now proceed to precisely define the semantic representations of nouns and relational words as density matrices. For what follows, I assume that every word is already associated with some statistical ensemble $S=\{(p_i,\ket{s_i})\}_i$, where $\ket{s_i}$ is a state vector corresponding to a particular meaning that might occur in the current context with probability $p_i$. 
We can think of this ensemble as the product of a \textit{preparation} step, a topic that we are going to examine in detail in Chapter \ref{ch:wsdexp}. 

\index{nouns, as density matrices|(}
\index{density matrices!for nouns|(}

\paragraph{Nouns} Nouns are represented as operators describing pure (if they are unambiguous) or mixed (if they have more than one meanings) states as below:

\begin{equation}
\footnotesize

\InputIfFileExists{./tikz/cpm-nouns.tikz}{}{\input{./tikz/cpm-nouns.tikz}}

\normalsize
\end{equation}
\vspace{0.2cm}

\noindent where $|S|$ is the size of the statistical ensemble associated with the word.

\index{density matrices!for nouns|)}
\index{nouns, as density matrices|)}

\index{verbs, as density matrices|(}
\index{density matrices!for verbs|(}

\paragraph{Relational words} We think of a relational word as a $N$-partite system that originally lives in some tensor product of Hilbert spaces $\mathcal{H}_1\ten \hdots \ten \mathcal{H}_N$. In case the word is ambiguous, the system as usual is represented by a mixed state:

\begin{equation}
\footnotesize

\InputIfFileExists{./tikz/cpm-verbs.tikz}{}{\input{./tikz/cpm-verbs.tikz}}

\normalsize
\end{equation}
\vspace{0.2cm}

\index{density matrices!for verbs|)}
\index{verbs, as density matrices|)}

We have now all the pieces we need to compute the meaning of a sentence or a phrase. For a simple intransitive sentence, the composition has the following form:

\begin{equation}
\footnotesize

\InputIfFileExists{./tikz/cpm-intr.tikz}{}{\input{./tikz/cpm-intr.tikz}}

\normalsize
\end{equation}
\vspace{0.2cm}

With the help of diagrammatic calculus, it is now clear why partial trace\index{partial trace} has the role of the composition function in the new setting. We can think of $\rho(s)\ten 1_S$ as our ``observable'', which, when measured with respect to the composite system representing the verb, will return a new reduced state of $\rho(v)$ on the right-hand subsystem. This reduced $\rho(v)$ is calculated as:

\begin{equation}
 \rho_{IN} = \op{Tr}(\rho(v)\circ (\rho(s)\ten 1_S))
\end{equation}

\noindent and serves as the ``meaning'' of the intransitive sentence. The case of an adjective-noun compound is quite similar, with the difference that now the partial trace takes place on the right:

\vspace{-0.3cm}
\begin{equation}
\footnotesize

\InputIfFileExists{./tikz/cpm-adj.tikz}{}{\input{./tikz/cpm-adj.tikz}}

\normalsize
\end{equation}

\begin{equation}
 \rho_{AN} = \op{Tr}(\rho(adj)\circ (1_N \ten \rho(n)))
\end{equation}

Finally, the meaning of a transitive sentence is computed by taking two partial traces, one for the subject dimension of the verb and one of the object dimension:

\vspace{-0.3cm}
\begin{equation}
\label{fig:cpm-trans}
\footnotesize

\InputIfFileExists{./tikz/cpm-trans.tikz}{}{\input{./tikz/cpm-trans.tikz}}

\normalsize
\end{equation}

This results in the following computation:

\begin{equation}
 \rho_{TS} = \op{Tr}(\rho(v)\circ (\rho(s) \ten 1_S \ten \rho(o)))
\end{equation}

\section{Density matrices and Frobenius algebras}
\label{sec:frob-density}

\index{Frobenius algebras!on density matrices|(}
\index{density matrices!and Frobenius algebras|(}
\index{Frobenius algebras!dagger|(}
\index{dagger Frobenius algebras|(}

In my exposition so far I used the assumption that a verb tensor has been faithfully constructed according to its grammatical type (that is, a transitive verb tensor is of order 3, an intransitive verb tensor of order 2, and so on). In this section I will show how the Frobenius framework of Chapter \ref{ch:frobverbs} can be applied to our new quantum setting. Note that while the dimensionality reduction gained by applying the Frobenius operators on our original framework was a welcome extra convenience, for the quantum formulation it becomes a strictly mandatory implementation strategy: doubling an order-3 verb tensor when using a 100-dimensional \textit{real-valued} vector space would require $100^6\times 8 = 8\times 10^{12}$ bytes, that is, 8 terabytes for a single density matrix. However, when one starts from a tensor of order 2 this number reduces to 800 megabytes, which, while still high, is definitely manageable.\index{density matrices!doubling of dimensions}\index{doubling of dimensions} Furthermore, the Frobenius framework will also allow us to continue using the intuitive reasoning previously discussed in Chapters \ref{ch:frobverbs} and \ref{ch:extend} regarding the meaning of relational and functional words. 

In the context of a dagger monoidal category, a \textit{dagger Frobenius algebra} is a Frobenius algebra in which $\Delta^\dagger=\mu$ and $\iota^\dagger = \zeta$. Extending our discussion in \S\ref{sec:frobenius} from \textbf{FVect} to \textbf{FHilb}, an orthonormal basis $\{\ket{i}\}$ in a finite-dimensional Hilbert space induces a special dagger Frobenius algebra with $\Delta:=\ket{i}\to\ket{i}\ten\ket{i}$ and $\iota:=\ket{i}\to 1$, while $\mu,\zeta$ are obtained by taking the daggers of these, respectively. The image of this algebra in \textbf{CPM(FHilb)} is again a special dagger Frobenius algebra, which, as shown in Table \ref{tbl:cpm}, can be constructed by simply doubling the original maps. Copying a dimension from a density matrix, then, takes the following form:

\begin{equation}
  \footnotesize
  
\InputIfFileExists{./tikz/cpm-frob.tikz}{}{\input{./tikz/cpm-frob.tikz}}

  \normalsize
\end{equation}

Similarly, deletion of a dimension is performed as follows:

\begin{equation}
  \footnotesize
  
\InputIfFileExists{./tikz/cpm-frob1.tikz}{}{\input{./tikz/cpm-frob1.tikz}}

  \normalsize
\end{equation}

Assume now that we compute a density matrix of reduced dimensionality for an intransitive verb applying a variation of the argument summing\index{density matrices!argument summing}\index{argument summing!for density matrices} procedure described in \S\ref{sec:tensor-rel}, as below:

\begin{equation}
   \ol{verb}_{IN} = \sum_i \rho(subj_i)
\end{equation}

\noindent where $\rho(subj_i)$ denotes the density matrix of the $i$th subject of that verb in the training corpus. The expansion of the density matrix to a tensor of order 4 to make it conform with the grammatical type $n^r\cdot s$ of the verb\footnote{Remember that in \textbf{FHilb} this would be a vector in $S^*\ten N\ten N^*\ten S$.} and the subsequent composition with the subject proceeds as:

\begin{equation}
  \footnotesize
  
\InputIfFileExists{./tikz/cpm-frobintr1.tikz}{}{\input{./tikz/cpm-frobintr1.tikz}}

  \normalsize
\end{equation}
\vspace{0.3cm}

Note that linear-algebraically this is nothing more than the point-wise multiplication of the two density matrices $\rho(s)\odot\rho(v)$, which is exactly the behaviour of the original Frobenius formulation on tensors acting on a single argument (\S\ref{sec:ditransitive}). Indeed, all the properties discussed in Chapter \ref{ch:frobverbs} that result from using Frobenius operators on simple meaning vectors are directly transferable to the new quantum setting. I will also examine the case of a transitive verb, the meaning of which can be defined as below:

\begin{equation}
   \ol{verb}_{TR} = \sum_i \rho(subj_i) \ten \rho(obj_i)
\end{equation}

\noindent with $i$ iterating through all occurrences of the verb in the training corpus, and $\rho(n)$ representing the density matrix produced for noun $n$. The application of the Copy-Subject model, then, will get the following form:

\vspace{-0.3cm}
\begin{equation}
  \footnotesize
  
\InputIfFileExists{./tikz/cpm-copysbj1.tikz}{}{\input{./tikz/cpm-copysbj1.tikz}}

  \normalsize
\end{equation}

From a linear algebraic perspective, we again faithfully recover the meaning of the original Copy-Subject model,\index{Copy-Subject model!on density matrices} since the density matrix of the subject contributes equally to the final result with the density matrix of the verb phrase:

\begin{equation}
  \rho_{CS} = \rho(s)\odot \op{Tr}(\rho(v)\circ(1_N\ten\rho(o)))
\end{equation}

Although I will not show it here, the computation proceeds similarly for the Copy-Object model\index{Copy-Object model!on density matrices} with the difference that the dimension of the result is now attached to the object loop. The formula is given below:

\begin{equation}
  \rho_{CO} = \op{Tr}(\rho(v)\circ(\rho(s)\ten 1_N)) \odot \rho(o)
\end{equation}

The topic of applying Frobenius algebras on density matrices is not by any means exhausted in this  section. In work that has been developed independently of (and about the same time with) the research presented in this chapter, Piedeleu \cite{piedeleu} explores an alternative non-commutative Frobenius algebra for a density matrix formulation coming from the work of Coecke and Spekkens \cite{coecke2012picturing}. His analysis suggests that, for certain cases, this non-commutative version seems to be better suited to the density matrix formalism, compared to the algebra presented in this section. Furthermore, there are many other structures that adhere to the definition of a Frobenius algebra in the context of a CPM construction;\index{Frobenius algebras!alternative structures for CPM}
this creates many opportunities for interesting combinations, the linguistic intuition of which still remains to be investigated.

\index{dagger Frobenius algebras|)}
\index{Frobenius algebras!dagger|)}
\index{density matrices!and Frobenius algebras|)}
\index{Frobenius algebras!on density matrices|)}

\section{Measuring ambiguity and entanglement}
\label{sec:entropy}

\index{ambiguity!measuring|(}

The quantum formulation offers to us a natural way to measure the degree of ambiguity expressed by a density matrix, in the form of \textit{Von Neumann entropy}.\index{Von Neumann entropy} For a density matrix $\rho$ with eigen-decomposition $\rho = \sum e_i \ket{e_i}\bra{e_i}$, this is given as:

\begin{equation}
  S(\rho) = -\op{Tr}(\rho\op{ln}\rho) = -\sum_i e_i \op{ln} e_i
  \label{equ:entropy}
\end{equation}

In the case of a pure state, entropy is always zero, while for a \textit{maximally mixed state}\index{maximally mixed states} it gets its maximum value, which is $\op{ln} D$, with $D$ denoting the dimension of the underlying Hilbert space. Intuitively, then, $S(\rho)$ provides an indication of the extent to which the density matrix $\rho$ deviates from a pure state. 

Recall that in our formalism a pure state corresponds to an \textit{unambiguous} word with a single definition. For density matrices representing words, then, Von Neumann entropy offers a measure of the degree of lexical ambiguity that those words express. Even more importantly, we have now a way to measure the degree of ambiguity in ``composite'' density matrices representing phrases or sentences. Returning to the example given in the introduction of this chapter, measuring the level of ambiguity in the sentence ``a fisherman was waiting by the bank'' should return a number much lower than the one we get from the sentence ``a man was waiting by the bank''. This can be an insightful tool towards further research that aims at investigating the presence of ambiguity in compositional settings, and the various ways it affects the result of composition.

\index{ambiguity!measuring|)}

\index{entanglement|(}
\index{density matrices!and entanglement|(}

Regarding our discussion in \S\ref{sec:entanglement}, Von Neumann entropy also provides a canonical way for measuring the level of entanglement in a density matrix representing a relational word. Take $\ket{\Psi}\in A\ten B$ to be the state of an \textit{unambiguous} relational word and $\rho_{AB} = \ket{\Psi}\bra{\Psi}$ the associated density matrix (expressing a pure state); then, the entanglement of $\rho_{AB}$ is defined as the Von Neumann entropy\index{Von Neumann entropy} of one of the two reduced matrices:

\vspace{-0.4cm}
\begin{equation}
  E(\ket{\Psi}) = S(\rho_A) = -\op{Tr}(\rho_A \op{ln} \rho_A) = -\op{Tr}(\rho_B \op{ln} \rho_B) = S(\rho_B)
\end{equation}

\noindent with $\rho_A=\op{Tr}_B(\rho_{AB})$ to be the partial trace of $\rho_{AB}$ over the basis of $B$, and vice versa for $\rho_B$. Note that this is zero only when the initial state $\ket{\Psi}$ is a pure state---which would lead to the separability problems\index{separability} discussed in \S\ref{sec:entanglement}. The form of a separable intransitive verb, for example, is the following:

\vspace{-0.4cm}
\begin{equation}
  \footnotesize
  
\InputIfFileExists{./tikz/separ-dm.tikz}{}{\input{./tikz/separ-dm.tikz}}

  \normalsize
\end{equation}
\vspace{0.2cm}

\noindent which, when composed with a subject noun, returns the following invariant outcome:

\vspace{-0.4cm}
\begin{equation}
  \footnotesize
  
\InputIfFileExists{./tikz/separ-dm1.tikz}{}{\input{./tikz/separ-dm1.tikz}}

  \normalsize
\end{equation}
\vspace{0.2cm}

Computing the level of entanglement in a density matrix expressing a mixed state (in our setting, an ambiguous word) is a much more involved process. A mixed state $\rho_{AB} \in A\ten B$ is separable if it can be factorized to the following form:

\begin{equation}
  \rho_{AB} = \sum_i p_i \rho^A_i\ten\rho^B_i
\end{equation}

\noindent where $p_i>0$ with $\sum_i p_i = 1$, and $\rho^A,\rho^B$ are states of the respective subsystems. Unfortunately, finding such a factorization is an NP-hard problem, for which there exists no single standard solution. Possibilities include \textit{distillable entanglement} and \textit{entanglement cost}, both of which are very difficult to compute. For a concise survey of various entanglement measures, I refer the reader to the study of Plenio amd Virmani \cite{plenio2007}.

\index{density matrices!and entanglement|)}
\index{entanglement|)}

\section{A demonstration}
\label{sec:amb-demo}

\index{density matrices!demonstration|(}
\index{demonstration of quantum model|(}

\newcommand{\vc}[2]{\left(\begin{smallmatrix}#1\\#2\end{smallmatrix}\right)}
\newcommand{\lv}[2]{\left(\begin{array}{c}#1\\#2\end{array}\right)}
\newcommand{\mm}[4]{\left(\begin{smallmatrix}#1 & #2\\#3 & #4 \end{smallmatrix}\right)}

In this section I would like to present a small concrete example that hopefully will help readers to get a better understanding of the model. I assume a two-dimensional space with basis $\vc{0}{1}$ associated with `finance' and $\vc{1}{0}$ associated with `water'. We will work with the following vectors created from a hypothetical corpus:
  
\begin{gather*}
  \ket{bank}_{f} = \vc{1}{8}~~~\ket{bank}_{r} = \vc{9}{2}~~~   
  \ket{river} = \vc{7}{1}~~~\ket{fish} = \vc{5}{2}~~~\ket{money} = \vc{2}{7}
\end{gather*}
  
Note that `bank' is an ambiguous word with two meanings, one related to finance and one to water, whereas all other words are assumed unambiguous with only one definition. Normalizing those vectors will give:

\begin{gather*}
  \ket{bank}_{f} = \vc{0.12}{0.99}~~~\ket{bank}_{r} = \vc{0.98}{0.22}~~~   
  \ket{river} = \vc{0.99}{0.14}\\ 
  \ket{fish} = \vc{0.93}{0.37}~~~\ket{money} = \vc{0.27}{0.96}
\end{gather*}

Let us now prepare a density matrix for the ambiguous word `bank', assigning equal probability to each one of the two possible meanings:
  
\begin{equation*}
  \rho_{bank} = \tfrac{1}{2} \ket{bank}_{f}\bra{bank}_{f} + \tfrac{1}{2} \ket{bank}_{r}\bra{bank}_{r} 
    = \mm{0.48}{0.17}{0.17}{0.52}
\end{equation*}

We might want to measure the level of ambiguity in $\rho_{bank}$. Computing the Von Neumann entropy\index{Von Neumann entropy} according to Eq. \ref{equ:entropy} gives:

\begin{equation*}
  S(\rho_{bank}) = -\op{Tr}(\rho_{bank}\op{ln}\rho_{bank}) = 0.64
\end{equation*}

Recall that the maximum entropy for a two-dimensional space is $\op{ln} 2 = 0.69$. So in our example `bank' is quite ambiguous, but not ``maximally'' ambiguous; in order to get a maximally mixed state, $\ket{bank}_f$ and $\ket{bank}_r$ should correspond to our basis vectors, in which case we get $\rho_{bank}=\mm{0.5}{0}{0}{0.5}$ and $S(\rho_{bank})=0.69$. 

Let us now proceed to composing a density matrix for the compound noun `river bank' using the Frobenius framework of \S\ref{sec:frob-density}. We will treat the (unambiguous) word `river' as an adjective, and we will compose it with the (ambiguous) word `bank' as follows:

\begin{equation*}
   \rho_{river~bank} = \ket{river}\bra{river} \odot \rho_{bank} = \mm{0.98}{0.14}{0.14}{0.02} \odot \mm{0.48}{0.17}{0.17}{0.52} = \mm{0.47}{0.02}{0.02}{0.01}
\end{equation*}

Measuring the entropy of the composite matrix gives $S(\rho_{river~bank})=0.40$, which directly reflects the fact that the meaning of the compound noun is much clearer than that of word `bank' alone, as expected. Furthermore, we would expect that `river bank' is semantically closer to `fish' than to `money'. Recall that for any two operators $\rho_1$ and $\rho_2$ representing words, the trace of $\rho_1\rho_2$ gives a number that can be interpreted as the probability of one word occurring in the vicinity of the other. Applying this notion of similarity between `river bank' and `fish' results in the following computation:


\begin{equation*}
  \op{Tr}(\rho_{river~bank} \circ \left(\ket{fish}\bra{fish})\right)
     =  \op{Tr}\left[\mm{0.47}{0.02}{0.02}{0.01}\mm{0.86}{0.35}{0.35}{0.14}\right] = 0.43
\end{equation*}

A similar computation for the similarity of `river bank' and `money' returns 0.06. Now assume an ambiguous vector $\ov{bank}=\ov{bank}_f+\ov{bank}_r=\vc{10}{10}$, and let us calculate the similarities between the same pairs of words when using directly the distributional vectors:

\begin{gather*}
  \op{cos}(\ov{river}\odot\ov{bank},\ov{fish}) = \op{cos}\left[\vc{70}{10},\vc{5}{2}\right] = 0.97 \\
  \op{cos}(\ov{river}\odot\ov{bank},\ov{money}) = \op{cos}\left[\vc{70}{10},\vc{2}{7}\right] = 0.41 \\
\end{gather*}

Evidently, these numbers are not so sensible as the ones returned by the density matrix model. Although this toy example cannot be possibly used as an indication for the relative performance of the models, in general it is reasonable for one to expect that the richer representation imposed by the quantum model would be able to reflect better the semantic relationships between the various words, and in turn, of the various phrases and sentences after word composition.

\index{demonstration of quantum model|)}
\index{density matrices!demonstration|)}

\section{A more convenient graphical notation}

\index{density matrices!graphical notation|(}
\index{graphical language!for density matrices|(}

For complex sentences, the hemi-circular diagrammatic notation imposed by the special requirements of the CPM structure will quickly become inconvenient and non-informative. Following a suggestion by Bob Coecke, in this section I will show how we can almost achieve our neat original graphical representation by a change of perspective. Recall that this hemi-circular form of the diagrams is a direct consequence of the fact that the tensor product of completely positive maps imposes the embedding of one object into another; as a result, wires that belong to the same object end up to be far apart. We can avoid this problem by introducing a third dimension in our diagrammatic notation, which will allow us to represent the operators in the following manner:

\begin{equation}
  \footnotesize
  
\InputIfFileExists{./tikz/boldwire1.tikz}{}{\input{./tikz/boldwire1.tikz}}

  \normalsize
\end{equation}
\vspace{0.3cm}

Drawing in three dimensions allows the luxury of placing objects and wires \textit{in front} of other objects, as opposed to the strict side-by-side juxtaposition we were restricted until now. Under this new perspective, the two copies of wires that constitute the $\eta$- and $\epsilon$-maps are not any more juxtaposed as before, but they overlap as follows (imagine that one copy has been placed in front of the other):

\begin{equation}
  
\InputIfFileExists{./tikz/boldwire2.tikz}{}{\input{./tikz/boldwire2.tikz}}

\end{equation}

The same is true for the Frobenius operators:

\begin{equation}
  \footnotesize
  
\InputIfFileExists{./tikz/frob-density-new.tikz}{}{\input{./tikz/frob-density-new.tikz}}

  \normalsize
\end{equation}
\vspace{0.2cm}

Composition in our new notation takes this form:

\begin{equation}
  \footnotesize
  
\InputIfFileExists{./tikz/boldwire3.tikz}{}{\input{./tikz/boldwire3.tikz}}

  \normalsize
\end{equation}
\vspace{0.2cm}

At this point it is not difficult to use a little visual abstraction, and essentially recover our previous notation:

\begin{equation}
\label{fig:boldwire}
  \footnotesize
  
\InputIfFileExists{./tikz/boldwire4.tikz}{}{\input{./tikz/boldwire4.tikz}}

  \normalsize
\end{equation}
\vspace{0.2cm}

A diagram such as (\ref{fig:boldwire}) above reflects the nature of the underlying model to a much higher degree than the original notation. First, it emphasizes the most prominent characteristic of the new setting, which is the fact that everything is \textit{doubled}. More importantly, it allows us to continue thinking about words as distinct systems that interact with each other through processes, a quality that is lost in the complex labyrinthine figures similar to that in (\ref{fig:cpm-trans}). 

\index{graphical language!for density matrices|)}
\index{density matrices!graphical notation|)}


\section{Discussion}
\label{sec:quantum-discussion}

The quantum formulation presented in this chapter provides perhaps the most natural way to incorporate the notions of homonymy and polysemy in the categorical framework of \cite{Coeckeetal}. The concept of an ambiguous word vector has been replaced by that of a density matrix, which expresses a probability distribution over the potential meanings of a word. While a density matrix expresses our uncertainty regarding the potential meaning of a homonymous word, smaller semantic deviations in meaning that correspond to cases of polysemy are left to be resolved by superpositions taking place over the pure states. In general, the probabilistic view imposed by the density matrix formalism can be seen as a step towards more robust tensor representations for relational words and, in turn, more effective compositional models. 

Furthermore, the new perspective creates opportunities for novel interesting research. In \S\ref{sec:entropy} we briefly discussed the notion of Von Neumann entropy, and how this can be used to quantify the level of ambiguity in a single word or a composite text segment. This subject is closely related to another interesting question: in what ways can \textit{entanglement} affect the result of composition? Or, to put things into a more practical perspective: How can we exploit this phenomenon in order to design more effective CDMs? Another very interesting characteristic of the new model is that it can accommodate more than one notion of ambiguity; this is a direct consequence of the fact that the CPM construction can be applied to \textit{any} dagger compact closed category, even to \textbf{CPM(FHilb)} itself.\index{density matrices!recursive application} In the new category, words will be represented by density matrices of density matrices; imagine, for example, that additionally to the first level of mixed states which represent a probability distribution over possible meanings of a word, we would like a second level that expresses our uncertainty regarding the syntactical use of that word (for example, `fly' can be a noun or a verb).

The practical aspects of the material presented here will be the subject of Chapter \ref{ch:wsdexp}. Specifically, in \S\ref{sec:disamb} and \S\ref{sec:wsdtensor} we will see how to apply standard word sense induction and disambiguation techniques in order to create unambiguous tensors representing each distinct meaning of a homonymous word. Putting together those tensors into a statistical ensemble that can be used for preparing a density matrix for the specific word is straightforward. However, the space complexity problems posed by doubling\index{doubling of dimensions}\index{density matrices!doubling of dimensions} the order of tensors (briefly discussed in \S\ref{sec:frob-density}) make a large-scale evaluation of the model not possible for the moment. In the second part of Chapter \ref{ch:wsdexp} we will conveniently side-step this difficulty, by evaluating our generic hypothesis regarding the separation of homonymy from polysemy directly on Eq. \ref{equ:maindis}---that is, by simplifying the process so that the disambiguation of a word takes place on the spot given the context, and the most probable meaning vector for each word is selected for the composition in a separate step. The results strongly suggest that the assumption upon which the quantum model was built is correct; a direct practical evaluation of the model would require the application of the Frobenius framework as proposed in \ref{sec:frob-density}, state vectors of reduced dimensionality compared to standard practice, and a carefully designed task capable of demonstrating the advantages of the formulation.

I would like to close this chapter by pointing out that the research presented here is not the only one that uses density matrices for linguistic purposes. In the context of the categorical framework of Coecke et al. \cite{Coeckeetal}, Piedeleu \cite{piedeleu} independently developed a model that handles lexical ambiguity in about the same way as the current exposition; this parallel development led to a subsequent joint paper, co-authored with Piedeleu, Coecke and Sadrzadeh \cite{piedeleu2015}. In yet another parallel development, Balk{\i}r \cite{balkir} uses a different form of density matrices in order to provide a similarity measure that can be used for evaluating hyponymy-hypernymy relations. Blacoe et al. \cite{blacoe-quantum} describe a distributional (but not compositional) model of meaning based on density matrices created by grammatical dependencies. Finally, and in a more generic level not directly associated to density matrices, the application of ideas from quantum theory to language has been proved in the past a very popular field of research. Some interesting examples include the work of Bruza et al. \cite{bruza2009} and Widdows \cite{widdows2003orthogonal}.

\mbox{}\newpage

\part{Practice}
\label{prt:practice}
\mbox{}\newpage
\chapter{Evaluating the Frobenius Models}
\label{ch:frobexp}

\begin{chabstract}
We will now turn our attention to investigating the theoretical models presented in Part \ref{prt:theory} from a practical perspective, providing insights about concrete implementations and discussing the inevitable difficulties that this transition from theory to practice poses. The aim of the current chapter is to provide a generic evaluation of the Frobenius framework of Chapter \ref{ch:frobverbs}, whereas Chapter \ref{ch:wsdexp} deals with the more extensive topic of implementing and evaluating a prior disambiguation step in CDMs. Material is based on \cite{kartsaklis2014,kartsaklis:2013:EMNLP,kartsaklis2012}.
\end{chabstract}

\noindent
Compositional-distributional models of meaning constitute a relatively new concept in NLP research, and they inevitably lack any established evaluation procedures.\index{CDMs!evaluation procedures}\index{evaluation of CDMs} In many cases (especially in deep learning settings), the CDM is trained in a way that maximizes the performance with a specific task in mind, e.g. paraphrase detection \cite{socher2011} or sentiment analysis \cite{socher2012}, and is evaluated accordingly. This means that a vector representing the meaning of a sentence is much less useful for tasks other than the one it was trained for. As claimed before, though, one of the goals of this thesis is to study CDMs in a level more generic than that, independently of specific training procedures and, inevitably, of specific tasks; therefore, what we seek is the most ``canonical'' way to evaluate a CDM, which has to be detached from any assumptions regarding the usage of the produced sentence vectors. For a model whose purpose is to assign a certain form of meaning to a sentence, it is reasonable to claim that this task must be based on a direct evaluation of the semantic similarity between pairs of sentences or phrases.\index{sentence similarity, for evaluating CDMs}\index{CDMs!and sentence similarity}

The above reasoning presupposes something important: the existence of a set of gold-standard similarity scores, against which the computer performance can be evaluated. Usually these gold-standard scores come from a number of human annotators, the duty of which is to provide a number that, according to their opinion, reflects the semantic similarity of two sentences in a specific scale (e.g. from 1=dissimilar to 7=identical). The performance of the computer model, then, can be computed as some form of correlation\index{correlation} (Spearman's $\rho$ or Pearson's $r$) between the list of computer scores and the list of human annotations. This form of task and specific variations of it will be one of our main evaluation tools in this part of the thesis.


The models that are going to participate in the experiments are presented in Table \ref{tbl:models}. The Relational model corresponds to the non-scalable approach of \cite{GrefenstetteThesis2013}, where a transitive sentence is represented by a matrix. The table also includes a non-compositional baseline (Only Verbs), in which the meaning of a sentence or phrase is just the distributional vector of the head verb. Furthermore, I test the performance of the two vector mixture models of \cite{Lapata}, in which the meaning of a sentence is computed as the point-wise addition or multiplication of the vectors for the words therein.\index{vector mixtures!as baselines} 

\renewcommand{\arraystretch}{1.3}
\begin{table}[h!]
\begin{center}
\small
\begin{tabular}{lc}
  \hline
  \textbf{Model} & \textbf{Formula} \\
  \hline\hline
  Only Verbs & $\ov{verb}$ \\
  \hline
  Additive & $\ov{subj} + \ov{verb} + \ov{obj}$ \\
  Multiplicative & $\ov{subj} \odot \ov{verb} \odot \ov{obj}$ \\
  \hline
  Relational & $\ol{verb} \odot (\ov{subj} \ten \ov{obj})$\\
  Copy-Subject & $\ov{subj} \odot (\ol{verb} \times \ov{obj})$ \\
  Copy-Object & $\ov{obj} \odot (\ol{verb}^{\mathsf{T}} \times \ov{subj})$ \\
  Frobenius additive & Copy-Subject $+$ Copy-Object\\
  \hline
\end{tabular}
\end{center}
\caption{Compositional models for transitive verbs.}
\label{tbl:models}
\end{table}
\renewcommand{\arraystretch}{1.0}

\section{Head verb disambiguation}
\label{sec:kintsch}

\index{head verb disambiguation|(}
\index{evaluation of CDMs!head verb disambiguation|(}

In one of the most common tasks for testing compositionality in distributional models of meaning, the sentence similarity process described in the introduction of this chapter is laid out around ambiguous verbs. Originally introduced by Kintsch \cite{kintsch2001predication}, and later adopted by Mitchell and Lapata \cite{Lapata} and others, the main question that the task aims to answer is the following: Given an ambiguous verb such as `file', to what extent can the presence of an appropriate context disambiguate its intended meaning? The context (a subject/object pair for a transitive verb) is composed with two ``landmark'' verbs corresponding to different meanings (for example, `smooth' and `register') in order to create simple sentences. Then, each one of these sentences is compared with the sentence produced by composition of the main verb (`file') with the same context. Therefore we have two sentence pairs of the following form:

\begin{exe}
  \ex \begin{xlist}
    \item\label{ex:pair1} Woman files application / Woman registers application
    \item\label{ex:pair2} Woman files application / Woman smooths application
  \end{xlist}  
\end{exe}

The assumption behind this methodology is that a good compositional model should be able to reflect the fact that the sentences of pair (\ref{ex:pair1}) are closer to each other than the sentences of pair (\ref{ex:pair2}). In other words, we have a task that aims to disambiguate the head verb of a sentence, disguised as a sentence similarity task. Following a tradition that started in \cite{GrefenSadr1} and aims for categorical models to be evaluated on at least one form of this task, I will start my experiments by testing the models of Table \ref{tbl:models} on the dataset of transitive sentences introduced by Grefenstette and Sadrzadeh in \cite{GrefenSadr1} (to which I will refer as G\&S 2011 for the rest of the thesis). The size of this dataset is 200 pairs of sentences (10 main verbs $\times$ 2 landmarks $\times$ 10 contexts). 

\index{evaluation of CDMs!head verb disambiguation|)}
\index{head verb disambiguation|)}

\subsection{Semantic space}
\label{sec:semspace}

\index{semantic spaces!construction|(}

For this task\footnote{The experiments presented in this thesis took place over the course of three years, and inevitably the reader will notice some inconsistency between the various settings from task to task.}
I train my vectors using ukWaC \cite{ukwac},\index{ukWaC corpus} a corpus of English text with 2 billion words (100m sentences). The basis of the vector space consists of the 2,000 content words (nouns, verbs, adjectives, and adverbs) with the highest frequency, excluding a list of stop words. Furthermore, the vector space is lemmatized and unambiguous regarding syntactic information; in other words, each vector is uniquely identified by a $\langle$lemma,pos-tag$\rangle$ pair, which means for example that `book' as a noun and `book' as a verb are represented by different meaning vectors. The weights of each vector are set to the ratio of the probability of the context word given the target word to the probability of the context word overall, as follows:

\begin{equation}
   v_i(t) = \frac{p(c_i|t)}{p(c_i)} = \frac{\text{count}(c_i,t) \cdot \text{count}(total)}{\text{count}(t) \cdot \text{count}(c_i)}
\label{equ:weights}
\end{equation}

\noindent where $\text{count}(c_i,t)$ refers to how many times $c_i$ appears in the context of $t$ (that is, in a five-word window on either side of $t$) and $\text{count}(total)$ is the total number of word tokens in the corpus. The choice of the particular weighting scheme (which is just PMI without the logarithm) was based on the study of Bullinaria and Levy \cite{bullinaria2007}, where it has been found very effective in various semantic similarity tasks.

\index{semantic spaces!construction|)}

%

\subsection{Results}

Each entry of the G\&S 2011 dataset has the form $\langle$subject, verb, object, high-sim landmark, low-sim landmark$\rangle$. The context is combined with the verb and the two landmarks, creating three simple transitive sentences. The main-verb sentence is paired with both the landmark sentences, and these pairs are randomly presented to human evaluators, the duty of which is to evaluate the similarity of the sentences within a pair in a scale from 1 to 7. The scores of the compositional models are the cosine distances between the composite representations of the sentences of each pair. As an overall score for each  model, I report its Spearman's $\rho$ correlation with the human judgements.\footnote{Unless stated otherwise, in the correlation-based experiments of this thesis each human judgement is treated as a separate entry in the correlation list. Note that this in general results in much lower numbers than first averaging human judgements, and then getting the correlation with computer scores. The relative performance of a model though with regard to other models is almost identical in both methodologies.} The results are shown in Table \ref{tbl:res-gsfrob}.


\begin{table}[h]
\begin{center}
\small
\begin{tabular}{l|c}
\hline
\textbf{Model} & \textbf{Spearman's $\rho$} \\
\hline\hline
Only verbs     & 0.198 \\
\hline
Additive       & 0.127 \\
Multiplicative & 0.137 \\
\hline
Relational	   & 0.219 \\
Copy-Subject   & 0.070 \\
Copy-Object    & \textbf{0.241} \\
Frobenius additive  & 0.142 \\ 
\hline
Human agreement & 0.600 \\
\hline
\end{tabular} 
\end{center} 
\normalsize
\caption[Head verb disambiguation results on the G\&S 2011 dataset.]{Head verb disambiguation results on the G\&S 2011 dataset. Difference between Copy-Object and Relational is s.s. with $p<0.001$.}
\label{tbl:res-gsfrob}
\end{table}

The most successful model is the Copy-Object Frobenius variation,\index{Copy-Object model!performance in head verb disambiguation task} which suggests two things. First, it provides an indication that the object of an ambiguous verb can be more important for disambiguating that verb than the subject. Intuitively, we can imagine that the crucial factor for disambiguating between the verbs ``write'' ``publish'' and ``spell'' is more the object than the subject: a book or a paper can be both published and written, but a letter or a shopping list can only be written. Similarly, a word can be spelled and written, but a book can only be written. The subject in all these cases is not so important. Furthermore, the superiority of the Copy-Object model over the Relational model (in which a sentence is represented as a matrix) means that the reduction of the dimensions imposed by the Frobenius operators has a positive effect, producing a compact vector that reflects better the semantic relationships between the various sentences. As a last comment, most of the Frobenius models perform better than the two vector mixtures, which suffer from the additional disadvantage that they completely ignore the word order.

\section{Verb phrase similarity}
\label{sec:phrase-sim}

\index{verb phrase similarity|(}
\index{evaluation of CDMs!verb phrase similarity|(}

The head verb disambiguation task of \S \ref{sec:kintsch} implicitly correlates the strength of the disambiguation effect that takes place during composition with the quality of composition; essentially it assumes that the stronger the disambiguation, the better the compositional model that produced this side-effect. Unfortunately, the extent to which this assumption is valid or not is not quite clear. 

In order to avoid this problem, this second experiment is based on a phrase similarity task introduced by Mitchell and Lapata \cite{lapata2010} as part of a larger test-bed. In contrast to the task of \S \ref{sec:kintsch}, this one does not involve any assumptions about disambiguation, and thus it seems like a more genuine test of models aiming to provide appropriate phrasal or sentential semantic representations; the only criterion is the degree to which these models correctly evaluate the \textit{similarity} between pairs of sentences or phrases. I work on the verb phrase part of the dataset (which from now on I will denote as M\&L 2010), consisting of 72 short verb phrases (verb-object structures). These 72 phrases have been paired in three different ways to form groups exhibiting various degrees of similarity: the first group contains 36 pairs of highly similar phrases (e.g. \textit{produce effect}-\textit{achieve result}), the pairs of the second group are of medium similarity (e.g.  \textit{write book}-\textit{hear word}), while a last group contains low-similarity pairs (\textit{use knowledge}-\textit{provide system})---a total of 108 verb phrase pairs. The task is again to compare the similarity scores given by the various models for each phrase pair with those of human annotators. 

\index{evaluation of CDMs!verb phrase similarity|)}
\index{verb phrase similarity|)}

\subsection{Computing the meaning of verb phrases}
\label{sec:verbphrase}

\index{meaning!of verb phrases|(}
\index{verb phrases, meaning|(}

Before proceeding to the results, I will first work out the closed form formulas giving the meaning of a verb phrase for the various categorical models. I remind the reader that in the canonical case of a full (order-3) verb tensor, the verb phrase meaning is simply given by:

\begin{equation}
  \ol{vp} = \ol{verb} \times \ov{obj}
\end{equation}

\noindent
which returns a matrix. For the Frobenius models, let us assume a verb matrix $\ol{verb}=\sum_{ij}v_{ij}\ov{n_i}\ten \ov{n_j}$ and an object vector $\ov{obj} = \sum_j o_j \ov{n_j}$. Then, for the Relational case we have the following:

\begin{eqnarray}
  (1_{N\ten N \ten N} \ten \epsilon^l_N)\circ (\Delta_N\ten \Delta_N \ten 1_N)
  (\ol{verb} \ten \ov{obj})& = & \sum\limits_{ij} v_{ij} \langle \ov{n_j}|\ov{obj} \rangle \ov{n_i} \ten \ov{n_i} \ten \ov{n_j} \nonumber \\ 
  & = & \sum\limits_{ij} v_{ij} o_j \ov{n_i} \ten \ov{n_i} \ten \ov{n_j}
\end{eqnarray} 

\noindent
which is just the vector $\ol{verb} \times \ov{obj}$ expanded to a cube. When one follows the Copy-Subject model,\index{Copy-Subject model!on verb phrases} the verb phrase meaning is again the same vector, but this time expanded to a matrix:

\begin{eqnarray}
\label{equ:copysbj-vp}
(1_{N\ten N} \ten \epsilon^l_N) \circ (\Delta_{N}\ten 1_N\ten 1_N)(\ol{verb} \ten \ov{obj}) & = &
\sum\limits_{ij} v_{ij} \langle \ov{n_j}|\ov{obj} \rangle \ov{n_i} \ten \ov{n_i} \nonumber \\ 
 & = & \sum\limits_{ij} v_{ij} o_j \ov{n_i} \ten \ov{n_i} 
\end{eqnarray}

Finally, the Copy-Object case returns a matrix as follows:\index{Copy-Object model!on verb phrases}

\begin{eqnarray}
  (1_{N\ten N} \ten \epsilon^l_N) \circ (1_N\ten\Delta_N\ten 1_N)(\ol{verb} \ten \ov{obj}) & = & \sum\limits_{ij} v_{ij} \langle \ov{n_j}|\ov{obj} \rangle \ov{n_i} \ten \ov{n_j} \nonumber \\ 
  & = & \sum\limits_{ij} v_{ij} o_j \ov{n_i} \ten \ov{n_j}
\end{eqnarray} 

Note that this is essentially the point-wise product (also known as Hadamard product) of the verb matrix with a matrix whose rows correspond to the vector of the object:

\singlespace
\begin{equation}
  \left(\begin{array}{ccc} v_{11} & \hdots & v_{1n} \\
         \vdots & \ddots & \vdots \\
         v_{n1} & \hdots & v_{nn} \end{array}\right)
         \odot
   \left(\begin{array}{ccc} o_{1} & \hdots & o_{n} \\
         \vdots & \ddots & \vdots \\
         o_{1} & \hdots & o_{n} \end{array}\right) 
\end{equation}
\onehalfspace

\index{verb phrases, meaning|)}
\index{meaning!of verb phrases|)}

\subsection{Results}
\label{sec:verbphrase-results}

For this experiment I used the same semantic space as in \S\ref{sec:semspace}. This time (and in the absence of any ambiguity) the results presented in Table \ref{tbl:res-mlfrob} show top performance from the Copy-Subject variation. Recall from Eq. \ref{equ:copysbj-vp} that in this model a vector is prepared for each verb phrase ``categorically'', by matrix-multiplying the verb matrix with the object vector; on the other hand, in the Copy-Object model the object vector is merged (by point-wise multiplication) with the object dimension of the verb matrix, as follows:

\begin{equation}
  \footnotesize
  
\InputIfFileExists{./tikz/frob-vp.tikz}{}{\input{./tikz/frob-vp.tikz}}

  \normalsize
\end{equation}
\vspace{0.2cm}

We can think of the matrix produced by the Copy-Object case as representing the meaning of an \textit{incomplete sentence}, which can take definite shape only after the introduction of the missing subject. From this perspective, the better performance of the Copy-Subject model is justified: the verb phrase here has the role of a \textit{complete theme}, so the categorically prepared composite vector reflects the meaning of the verb phrase in a more self-contained way.\index{Copy-Subject model!performance on verb phrase similarity task}

\begin{table}[h]
\begin{center}
\small
\begin{tabular}{l|c}
\hline
\textbf{Model} & \textbf{Spearman's $\rho$} \\
\hline\hline
Only verbs     & 0.310 \\
\hline
Additive       & 0.291 \\
Multiplicative & 0.315 \\
\hline
Copy-Subject/Relational	   & \textbf{0.340} \\
Copy-Object    & 0.290 \\
Frobenius additive       & 0.270 \\
\hline
Human agreement & 0.550 \\
\hline
\end{tabular} 
\end{center} 
\normalsize
\caption[Verb phrase similarity results on the M\&L 2010 dataset.]{Verb phrase similarity results on the M\&L 2010 dataset. Difference between Copy-Subject and Multiplicative is s.s. with $p<0.001$.}
\label{tbl:res-mlfrob}
\end{table}

\section{Classification of definitions to terms}
\label{sec:term-definition}

\index{term-definition classification|(}
\index{evaluation of CDMs!term-definition classification|(}

The sentence comparison methodology I adopted in this chapter suffers from an important problem: \textit{subjectivity}. Humans rarely agree on their judgements, whereas in many cases their evaluations differ drastically. In the paper that introduced the Frobenius models \cite{kartsaklis2012} (joint work with Sadrzadeh and Pulman), I and my co-authors propose a way to cope with this difficulty. The suggestion was based on the idea that the meaning of certain phrases is expected to be close to the meaning of certain words, as in the case of a dictionary entry: the meaning vector for a definition, such as ``woman who rules a country'', must be close to the word vector of the term (`queen'), and far away from vectors of other irrelevant words (e.g. `window'). A setting like this eliminates the necessity\index{evaluation of CDMs!eliminating the need for gold-standard scores} of gold-standard scores, since the evaluation can be approached as a normal classification task: each term is assigned to the closest definition according to the cosine distance of their vectors, and the result is expressed by the usual means of information retrieval: precision/recall and F-score, or accuracy. This idea also appears in the work of Turney \cite{turney2012}, where bigrams corresponding to noun compounds (e.g. `dog house') are compared with single nouns (`kennel').

In this section I describe two relevant experiments which were based on a dataset I created in collaboration with M. Sadrzadeh.\footnote{This and all other datasets introduced in the present thesis are available at \texttt{http://www.cs. ox.ac.uk/activities/compdistmeaning/}.} The dataset consists of 112 terms (72 nouns and 40 verbs) that have been extracted from Oxford Junior Dictionary \cite{OxfordJun} together with their main definitions. For each term we added two more definitions, either by using  entries of WordNet for the term or by simple paraphrase of the main definition using a thesaurus, getting a total of three definitions per term. In all cases a definition for a noun term is a noun phrase, whereas the definitions for the verb terms consist of verb phrases. A sample of the dataset entries can be found in Table \ref{tbl:dfnsample}. 

\begin{table}[h!]
\begin{center}
\footnotesize
\begin{tabular}{l|lll}
 \hline
 \textbf{Term}  & \textbf{Main definition} & \textbf{Alternative definition 1} & \textbf{Alternative definition 2} \\
 \hline\hline
 blaze & large strong fire & huge potent flame & substantial heat \\
 husband & married man & partner of a woman & male spouse \\
 foal  & young horse & adolescent stallion & juvenile mare \\
 horror & great fear & intense fright & disturbing feeling \\
 \hline
 apologise & say sorry & express regret or sadness & acknowledge shortcoming or failing \\
 embark    & get on a ship & enter boat or vessel & commence trip \\
 vandalize & break things & cause damage & produce destruction \\
 regret    & be sad or sorry & feel remorse & express dissatisfaction \\
 \hline 
\end{tabular}
\caption[A sample of the dataset for the term/definition classification task.]{A sample of the dataset for the term/definition classification task (noun terms at the top part, verb terms at the lower part).}
\label{tbl:dfnsample}
\end{center}
\normalsize
\end{table}

As is evident from the samples, besides the elimination of gold-standard scores the specific setting has another important advantage for our purposes: abandoning the restricted syntactic structure of the previous experiments provides us the opportunity to apply in practice a number of suggestions from Chapter \ref{ch:extend} regarding the treatment of functional words. Specifically, in order to create composite vectors for the definitions I prepare tensors for prepositions (\S\ref{sec:preposition}), phrasal verbs (\S\ref{sec:phrasal-verbs}), and conjunctions (\S\ref{sec:coordination}), as described in the referred sections.

\index{meaning!of verb phrases|(}
\index{verb phrases, meaning|(}

Note that the nature of the experiment imposes a treatment of verb phrases different than that of \S\ref{sec:verbphrase}. Since now our goal is to compare the meaning of phrases with that of words, as expressed by their distributional vectors, both noun- and verb phrases must live in our basic vector space $N$. Recall from the discussion in Chapter \ref{ch:framework} that this is the case only for atomic types, which means that now, not only a noun phrase, but also a verb phrase must be treated accordingly. Assuming the atomic type $j$ for a verb phrase, a verb that expects a noun at object position in order to return a verb phrase gets the type $j\cdot n^r$, which means that its tensor is a matrix in $N\ten N$. Following the argument summing procedure, we can construct these by summing over all distributional vectors of objects with which the verb occurs in the corpus, and then expand this vector to a matrix as usual; that is, 

\begin{equation}
 \overline{verb} = \Delta\left(\sum_i \ov{obj_i}\right)
\end{equation} 

\index{verb phrases, meaning|)}
\index{meaning!of verb phrases|)}

This is referred to as the Frobenius model, and is compared with the additive and multiplicative models. The semantic space used in this task follows exactly the setting of \S\ref{sec:semspace}, with the important difference that the training was based on the BNC corpus,\index{British National Corpus (BNC)} which is magnitudes smaller than ukWaC.\index{ukWaC corpus}

In the first experiment, each term corresponds to a distinct class; in order to classify a definition to a specific term, we compare the composite vector of that definition with the distributional vectors of all terms in the dataset, and we select the term that gives the highest similarity. The results are reported in Table \ref{tbl:dfnresults} as the weighted average of F1-scores. The performance on the noun terms is evaluated separately from the performance on the verb terms, since a mixing of the two sets would be inconsistent.

\begin{table}[h]
\begin{center}
\small
\begin{tabular}{l|ccc||ccc}
 \hline
     & \multicolumn{3}{c||}{\textbf{Nouns}} & \multicolumn{3}{|c}{\textbf{Verbs}} \\
   \textbf{Model} & P & R & F1 & P & R & F1 \\ 
   \hline\hline
 Additive & 0.21 & 0.17 & 0.16 & 0.28 & 0.25 & 0.23 \\
 Multiplicative & 0.21 & 0.22 & 0.19 & 0.31 & 0.30 & 0.26 \\
 \hline 
 Frobenius & 0.22 & 0.24 & \textbf{0.21} & 0.32 & 0.28 & \textbf{0.27} \\
 \hline 
\end{tabular}
\caption{Results of the term/definition comparison task.}
\label{tbl:dfnresults}
\end{center}
\normalsize
\end{table}

The Frobenius model delivers again the best performance, although the difference from the multiplicative model in the case of verbs is small. All models perform better on the verb terms than the noun part of the dataset, yet in general F-scores tend to be low. In order to get a better insight regarding the potential of the model, we perform a second experiment where the direction of comparison is reversed: now we compare each term with every \textit{main} definition, constructing a list of definitions ranked by similarity, then examine the position of the correct definition inside this list. The detailed results are shown in Table \ref{tbl:dfnresults01}.

\begin{table}[h]
\begin{center}
\small
\begin{tabular}{l||l|ccc||ccc}
 \hline
  & & \multicolumn{3}{c||}{\textbf{Multiplicative}} & \multicolumn{3}{|c}{\textbf{Frobenius}} \\
  & \textbf{Rank} & Count & \% & \%* & Count & \% & \%* \\ 
 \hline\hline
  \multirow{4}{*}{\textbf{Nouns}}
  & 1     & 26 & 36.1 & 36.1  & 25 & 34.7 & 34.7 \\
  & 2-5   & 20 & 27.8 & 63.9  & 22 & 30.6 & 65.3 \\
  & 6-10  & 11 & 15.3 & 79.2  & 5  & 6.9  & 72.2 \\
  & 11-72 & 15 & 20.8 & 100.0 & 20 & 27.8 & 100.0\\
 \hline  
 \multirow{4}{*}{\textbf{Verbs}}
  & 1     & 15 & 37.5 & 37.5  & 8  & 20.0 & 20. 0\\
  & 2-5   & 10 & 25.0 & 62.5  & 13 & 32.5 & 52.5 \\
  & 6-10  & 6  & 15.0 & 77.5  & 4  & 10.0 & 62.5 \\
  & 11-40 & 9  & 22.5 & 100.0 & 15 & 37.5 & 100.0 \\
 \hline  
\end{tabular}
\caption{Results of the term/definition comparison task based on the rank of the main definition.}
\label{tbl:dfnresults01}
\end{center}
\normalsize
\end{table}

\vspace{-0.3cm}
This new perspective shows that for the noun-term set the Frobenius model returns the correct main definition in 25 of the 72 cases (34.7\%), whereas in 47 cases (65\%) the correct definition is in the top-five list for that term. The multiplicative model performs similarly, and better for the verb-term set. Furthermore, an error analysis revealed that some of the ``misclassified'' cases can also be considered as somehow ``correct''. For example, the definition originally assigned to the term `jacket' was `short coat'; however, the system ``preferred'' the definition `waterproof cover', which is also correct.  Some other interesting cases are presented in Table \ref{tbl:dfntbl}. 

\begin{table}[h]
\begin{center}
\small
\begin{tabular}{c|cc}
 \hline
 \textbf{Term}  & \textbf{Original definition} & \textbf{Assigned definition} \\
 \hline\hline
 rod    & fishing stick & round handle \\
 jacket & short coat & waterproof cover \\
 mud    & wet soil   & wet ground \\
 labyrinth & complicated maze & burial chamber\\
 \hline
\end{tabular}
\caption{A sample of ambiguous cases where the model assigned a different definition than the original.}
\label{tbl:dfntbl}
\end{center}
\normalsize
\end{table}

\index{evaluation of CDMs!term-definition classification|)}
\index{term-definition classification|)}

\section{Discussion}
\label{sec:frobexp-discussion1}

The experimental work presented in this chapter suggests that the Frobenius setting forms a robust implementation basis for the categorical framework of \cite{Coeckeetal}. Compared with the original implementation of \cite{GrefenSadr1} (Relational model), it improves the results while at the same time overcomes its shortcomings, which all stemmed from the fact that logical and concrete types did not match. Furthermore, this is despite the fact that in the original model a sentence is assigned a richer representation, with exponentially higher number of parameters. 

For consistency reasons, the results presented in this chapter were selected from work that uses the same experimental setting with the research I present in Chapter \ref{ch:wsdexp} (specifically, the experiments in \S\ref{sec:kintsch}, \S\ref{sec:phrase-sim} and \S\ref{sec:wsdfrob} follow the setting of \cite{kartsaklis:2013:EMNLP}). However, it should be noted that the same general performance pattern between the models has been verified in many other cases. For example, one parameter that can drastically affect the performance of a CDM is the choice of the vector space. In a recent comparative study I jointly conducted with Milajevs, Sadrzadeh and Purver \cite{milajevs2014}, we showed that neural word representations (briefly discussed in \S\ref{sec:neural-embeddings}) present performance superior to that of co-occurrence vectors for tensor-based settings and tasks similar to those I use in the current chapter. Using a vector space\footnote{The vector space is provided as part of the {\small \texttt{word2vec}} toolkit, which can be found online at {\small\texttt{https://code.google.com/p/word2vec/}}.} trained with the skip-gram model of Mikolov et al. \cite{mikolov2013distributed}, for example, we report results for the G\&S 2011 task as shown in Table \ref{tbl:res-gsfrobnn}.\index{head verb disambiguation!with neural embeddings}\index{neural word embeddings!and head verb disambiguation}

\begin{table}[h!]
\begin{center}
\small
\begin{tabular}{l|c}
\hline
\textbf{Model} & \textbf{Spearman's $\rho$} \\
\hline\hline
Only verbs     & 0.107 \\
\hline
Additive       & 0.149 \\
Multiplicative & 0.095 \\
\hline
Relational	   & 0.362 \\
Copy-Subject   & 0.131 \\
Copy-Object    & \textbf{0.456} \\
Frobenius additive  & 0.359 \\ 
\hline
Human agreement & 0.600 \\
\hline
\end{tabular} 
\end{center} 
\normalsize
\caption[Results on the G\&S 2011 dataset using neural embeddings.]{Results on the G\&S 2011 dataset using neural embeddings. Difference between Copy-Object and Relational is s.s. with $p<0.05$. Correlation is reported on \textit{averaged} human judgements.}
\label{tbl:res-gsfrobnn}
\end{table} 

The numbers clearly follow a pattern similar to that of Table \ref{tbl:res-gsfrob}. However, now the score of the Copy-Object model is very close to the current state-of-the-art for this dataset, which is 0.47 and has been reported by Tsubaki et al. \cite{tsubaki2013} who use a deep-learning model for composition. Furthermore, note that this time the performance of the Frobenius Additive and Relational models converges. This is encouraging, since it suggests that the vector produced by the sum of all Frobenius vectors is at least as expressive as the matrix of the Relational model in modelling the meaning of a declarative sentence. Similar observations (with small variations) have been reported by me and Sadrzadeh in \cite{kartsadrqpl2014}, where we used yet another form of vector space, based on singular value decomposition.

A second observation stemming from the current work is that in general the categorical models present higher performance than that of vector mixtures. Given the word order-forgetting nature of these models, this is quite encouraging. However, the results are less conclusive regarding this matter. As has been reported in the past \cite{blacoe2012}, models based on simple element-wise operations between vectors can pose an important challenge even for non-linear settings with the highest theoretical power. This fact will be more evident in Chapter \ref{ch:wsdexp}, where there will be cases in which the best performance in some experiment comes from a vector mixture model.

\chapter{Prior Disambiguation in Practice}
\label{ch:wsdexp}

\begin{chabstract}
This chapter proposes an algorithm for the implementation of a prior disambiguation step in tensor-based models of meaning, and it details a specific way to instantiate it. Based on this work, I proceed and experimentally verify the prior disambiguation hypothesis laid out in Chapter \ref{ch:ambiguity} in a number of natural language processing tasks. The experimental work is extended to every CDM class with positive results, and the outcome is discussed at a technical level. Material is based on \cite{kartsaklis:2013:CoNLL,kartsaklis:2013:EMNLP,kartsaklis:2014:ACL,cheng2014}
\end{chabstract}

\noindent
In Chapter \ref{ch:ambiguity}, I detailed a theory regarding the proper treatment of lexical ambiguity in CDMs, and specifically in tensor-based models of meaning. The purpose of the current chapter is to make the abstract ideas expressed there more concrete, and to provide practical evidence based on real-world tasks for their validity. 
%
Although tensor-based models and, specifically, the categorical framework of \cite{Coeckeetal}, still remain our main point of interest, the experiments and the discussion that will follow are not restricted solely to these cases. I am particularly interested to investigate how the proposed prior disambiguation methodology affects every model class of the taxonomy presented in \S \ref{sec:rev-taxonomy}---that is, vector mixtures, partial tensor-based models, tensor-based models, and deep learning models. The discussion that follows the experimental work attempts a meaningful comparison of each CDM's behaviour, and hopefully draws some useful conclusions based on the special nature of each CDM class.

Recall that the central point of the methodology discussed in Chapter \ref{ch:ambiguity} was that each word is assigned to a number of discrete meanings, which ideally correspond to lexicographic definitions of homonymous usages for that particular word. I will start by detailing a method for achieving that at the level of simple vectors; later, in \S\ref{sec:wsdtensor}, I propose an algorithm to extend this to relational tensors of any arity.

\section{Generic word sense induction and disambiguation}
\label{sec:disamb}

\index{word sense induction|see {WSI}}
\index{word sense disambiguation|see {WSD}}

\index{WSI|(}

The notion of word sense induction, that is, the task of detecting the different meanings under which a word appears within a text, is related to that of distributional semantics through the distributional hypothesis---that the meaning of a word is always context-dependent. The assumption is, for example, that the word `bank' as a financial institution would appear in quite different contexts than as land alongside a river. If we had a way to create a vectorial representation for the contexts in which a specific word occurs, then, a clustering algorithm could be applied in order to create groupings of these contexts that hopefully reveal different usages of the word---different meanings---in the training corpus. 

This intuitive idea was first presented by Sch\"utze \cite{Schutze} in 1998, and more or less is the cornerstone of every unsupervised word sense induction and disambiguation method based on semantic word spaces up to today. The approach I use is a direct variation of this standard technique, close to the multi-prototype models of Reisinger and Mooney \cite{reisinger2010}. For what follows, I assume that each word in the vocabulary has already been assigned to an \textit{ambiguous} semantic vector by following typical distributional procedures, for example similar to the setting described in \S\ref{sec:semspace}. 

First, each context for a target word $w_t$ is represented by a \textit{context vector} of the form $\frac{1}{n}\sum_{i=1}^n\ov{w_i}$, where $\ov{w_i}$ is the semantic vector of some other word $w_i \neq w_t$ in the same context. Next, I apply the hierarchical agglomerative clustering method on this set of vectors in order to discover the latent senses of $w_t$. Ideally, the contexts of $w_t$ will vary according to the specific meaning in which this word has been used. The above procedure returns a number of clusters, each consisting of context vectors in the sense described above; the \textit{centroid} of each cluster, then, can be used as a vectorial representation of the corresponding sense/meaning. Thus, in this model each word $w$ is initially represented by a tuple $\langle \ov{w}, S_w \rangle$, where $\ov{w}$ is the ambiguous semantic vector of the word as created by the usual distributional practice, and $S_w$ is a set of \textit{sense vectors} (centroids of context vectors clusters) produced by the above procedure. This completes the \textit{word sense induction} (WSI) step.

\index{WSI|)}
\index{WSD|(}

Being equipped with an unambiguous semantic space created as above, we can now accomplish the disambiguation of an arbitrary word $w$ under a context $C$ as follows: we create a context vector $\ov{c}$ for $C$ as above, and we compare it with every sense vector of $w$; the word is assigned to the sense corresponding to the closest sense vector. Specifically, if $S_w$ is the set of sense vectors for $w$, $\ov{c}$ the context vector for $C$, and $d(\ov{v},\ov{u})$ our vector distance measure, the preferred sense $\hat{s}$ is given by:

\begin{equation}
  \hat{s} = \underset{\ov{v_s} \in S_w}{\arg\min}~d(\ov{v_s},\ov{c})
 \label{equ:wsd}
\end{equation}


Eq. \ref{equ:wsd} above offers to us a practical way for addressing the space complexity problems related to density matrices, as these were briefly discussed in \S\ref{sec:frob-density}; instead of having a massive operator expressing a probability distribution over all the different potential meanings of a word, it allows us to select on the spot the single most probable meaning tensor, using the local context as an oracle. For the experiments that follow later in this chapter, this method will be used as the homonymy-related function $\phi$ in Eq. \ref{equ:maindis}. 

\index{WSD|)}

%
%

%

\section{Disambiguating tensors}
\label{sec:wsdtensor}

\index{tensors!disambiguation|(}
\index{WSI!on tensors|(}
\index{prior disambiguation|(}

The procedure of \S\ref{sec:disamb} above takes place at the vector level (as opposed to tensors of higher order), so it provides a natural way to create sets of meaning vectors for ``atomic'' words of the language, that is, for nouns. In this section I generalize the WSI and WSD methodology to relational words that live in tensor product spaces. Recall that the training of a relational word tensor is based on the set of contexts where this word occurs. Hence, the problem of creating disambiguated versions of tensors can be recast to that of further breaking down the set of contexts in such a way that each subset reflects a different sense of the word in the corpus. If, for example, $S$ is the whole set of sentences for a word $w$ that occurs in the corpus under $n$ different senses, then the goal is to create $n$ subsets $S_1,\hdots S_n$ such that $S_1$ contains the sentences where $w$ appears under the first sense, $S_2$ the sentences where $w$ occurs under the second sense, and so on. Each one of these subsets can then be used to train a tensor for a specific sense of the target relational word.

It turns out that the required partitioning of the context set is something we have already achieved during the WSI stage. Recall that
the clustering step of the procedure detailed in \S \ref{sec:disamb} provides us with $n$ clusters of context vectors for a target word. Since in our case each context vector corresponds to a distinct sentence, the output of the clustering scheme can also be seen as $n$ subsets of sentences, where the word appears under different meanings. This one-to-one correspondence of the context clusters with sentence sets can be used for learning unambiguous versions of verb tensors. Let $S_1 \hdots S_n$ be the sets of sentences returned by the clustering step for a verb. Then, the verb tensor for the $i$th sense is:

\begin{equation}
  \overline{verb}_{i} = \sum\limits_{s \in S_i} (\ov{subj_s} \otimes \ov{obj_s})
  \label{equ:reldis}
\end{equation}

\noindent where $subj_s$ and $obj_s$ refer to the subject/object pair that occurred with the verb in sentence $s$. We have now re-stated the ``argument summing'' procedure of Eq. \ref{equ:weightrel} in a way that creates unambiguous versions of relational tensors. Note that this approach can be generalized to any arity $n$ as follows:

\begin{equation}
\overline{word}_i = \sum\limits_{s \in S_i} \bigotimes\limits_{k=1}^{n} \ov{arg_{k,s}}
\end{equation}

\noindent where $arg_{k,s}$ denotes the $k$th argument of the target word in sentence $s$. The disambiguation process for the words in a sentence is summarized in pseudo-code form in Algorithm \ref{alg:wsd}.\index{prior disambiguation!algorithm} 

\begin{algorithm}[b!]
\caption{The disambiguation process. The $\oplus$ symbol in lines \ref{lin:concat1} and \ref{lin:concat2} denotes list concatenation.}
\label{alg:wsd}
\small
\begin{algorithmic}[1]
\STATE {\bf Algorithm} {\sc Wsd}$(M=\{\langle w_1,\ov{w_1},C_1,S_1\rangle, \dots, \langle w_n,\ov{w_n},C_n,S_n\rangle\})$ {\bf returns} {\it list}
\STATE ; $w_i$: the $i$th word in the sentence
\STATE ; $\ov{w_i}$: ambiguous vector for word $w_i$
\STATE ; $C_i=\{\text{cl}_1, \dots, \text{cl}_k\}$: set of clusters for word $w_i$
\STATE ; $S_i=\{s_1,\dots,s_k\}$: set of contexts for word $w_i$ in 1-to-1 correspondence with $C_i$ 
\STATE ; $list$: an ordered list of the disambiguated word vectors/tensors
\STATE ~
\STATE $list \gets [~]$
\FOR{{\bf each} $\langle w_t,\ov{w_t},C_t,S_t \rangle$ {\bf in} $M$} 
\STATE $\ov{c} = \frac{1}{n}\sum\limits_{w\neq w_t} \ov{w}$
\STATE $\text{cl} = \underset{\text{cl} \in C_t}{\arg\min}~d(\op{centroid}(\text{cl}),\ov{c})$
\IF{$w_t$ {\bf is} noun}
\STATE\label{lin:concat1} $list \gets list \oplus [\op{centroid}(\text{cl})]$
\ELSE
\STATE $\ol{tensor} = \sum\limits_{\text{cnt} \in s_{\text{cl}}}\bigotimes\limits_{arg\in \text{cnt}} \ov{arg}$
\STATE\label{lin:concat2} $list \gets list \oplus [\ol{tensor}]$
\ENDIF
\ENDFOR
\RETURN{$list$}
\normalsize
\end{algorithmic}
\end{algorithm}

\index{prior disambiguation|)}
\index{WSI!on tensors|)}
\index{tensors!disambiguation|)}

As is evident from the discussion so far, clustering is an essential part of every unsupervised WSI method---actually, the heart of it. Before I proceed to start detailing the experimental work in \S\ref{sec:wsdfrob}, I would like to discuss how I approached this important topic in some depth. The following exposition builds on my first paper related to the prior disambiguation hypothesis, co-authored with Sadrzadeh and Pulman \cite{kartsaklis:2013:CoNLL}.

\section{Clustering}
\label{sec:clustering}

The choice of a clustering algorithm is an important decision in the overall process, and involves a selection ranging from approaches such as the well-known $k$-means algorithm and hierarchical clustering to more advanced probabilistic methods like Bayesian mixture models. Deciding which approach is the most appropriate for a specific case might depend on a number of factors, for example the size and the form of the dataset or the available computational resources. In practice it has been found that although Bayesian approaches tend to give better results, simpler methods like $k$-means usually follow very closely. Brody and Lapata \cite{brody2009bayesian}, for example, report an F-score  of 87.3 on the {\sc Semeval} 2007 task for their sense induction approach based on Latent Dirichlet Allocation, not much higher than the 84.5 F-score reported by Pedersen \cite{pedersen2007umnd2} for the same task, who creates meaning vectors using $k$-means. 


\index{hierarchical agglomerative clustering|see {HAC}}

Since my intention here is not to construct a cutting-edge word sense induction system, but rather to provide a proof of concept for the benefits of prior disambiguation in CDMs that would not be tied to any specific clustering algorithm, I will use a generic iterative bottom-up approach known as \textit{hierarchical agglomerative clustering} (HAC).\index{HAC} Hierarchical clustering has been invariably applied for unsupervised word sense discrimination on a variety of languages, generally showing good performance---see, for example, the comparative study of \cite{broda2012evaluation} for English and Polish. Compared to $k$-means clustering, this approach has the major advantage that it does not require us to define in advance a specific number of clusters. Compared to more advanced probabilistic techniques, such as Bayesian mixture models, it is much more straightforward and simple to implement.\footnote{Actually, informal experimentation with robust probabilistic methods, such as Dirichlet process Gaussian mixture models, revealed no significant benefits for my purposes.} 

\begin{figure}[t]
\centering
\includegraphics[scale=0.40,clip=true,trim=4.7cm 1.05cm 3.7cm 1cm]{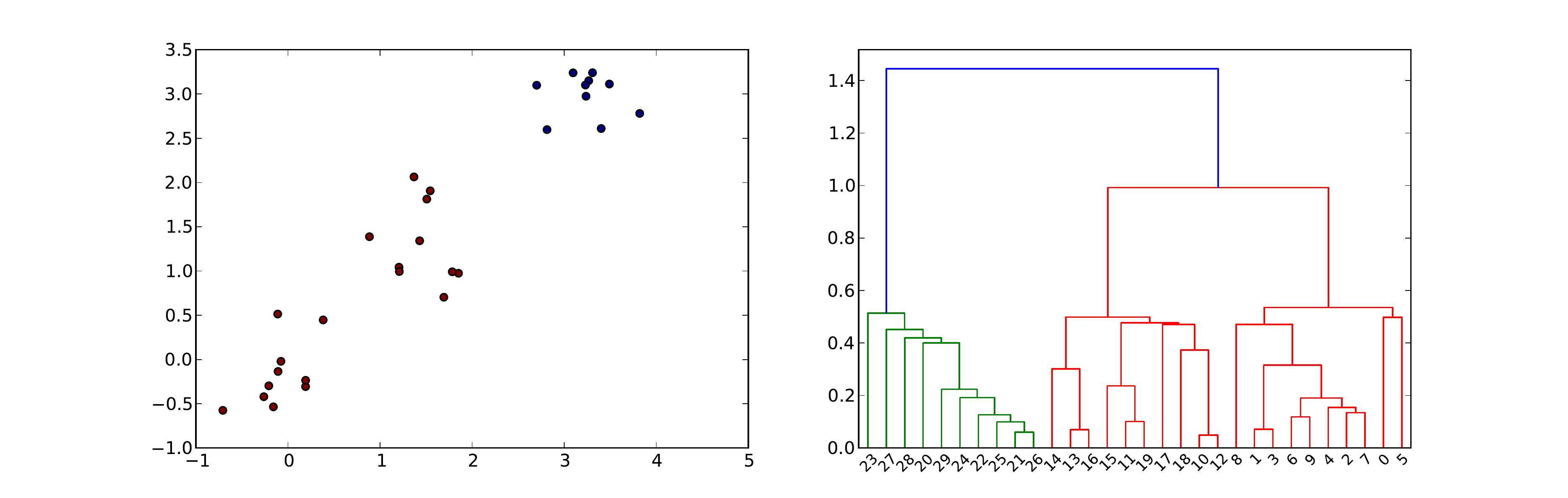}

\caption{Hierarchical agglomerative clustering.}
\label{fig:dendrogram}
\end{figure}

HAC is a bottom-up method of cluster analysis, starting with each data point (context vector in our case) forming its own cluster; then, in each iteration the two closest clusters are merged into a new cluster, until all points are finally merged under the same cluster. This process produces a \textit{dendrogram} (i.e. a tree diagram),\index{dendrogram} which essentially embeds every possible clustering of the dataset. As an example, Figure \ref{fig:dendrogram} shows a small dataset produced by three distinct Gaussian distributions, and the dendrogram derived by the above algorithm. Implementation-wise, I use the \texttt{fastcluster} library \cite{fastcluster}, which provides $\text{O}(n^2)$ implementations for all the methods relevant to this work, and comes with interfaces for R and Python.\footnote{I am grateful to Daniel M\"ullner for his help and the useful suggestions regarding the proper use of his library.}

\subsection{Choosing the number of clusters}
\label{sec:vrc}

\index{number of clusters, choosing|(}
\index{HAC!choosing the number of clusters|(}
\index{variance ration criterion|see {VRC}}
\index{VRC|(}

Detecting the number of clusters that gives the most natural partitioning of the data is not an easy task, and is the reason that many word sense induction systems, like those of \cite{Schutze} and \cite{brody2009bayesian}, adopt a fixed number of senses. Of course, such a solution seems artificial and unintuitive---a good WSI system should be able to automatically detect the correct number of meanings from the underlying data. Our choice of clustering method actually makes this task easier, since the dendrogram produced by HAC is a compact representation of all possible clusterings. However, one still needs to decide where exactly to ``cut'' the tree in order to get the best possible partitioning of the data.

Taking as an example the dendrogram of Figure \ref{fig:dendrogram}, the optimal partitioning of 3 clusters is given by cutting the tree at about 0.8. However, cutting the tree at 1.2 will give us two clusters; even worse, if we cut the tree at 0.2 we will produce 17 clusters, a way off the natural partitioning of the dataset. Although the right answer to this problem might depend on the specific level of granularity we wish to get from our clusters, we can safely assume that the optimal partitioning is the one that provides the most compact and maximally-separated clusters. One way to measure the quality of a clustering based on this criterion is the Cali\'nski/Harabasz index \cite{calinski1974}, also known as \textit{variance ratio criterion} (VRC). Given a set of $N$ data points and a partitioning of $k$ disjoint clusters, VRC is computed as follows:

\begin{equation}
   VRC_k = \frac{\op{Tr}(\textbf{B})}{\op{Tr}(\textbf{W})} \times \frac{N-k}{k-1}
\end{equation}

Here, \textbf{B} and \textbf{W} are the inter-cluster and inner-cluster dispersion matrices, respectively, defined as:

\begin{equation}
  \textbf{W} = \sum\limits_{i=1}^k \sum\limits_{l=1}^{N_i} (\ov{x_i}(l) - \bar{x_i}) (\ov{x_i}(l) - \bar{x_i})^{\mathsf{T}}
\end{equation}

\begin{equation}
  \textbf{B} = \sum\limits_{i=1}^k N_i (\bar{x_i}-\bar{x})(\bar{x_i}-\bar{x})^{\mathsf{T}}
\end{equation}

\noindent where $N_i$ is the number of data points assigned to cluster $i$, $\ov{x_i}(l)$ is the $l$th point assigned to this cluster, $\bar{x_i}$ is the centroid of $i$th cluster (the mean), and $\bar{x}$ is the data centroid of the overall dataset. Given the above formulas, the trace of \textbf{B} is the sum of inter-cluster variances, while the trace of \textbf{W} is the sum of inner-cluster variances. A good partitioning should have high values for \textbf{B} (which is an indication for well-separated clusters) and low values for \textbf{W} (an indication for compact clusters), so the higher the quality of the partitioning the greater the value of the  ratio.

Compared to other criteria, VRC has been found to be one of the most effective approaches for clustering validity---see for example the comparative studies of \cite{milligan1985} and  \cite{vendramin2009}. Furthermore, it has been previously applied to word sense discrimination successfully, returning the best results among a number of other measures \cite{savova2006}. In the experiments that will follow, I compute the VRC for a number of different partitionings (ranging from 2 to 10 clusters), and I keep the partitioning that results in the highest VRC value as the optimal number of senses for the specific word. Note that since the HAC dendrogram already embeds all possible clusterings, the cutting of the tree in order to get a different partitioning is performed in constant time. This is quite different, for example, from $k$-means or other methods based on an iterative expectation-maximization approach, where in order to get a different clustering one needs to define the number of clusters beforehand and run the algorithm again.

While this method of automatically detecting the number of clusters is efficient and effective, it has one limitation: VRC needs at least two clusters in order to be computable, so this is the lowest number of senses that a word can get in our setting. Although on the surface this seems unintuitive, note that my previous comments in \S\ref{sec:priordis} regarding the distinction between cases of homonymy and cases of polysemy are also relevant here: when a word is completely unambiguous (so normally it would be represented by a single vector instead of two), we can safely assume that the two sense vectors produced from the partitioning imposed by VRC will be very close to each other as well as to the original ambiguous word vector, so the effect for the sense selection process as expressed by Eq. \ref{equ:wsd} would be minimal in any case.

\index{VRC|)}
\index{HAC!choosing the number of clusters|)}
\index{number of clusters, choosing|)}

\subsection{Tuning the parameters}
\label{sec:wsdtuning}

The parameters of the clustering scheme are optimized on the noun set of {\sc Semeval} 2010 Word Sense Induction \& Disambiguation Task \cite{manandhar2010semeval}. Specifically, when using HAC one has to decide how to measure the distance between the clusters, which is the merging criterion applied in every iteration of the algorithm, as well as the measure between the data points, i.e. the individual vectors. A number of empirical tests I conducted limited the available options to two inter-cluster measures: complete-link and Ward's methods. In the complete-link method the distance between two clusters $X$ and $Y$ is the distance between their two most remote elements:\index{complete-link}\index{HAC!complete-link}

\begin{equation}
   D(X,Y) = \max\limits_{x \in X, y \in Y} d(x,y)
\end{equation}

In Ward's method,\index{Ward's method}\index{HAC!Ward's method} two clusters are selected for merging if the new partitioning exhibits the minimum increase in the overall intra-cluster variance. The cluster distance is given by:

\begin{equation}
  D(X,Y) = \frac{2 \vert X\vert \vert Y\vert}{\vert X\vert+\vert Y\vert} \Vert \ov{c_X} - \ov{c_Y} \Vert ^2
\end{equation}

\noindent where $\ov{c_X}$ and $\ov{c_Y}$ are the centroids of $X$ and $Y$.

These \textit{linkage} methods have been tested in combination with three vector distance measures: Euclidean\index{Euclidean distance}, cosine\index{cosine distance}, and Pearson's correlation\index{correlation} (6 models in total). The metrics were chosen to represent progressively more relaxed forms of vector comparison, with the strictest form to be the Euclidean distance and correlation as the most relaxed. For sense detection I use Eq. \ref{equ:wsd}, considering as context the whole sentence in which a target word appears. The distance metric used for the disambiguation process in each model is identical to the metric used for the clustering process, so in the Ward/Euclidean model the disambiguation is based on the Euclidean distance, in complete-link/cosine model on the cosine distance, and so on. I evaluate the models using V-measure,\index{V-measure} an entropy-based metric that addresses the so-called \textit{matching problem} of F-score \cite{rosenberg2007v}. Table \ref{tbl:semeval} shows the results.

\begin{table}[h!]
\small
\begin{center}
\begin{tabular}{l|ccc}
\hline
\textbf{Model} & \textbf{V-Measure} & \textbf{Avg clusters} \\
\hline\hline
Ward/Euclidean & 0.05 & 1.44 \\
\textbf{Ward/Correlation} & \textbf{0.14} & \textbf{3.14} \\
Ward/Cosine & 0.08 & 1.94 \\
\hline
Complete/Euclidean& 0.00 & 1.00 \\
Complete/Correlation& 0.11 & 2.66 \\
Complete/Cosine& 0.06 & 1.74 \\
\hline
Most frequent sense & 0.00 & 1.00 \\
1 cluster/instance & 0.36 & 89.15 \\
\hline
Gold standard & 1.0 & 4.46 \\
\hline
\end{tabular}
\normalsize
\caption{Results on the noun set of {\sc Semeval} 2010 WSI\&D task.}
\label{tbl:semeval}
\end{center}
\end{table}

Ward's method in combination with correlation distance provided the highest V-measure, followed by the combination of complete-link with (again) correlation. Although a direct comparison of my models with the models participating in this task would not be quite sound (since these models were trained on a special corpus provided by the organizers), it is nevertheless enlightening to mention that the 0.14 V-measure places the Ward/Correlation model at the 4th rank among 28 systems for the noun set of the task, while at the same time provides a reasonable average number of clusters per word (3.14), close to that of the human-annotated gold standard (4.46). Compare this, for example, with the best-performing system that achieved a V-measure of 0.21, a score that was largely due to the fact that the model assigned the unrealistic number of 11.54 senses per word on average (since V-measure tends to favour higher numbers of senses, as the baseline \textit{1 cluster/instance} shows in Table \ref{tbl:semeval}).\footnote{The results of {\sc Semeval} 2010 can be found online at \texttt{http://www.cs.york.ac.uk/seme\-val2010{\scriptsize \_}WSI/task{\scriptsize \_}14{\scriptsize\_}ranking.html}.}

\begin{figure}
\centering
\subfigure[Single-link]{
\includegraphics[scale=0.30,clip=true,trim=2.5cm 2.0cm 1cm 1.5cm]{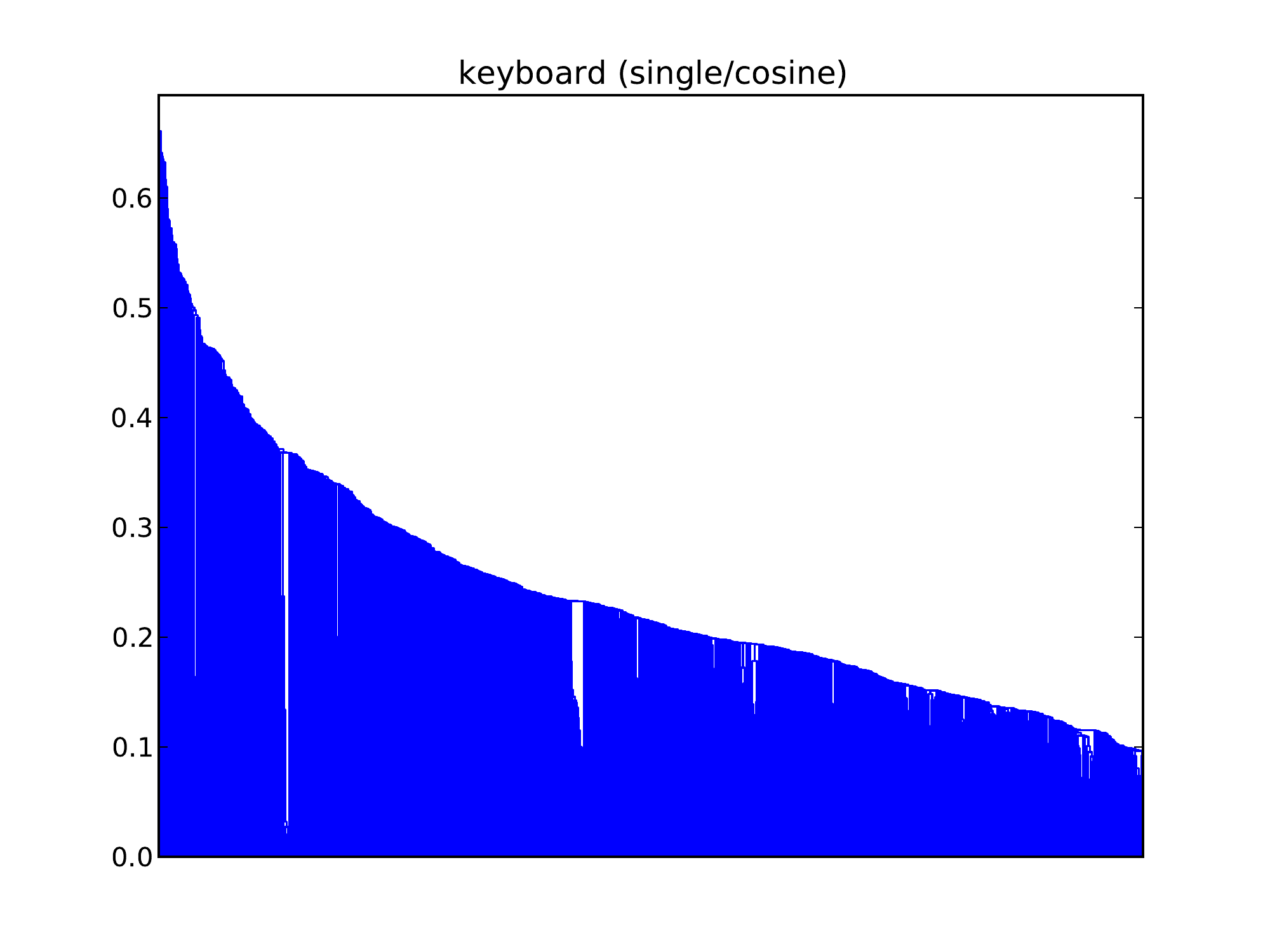}}~
\subfigure[Average-link]{
\includegraphics[scale=0.30,clip=true,trim=2.5cm 2.0cm 1cm 1.5cm]{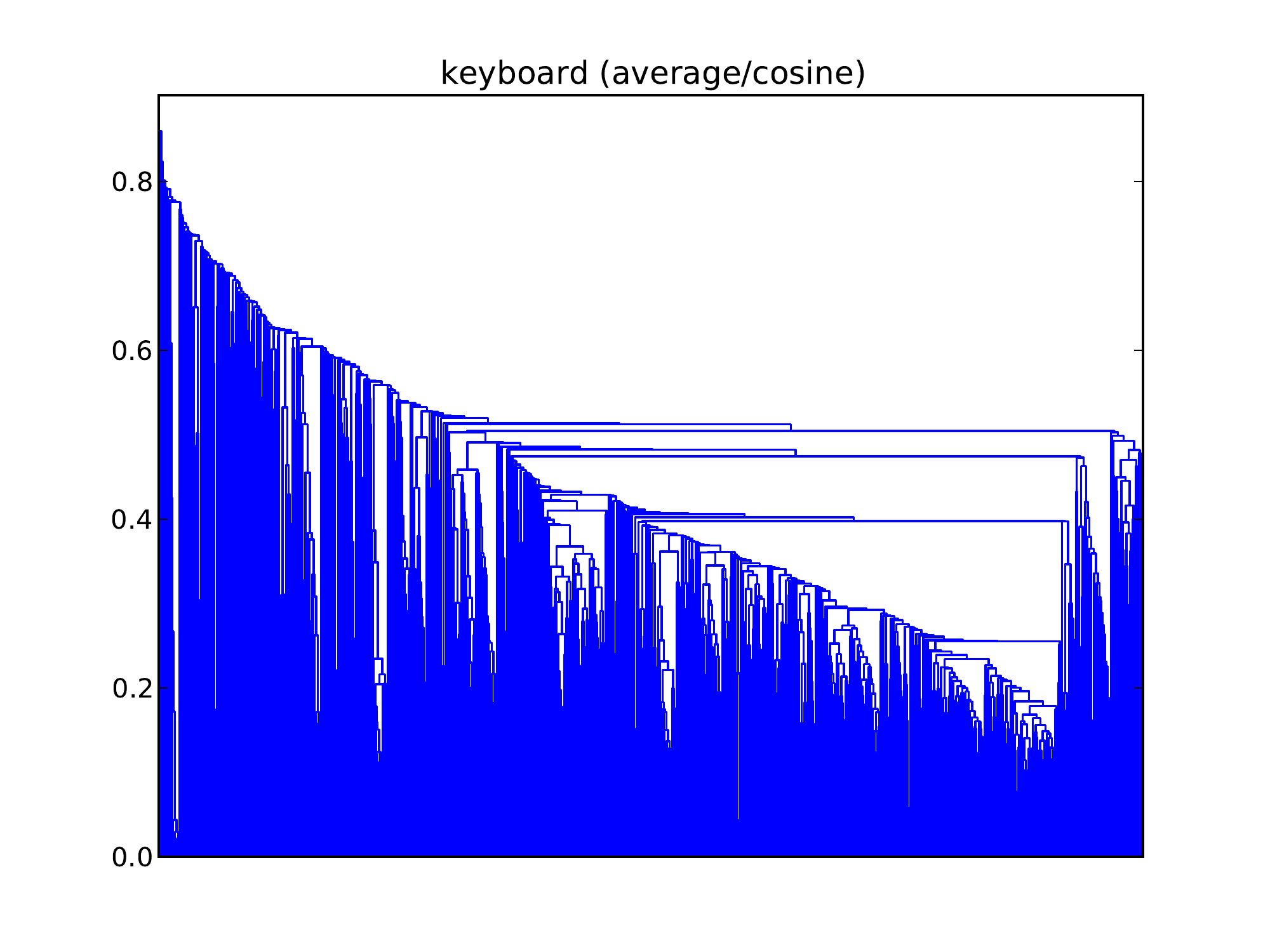}} \\
\subfigure[Ward's method]{
\includegraphics[scale=0.30,clip=true,trim=2.5cm 2.0cm 1cm 1.5cm]{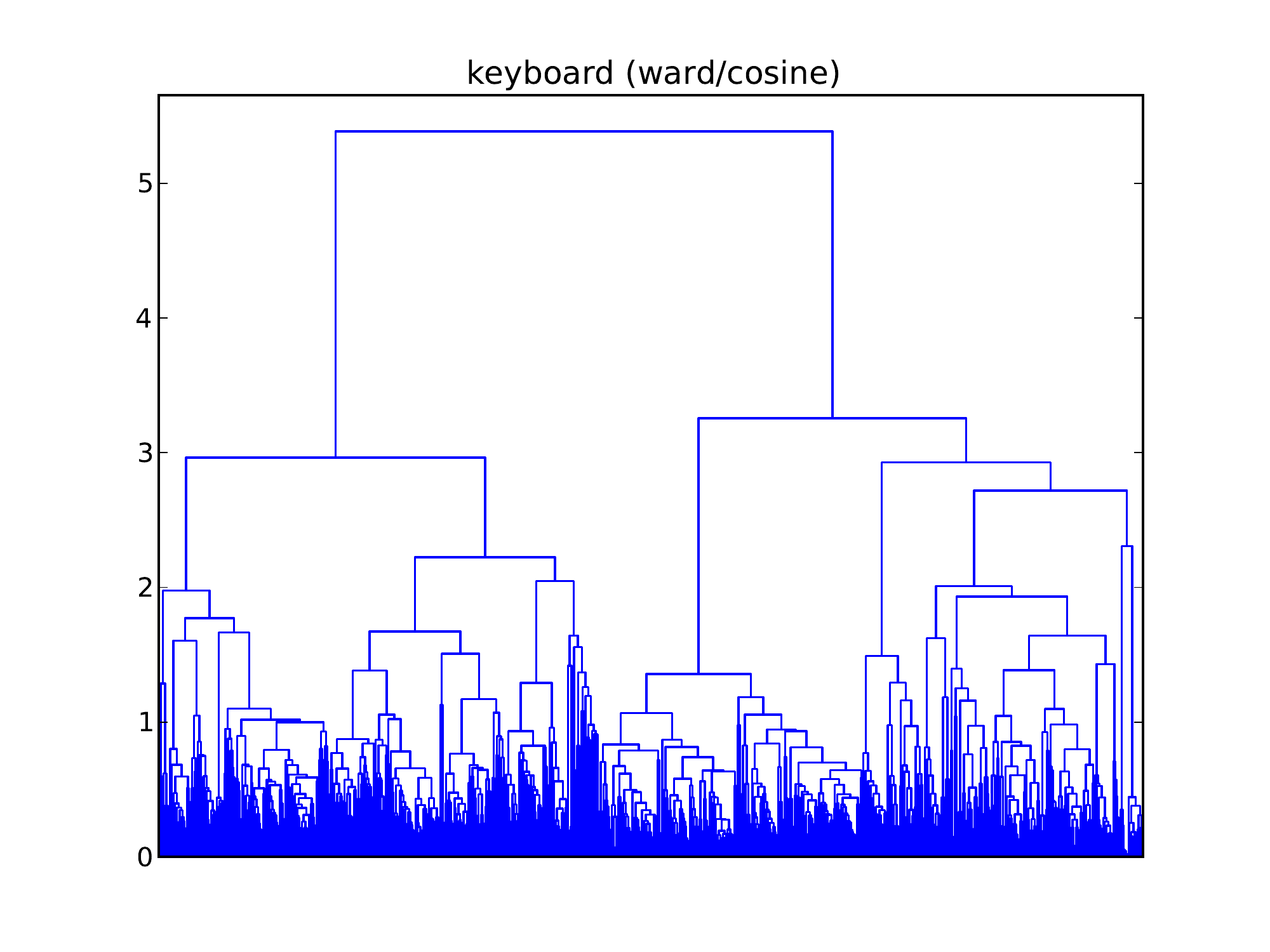}}
\subfigure[Complete-link]{
\includegraphics[scale=0.30,clip=true,trim=2.5cm 2.0cm 1cm 1.5cm]{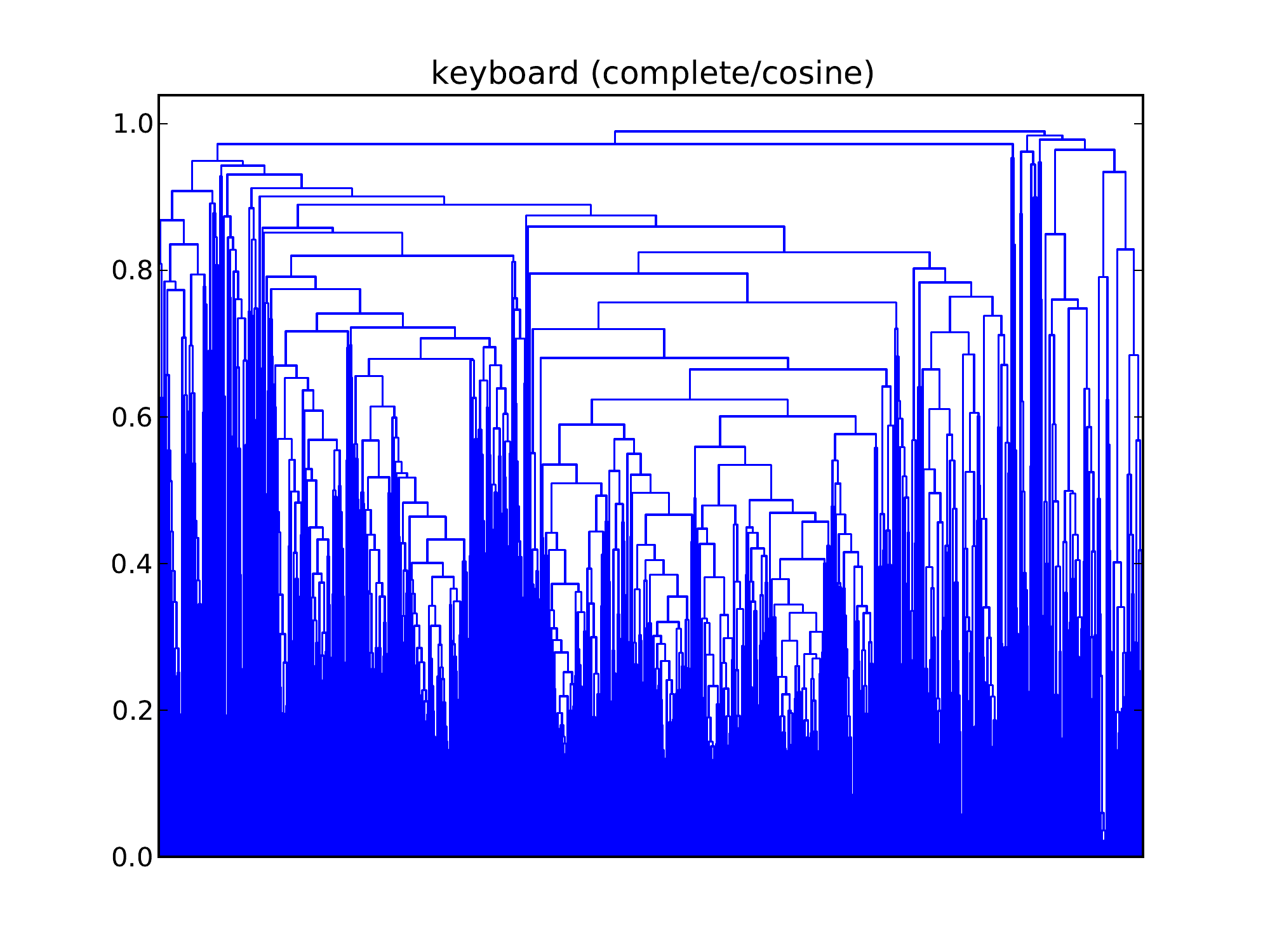}}
\caption{Dendrograms produced for word `keyboard' according to 4 different linkage methods.}
\label{fig:dendrograms}
\end{figure}

Table \ref{tbl:sensedemo} provides an example of the results, showing the senses for a number of words learned by the best model of Ward's method  and correlation measure. Each sense is visualized as a list of the most dominant words in the cluster, ranked by their {\sc Tf-Idf} values. Furthermore, Figure \ref{fig:dendrograms} shows the dendrograms produced by four linkage methods for the word `keyboard', demonstrating the superiority of Ward's method.

\begin{table}
\begin{center}
\footnotesize
\begin{tabular}{p{13cm}}
  \textbf{keyboard:} 2 senses \\
  \hline
  \leftskip 0.4cm\parindent -0.4cm
 \textsc{Computer}: program dollar disk power enter port graphic card option select language drive pen application corp external editor woman price page design sun cli amstrad lock interface lcd slot notebook \\
 \leftskip 0.4cm\parindent -0.4cm
 \textsc{Music}: drummer instrumental singer german father fantasia english generation wolfgang wayne cello body join ensemble mike chamber gary saxophone sax ricercarus apply form son metal guy clean roll barry orchestra \\
  \hline
  \\
  \textbf{vessel:} 2 senses \\
  \hline
  \leftskip 0.4cm\parindent -0.4cm
  \textsc{Naval}: fishing state member ship company british united sea kingdom register flag owner registration community crew law boat board merchant cargo operator operate act quota man national naval fleet \\
 \leftskip 0.4cm\parindent -0.4cm
  \textsc{Medical}: disease coronary grade nerve muscle vascular infarction tumour non-bleeding dilate myocardial probe patency successful angina infarct lymph haemorrhage chronic capillary stain bleeding heater clot mucosal constrict thrombosis \\
  \hline
  \\
  \textbf{tank:} 3 senses \\
  \hline
  \leftskip 0.4cm\parindent -0.4cm
  \textsc{Generic}: channel upper wheel cable cupboard flight gate standard haul rail pound acid cistern programme rope tender gauge insulation incline certain set round passenger insulate ballast garden safety lead thin frame philip lift joint flotation \\
  \leftskip 0.4cm\parindent -0.4cm
  \textsc{Fish Tank}: divider swordtail well-planted tanganyikan hagen female fish-only tang gravity-fed infusorium mulm ledsam spawner scavenger devoid pair misbehave malawis cleanest synodontis bear growing-on piranha dirty nett bristleworm  \\
  \leftskip 0.4cm\parindent -0.4cm
  \textsc{Military}: syrian pact supreme ural station sarajevo campaign parliament israeli guerrilla secretary soviet-made emirates launcher jna cordon deployment press export times dramatic moscow mikhail motorize defender rumour uae bosnian \\
  \hline
\normalsize
\end{tabular}
\vspace{-0.3cm}
\caption{Derived senses for various words.}
\label{tbl:sensedemo}
\end{center}
\end{table}

\section{Experiments on Frobenius models}
\label{sec:wsdfrob}

I will now start presenting a series of experiments, the purpose of which is to evaluate the prior disambiguation hypothesis in concrete NLP tasks. In this section I focus on the Frobenius models of Chapter \ref{ch:frobverbs}, so naturally my evaluation methodology follows a pattern similar to that of Chapter \ref{ch:frobexp}. Specifically, I perform experiments based on (a) two verb disambiguation tasks similar to the one presented in \S\ref{sec:kintsch}; (b) the  verb phrase similarity task of \S\ref{sec:phrase-sim}; and (c) a transitive version of that task, where the verb-object structures have been extended with an appropriate subject. 

In all cases the prior disambiguation methodology follows Algorithm \ref{alg:wsd}. First, I apply the word sense induction procedure of \S\ref{sec:disamb} in order to induce meaning clusters for every word in my datasets. As discussed in \S\ref{sec:wsdtensor}, these clusters correspond to distinct sets of sentences in which the verb occurs under different meanings. Each one of these sets is used for training an ``unambiguous'' tensor according to Eq. \ref{equ:reldis}, which represents a distinct usage of the verb in the corpus. The application of the prior disambiguation step then proceeds as follows: For each word in the sentence, the optimal sense is selected according to Eq. \ref{equ:wsd}. If this word is a noun, the vector used in the composition is the centroid of the cluster that corresponds to the optimal sense, while for verbs we pick the corresponding unambiguous tensor created as above. Composition proceeds according to the models and equations of Table \ref{tbl:models} on the unambiguous word representations. For ambiguous models, a tensor is trained for each verb by using the union of all sentence sets, and composition takes place as usual between those tensors and the original ambiguous word vectors for subjects and objects. All experiments that follow use the semantic space I have introduced in \S\ref{sec:semspace}, and have been presented in an EMNLP 2013 paper I co-authored with Sadrzadeh \cite{kartsaklis:2013:EMNLP}.

\subsection{Head verb disambiguation}
\label{sec:frob-verbdis}

\index{head verb disambiguation|(}
\index{prior disambiguation!head verb disambiguation|(}

Since this time the hypothesis under test is directly related to disambiguation, I will perform the head verb disambiguation task of Kintsch on two different datasets: the G\&S 2011 dataset from \S\ref{sec:kintsch}, and an additional dataset (to which I will refer as K\&S 2013) that has been created by Mehrnoosh Sadrzadeh in collaboration with Edward Grefenstette, and introduced for the first time in relevant work I conducted jointly with Sadrzadeh and Pulman \cite{kartsaklis:2013:CoNLL}. The main difference between the two datasets is that in the former the verbs and their alternative meanings have been selected automatically using the JCN metric of semantic similarity \cite{jcn}, while in the latter the selection was based on human judgements from the work of Pickering and Frisson in psycholinguistics \cite{pickering2001processing}. One consequence of the selection process was that while in the first dataset many verbs are clear cases of polysemy (e.g. `say' with meanings \textit{state} and \textit{allege} or `write' with meanings \textit{publish} and \textit{spell}), the landmarks in the second dataset correspond to clearly separated meanings (e.g. `file' with meanings \textit{register} and \textit{smooth} or `charge' with meanings \textit{accuse} and \textit{bill}), i.e. they express homonymous cases. Furthermore, subjects and objects of this latter dataset are modified by appropriate adjectives. The important point for our purposes is that both datasets together cover a wide range of ambiguity cases, which can help us to draw useful results regarding the performance of the tested models.

The results for the G\&S 2011 dataset are shown in Table \ref{tbl:gands}, expressed once more as the Spearman's correlation with human judgements.\footnote{For all tables in this section, $\ll$ and $\gg$ denote highly statistically significant differences with $p<0.001$.} The Frobenius models present much better performance than the vector mixture ones, with the disambiguated version of the Copy-Object model significantly higher than the Relational model. In general, the disambiguation step improves the results of all the Frobenius models; however, the effect is drastically reversed for the vector mixtures. 

\index{prior disambiguation!head verb disambiguation|)}
\index{head verb disambiguation|)}

\begin{table}[h]
\small
\begin{center}
\begin{tabular}{l|ccc}
 \hline
 \textbf{Model} & \textbf{Ambiguous} & & \textbf{Disambiguated} \\
 \hline\hline
 Only verbs & 0.198 & $\gg$ & 0.132 \\
 \hline
 Multiplicative & 0.137 & $\gg$ & 0.044 \\
 Additive       & 0.127 & $\gg$ & 0.047 \\
 \hline
 Relational     & 0.219 & $<$ & 0.223 \\
 Copy-Subject   & 0.070 & $\ll$ & 0.122 \\
 Copy-Object & \textbf{0.241} & $\ll$ & \textbf{0.262} \\
 Frobenius additive & 0.142 & $\ll$ & 0.214 \\   
 \hline
 Human agreement & \multicolumn{3}{c}{0.599} \\
 \hline
\end{tabular}
\caption[Prior disambiguation results for Frobenius models (G\&S dataset).]{Results for the G\&S 2011 dataset. The difference between the Copy-Object and Relational model is s.s. with $p<0.001$.}
\label{tbl:gands}
\end{center}
\end{table}

The effect of disambiguation is clearer for the K\&S 2013 dataset (Table \ref{tbl:pandf}). The longer context in combination with genuinely ambiguous verbs produces two effects: first, disambiguation is now helpful for all models, either vector mixtures or tensor-based; second, the disambiguation of just the verb (Only Verbs model), without any interaction with the context, is sufficient to provide the best score (0.22) with a difference statistically significant from the second model (0.19 for disambiguated additive). In fact, further composition of the verb with the context decreases performance, confirming results I reported in previous work \cite{kartsaklis:2013:CoNLL} for vectors trained using BNC. Given the nature of the specific task, which is designed around the ambiguity of the verb, this result is not surprising: a direct disambiguation of the verb based on the rest of the context should naturally constitute the best method to achieve top performance---no composition is actually necessary for this task to be successful.\index{head verb disambiguation!and composition} 

\begin{table}
\small
\begin{center}
\begin{tabular}{l|ccc}
 \hline
 \textbf{Model} & \textbf{Ambiguous} & & \textbf{Disambiguated} \\
 \hline\hline
 Only verbs   & \textbf{0.151} & $\ll$ & \textbf{0.217} \\
 \hline
 Multiplicative & 0.131 & $<$ & 0.137 \\
 Additive       & 0.085 & $\ll$ & 0.193 \\
 \hline
 Relational     & 0.036 & $\ll$ & 0.121 \\
 Copy-Subject   & 0.035 & $\ll$ & 0.117 \\
 Copy-Object    & 0.033 & $\ll$ & 0.095 \\
 Frobenius additive & 0.036 & $\ll$ & 0.114 \\ 
 \hline
 Human agreement & \multicolumn{3}{c}{0.383} \\
 \hline 
\end{tabular}
\caption[Prior disambiguation results for Frobenius models (K\&S dataset).]{Results for the K\&S 2013 dataset. The difference between Only Verbs and Additive model is s.s. with $p<0.001$}
\label{tbl:pandf}
\end{center}
\end{table}

\subsection{Verb phrase/sentence similarity}
\label{sec:frob-vpsim}

\index{verb phrase similarity|(}
\index{prior disambiguation!and verb phrase similarity|(}

The second set of experiments is based on the phrase similarity task I introduced in \S\ref{sec:phrase-sim}. Recall from the discussion in \S\ref{sec:verbphrase} that since now we deal with verb phrases, the Relational model is reduced to Copy-Subject. The results presented in Table \ref{tbl:mandl} show that the effects of disambiguation for this task are quite impressive: the differences between the scores of all disambiguated models and those of the ambiguous versions are highly statistically significant (with $p<0.001$), while  4 of the 6 models present an improvement greater than 10 units of correlation. The models that benefit the most from disambiguation are the vector mixtures; both of these approaches perform significantly better than the best Frobenius model (which is Copy-Object again). 

\begin{table}[h]
\small
\begin{center}
\begin{tabular}{l|ccc}
 \hline
 \textbf{Model} & \textbf{Ambiguous} & &\textbf{Disambiguated} \\
 \hline\hline
 Only verbs & 0.310 & $\ll$ & 0.420 \\
 \hline
 Multiplicative & 0.315 & $\ll$ & \textbf{0.448} \\
 Additive       & 0.291 & $\ll$ & 0.436 \\
 \hline
 Relational/Copy-Subject  & \textbf{0.340} & $\ll$ & 0.367 \\
 Copy-Object    & 0.290 & $\ll$ & 0.393 \\
 Frobenius additive & 0.270 & $\ll$ & 0.336 \\
 \hline
 Human agreement & \multicolumn{3}{c}{0.550} \\
 \hline 
\end{tabular}
\caption[Prior disambiguation results for Frobenius models (M\&L dataset).]{Results for the M\&L 2010 task. Difference between Multiplicative and Additive is not s.s; difference between Multiplicative and Only Verbs is s.s. with $p<0.001$.}
\label{tbl:mandl}
\end{center}
\end{table}

In a variation of the above experiment, my goal is to examine the extent to which the picture of Table \ref{tbl:mandl} can change for the case of text structures longer than verb phrases. In order to achieve this, I extend each one of the 72 verb phrases to a full transitive sentence by adding an appropriate subject such that the similarity relationships of the original dataset are retained as much as possible, so the human judgements for the verb phrase pairs could as well be used for the transitive cases. I worked pair-wise: for each pair of verb phrases, I first selected one of the 5 most frequent subjects for the first phrase; then, the subject of the other phrase was selected by a list of synonyms of the first subject in a way that the new pair of transitive sentences constitutes the least more specific version of the given verb-phrase pair. So, for example, the pair \textit{produce effect/achieve result} became \textit{drug produce effect/medication achieve result}, while the pair \textit{pose problem/address question} became \textit{study pose problem/paper address question}. 

The restrictions of the verb-phrase version do not hold for the transitive case, so here the evaluation takes place on the full set of models (Table \ref{tbl:mandltr}). Once more disambiguation produces better results in all cases, with highly statistically significant differences for all models. The most evident observation is that as the length of the text segments increases, the performance of vector mixtures and tensor-based models seems to converge. Indeed, note how the numbers of the vector mixture models are significantly lower compared to the verb phrase task.

\begin{table}
\small
\begin{center}
\begin{tabular}{l|ccc}
 \hline
 \textbf{Model} & \textbf{Ambiguous} & & \textbf{Disambiguated} \\
 \hline\hline
 Only verbs & 0.310 & $\ll$ & 0.341 \\
 \hline
 Multiplicative & 0.325 & $\ll$   & 0.404 \\
 Additive       & 0.368 & $\ll$ & \textbf{0.410} \\
 \hline
 Relational     & \textbf{0.368} & $\ll$ & 0.397 \\
 Copy-Subject   & 0.310 & $\ll$ & 0.337 \\
 Copy-Object    & 0.321 & $\ll$ & 0.368 \\
 Frobenius additive & 0.309 & $\ll$ & 0.390 \\
 \hline
 Human agreement & \multicolumn{3}{c}{0.550} \\
 \hline 
\end{tabular}
\caption[Prior disambiguation results for Frobenius models (K\&S 2013 dataset).]{Results for the transitive version of M\&L 2010 task.}
\label{tbl:mandltr}
\end{center}
\end{table}

\index{prior disambiguation!and verb phrase similarity|)}
\index{verb phrase similarity|)}

\subsection{Discussion}
\label{sec:frobexp-discussion}

\begin{figure}[b!]
  \begin{minipage}{0.5\linewidth}
  \includegraphics[scale=0.5]{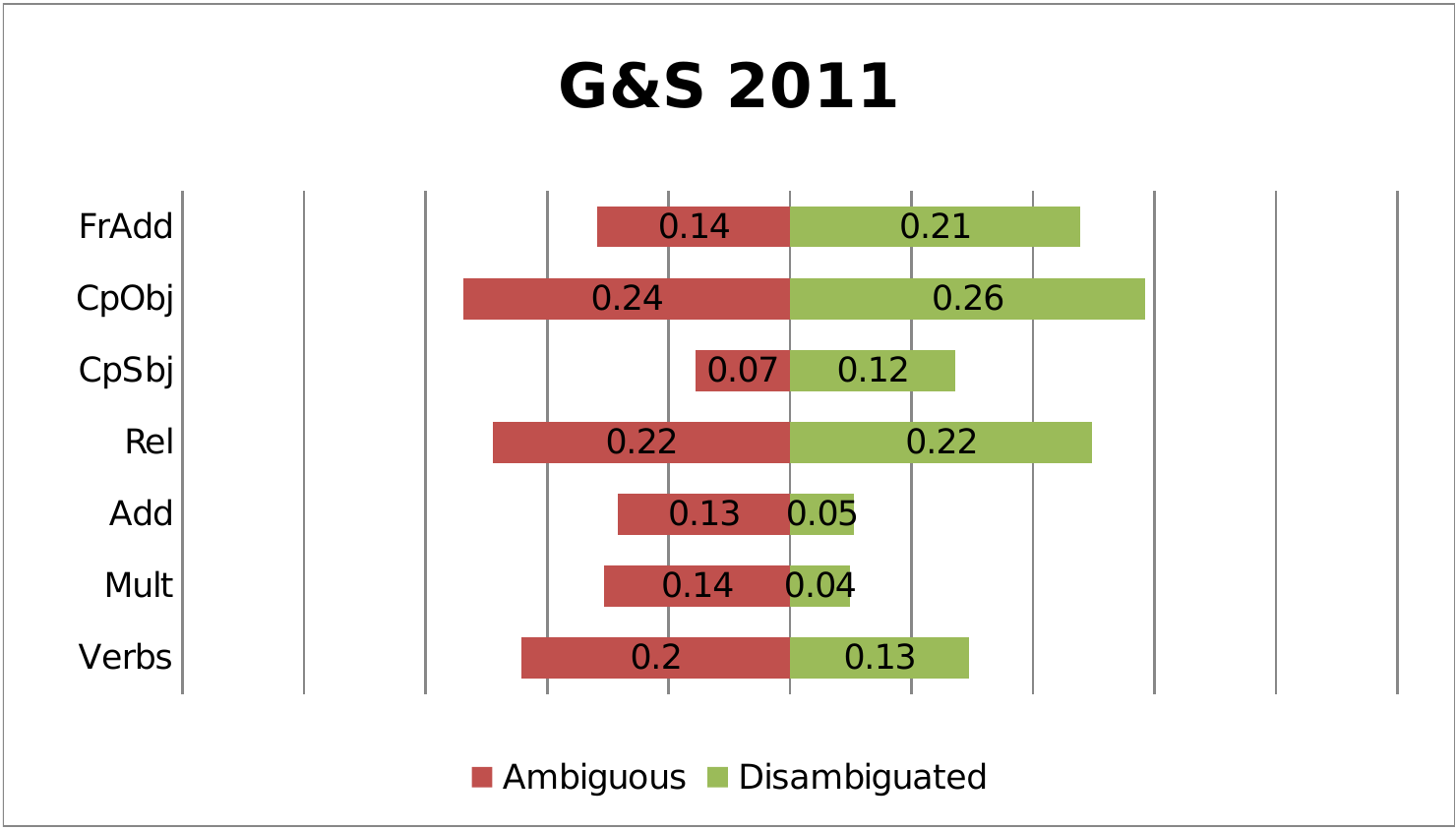}
  \end{minipage}
    \begin{minipage}{0.5\linewidth}
  \includegraphics[scale=0.5]{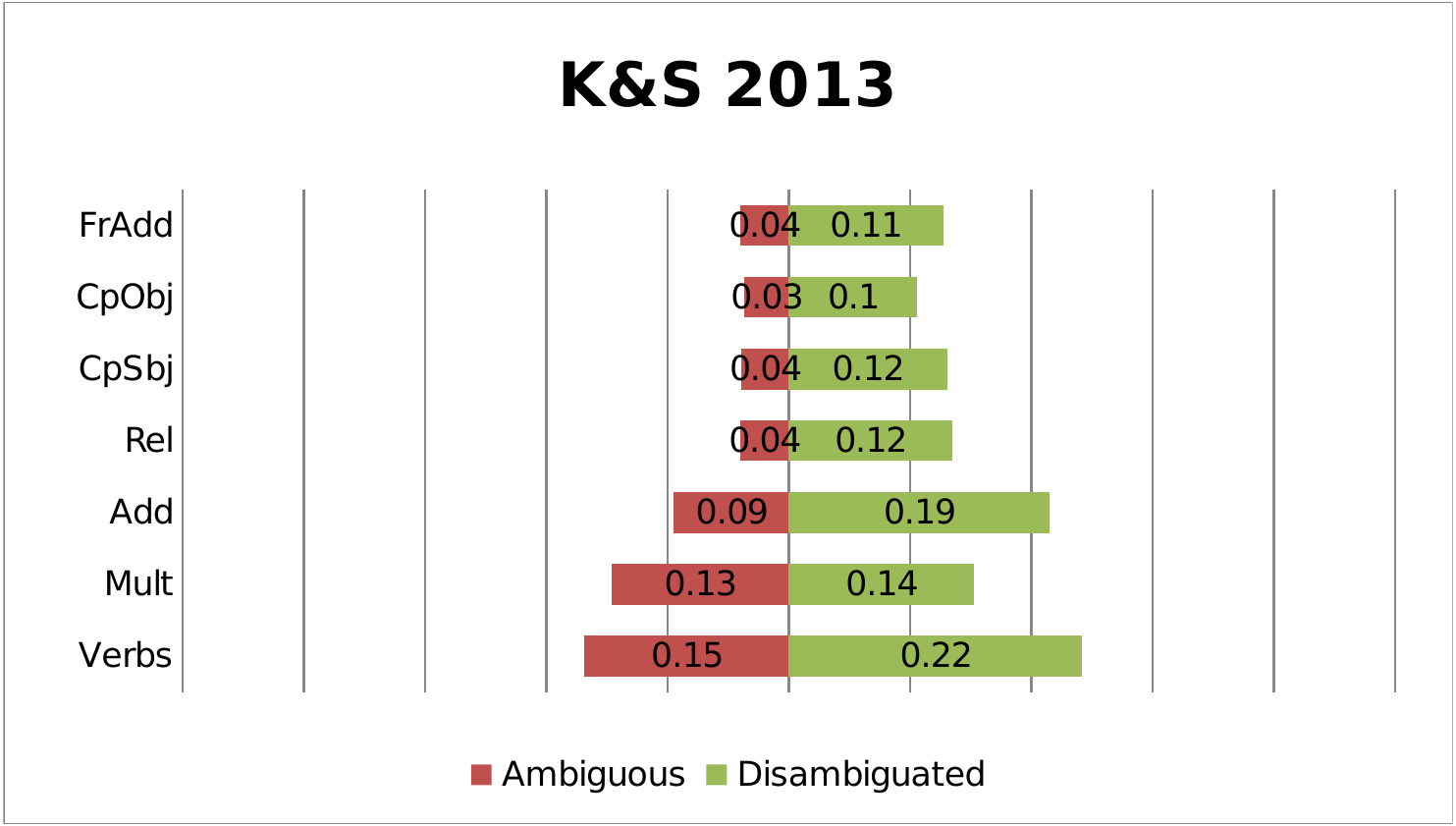}
  \end{minipage}
  
  \begin{minipage}{0.5\linewidth}
  \includegraphics[scale=0.5]{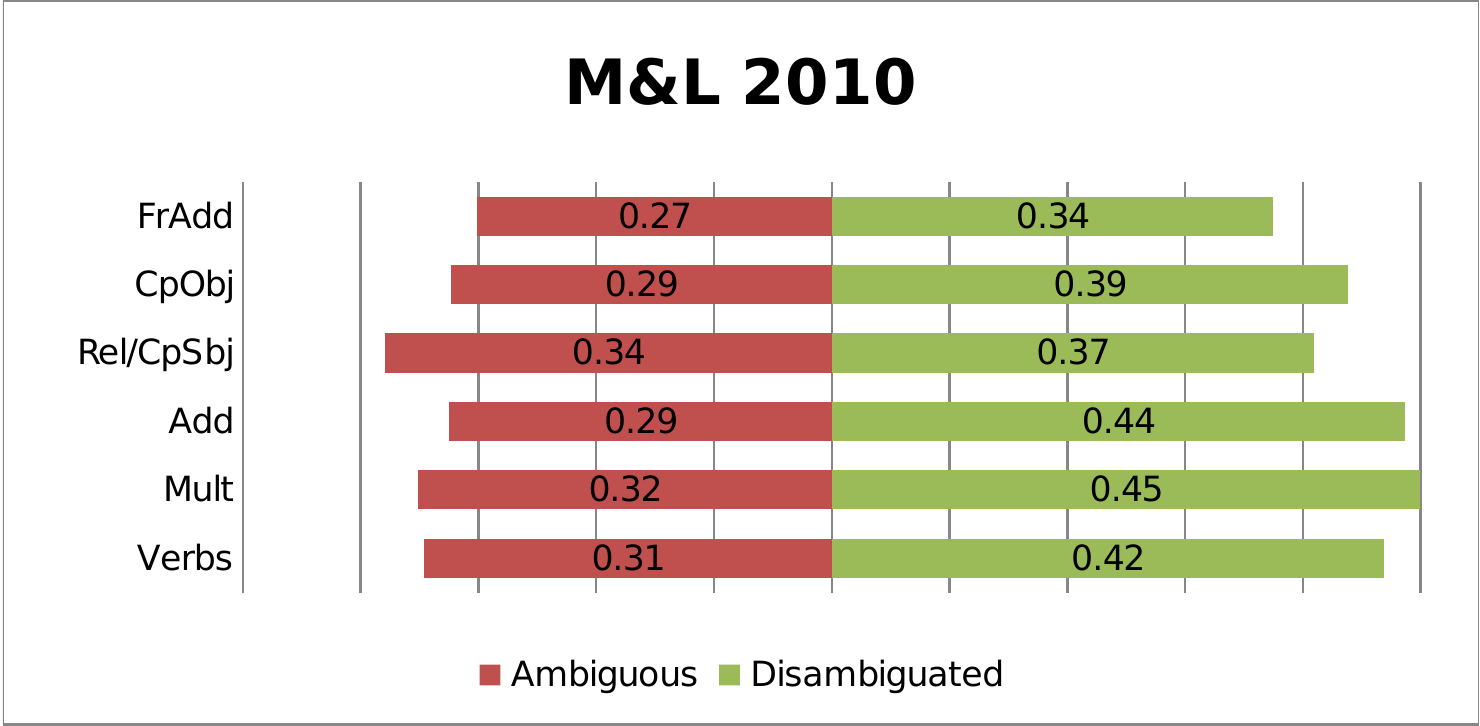}
  \end{minipage}
    \begin{minipage}{0.5\linewidth}
  \includegraphics[scale=0.5]{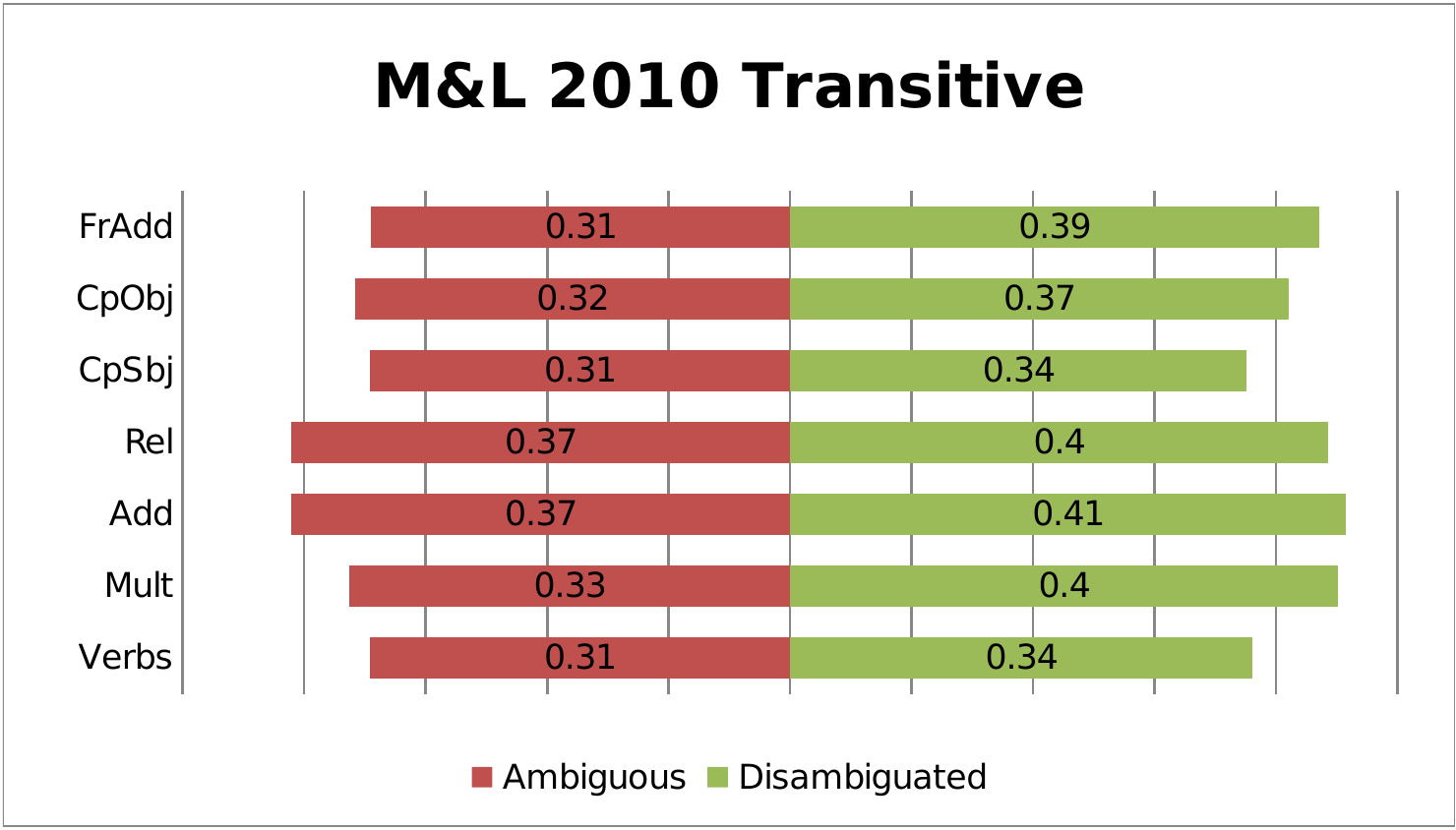}
  \end{minipage}
  
  \caption{Prior disambiguation in diagrams.}
  \label{fig:diagrams}
\end{figure}

The experimental work presented here allows us to say with enough confidence that disambiguation as an additional step prior to composition is very beneficial for Frobenius models. Furthermore, it confirms and strengthen previous work \cite{reddy2011,kartsaklis:2013:CoNLL} that showed better performance of disambiguated vector mixture models compared to their ambiguous versions. Fig. \ref{fig:diagrams} provides a summary of the results in the form of diverging stacked bar graphs. 

The positive effect of disambiguation is more evident for the vector mixture models (especially for the additive model) than for the Frobenius ones. Perhaps this is not surprising: composite representations created by element-wise operations are averages, and the provision of more accurate starting points by a prior step of disambiguation can make a great difference.\index{vector mixtures!and prior disambiguation}

From a task perspective, the effect of disambiguation was much more definite in the phrase/sentence similarity task. This observation is really interesting, since the words of that dataset were \textit{not} selected in order to be ambiguous in any way. The superior performance of the disambiguated models, therefore, implies that the proposed methodology can improve tasks based on phrase or sentence similarity \textit{regardless} of the level of ambiguity in the vocabulary. For these cases, the proposed disambiguation algorithm acts as a fine-tuning process, the outcome of which seems to be always positive; it can only produce better composite representations, not worse.\index{prior disambiguation!as fine-tuning process} 

%

For disambiguation-based tasks similar to those of \S\ref{sec:frob-verbdis}, the effect of prior disambiguation seems more positive for homonymous cases than for polysemous ones; hence the inferior performance of disambiguated models in the G\&S 2011 dataset, compared to the K\&S 2013 dataset. In fact, the G\&S 2011 dataset was the only one where disambiguation was not helpful for some cases (specifically, for vector mixtures). Although the better performance under the presence of genuinely ambiguous words is rather expected, a comparison of the results between the two datasets should also take into account that the automatic selection of landmark verbs using the JCN metric (as done with the G\&S 2011 dataset) was not very balanced for certain cases. Note, for example, that the bare baseline of comparing just \textit{ambiguous} versions of verbs (without any composition) in that dataset already achieves a very high correlation of 0.20 with human judgements (Table \ref{tbl:gands}).\footnote{The reported number for this baseline by \cite{GrefenSadr1} was 0.16 using vectors trained from BNC.} This number is only 0.15 for the K\&S 2013 dataset, due to the more effective verb selection procedure. The Only Verbs baseline can be seen as a ``sanity test'' for the specific task, given its design; putting composition aside, one would expect that comparing \textit{disambiguated} versions of the verbs alone will always produce better results than those produced by comparing ambiguous ones. This sanity test is passed only by the K\&S 2013 dataset.

Finally, the results presented in \S \ref{sec:frob-vpsim} imply that the effectiveness of vector mixture and Frobenius models depends to some extent on the length of the text segment: when more words are involved, vector mixture models tend to be less effective; on the contrary, the performance of tensor-based models seems to be proportional to the length of the phrase or sentence---the more, the better. These observations comply with the nature of the approaches: ``averaging'' larger numbers of points results in more general (hence less accurate) representations; on the other hand, a larger number of arguments makes a function (such as a verb) more accurate.

\section{Experiments on full tensors}
\label{sec:wsdfull}

\index{prior disambiguation!on full tensors|(}

The experiments in \S\ref{sec:wsdfrob} provide evidence that a prior disambiguation step on word vectors before the composition is beneficial for vector mixture models and partial tensor-based models that follow the Frobenius framework of Chapter \ref{ch:frobverbs}. The purpose of the current section is to generalize the results for the case of full tensor-based models, where the tensor of a verb follows faithfully the grammar type without any application of Frobenius operators. In order to do that, I apply the linear regression method described in \S\ref{sec:frob-regression} for constructing ambiguous and disambiguated tensors of order 2 (that is, matrices) for verbs taking one argument. In principle, the methods discussed here are directly applicable to tensors of higher order, assuming one has some way to construct them efficiently.\footnote{For example, Grefenstette et al. \cite{GrefSadrBarIWCS13} create order-3 tensors for transitive verbs using a multi-step procedure based on linear regression.}

We will concentrate on elementary verb phrases of the form \textit{verb-object} (e.g. `play football', `admit student'), since in general objects comprise stronger contexts for disambiguating the usage of a verb. In order to create a matrix for, say, the verb `play', I first collect all instances of the verb occurring with some object in the training corpus, and then I create non-compositional vectors for these elementary sentences (to which I will refer as \textit{holistic} vectors) following exactly the same methodology as if they were words. We now have a dataset with instances of the form $\langle \ov{obj_i}, \ov{play~obj_i} \rangle$ (e.g. the vector of `flute' paired with the holistic vector of `play flute', and so on), that can be used to train a linear regression model in order to produce an appropriate matrix for verb `play'. As discussed in \S\ref{sec:frob-regression}, the premise of a model like this is that the multiplication of the verb matrix with the vector of a new object will produce a result that approximates the distributional behaviour of all these elementary two-word exemplars used in training. The work of this section has been presented in the context of ACL 2014 \cite{kartsaklis:2014:ACL} (joint with Kalchbrenner and Sadrzadeh).


\subsection{Experimental setting}
\label{sec:setting}

\index{CDMs!and linear regression|(}
\index{linear regression, as composition method|(}

Linear regression in general can impose long training times, so the word vectors that are going to be used as input to the algorithm must be relatively small. For this reason, this time I use a 300-dimensional semantic space that has been produced by application of SVD.\index{singular value decomposition} In order to compute the weights of the vectors, I used local mutual information (LMI), that is, the original counts multiplied by PMI. From almost every other aspect the creation of the vector space follows the methodology of \S\ref{sec:semspace}. In order to create the vector space of the holistic verb phrase vectors, I first collected all instances where a verb participating in the experiments appeared at least 100 times in a verb-object relationship with some noun in the corpus. As context of a verb phrase I considered any content word that falls into a 5-word window from either side of the verb \textit{or} the object. For the 68 verbs participating in the experiments, this procedure resulted in 22,000 verb phrases, a vector space that again was projected into 300 dimensions using SVD.\index{distributional hypothesis!on multi-word expressions}

For each verb I use linear regression with gradient descent directly applied on matrices \textbf{X} and \textbf{Y}, where the rows of \textbf{X} correspond to vectors of the nouns that appear as objects for the given verb and the rows of \textbf{Y} to the holistic vectors of the corresponding verb phrases. Our objective function then becomes:

\begin{equation}
 \hat{\textbf{W}} =  \underset{\textbf{W}}{\arg\min} \frac{1}{2m} \left( \Vert \textbf{W} \textbf{X}^{\mathsf{T}} - \textbf{Y}^{\mathsf{T}} \Vert^2 + \lambda \Vert \textbf{W} \Vert^2 \right)
\end{equation}

\noindent where $m$ is the number of training examples and $\lambda$ a regularization parameter. The  matrix \textbf{W} is used as the tensor for the specific verb.

\index{linear regression, as composition method|)}
\index{CDMs!and linear regression|)}

\subsection{Supervised disambiguation}
\label{sec:regr-sup}

\index{WSD!supervised|(}

The first experiment tests the effectiveness of a prior disambiguation step for a tensor-based model using supervised learning. The goal is to create composite vectors for a number of elementary verb phrases of the form \textit{verb-object} with and without an explicit disambiguation step, and evaluate which model approximates better the holistic vectors of these verb phrases. 

\begin{table}[b!]
  \small
  \centering
  \begin{tabular}{l|ll}
   \hline
    \textbf{Verb}  & \textbf{Meaning 1} & \textbf{Meaning 2} \\
   \hline\hline
    break & violate (56) & break (22) \\
    catch & capture (28) & be on time (21) \\
    play  & musical instrument (47) & sports (29) \\
    admit & permit to enter (12) & acknowledge (25) \\
    draw  & attract (64) & sketch (39)  \\
    \hline
  \end{tabular}
  \caption[List of ambiguous verbs for the supervised task in the linear regression experiment.]{Ambiguous verbs for the supervised task. The numbers in parentheses refer to the collected training examples for each case.}
  \label{tbl:ambverbs}
\end{table}

The verb phrases of our dataset are based on the 5 ambiguous verbs of Table \ref{tbl:ambverbs}. Each verb has been combined with two different sets of nouns that appear in a verb-object relationship with that verb in the corpus (a total of 343 verb phrases). The nouns of each set have been manually selected in order to explicitly represent a different meaning of the verb. As an example, in the verb `play' I impose the two distinct meanings of using a musical instrument and participating in a sport; so the first set of objects  contains nouns such as `oboe', `piano', `guitar', and so on, while in the second set we see nouns such as `football', 'baseball'' etc. 

In more detail, the creation of the dataset was done in the following way: First, all verb entries with more than one definition in the Oxford Junior Dictionary \cite{OxfordJun} were collected into a list. Next, a hired linguist (who was a native speaker of English) annotated the semantic difference between the definitions of each verb in a scale from 1 (similar) to 5 (distinct). Only verbs with definitions exhibiting completely distinct meanings (marked with 5) were kept for the next step. For each one of these verbs, a list was constructed with all the nouns that appear at least 50 times under a verb-object relationship in the corpus with the specific verb. Then, each object in the list was manually annotated as \textit{exclusively} belonging to one of the two senses; so, an object could be selected only if it was related to a single sense, but not both. For example, `attention' was a valid object for the \textit{attract} sense of verb `draw', since it is unrelated to the \textit{sketch} sense of that verb. On the other hand, `car' is not an appropriate object for either sense of `draw', since it could appear under both of them in different contexts. The verbs of Table \ref{tbl:ambverbs} were the ones with the highest numbers of exemplars per sense, creating a dataset of significant size for the intended task (each holistic vector is compared with 343 composite vectors).

The experiment proceeds as follows: I apply linear regression in order to train verb matrices using jointly the object sets for both meanings of each verb, as well as separately---so in this latter case we get two matrices for each verb, one for each sense. For each verb phrase, I create a composite vector by matrix-multiplying the verb matrix with the vector of the specific object. Then I use 4-fold cross validation to evaluate which version of composite vectors (the one created by the ambiguous tensors or the one created by the unambiguous ones) approximates better the holistic vectors of the verb phrases in our test set. This is done by comparing each holistic vector with all the composite ones, and then evaluating the rank of the correct composite vector within the list of results. 

In order to get a proper mixing of objects from both senses of a verb in training and testing sets, I set the cross-validation\index{cross-validation} process as follows: I first split both sets of objects in 4 parts. For each fold then, our training set is comprised of $\frac{3}{4}$ of set \#1 plus $\frac{3}{4}$ of set \#2, while the test set consists of the remaining $\frac{1}{4}$ of set \#1 plus $\frac{1}{4}$ of set \#2. The data points of the training set are presented in the learning algorithm in random order.

I measure approximation in three different metrics. The first one, accuracy, is the strictest, and evaluates in how many cases the composite vector of a verb phrase is the closest one (the top one in the result list) to the corresponding holistic vector. A more relaxed and perhaps more representative method is to calculate the \textit{mean reciprocal rank} (MRR),\index{mean reciprocal rank (MRR)} which is given by:

\begin{equation}
 \text{MRR} = \frac{1}{m} \sum\limits_{i=1}^m\frac{1}{\textit{rank}_i}
\end{equation}

\noindent where $m$ is the number of objects and \textit{rank}$_i$ refers to the rank of the correct composite vector for the $i$th object. Finally, a third way to evaluate the performance of each model is to simply calculate the average cosine similarity between every holistic vector and its corresponding composite vector:

\begin{equation}
  \text{SIM}_{avg} = \frac{1}{m} \sum\limits_{i=1}^{m} \cos(\ov{verb~obj}_i,\overline{verb} \times \ov{obj}_i)
\end{equation}
\normalsize
\vspace{-0.3cm}

The results are presented in Table \ref{tbl:exp1results}, reflecting a clear superiority ($p < 0.001$ for average cosine similarity) of the prior disambiguation method for every verb and every metric.

\begin{table}[h]
  \small
  \centering
  \begin{tabular}{l|cc|cc|cc}
    \hline
     & \multicolumn{2}{|c|}{\textbf{Accuracy}} & \multicolumn{2}{|c|}{\textbf{MRR}} & \multicolumn{2}{|c}{\textbf{Avg Sim}} \\
    \hline
     & Ambiguous & Disamb. & Ambiguous & Disamb. & Ambiguous & Disamb. \\
    \hline\hline
    break & 0.19 & 0.28 & 0.41 & 0.50 & 0.41 & 0.43 \\
    catch & 0.35 & 0.37 & 0.58 & 0.61 & 0.51 & 0.57 \\
    play  & 0.20 & 0.28 & 0.41 & 0.49 & 0.60 & 0.68 \\
    admit & 0.33 & 0.43 & 0.57 & 0.64 & 0.41 & 0.46 \\
    draw  & 0.24 & 0.29 & 0.45 & 0.51 & 0.40 & 0.44 \\
    \hline
  \end{tabular}
  \caption[Results for the supervised task of the linear regression experiment]{Results for the supervised task.}
  \label{tbl:exp1results}
\end{table}

\index{WSD!supervised|)}

\subsection{Unsupervised disambiguation}
\label{sec:regr-unsup}

\index{WSD!unsupervised|(}

In \S \ref{sec:regr-sup} I used a controlled procedure to collect genuinely ambiguous verbs and I trained my models from manually annotated data. In this section I outline how the process of creating tensors for distinct senses of a verb can be automated, and I test this idea on the generic verb phrase similarity task introduced in \S\ref{sec:phrase-sim}. 

The first thing we have to do is to use the unsupervised learning method detailed in \S\ref{sec:clustering} in order to detect the latent senses of each verb in the corpus. The next step is to classify every noun that has been used as an object of that verb to the most probable verb sense, and then use these sets of nouns as before for training tensors for the various verb senses. Being equipped with a number of sense clusters created as above for every verb, the classification of each object to a relevant sense is based on the cosine distance of the object vector from the centroids of the clusters (again, a direct application of Eq. \ref{equ:wsd}). Every sense with less than 3 training exemplars is merged to the dominant sense of the verb. The union of all object sets is used for training a single unambiguous tensor for the verb. As usual, data points are presented to the learning algorithm in random order, and no objects in the test set are used for training.


The experiment has the following form: For every pair of verb phrases, I construct composite vectors for the phrases and then I evaluate their cosine similarity. For the ambiguous regression model, the composition is done by matrix-multiplying the ambiguous verb matrix (learned by the union of all object sets) with the vector of the noun. For the disambiguated version, I first detect the most probable sense of the verb given the noun, again by comparing the vector of the noun with the centroids of the verb clusters; then, I matrix-multiply the corresponding unambiguous tensor created exclusively from objects that have been classified as closer to this specific sense of the verb with the vector of the noun. I also test the usual baselines (Only Verbs, Additive, Multiplicative).


\begin{table}[b!]
  \small
  \centering
  \begin{tabular}{l|c}
    \hline
    \textbf{Model} & \textbf{Spearman's $\rho$} \\
    \hline\hline
    Only Verbs &   0.331    \\
    Additive & 0.379 \\
    Multiplicative & 0.301 \\
    Linear regression (ambiguous) & 0.349 \\
    Linear regression (disambiguated) & \textbf{0.399} \\
    \hline
    Holistic verb phrase vectors & 0.403 \\
    \hline
    Human agreement & 0.550 \\
    \hline
  \end{tabular}
  \caption[Results for the unsupervised task of the linear regression experiment.]{Results for the phrase similarity task. The difference between the ambiguous and the disambiguated version is s.s. with $p < 0.001$.}
  \label{tbl:exp2results}
\end{table}    

The results are presented in Table \ref{tbl:exp2results}, where again the version with the prior disambiguation step shows performance superior to that of the ambiguous version. There are two interesting observations that can be made on the basis of Table \ref{tbl:exp2results}. First and foremost, the regression model is based on the assumption that the holistic vectors of the exemplar verb phrases follow an ideal distributional behaviour that the model aims to approximate as close as possible. The results of Table \ref{tbl:exp2results} confirm this: using just the holistic vectors of the corresponding verb phrases (no composition is involved here) returns the best correlation with human annotations (0.403), providing a proof that the holistic vectors of the verb phrases are indeed reliable representations of each verb phrase's meaning. Next, observe that the prior disambiguation model approximates this behaviour very closely (0.399) on unseen data, with a difference \textit{not} statistically significant. This is very important, since a regression model can only perform as well as its training dataset allows it---and in our case this is achieved to a very satisfactory level.

\index{WSD!unsupervised|)}
\index{prior disambiguation!on full tensors|)}

\section{Prior disambiguation and deep learning}
\label{sec:wsd-nn}

\index{prior disambiguation!and deep learning|(}
\index{deep-learning models!and prior disambiguation|(}

Throughout this chapter I have experimentally investigated (and verified) the prior disambiguation methodology in almost every compositional model class in the hierarchy of Fig. \ref{fig:rev-taxonomy}. This section deals with the last (and theoretically most powerful) class, that of deep learning models. In joint work with Cheng and Grefenstette \cite{cheng2014}, we show that one can gain certain benefits even in the case where composition is achieved by the means of a recursive neural network, as this was described in \S\ref{sec:deeplearning}. The effectiveness of prior disambiguation on deep learning compositional models is tested on the same three datasets of \S\ref{sec:wsdfrob}, used for evaluating the disambiguated Frobenius models. In terms of neural compositional models, we implement a RecNN and a RAE (more details about these models can be found in \S\ref{sec:deeplearning}).\index{recursive neural networks}\index{recursive auto-encoders} Furthermore, we use additive and multiplicative models as baselines. 

For each dataset and each model, the evaluation is conducted in two ways. First, we measure the Spearman's correlation between the computed cosine similarities of the composite sentence vectors and the corresponding human scores, as in the previous sections of this chapter. Second, we apply a more relaxed evaluation, based on a binary classification task. Specifically, we use the human score that corresponds to each pair of sentences in order to decide a label for that pair (1 if the two sentences are highly similar and 0 otherwise), and we use the training set that results from this procedure as input to a logistic regression classifier. As a measure of the matching rate, the 4-fold cross validation accuracy is reported. The results for each dataset and experiment are listed in  Tables~\ref{tbl:nn-gs1}--\ref{tbl:nn-ml2}.

\begin{table}[ht!]
  \centering
    \small
\begin{tabular}{l|ccc}
  \hline
  \multicolumn{4}{c}{\textbf{Spearman's correlation}} \\
  \hline
    & Ambiguous & Disambig. every word & Disamb. only verbs \\
  \hline
  Additive & \textbf{0.221} & 0.071 & 0.105 \\
  Multiplicative & 0.085 & 0.012 & 0.043 \\
  RecNN & 0.127 & 0.119 & 0.128 \\
  RAE & 0.124 & 0.098 & 0.126 \\
  \hline\hline
  \multicolumn{4}{c}{\textbf{Cross validation accuracy}} \\
  \hline
  Additive  & 63.07\% & 63.08\% & 62.48\% \\
  Multiplicative & 61.89\% & 59.20\% & 60.11\% \\
  RecNN & 62.66\% & 63.53\% & \textbf{66.19\%} \\
  RAE & 63.04\% & 60.51\% & 65.17\% \\
  \hline
\end{tabular}
  \caption[Prior disambiguation results for deep learning models (G\&S 2011 dataset).]{Results for G\&S 2011 dataset.}\label{tbl:nn-gs1}
\end{table}

\begin{table}[ht!]
  \centering
    \small
\begin{tabular}{l|ccc}
  \hline
  \multicolumn{4}{c}{\textbf{Spearman's correlation}} \\
  \hline
    & Ambiguous & Disambig. every word & Disamb. only verbs \\
  \hline
  Additive & 0.132 & \textbf{0.152} & 0.147 \\
  Multiplicative & 0.049 & 0.129 & 0.104 \\
  RecNN & 0.085 & 0.098 & 0.101 \\
  RAE & 0.106 & 0.112 & 0.123 \\
  \hline\hline
  \multicolumn{4}{c}{\textbf{Cross validation accuracy}} \\
  \hline
  Additive & 49.28\% & 51.51\% & 51.04\% \\
  Multiplicative & 49.76\% & 52.37\% & 53.06\% \\
  RecNN & 51.37\% & 52.64\% & \textbf{59.26\%} \\
  RAE & 50.92\% & 53.35\% & 59.17\% \\
  \hline
\end{tabular}
  \caption[Prior disambiguation results for deep learning models (K\&S 2013 dataset).]{Results for K\&S 2013 dataset.}\label{tbl:nn-ka2}
\end{table}

\begin{table}[ht!]
  \centering
    \small
\begin{tabular}{l|ccc}
  \hline
  \multicolumn{4}{c}{\textbf{Spearman's correlation}} \\
  \hline
    & Ambiguous & Disamb. every word & Disamb. only verbs \\
  \hline
  Additive & 0.379 & \textbf{0.407} & 0.382 \\
  Multiplicative & 0.301 & 0.305 & 0.285 \\
  RecNN & 0.297 & 0.309 & 0.311 \\
  RAE & 0.282 & 0.301 & 0.303 \\
  \hline\hline
  \multicolumn{4}{c}{\textbf{Cross validation accuracy}} \\
  \hline
  Additive  & 56.88\% & 59.31\% & 58.26\% \\
  Multiplicative & 59.53\% & 59.24\% & 57.94\% \\
  RecNN & 60.17\% & 61.11\% & \textbf{61.20\%} \\
  RAE & 59.28\% & 59.16\% & 60.95\% \\
  \hline
\end{tabular}
  \caption[Prior disambiguation results for deep learning models (M\&L 2010 dataset).]{Results for M\&L 2010 dataset.}\label{tbl:nn-ml2}
\end{table}

Once more the results are quite promising, since they suggest that disambiguation as an extra step prior to composition can bring at least marginal benefits to deep learning compositional models. Note that the performance of the models follows a pattern very similar to that observed in \S\ref{sec:wsdfrob}, where we evaluated the prior disambiguation hypothesis on the Frobenius models. Specifically, the effect of disambiguation is clearest for the M\&L 2010 dataset, which does not contain any words chosen to be ambiguous on purpose; in both evaluations we carried out, disambiguation has a positive effect for the subsequent composition. For one more time, prior disambiguation acts as a useful pre-processing step, which might improve the final outcome (if the sentence has ambiguous words) or not (if all words are unambiguous), but never decrease the performance.\index{prior disambiguation!as fine-tuning process} The effect of disambiguation seems also to be quite clear for the K\&S 2013 dataset, whereas the result for the G\&S 2011 dataset, although positive, is less definite due to the reasons discussed in \S\ref{sec:frobexp-discussion}. 

\index{deep-learning models!and prior disambiguation|)}
\index{prior disambiguation!and deep learning|)}

\section{Assessing the results}
\label{sec:wsd-gendiscus}

The work of this chapter clearly suggests that explicitly dealing with lexical ambiguity in the context of a CDM is a safe way to increase the performance of the model. Especially encouraging is the fact that the prior disambiguation methodology seems to work very well in general cases, such as the M\&L 2010 task, where the majority of words  are unambiguous or express small deviations in their meaning. This confirms the hypothesis I posed in \S\ref{sec:priordis}, regarding the potential effect of the prior disambiguation step on polysemous or unambiguous words: since the sense vectors of such a word are very similar to each other (and to the original ambiguous vector), the extent to which the final composite vector is affected remains low. As we move from polysemous to homonymous cases and deviations in meaning get higher, the prior disambiguation step starts to contribute positively to the composite representation of the phrase or sentence.

\begin{figure}[h!]
 \centering
 \includegraphics[scale=0.75]{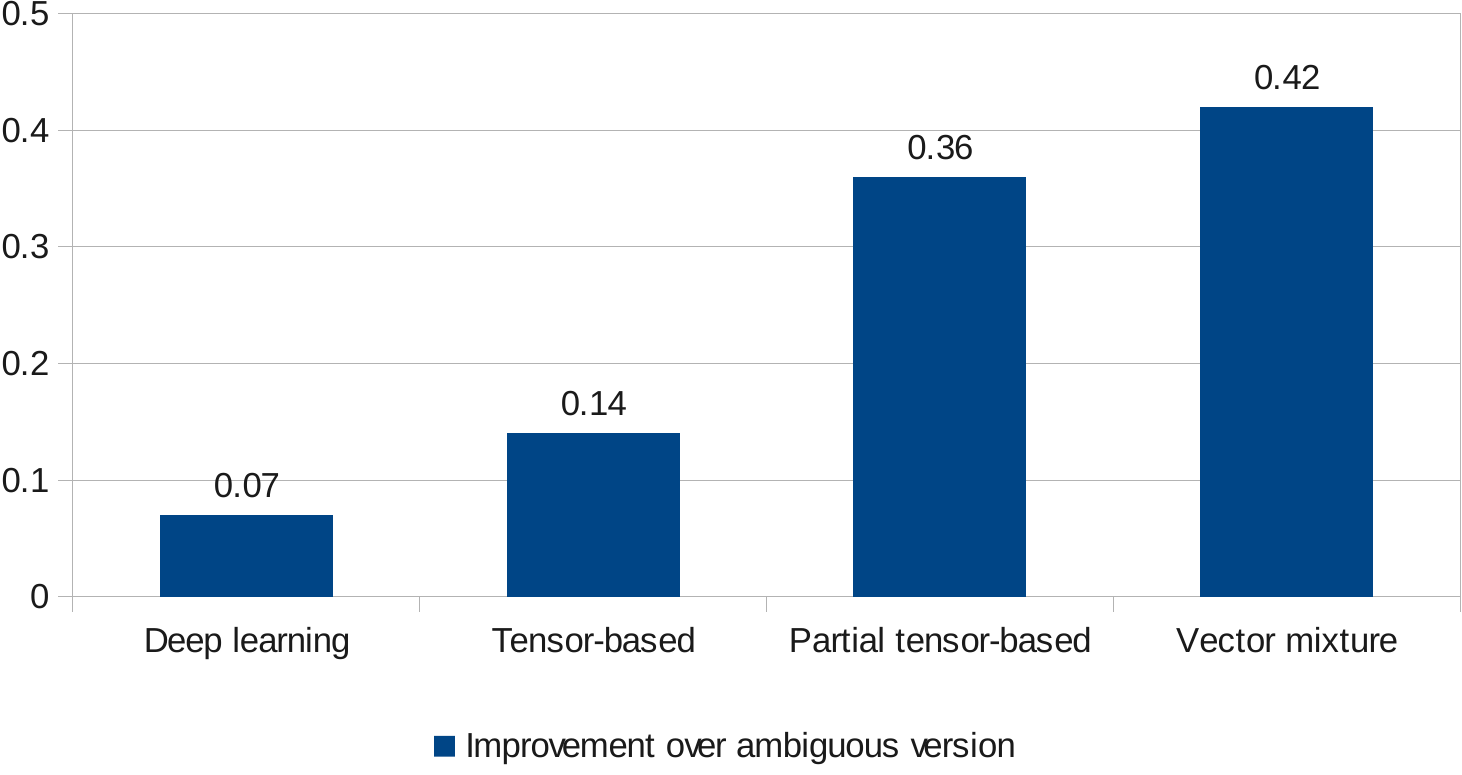}
 \caption{General improvement due to prior disambiguation per CDM class for the M\&L 2010 dataset.}
 \label{fig:wsd-impr}
\end{figure}

Interestingly, this observation seems to hold for every class of CDM in the hierarcy of \S\ref{sec:rev-taxonomy}, although up to a different degree. Table \ref{tbl:vp-comp} provides a summary of the results across models for the M\&L 2010 task, which shows that the improvement we get in each case is inversely proportional to the theoretical power of the model. This is more clear in Fig. \ref{fig:wsd-impr}, where this improvement is expressed as percentage over the performance of the ambiguous version.

\begin{table}[h!]
  \small
  \centering
  \begin{tabular}{l|ccc}
    \hline
    \textbf{Model} & \textbf{Ambiguous} & & \textbf{Disambiguated} \\
    \hline\hline
    Deep learning & 0.282 & $\ll$ & 0.301 \\
    Tensor-based & 0.349 & $\ll$ & 0.399 \\
    Partial-tensor based & 0.290 & $\ll$ & 0.393 \\
    Vector mixture & 0.315 & $\ll$ & 0.448 \\
    \hline
  \end{tabular}
  \caption[Summary of disambiguation results for M\&L 2010 dataset across CDM classes.]{Summary of disambiguation results for M\&L 2010 dataset across CDM classes. The numbers of each class refer to Spearman's correlation for models maximally benefited from the prior disambiguation step.}
  \label{tbl:vp-comp}
\end{table}


While the effect of prior disambiguation is minimal for the deep learning model, it is doubled for the tensor-based approach of \S\ref{sec:wsdfull}. Recall from the discussion in \S\ref{sec:intuition} that non-linear models are in a position to drastically transform the phrase or sentence space in a way that reflects better any underlying relationships. In principle, this means that a non-linear compositional model approximates the concept of continuous meaning (as expressed by a function that can directly act on ambiguous vectors efficiently without any prior discrimination of the senses) to a much higher degree than a linear one. 

%

\index{prior disambiguation!as means of non-linearity|(}

\begin{figure}[b!]
 \centering
  \includegraphics[scale=0.7]{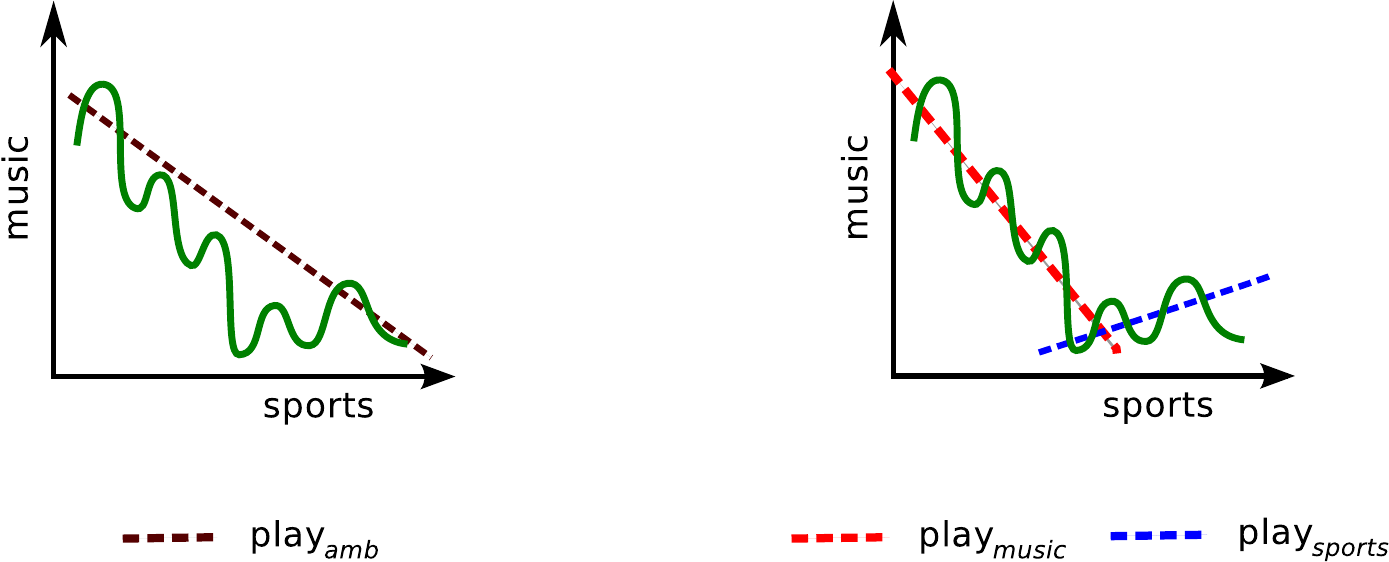}
  \caption{Prior disambiguation as a means of providing non-linearity to linear models.}
  \label{fig:tensor-nonlinear}
\end{figure}

This provides us some intuition for the success of prior disambiguation on tensor-based models: replacing a single linear model with two or more, offers a means of non-linearity that improves the performance of the model. This idea is demonstrated in Fig. \ref{fig:tensor-nonlinear}. Imagine that the green curve describes verb-object structures for the verb `play', such as ``play guitar'', ``play football'', and so on; naturally, we expect phrases of the former kind to be closer to base `music', and of the latter form to base `sports'. The dashed line in the left-hand diagram represents a single ambiguous tensor for verb `play' that aims to approximate this non-linear behaviour of the verb phrases. However, this can be done efficiently only by creating a separate linear model for every meaning, as in the right-hand diagram. Of course, the ability of a deep learning model to fit on such a curve would be greater in any case.

\index{prior disambiguation!as means of non-linearity|)}

Prior disambiguation offers even greater benefits for partial tensor-based models and vector mixtures, as it is shown in Fig. \ref{fig:wsd-impr}. The composition in both of these classes is based (only partially for the former class) on some form of averaging of the vectors of the words of the sentence or phrase. As discussed before, the contribution of the prior disambiguation step in this case is that it prevents the model from using very generic inputs, which will only produce even more generic and non-informative points at the sentence space as results.


\chapter{Future Work and Conclusions}
\label{ch:conclusions}

In this dissertation I presented various ways to improve and extend the categorical model of Coecke, Sadrzadeh and Clark \cite{Coeckeetal}. The implementation framework based on the Frobenius algebras constitutes a scalable and linguistically motivated base that provides extensive coverage of the language and a test-bed for creative experimentation. Furthermore, although in my exposition I mostly used as a starting point the simple argument summing procedure of \cite{GrefenSadr1}, the framework in principle is independent of any specific technique of creating the concrete verb tensors, assuming of course that the adopted approach will respect the multi-linear nature of the model. In \S\ref{sec:frob-regression}, for example, I briefly discussed how one can produce the required multi-linear maps using regression. 

The work contributed by this project, in conjunction with relevant research coming from other sources, allow us for the first time to abandon the fixed simple structures on which the categorical model has been evaluated until now and move on towards a large-scale evaluation conducted on sentences of arbitrary complexity. This is one of our first priorities for the future. An appropriate generic task towards this purpose is paraphrase detection, which has been used in the past for evaluating the performance of other classes of CDMs \cite{socher2012}, thus it allows the first cross-model direct comparison.

Admittedly, there are a lot of issues that still need to be addressed in the context of tensor-based models. One of the biggest problems for the moment is the proper treatment of logical aspects of language, such as negation, for which there still exists no satisfactory solution. Furthermore, concepts such as quantification seem to be incompatible with vector space semantics: a quantifier operates on a number of individual entities by counting or enumerating them: all men, some women, three cats, at least four days. This fits nicely in the logical view of formal semantics, but it is not clear how it can be applied in a distributional setting. For the next few years, it is expected that these issues will receive a lot of attention from researchers working on the theoretical aspects of CDMs. As an example, Rypacek and Sadrzadeh \cite{rypacek} made a first step towards modelling quantifiers in the context of the categorical framework. 

The quantum formulation of \S\ref{sec:quantum} provides many opportunities for further work, both at the theory and the practice side. For the latter case, the plan is to attempt a first evaluation of the setting using real-valued state vectors. In order to avoid the inevitable explosion of dimensionality that comes with density matrices, an experiment like that would have to be restricted to compositions between two words, such as the verb-object constructs used in \S\ref{sec:wsdfull}. The Frobenius framework will be once more a valuable ally towards this purpose; recall from \S\ref{sec:frob-density} that compositions of this kind reduce to the point-wise multiplication between the density matrices of the two words, significantly mitigating space complexity problems.

The new quantum perspective opens many new doors for exciting new research at the theory side as well. One interesting direction for us is to further investigate the role of entanglement in linguistics, a topic that has been briefly discussed in \S\ref{sec:entanglement}, but when seen in the context of quantum mechanics takes a new richer meaning (\S\ref{sec:entropy}). Furthermore, as discussed in \S\ref{sec:frob-density}, a very interesting topic with both theoretical and practical aspects would be the further investigation of alternative structures that conform to the definition of a Frobenius algebra in \textbf{CPM(FHilb)}, and the analysis and evaluation of the results from a linguistic perspective. Last, but not least, note that under certain conditions the density matrix formalism imposes a form of logic that potentially can provide solutions to the aforementioned problems related to logical aspects of language. Further work is needed for making these insights concrete and for evaluating their true potential.

An important general outcome of this thesis which is of great interest for any class of CDMs, is that explicitly dealing with lexical ambiguity improves the compositional process. Especially for multi-linear models, a prior disambiguation step can bring substantial improvements, due to the fact that it provides some means for them to approximate possibly non-linear relationships in the sentence space. Although the effect is smaller, the experimental work suggests that there exist certain benefits even in the case of deep learning models, despite their higher theoretical power. A reasonable future step is to investigate the hypothesis that incorporating the disambiguation step as an additional layer \textit{within} a deep learning compositional architecture can be an even more effective way to improve the quality of the produced vectors. In the simplest case, this can have the form of a \textit{competitive layer}, an idea that has been applied before for creating unambiguous neural vector spaces by Neelakantan et al. \cite{neelakantan}, although not in a compositional setting. A more interesting approach is to model the unsupervised part of the architecture as a \textit{self-organized map} (SOM) \cite{kohonen1982}, which in general can provide more fine-grained clusters. 

\subsection*{A closing remark}


In the recent years, compositional distributional models of meaning have been evolved to a very active area of research, attracting a lot of attention from the NLP community. The chances are that this trend will continue in the future. Admittedly, there is a lot of ground yet to be covered---to the extent that, for many people, the ultimate goal of capturing the meaning of a sentence in a computer's memory might currently seem rather utopian, something of only theoretical interest for the researchers to play with. That would be wrong; this is a technology with great potential that can drastically influence and improve the practice of natural language processing, and my hope for the work presented in this dissertation is that it positively contributes towards this purpose.

\mbox{}\newpage

\singlespacing
\addcontentsline{toc}{chapter}{Bibliography}
\bibliography{refs}        
\bibliographystyle{plain}  

\cleardoublepage
\phantomsection
\mbox{}\newpage
\addcontentsline{toc}{chapter}{Index}
\printindex

\end{document}